THE UNIVERSITY OF CHICAGO

# MEMORY PLANNING FOR DEEP NEURAL NETWORKS

SUBMITTED TO
THE FACULTY OF THE DIVISION OF THE PHYSICAL SCIENCES
IN CANDIDACY FOR THE DEGREE OF
MASTER OF SCIENCE

DEPARTMENT OF COMPUTER SCIENCE

BY
MAKSIM LEVENTAL

CHICAGO, ILLINOIS
WINTER 2022

# TABLE OF CONTENTS

# LIST OF FIGURES

# LIST OF TABLES

# ACKNOWLEDGMENTS

Portions of this work were done at Facebook, Inc., between June 2021 and February 2022, in close collaboration with Bin Bao, Nick Korovaiko, Elias Ellison, Don Jang, and Peng Wu.

# ABSTRACT

We study memory allocation patterns in DNNs during inference, in the context of large-scale systems. We observe that such memory allocation patterns, in the context of multi-threading, are subject to high latencies, due to `mutex` contention in the system memory allocator. Latencies incurred due to such `mutex` contention produce undesirable bottlenecks in user-facing services. Thus, we propose a "memorization" based technique, `MemoMalloc`, for optimizing overall latency, with only moderate increases in peak memory usage. Specifically, our technique consists of a runtime component, which captures all allocations and uniquely associates them with their high-level source operation, and a static analysis component, which constructs an efficient allocation "plan". We present an implementation of `MemoMalloc` in the PyTorch deep learning framework and evaluate memory consumption and execution performance on a wide range of DNN architectures. We find that `MemoMalloc` outperforms state-of-the-art general purpose memory allocators, with respect to DNN inference latency, by as much as 40%.

# CHAPTER 1
# INTRODUCTION

Deep neural networks (DNNs) are ubiquitous as components of research and production systems; they are employed to fulfill tasks across a broad range of domains, including image classification [5], object detection [72], speech recognition [6], and content recommendation [18]. Traditionally, DNNs are deployed to multi-processor (or multi-core processor) server-class platforms, such as those found in commercial data centers and scientific high-performance clusters. This is because DNNs, generally, are resource-intensive, in terms of compute, memory, and network usage; see Table 1.1 for representative DNN workloads at Facebook, Inc., a large social media services company that employs DNNs in many of its products.

Indeed, as a result of latency constraints imposed by quality-of-service guarantees, data center deployments usually target CPU architectures (and corresponding memory hierarchies), as opposed to GPGPU architectures [42]. This is a consequence of the fact that CPUs are better suited for low latency applications, owing to their high clock speeds and synchronous execution model, as opposed to GPUs, which typically have lower clock speeds and an asynchronous execution model. Further, new DNN techniques, such as Transformers [10] and Mixture-of-Experts [53], lead to networks with billions, or even trillions [22], of floating-point parameters (called *weights*), thus indicating (current) upper bounds on potential memory consumption; for instance, training BERT networks (a transformer) requires up to 16TB of memory [57]. Applying such complex DNNs effectively in high traffic services necessitates managing system resources carefully. To be specific, managing memory usage is important, both for preventing failures (such as out-of-memory conditions), and, as we discuss in the following, reducing latencies.

In this work, we focus on the implications of memory management for execution performance in server-class deployments of DNNs. It is well-known that in multithreaded environments, with many non-uniform service requests, heap synchronization routines can lead to blocking that inhibits scaling performance gains [9]. Specifically, we refer to contention on locks (i.e., `mutex`es) held to enforce mutual exclusion on code that modifies the heap data structure (i.e., `malloc` and `free`). The standard mitigation of such issues is replacing system `malloc` with a caching allocator such as `jemalloc` [21], `tcmalloc` [24], or SuperMalloc [32]. Caching allocators such as these alleviate lock contention by maintaining many independent heaps, each with its own `mutex`es, and distributing memory requests among them, thereby reducing pressure on any single lock. These allocators can be effective for many workloads and memory allocation patterns, but they are not a panacea. In the case of diverse DNN workloads on servers, where a process may exhibit $2 \times 10^7$ `malloc` requests per second, distributed across 2,000 concurrent threads [26], it is still possible for a program to experience significantly reduced performance due to lock contention. For DNNs with many allocation requests, spanning a wide range of sizes, this can readily be observed (see Section 2.2).

| Category | Model Type | Model Size (# params) | Typical Batch Size | Max # Live Activations | Latency (constraint) |
|---|---|---|---|---|---|
| Ranking | Linear | 1 - 10M | 1 - 100 | >10K | ∼ 10 ms |
| | Embedding | >10 billion | 1 - 100 | >10K | ∼ 10 ms |
| Vision | ResNet50 | 25M | 1 (image) | 2M | N/A |
| | ResNeXt-101-32x4 | 43 - 829M | 1 (image) | 2.4 - 29M | N/A |
| | FasterRCNN | 6M | 1 (image) | 13.2M | N/A |
| | ResNeXt3D-101 | 21M | 1 (movie clip) | 58M | N/A |
| Language | Seq2seq | 100M - 1B | 1 - 8 tokens | >100K | ∼ 10 ms |

Table 1.1: Resource requirements of representative DNN inference workloads implemented on CPU. Reprinted with permission from [42].

It is important to note that DNNs allocate memory in addition to that needed for just their weights; substantial temporary memory is associated with buffers (known as *tensors*) that correspond to intermediate results created during the evaluation of *layers* of the DNN. We observe that even with reasonable input sizes, the intermediate tensors of `resnext101_32x8d` [69] comprise 27% of the total 13GB run-time memory, 57% (of 760MB) for `squeezenet1_0` [30], and 66% (of 2473MB) for `mnasnet0_75` [64]. Similar figures have been reported in prior work [44]. These intermediates are often short-lived (serving only to propagate results between sequential *operations*) and overlap with only a small subset of the lifetimes of other intermediates. Thus, the effective memory needed to materialize the entire collection of intermediates is often much less than the sum total of the individual memories. Given foreknowledge of all lifetimes and sizes of intermediate tensors, and a *strategy* for computing corresponding offsets, memory can be allocated statically (or, at worst, just prior to inference). More importantly, as it pertains to performance, this single batch allocation effectively eliminates lock contention. Such an approach is called *static memory planning*, or *static allocation*. Unfortunately, due to pointer aliasing and control flow, comprehensive and robust lifetime and size data are difficult to derive statically (i.e., correctly, completely, and prior to any execution).

Hence, to reduce allocations while satisfying peak memory usage constraints, we propose a hybrid static-runtime memory management solution, called `MemoMalloc`, that makes use of both the statically known structure of the neural network and a single profiling pass. Specifically, our method uses a convenient representation of the neural network, along with lightweight stack tracing and pointer tagging, to reconstruct the lifetimes, sizes, and aliasing relationships of all intermediate tensors completely and accurately. Our system then constructs memory plans using one of several performant strategies. We present an implementation of the technique in the PyTorch [43] deep learning framework and evaluate our implementation on a large and representative set of DNNs. In terms of execution performance (as measured by latency) our solution outperforms PyTorch+`jemalloc` (i.e., PyTorch backed by the state-of-the-art caching allocator `jemalloc`). Specifically, across almost all input sizes and threading configurations (in terms of the number of threads) we observe, on average 20% lower inference latencies, and at best 40% lower latencies.

In summary, the principal contributions of this work are:

1. A study of the memory allocation patterns of a wide range of DNN architectures.

2. A study of several different exact and heuristic static allocation strategies, as they pertain to DNNs.

3. An implementation and evaluation of `MemoMalloc`, a system for managing memory for DNNs, which outperforms `jemalloc`.

The remainder of this thesis is organized as follows: Chapter 2 gives necessary background on representations of DNNs and memory allocators, along with a discussion of worst-case results concerning caching allocators and DNNs. Chapter 3 discusses our implementation, with a particular focus on how we resolve aliases exactly and performantly. Chapter 4 presents a thorough evaluation of our implementation, across various representative DNN architectures and workloads (in terms of input sizes and threading environment). Chapter 5 discusses the evaluation and the insights garnered thereof. Finally, Chapter 6 reviews prior work in this area and Chapter 7 concludes and discusses future work, including dynamics, training, GPUs, and applications to edge device deployments.

# CHAPTER 2
# BACKGROUND

We review the necessary background for our work. This includes a discussion of how DNNs are represented in deep learning frameworks (i.e., PyTorch) as it pertains to our manipulation of those representations. We then discuss the memory allocation issues addressed by caching allocators (including an empirical study of worst-case performance). Finally, we define memory planning formally and introduce the memory planning strategies that inform the design of the static memory planning component of `MemoMalloc`.

## 2.1 Representations of DNNs

Deep neural networks are typically specified using high-level frameworks that can be compiled into low-level platform and hardware specialized code. For example, TVM [16] generates highly optimized, hardware-specific code for various hardware backends by efficiently exploring the space of possible DNN transformations (specifically, with respect to kernel fusion). Such transformations are carried out on a representation of the DNN (Relay [50] of TVM, HLO of TensorFlow [34], TorchScript [2] of PyTorch) that captures the data and control flow dependencies between individual layers, as well as attributes of the data (i.e., tensors), such as type (e.g., `float32`, `int`, or `bfloat16`), memory layout (e.g., *contiguous*, *strided*, or *sparse*), and *shape*. Note that inputs to DNNs are characterized by their shape, i.e., the sizes of the dimensions of the input tensors, represented as arrays; a common shape corresponding to an image input for computer vision networks is $(N, C, H, W)$, with[1] corresponding size $N \times C \times H \times W \times size(\texttt{dtype})$, where $size(\texttt{dtype})$ is the width of the data type (e.g., 4 bytes for `float32`). This representation is called an *intermediate representation* (IR) since it functions as an intermediary between the high-level specification and the lower-level hardware characteristics.

TorchScript (TS) is a compiler infrastructure within the PyTorch deep learning framework that produces a type-annotated, static single assignment (SSA) IR (called TS IR). TorchScript is executed using an interpreter attached to a Just in Time (JIT) optimizer and compiler. There are two ways to generate TS IR from a PyTorch specified DNN:

- `torch.jit.trace`, which executes a forward pass iteration of a DNN and records the PyTorch *operators* (corresponding to the conceptual layers that comprise the DNN) that are invoked, thus "freezing" the code path of the DNN and hence eliminating control flow;

- `torch.jit.script`, which analyzes the Python abstract syntax tree representation of the DNN and *lowers* it to TS IR.

1. $N, C, H, W$ correspond to the batch size, number of channels, height, and width of the input, respectively.

```python
class Net(torch.nn.Module):
    def __init__(self):
        super(Net, self).__init__()
        self.linear = nn.Linear(4, 4)
        self.relu = nn.ReLU()

    def forward(self, x, h):
        y = self.linear(x) + h
        y = self.relu(y)
        return y
```

Listing 1: Example neural network

```
graph(%x : Tensor, %h : Tensor):
  %6: int = prim::Constant[value=1]()
  %linear_weight: Float(4, 4, strides=[4, 1])
    = prim::Constant[value=<Tensor>]()
  %linear_bias: Float(4, strides=[1]) = prim::Constant[value=<Tensor>]()
  %11: Float(3, 4, strides=[4, 1]) = aten::linear(
    %x, %linear_weight, linear_bias
  )
  %12: Tensor = aten::add(%11, %h, %6)
  %13: Tensor = aten::relu(%12)
  return (%13)
```

Listing 2: TS IR representation of neural network in Listing 1

In this work we exclusively make use of the `torch.jit.trace` path. Consider the example neural network, specified as a PyTorch model, presented in Listing 1. Given an input tensor with shape $(3, 4)$, it is "traced" to the TS IR presented in Listing 2.

Within TS IR, identifiers on the left-hand sides of assignments are called *values*, and identifiers on the right-hand sides are the operators invoked during execution. As prescribed by SSA semantics, each value is assigned only once, and thus the TS IR representation permits a one-to-one mapping with a directed, acyclic, control and *data flow* graph (hence, the pairing of operator and output are considered a *node* in this graph). Note, as well, that all values have type annotations of varying levels of specificity; for example (cf. Listing 2), the *concrete* annotation `Float(3, 4, strides=[4, 1])` uniquely determines the size of the intermediate tensor %11 as $3 \times 4 \times size(\texttt{Float}) = 48$ bytes (`strides=[4, 1]` indicates the tensor is arranged contiguously in memory) while the *abstract* annotation `Tensor` indicates value %12's type cannot be determined until runtime. The TS compiler has facilities for traversing and transforming these representations of DNNs. In particular one can implement graph rewrite passes that arbitrarily insert, remove, and rearrange nodes. We make use of these facilities in our implementation to augment the IR with memory allocation nodes that are then executed by the TS JIT and effectuate the memory plan (see Chapter 3).

## 2.2 Caching Allocators and Lock Contention

Caching allocators [8] address performance issues with memory allocation and deallocation, at runtime. Specifically total memory usage (i.e., reduction of internal and external fragmentation of allocated memory), cache locality of sequences of allocations, and overall latency in allocating memory for complex objects. They accomplish their goals by caching recent allocations (typically for configurable lengths of time called *decay times*) in order to reduce the number of expensive system calls (`sbrk` and `mmap`). An implicit concern of allocators is the performance overhead of the use of the allocator itself. An allocator that allocates optimally (either in terms of cache locality or total usage) but does so at the cost of excessive blocking times per allocation is of questionable value for typical users.

In the context of multi-threaded applications running on multiprocessor systems, blocking occurs during synchronization to prevent race conditions on the cache data structures. Caching allocators balance these costs (against those associated with fragmentation) by deploying multiple, independently managed caches (called *arenas*) and distributing allocation requests among them (thereby reducing request service and synchronization pressure on any one cache). In principle, this solution is in direct contradiction with the stated aim of reducing fragmentation: many independently managed caches managed by a single caching allocator degenerate to the same fragmentation pattern as many independent non-caching allocators managing their own subsets of system memory. Thus, care must be taken with respect to large allocations (typical of DNNs) to prevent severe fragmentation (i.e., mixing of small and large allocations in the same regions of memory).

"Per-thread" caching allocators, such as `jemalloc`, `tcmalloc`, and SuperMalloc, support thread-specific caching, in addition to maintaining multiple caches (called, appropriately, *thread caches*). That is to say, they maintain unique caches for each live thread executing on a system. This enables those allocations that can be serviced by the thread cache to happen without any synchronization and therefore very efficiently. This leads to very fast allocation in the common case, but also increases memory usage and fragmentation since a fixed number of objects can persist in each thread cache over the course of the entire execution of the program [33]. Effectively, this is the same failure mode (writ small) as that which betides conventional caching allocators operating many caches. To account for such fragmentation, thread caches are usually configured to be quite small; the default thread cache for `jemalloc` is 32KB in size. In addition, as in the case of DNN workloads, it is common to instantiate a manually managed arena for "oversized" allocations that has no thread cache at all; typical allocation size thresholds for this oversized arena are 1MB, 2MB, or 4MB.

To further illustrate the challenge posed by memory allocation patterns in the context of DNN workloads, with respect to latency, we perform a worst-case analysis; we exercise some common networks with `jemalloc` as the allocator with no thread cache and a single arena for all allocations. To be precise, we execute ten iterations of a forward pass on inputs sized

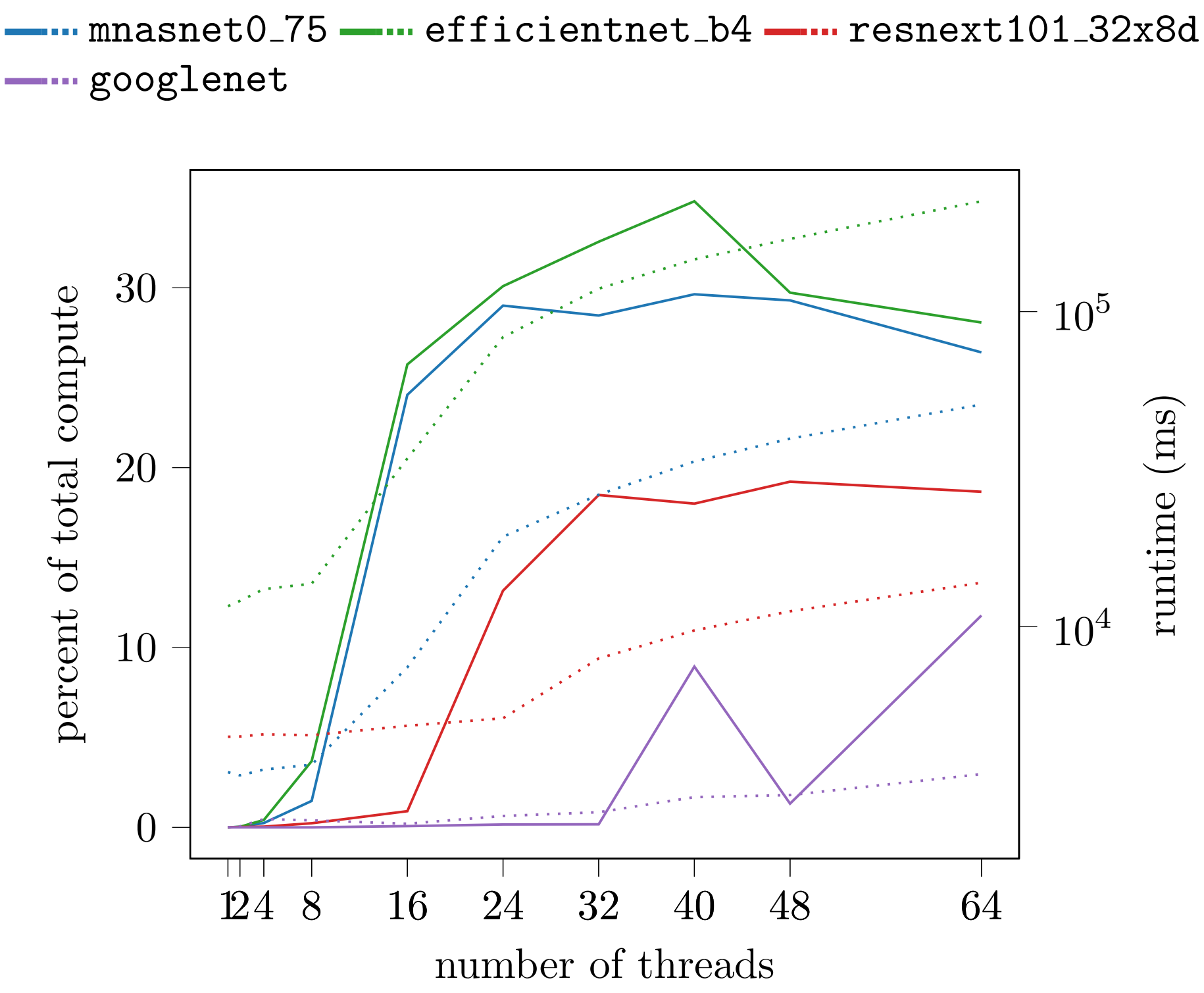

Figure 2.1: Total runtime (dashed line) and percent (solid line) of compute time spent in `malloc_mutex_lock_slow` as a function of the number of concurrent threads.

$(1, 3, 128, 128) \approx$ 192KB and record (using `perf`) time spent in `malloc_mutex_lock_slow` (a `jemalloc` utility function related to locking). See Figure 2.1. The result is that even at moderate concurrency (16 threads on our 32-core test platform; see Chapter 4) most iterations spend considerable time contending with locks. We can further investigate lock contention by collecting statistics on blocking wait times for lock acquisition (as recorded by `mutexes.ctl.total_wait_time` and `mutexes.ctl.max_wait_time`[2]). The results, shown in Figure 2.2, can be understood given consideration of the sizes and frequencies of the intermediate allocations made by these DNNs. We observe that the DNNs most affected make many allocations, most below 1MB (see Figure 2.3), and incur high request rates on `jemalloc` and locks related to those allocation sizes, evident from statistics on individual arena bins (`jemalloc` partitions arenas into bins of size $2^k$, and distributes allocations requests amongst those bins). We make use of this data to tune `jemalloc` during our evaluation (see Chapter 4).

2. http://jemalloc.net/jemalloc.3.html#tuning

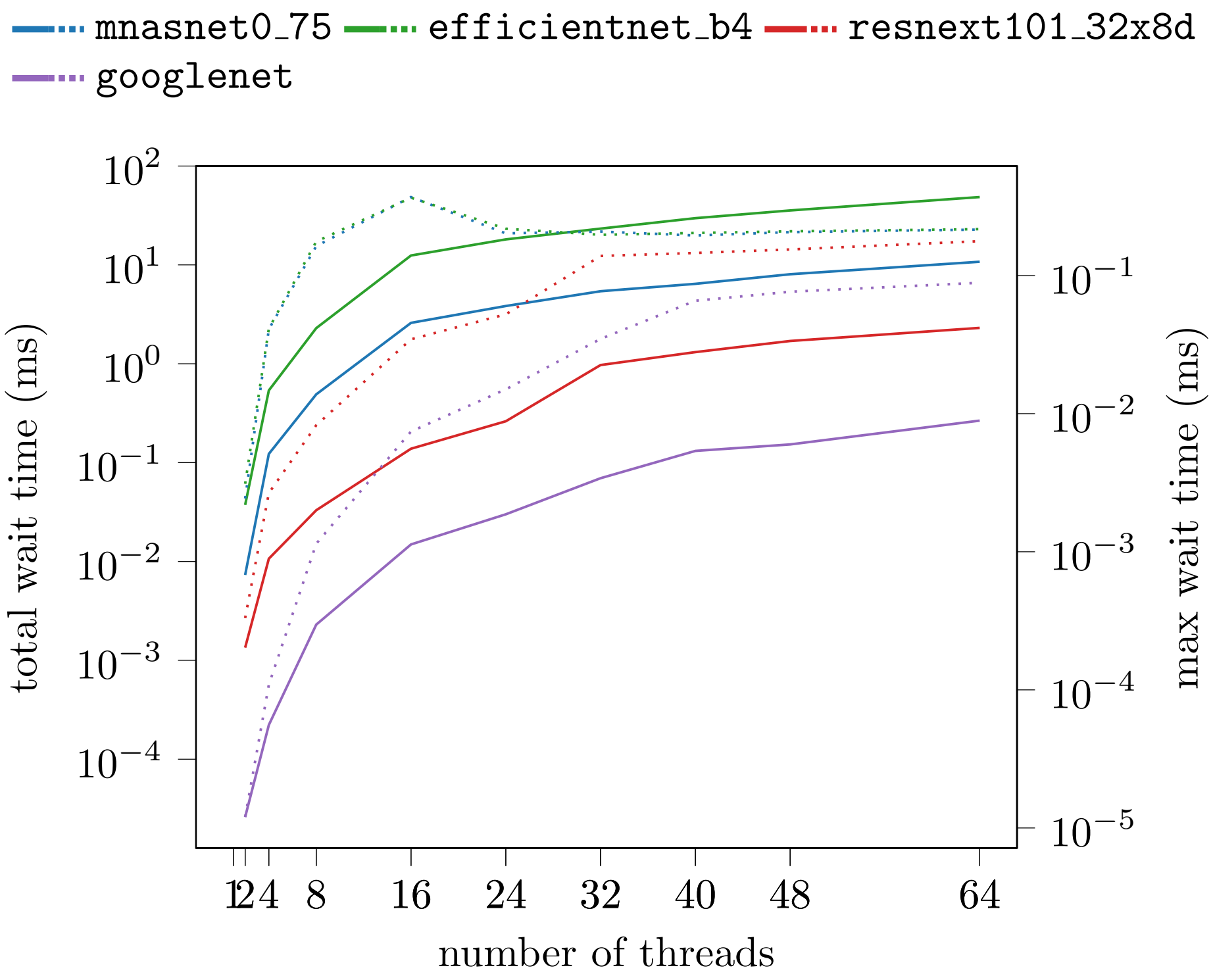

Figure 2.2: Maximum (dashed line) and total (solid line) lock wait times for the entire `jemalloc` arena.

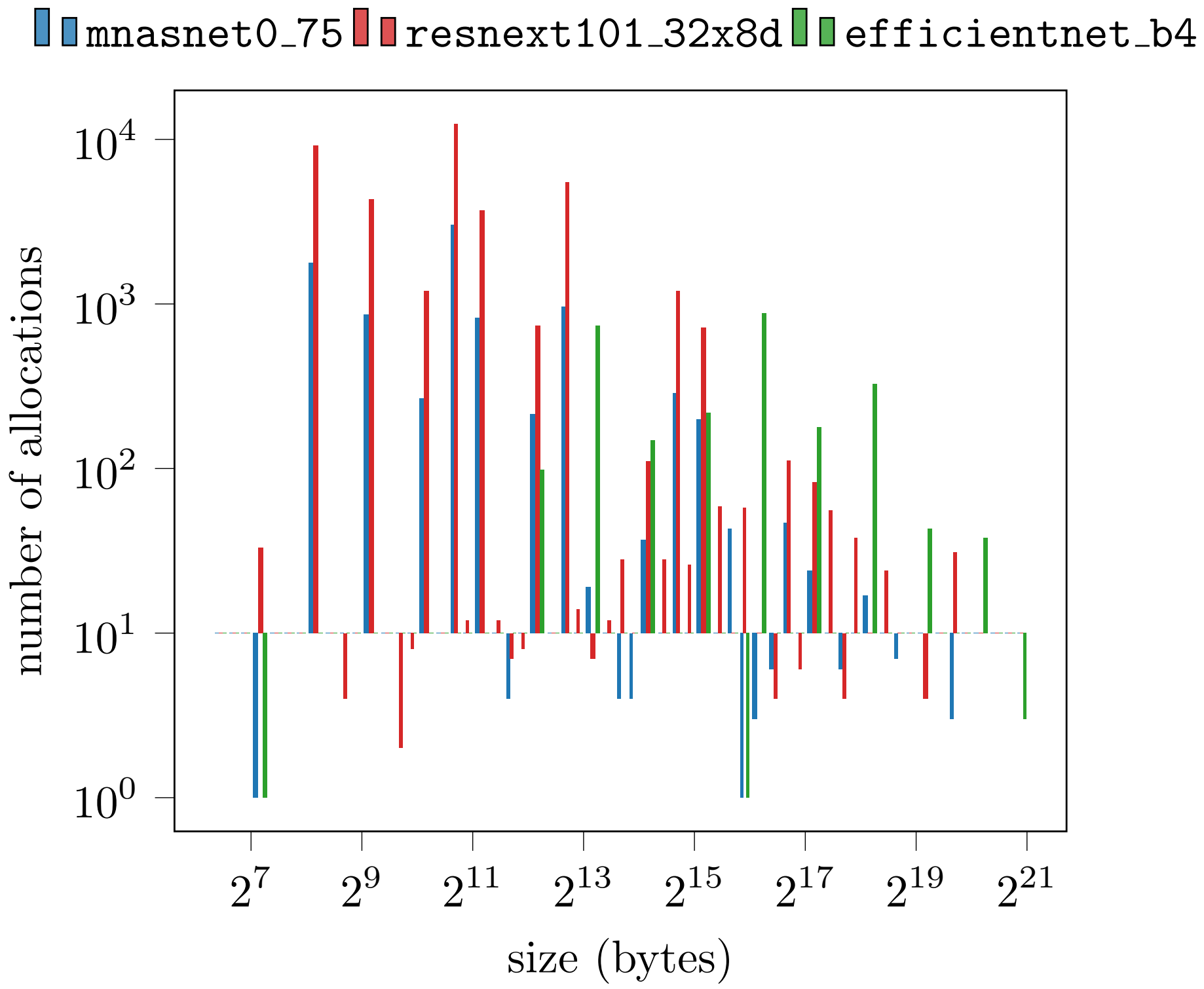

Figure 2.3: Distributions of intermediate allocations for various DNNs. Note that size is log scaled.

## 2.3 Memory Planning

In general, memory planning can be framed as an instance of the offline dynamic storage allocation (DSA) problem. To be precise, given static knowledge of all intermediate tensor sizes and lifetimes, we seek to determine the initial allocation size and the set of suitable offsets such that all intermediate tensors fit within the allocation. Therefore, the offsets can be computed by solving the mixed-integer program (MIP) formulation of offline DSA [52]:

$$\begin{aligned} &\min \texttt{total_mem} \\ &\text{s.t. } \texttt{offset}_i + \texttt{mem}_i \leq \texttt{total_mem} \end{aligned} \tag{2.1}$$

where tensors with overlapping lifetimes are constrained to be ordered in memory by

$$\begin{aligned} \texttt{offset}_i + \texttt{mem}_i &\leq \texttt{offset}_j + z_{ij} * \texttt{total_mem} \\ \texttt{offset}_j + \texttt{mem}_j &\leq \texttt{offset}_i + \left(1 - z_{ij}\right) * \texttt{total_mem} \end{aligned}$$

Here $z_{ij}$ are decision variables, defined as

$$z_{ij} := \begin{cases} 0 & \text{if } \texttt{offset}_i + \texttt{mem}_i \leq \texttt{offset}_j \\ 1 & \text{if } \texttt{offset}_j + \texttt{mem}_j \leq \texttt{offset}_i \end{cases}$$

that determine ordering (in address space) of allocations that overlap in lifetime.

While the offsets that comprise the solution to the MIP formulation are provably correct and optimal, the MIP is, in general, computationally intractable [37]. The best-known polynomial-time approximation is $2+\varepsilon$ by Buchsbaum [11], over the previously $3+\varepsilon$ best by Gergov [23]. There also exist simpler heuristics that generally perform well in terms of peak memory usage, fragmentation, and planning time. In this work, we consider five distinct memory planning strategies:

- `bump_allocation`, the baseline allocation strategy that consists of iterating through all allocations and maintaining a maximum offset, which is incremented ("bumped") for each new allocation;
- `mip` [52], i.e., offsets computed by solving the MIP optimization specified by Eqns. 2.1;
- `gergov` [23], Gergov's $3+\varepsilon$ approximation, based on constructing an infeasible solution and then transforming to a feasible solution using the Best Fit heuristic for interval graph coloring;
- `greedy_by_size` [44], that operates by sorting all intermediate allocations by size and then proceeding to assign offsets for overlapping (in lifetime) tensors according to a best fit criterion;
- `mincost_flow` [36], which frames the allocation problem as a minimum cost flow problem (with edges in the flow network corresponding to memory reuse).

We evaluate these strategies for the purposes of designing the memory planning component of `MemoMalloc` (see Section 3.2).

# CHAPTER 3
# IMPLEMENTATION

Our implementation consists of three components:

- A hybrid static analysis and profiling component that captures sizes and lifetimes of all memory allocations;
- A memory planner that constructs structured plans, consisting of an initial memory allocation and offsets for allocations associated with each operator of the DNN;
- A runtime component that effectuates the memory plan by computing runtime offsets and instantiating tensors, which are then consumed by operators.

We describe each component in turn.

## 3.1 Profiling

Recall the ultimate goal of our system: statically allocating all memory necessary for a forward pass iteration of a DNN. In order to accomplish this goal, it is necessary to describe accurately and uniquely all allocations made during a forward pass. Initial implementations involved recovering sizes of intermediate tensors wholly from the TS IR representation of a DNN. While practical and conceptually straightforward (involving propagating input shapes on tensors and computing tensor sizes from outputs of operators) it suffers from a critical flaw: since TS IR is a higher-level representation of the DNN than the kernel-level implementations, it does not capture all allocations made during the execution of the DNN (see Table 3.1). Primarily, this a product of operators that delegate to generic implementations; for example, a `max_pool2d` operation could appear in the TS IR as

```
%input.177 : Float(1, 512, 15, 15, strides=[...])
                        = aten::max_pool2d(%input.151, %4, %3, %3, %3, %6)
```

and reflect only a single output tensor, but whose actual implementation (see Listing 3) delegates to one of various specializations, and then, potentially, immediately frees parts of the results. Such implementation-dependent allocations are not reflected at the IR level and are fairly common. While it might be argued that such issues should be handled in a principled manner (e.g., by refactoring `max_pool2d_with_indices`) such delegation is necessary given the breadth of operators that PyTorch supports.[1]

Another complication involved in using TS IR to reconstruct all tensor lifetimes is the inherent aliasing of names; while TS is equipped with alias analysis infrastructure, it is, by necessity, conservative. For example, TS does not attempt to analyze aliasing of tensors that are inserted into containers (such as `Dict`, `List`, and `Tuple`). Nor is it able to precisely infer aliasing relationships between tensors that are never materialized but are actually *views*

1. Over 2,000 as of this writing.

```
Tensor max_pool2d(
    const Tensor& self,
    IntArrayRef kernel_size,
    IntArrayRef stride,
    IntArrayRef padding,
    IntArrayRef dilation,
    bool ceil_mode) {
  if (self.is_quantized()) {
    return at::quantized_max_pool2d(
      self, kernel_size, stride, padding, dilation, ceil_mode
    );
  }
  if (self.is_mkldnn()) {
    return at::mkldnn_max_pool2d(
      self, kernel_size, stride, padding, dilation, ceil_mode
    );
  }
  auto output_and_indices = at::max_pool2d_with_indices(
    self, kernel_size, stride, padding, dilation, ceil_mode
  );
  return std::get<0>(output_and_indices);
}
```

Listing 3: `max_pool2d` C++ implementation. Note, in the case of delegating to `at::max_pool2d_with_indices`, an immediate `free` occurs when `std::get<0>(output_and_indices)` is tail called.

on tensors (e.g., *slices* of tensors). In fact, memory planning in the context of this type of aliasing leads to "overplanning", i.e., overestimation of memory needs due to planning for tensors that do not correspond to unique allocations.

Note that the diametrically opposed alternative, namely a purely memorization-based approach that depends solely on the order of allocations, would be brittle with respect to relationships between operators and allocations. This is because such relationships are critical for adjusting memory plans post any optimization passes (such as those performed by an optimizing JIT) that occur after constructing a memory plan. Consider a "ResBlock" in a ResNet (see Figure 3.1) where control flow diverges after the `MaxPool` activation layer; since there is no total order of operations on distinct paths, a JIT compiler is free to reorder them. This has implications for the allocations performed by those operators. Consider the `Conv` + `BatchNormalization` pairs of operators, which make intermediate allocations of the same sizes but with differing lifetimes. If a given memory plan assigns memory addresses $[\texttt{offset}_1, \texttt{offset}_1 + \textit{size})$ to the intermediate tensor in Group 1, computed under the assumption that its lifetime covers (see Figure 3.2a) the lifetime of the intermediate tensor in Group 2 (with assigned memory addresses $[\texttt{offset}_2, \texttt{offset}_2 + size)$), then a reordering of those operations such that Group 1's `BatchNormalization` operator executes prior to Group 2's (see Figure 3.2b) would lead to an illegal address access by Group 2's `BatchNormalization` operator. This cannot be averted, since, at the time of allocation, a purely order-based solution could only distinguish allocations according to lifetime *start*s and tensor *size*s. In the structured approach (i.e., one that unambiguously associates allocations with operators), $\texttt{offset}_1$ and $\texttt{offset}_2$ would be effectively reordered along with their respective operators, thus avoiding any illegal memory accesses.

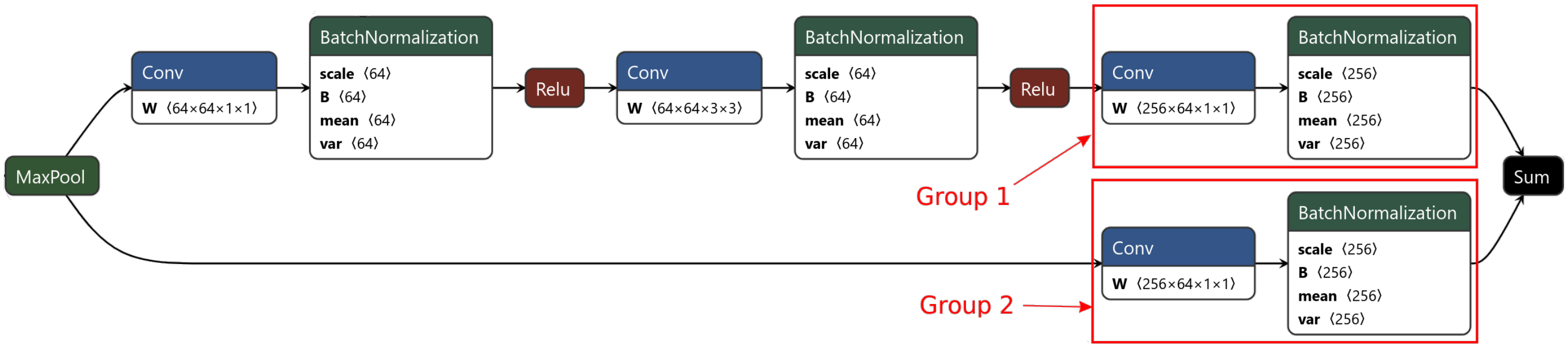

Figure 3.1: A "ResBlock" in a ResNet DNN, where the final `Conv` and `BatchNormalization` layers along both paths require allocations of the same size, but which can be made in arbitrary order (figure created using Netron [49]).

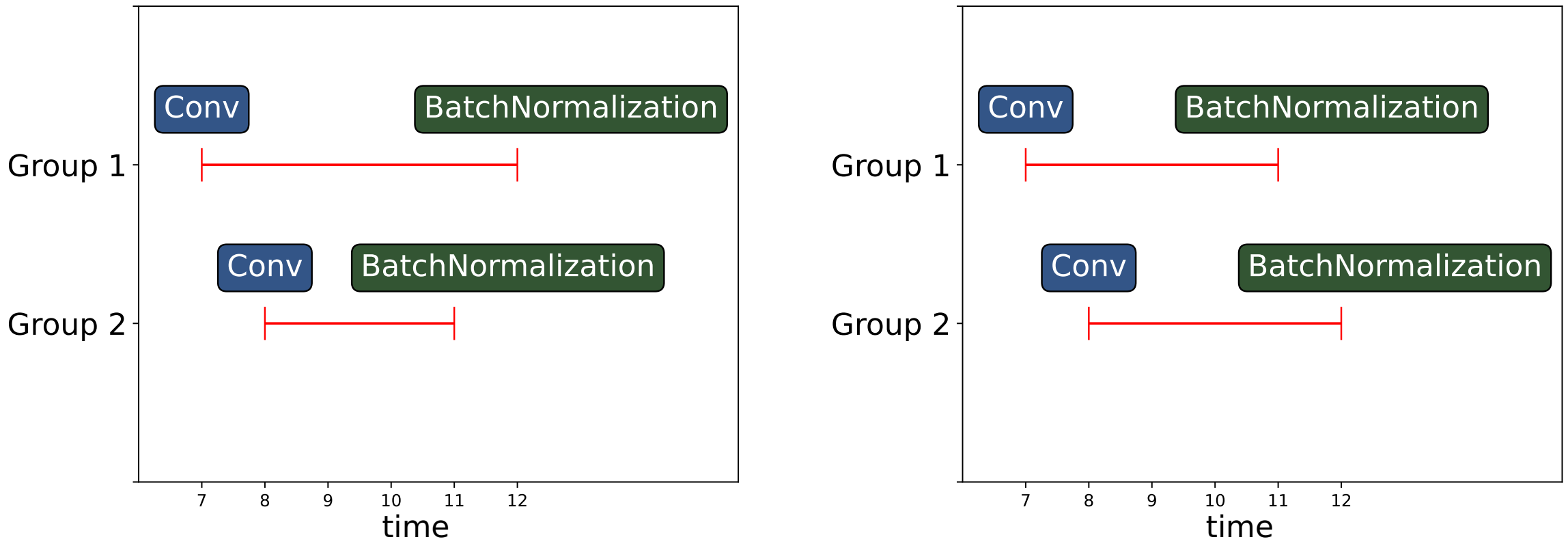

(a) Group 1's intermediate allocation covers Group 2's.

(b) Group 2's intermediate allocation outlives Group 1's.

Figure 3.2: Problematic orderings of operators. If a given memory plan assumes the ordering of operators in Figure 3.2a then a reordering such as that of Figure 3.2b leads to the `BatchNormalization` operator in Group 2 performing an illegal memory access (because its allocation should only "last" until time = 11).

As a result of all of these complexities, we refined our approach and designed a hybrid solution: we use profiling to capture all allocation sizes and lifetimes and avail ourselves of the TS IR representation of the DNN. We do so by instrumenting the allocator to record pointer values associated with sizes. We capture this information in tandem with lightweight stack tracing that establishes the provenance of an allocation (i.e., the operator and kernel within whose scope that allocation was made). The stack tracing is "lightweight" in the sense that it does not unwind the stack but maintains an auxiliary stack (which only records calls to functions in the `aten` namespace of the PyTorch library).

Table 3.1: Statistics on captured intermediate allocations (total number and total memory), by TS IR versus allocations captured by our profiling approach.

| Model | TS IR # | TS IR memory (MB) | Profiling # | Profiling memory (MB) |
|---|---|---|---|---|
| `mnasnet0_75` | 98 | 11 | 12,931 | 44 |
| `wide_resnet50_2` | 121 | 41 | 662 | 71 |
| `efficientnet_b4` | 379 | 50 | 57,238 | 190 |
| `resnext101_32x8d` | 240 | 87 | 3370 | 194 |
| `googlenet` | 138 | 11 | 788 | 24 |

One challenging aspect of this approach is in the capture of lifetime endpoints; since calls to `free` only receive a `void*` pointer (and no other metadata about the use of the memory pointed to), there is, in principle, no way to bracket the lifetime of a tensor (i.e., associate `mallocs` with corresponding `frees`). A naive solution could rely on pointer values themselves (in combination with a lookup table that records the size corresponding to a pointer) to make this identification, but this approach fails when the system allocator (that has been instrumented) reuses an address (which one hopes it often does!).

Instead, we employ a tagged pointer [41] approach. Specifically, we make use of the fact that, on `x86_64` architectures, pointers only occupy the lower six bytes of an 8-byte word (on `AArch64`, this feature is called Top Byte Ignore [3]). Making full use of the upper two bytes, we store a unique identifier, corresponding to each allocation (up to $2^{16}$ unique allocations) made during the profiling pass. This identifier is then used to uniquely identify `frees` with their corresponding `mallocs`. Note, `x86_64` requires pointers to be in "canonical form" before they are dereferenced (otherwise a "stack fault" is generated). We resolve this issue by encapsulating the tagged pointers in a smart pointer that canonicalizes (in a standards-compliant way) on dereference (see Listing 4). In addition to enabling us to determine tensor

```
inline void* canonicalize(void* ptr) {
  uintptr_t p2 = (((uintptr_t)ptr & ((1ull << 48) - 1)) |
                 ~(((uintptr_t)ptr & (1ull << 47)) - 1));
  return (void*)(p2);
}
```

Listing 4: Standards-compliant method of canonicalizing a tagged pointer. The first bitwise AND (`&`) clears the upper 16 bits of the pointer. Then, if bit 47 is 1, the bitwise OR (`|`) sets bits 47 through 63, but if bit 47 is 0, the bitwise OR is a no-op (since it is an OR with 0).

lifetimes, tagged pointers enable us to completely resolve aliases (by querying for this tag at operator and kernel boundaries). Using fully the resolved aliasing relationships, we can reconstruct relationships between operators and the kernels to which they delegate.

## 3.2 Memory Planner

After profiling to collect unambiguous tensor lifetimes and sizes, we statically plan memory allocation for subsequent forward pass iterations. In designing this aspect of the system, we considered the strategies discussed in Chapter 2. In order to evaluate the best planning strategy, we compared execution times and errors (relative to the optimum produced by the MIP). We observed that `greedy_by_size` generally achieves near-optimal results in terms of memory usage. We also evaluated the fragmentation incurred by various memory planning strategies (see Appendix 8.2) and observed that `greedy_by_size` generally has acceptable fragmentation. In addition to being efficient with respect to peak memory usage, `greedy_by_size` is performant enough to be executed prior to every forward pass of a DNN (see Figure 3.3). Our memory planner executes the `greedy_by_size` strategy by default but can be configured to use any of the other aforementioned planning strategies.

## 3.3 Runtime

After performing memory planning, we use the TS IR to "scope" the allocations to each operator, in order to preserve the structure of the allocations (i.e., groupings of allocations made in the service of carrying out an operation). On subsequent inference passes, we leverage that structure to assign offsets to tensors requested by operators. As already discussed, the alternative, simply assigning offsets on subsequent execution passes in some fixed order, was deemed to be brittle because it prevents plans from being transformed by IR passes that optimize the DNN, i.e., passes that potentially reorder operators and their concomitant allocations (see the discussion in Section 3.1). Our extension of TS IR (and the corresponding TS runtime) includes two new primitive operators:

- `prim::AllocateSlab`, borrowing terminology common in the allocator literature, is an operator that allocates all the memory that will be necessary for the duration of the inference pass of the DNN. It takes, as an attribute, the `total_size` and returns a `Storage` value (called `%memory`) backed by this allocation.

- `prim::AllocateTensor`, which takes, as attributes, the `size` and `offset` for the planned allocation that will be requested by the immediately subsequent operator and takes as input the `%memory` value. Internally, it functions in one of two ways: it either constructs a `Tensor` with manually set address (using pointer arithmetic to calculate $\texttt{offset}' = \texttt{offset} + \mathit{start}(\texttt{\%memory})$) if the subsequent operator can directly consume the allocation (i.e., it is an *out variant* operator) or it queues allocations that will be made implicitly by the operator (using, counterintuitively, a stack structure owned by an instance of `MemoMalloc`).

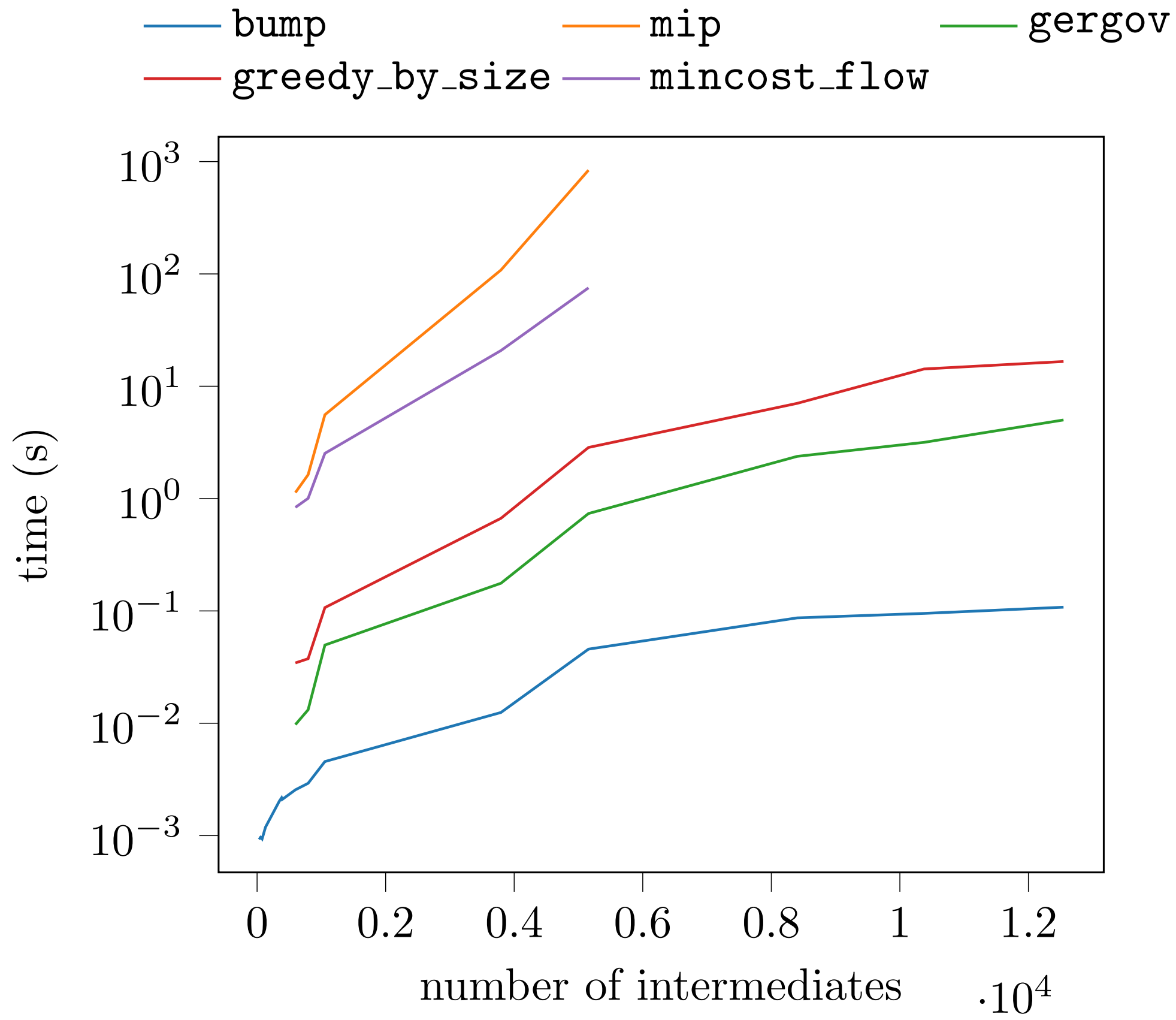

Figure 3.3: Runtimes for memory management strategies across various DNNs (i.e., various numbers of intermediate tensors). Note that `mip` and `mincost_flow` both time out for large numbers of intermediates.

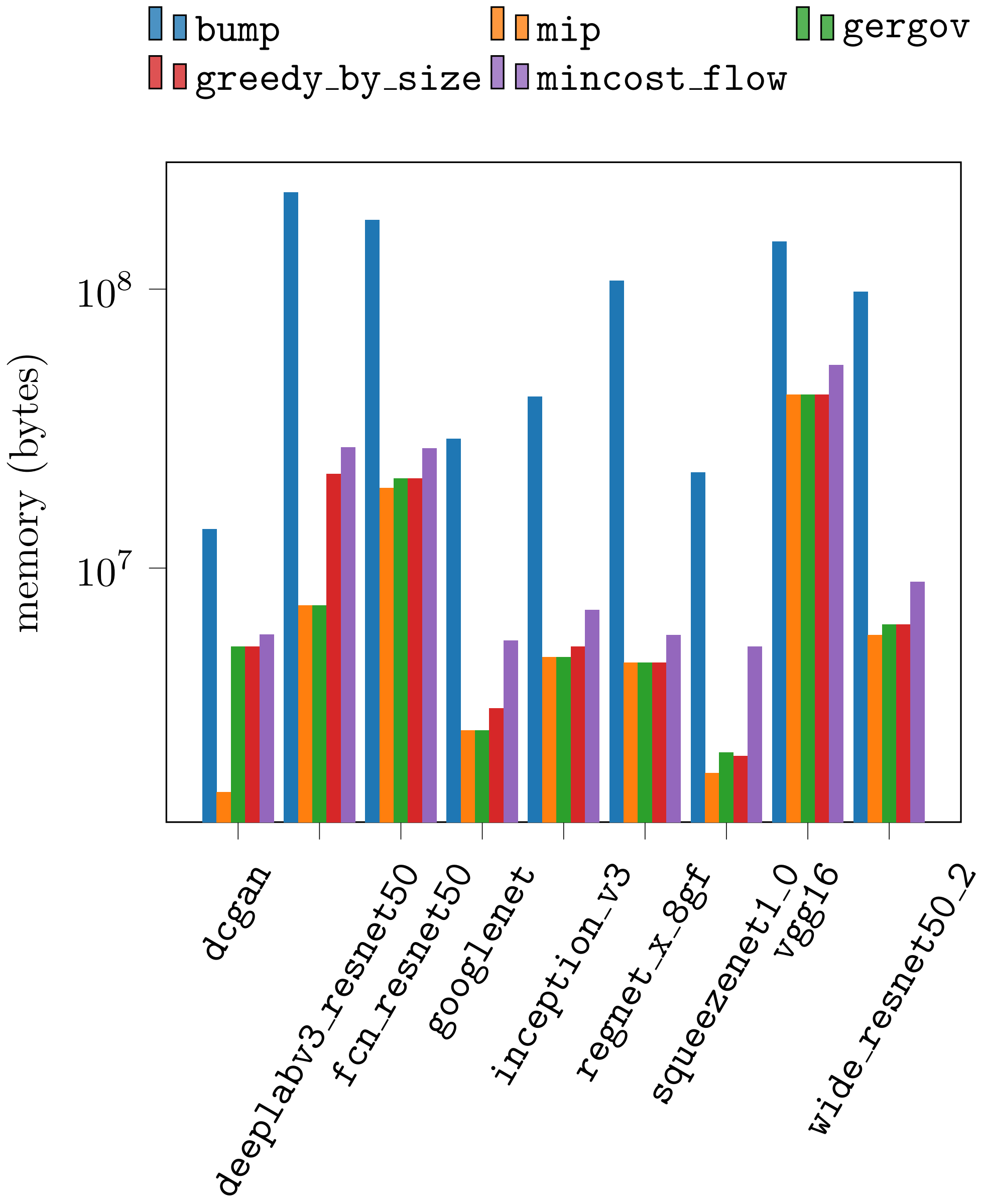

Figure 3.4: Peak memory usage for intermediate tensors for various DNNs, per memory planning strategy, for input shape $(1, 3, 128, 128)$.

```
graph(%w : Tensor, %x: Tensor, %h: Tensor):
  %memory: Storage = prim::AllocateSlab[total_size=1344]()
  %1: Tensor = prim::AllocateTensor[size=448, offset=0](%memory)
  %2: Tensor = aten::mm(%w, %x, %1)
  %3: Tensor = prim::AllocateTensor[size=448, offset=488](%memory)
  %4: Tensor = aten::add(%2, %h, %3)
  %5: Tensor = aten::relu(%4)
  return (%5)
```

Listing 5: Simple memory planning example.

See Listing 5 for a simple example. Note that tensors returned to the user (such as `%5` in Listing 5) are not managed since the solution aims to be orthogonal to other aspects of the PyTorch runtime (i.e., `MemoMalloc` should not own tensors that "escape" the DNN).

# CHAPTER 4
# EVALUATION

We evaluate our system (here denoted PyTorch+`MemoMalloc`) on several DNNs that are designed for various computer vision tasks; DCGAN [45] is used for representation learning; DeepLabv3 [14] and FCN [54] are used for semantic segmentation; GoogLeNet [62], WideResNet [71], VGG16 [58], InceptionV3 [63], RegNet [46], and SqueezeNet [31] are used for image classification. Due to shifting compute resources available, we made use of two test platforms over the course of our analysis (see Tables 4.3, 4.4), with slightly differing design matrices on each (see Tables 4.3, 4.4).

We evaluate our system against a baseline of PyTorch with memory managed by `jemalloc` (a common pairing in deployments of PyTorch). For PyTorch+`jemalloc`, we set the oversize arena (informed by our analysis in Section 2.2) threshold at 1MB, i.e., all allocations with sizes below 1MB are managed by `jemalloc` in the default way, making full use of the thread cache and $n \times 4$ arenas (where $n$ is the number of processor cores, including hyperthreading, on each test platform). For allocations greater than 1MB, the PyTorch+`jemalloc` configuration uses one arena with no thread cache and default decay rates. These configuration parameters are comparable to those typical of PyTorch deployments on server-class platforms [26]. For PyTorch+`MemoMalloc`, neither a caching allocator nor an oversize arena is used (i.e., only the single static allocation in combination with a memory plan).

We run each design configuration in a multithreaded fashion (with the number of threads being a design parameter). Each configuration performs `num_iterations` iterations of its forward pass on inputs with dimensions ranging in batch size and characteristic height/width (i.e., input images are square). Additionally, the configuration with `jemalloc` is run for a warmup period of 10 iterations. We repeat each configuration `num_repeats` times and collect the average execution time across all non-warmup iterations. We report the ratio of execution time between PyTorch+`jemalloc` and PyTorch+`MemoMalloc`. See tables 4.3, 4.4 for our design matrices. Note that since `batch_size` and `height_width` completely determine input size we group results by input sizes, i.e., $\texttt{input_size} = 4 \times 3 \times \texttt{batch_size} \times \texttt{height_width}^2$ (since all tensors are `float32` tensors, each element comprising 4 bytes, and all inputs have

Table 4.1: Test platform 1 characteristics.

| Component | Value |
|---|---|
| CPU | AMD(R) Threadripper(R) 3975WX 32-Cores (64 threads) |
| RAM | 128GB DDR4 |
| Hard drive | 1.9T Samsung MZVLB2T0HALB-000L7 |

Table 4.2: Test platform 2 characteristics.

| Component | Value |
|---|---|
| CPU | Intel(R) Xeon(R) Platinum 8339HC 24-Core (48 threads) |
| RAM | 376GB DDR4 |

Table 4.3: Design matrix for evaluation on test platform 1.

| Dimension | Values |
|---|---|
| batch_size | $[1, 4, 8]$ |
| height_width | $[128, 256]$ |
| num_threads | $[1, 32, 64, 128]$ |
| num_iterations | 10 |
| num_repeats | 10 |

Table 4.4: Design matrix for evaluation on test platform 2.

| Dimension | Values |
|---|---|
| batch_size | $[1, 4, 8]$ |
| height_width | $[64, 128, 256, 512]$ |
| num_threads | $[1, 2, 4, 8, 16, 32, 48, 64]$ |
| num_iterations | 32 |
| num_repeats | 32 |

3 channels). **We report input size in terms kilobytes** to reduce clutter on plots.

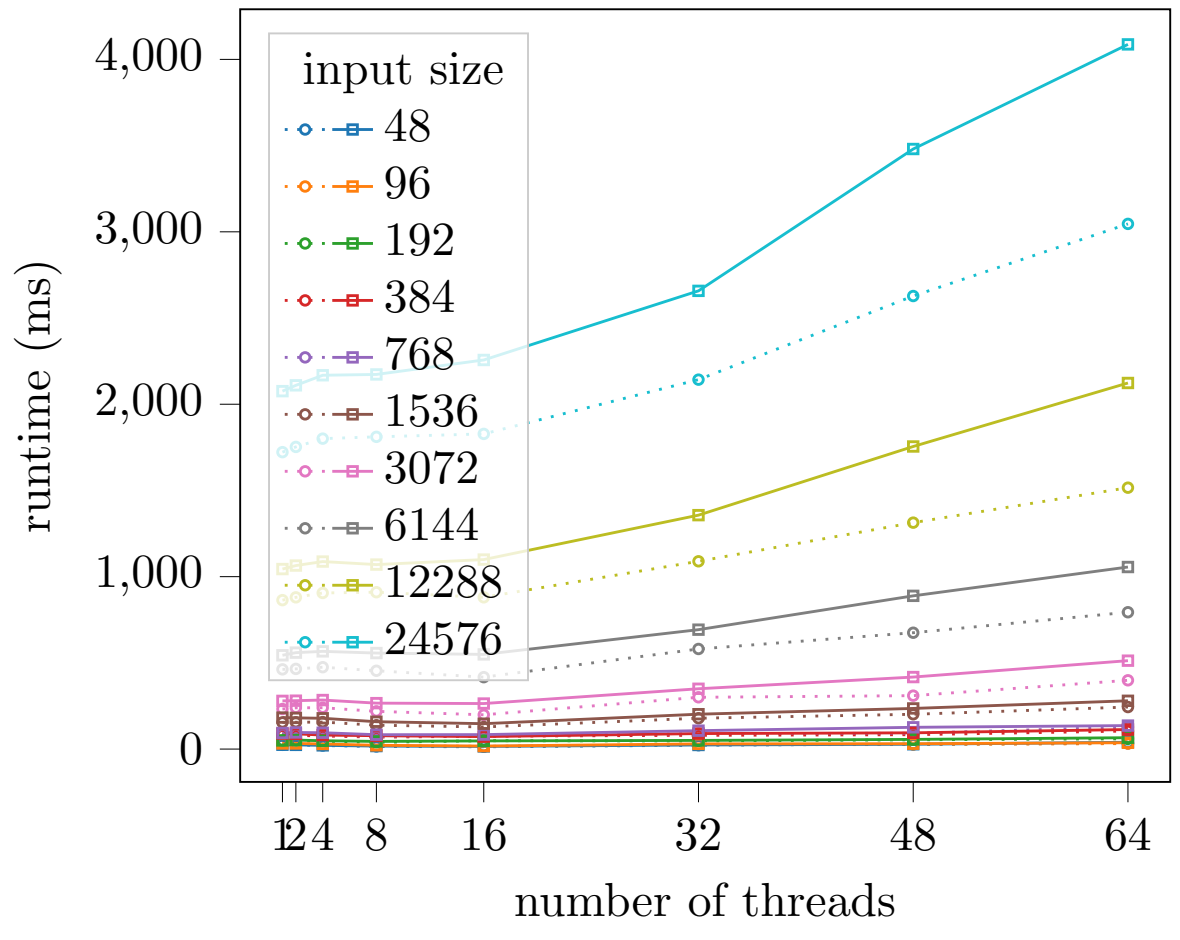

(a) Actual runtimes.

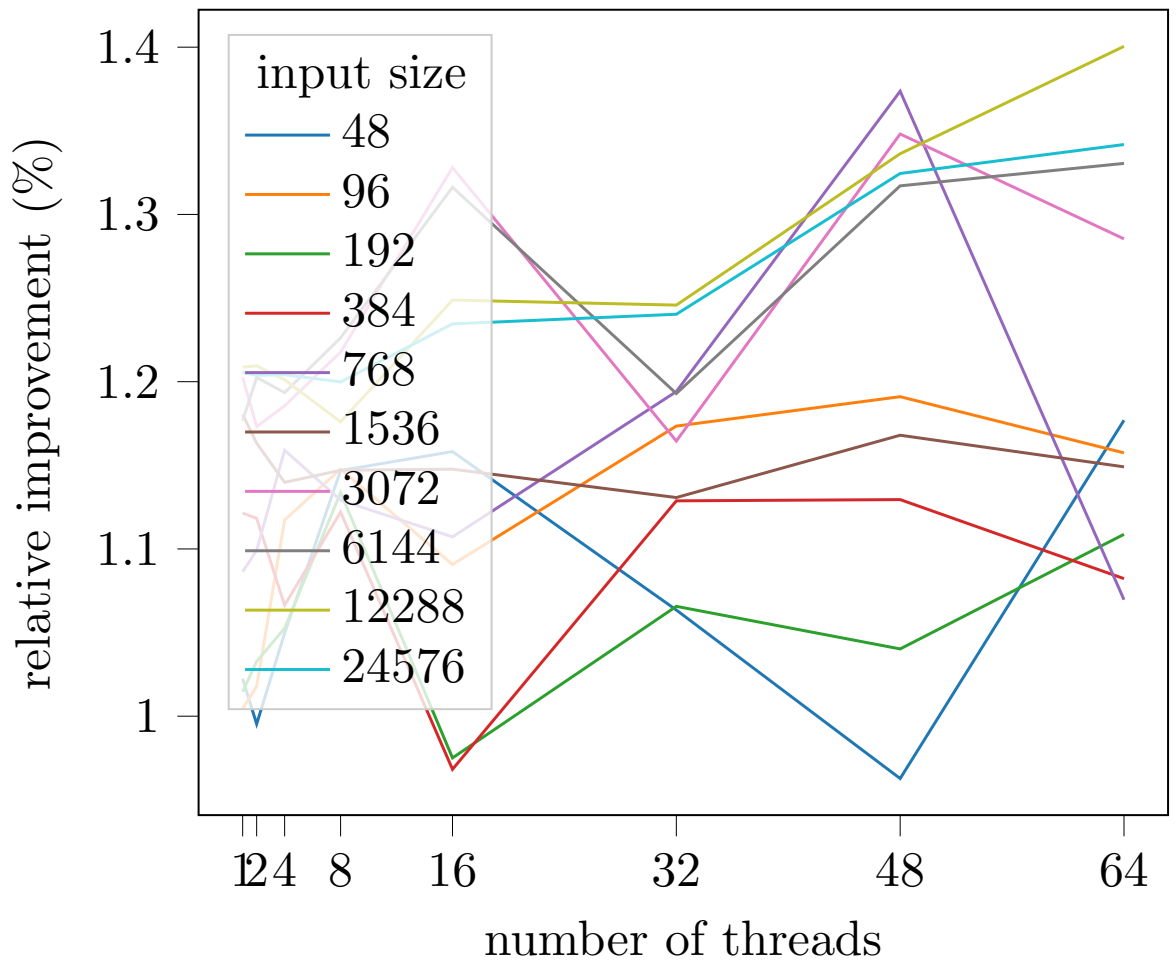

(b) Relative runtimes.

Figure 4.1: Evaluation on alexnet on platform 2.

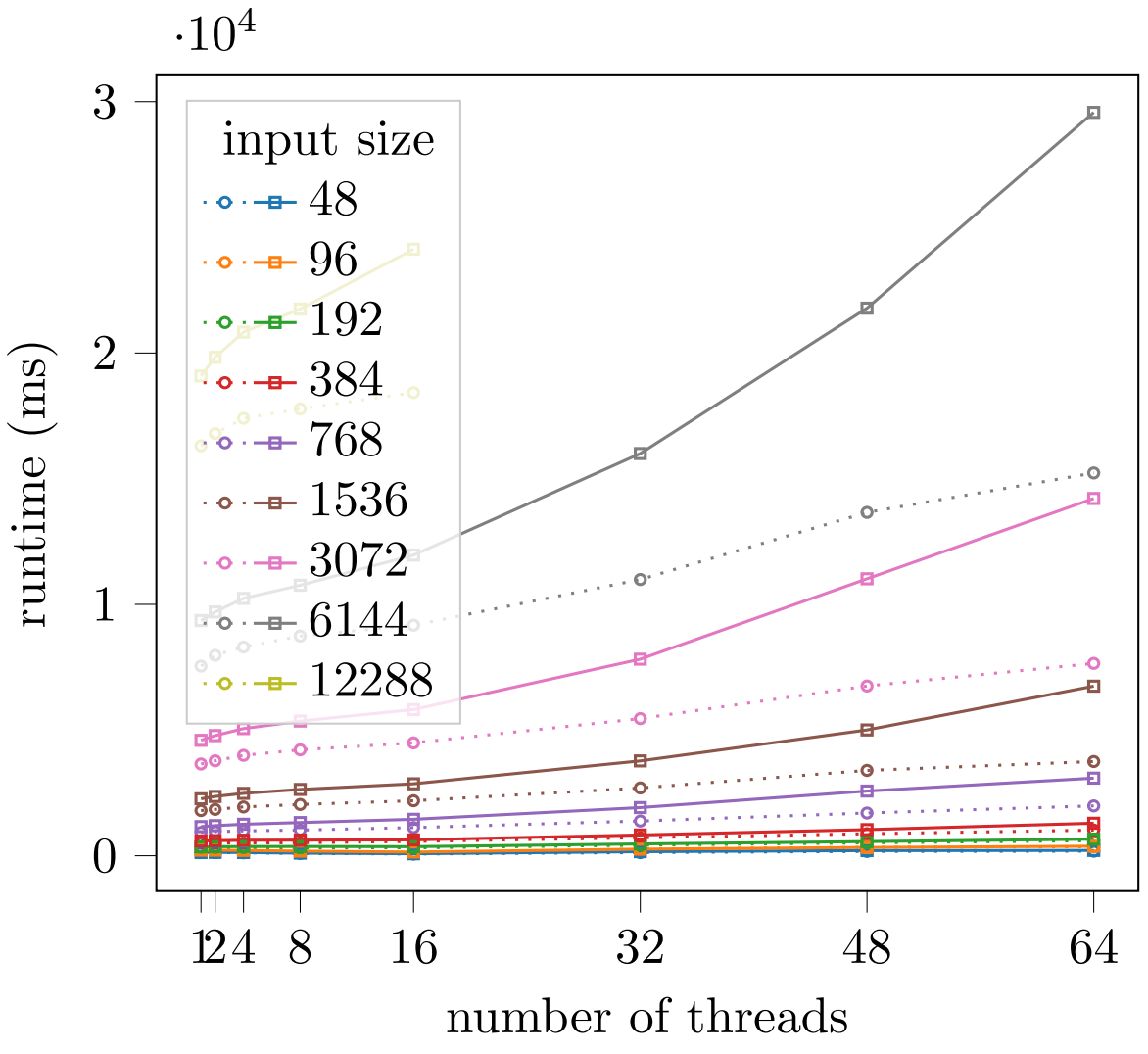

(a) Actual runtimes.

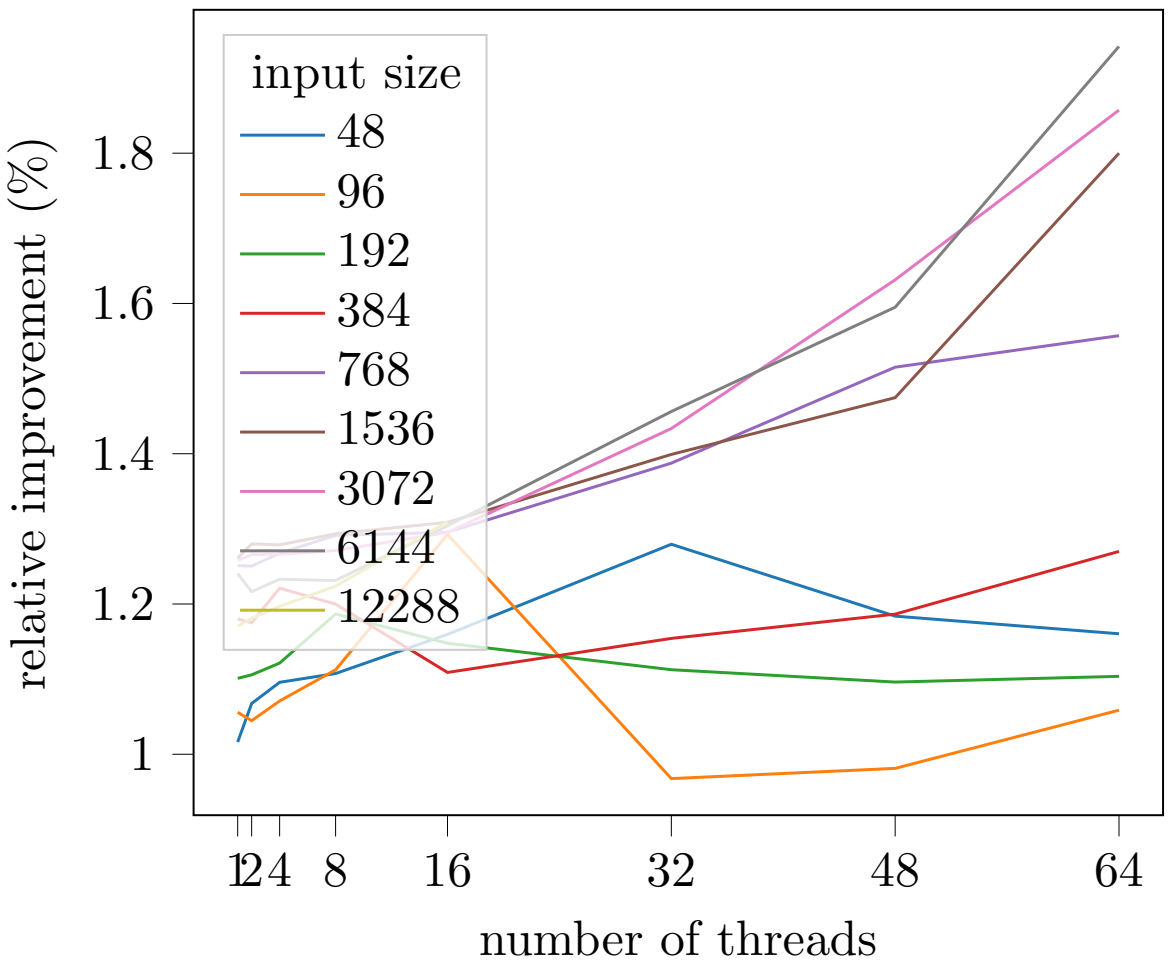

(b) Relative runtimes.

Figure 4.2: Evaluation on `densenet161` on platform 2.

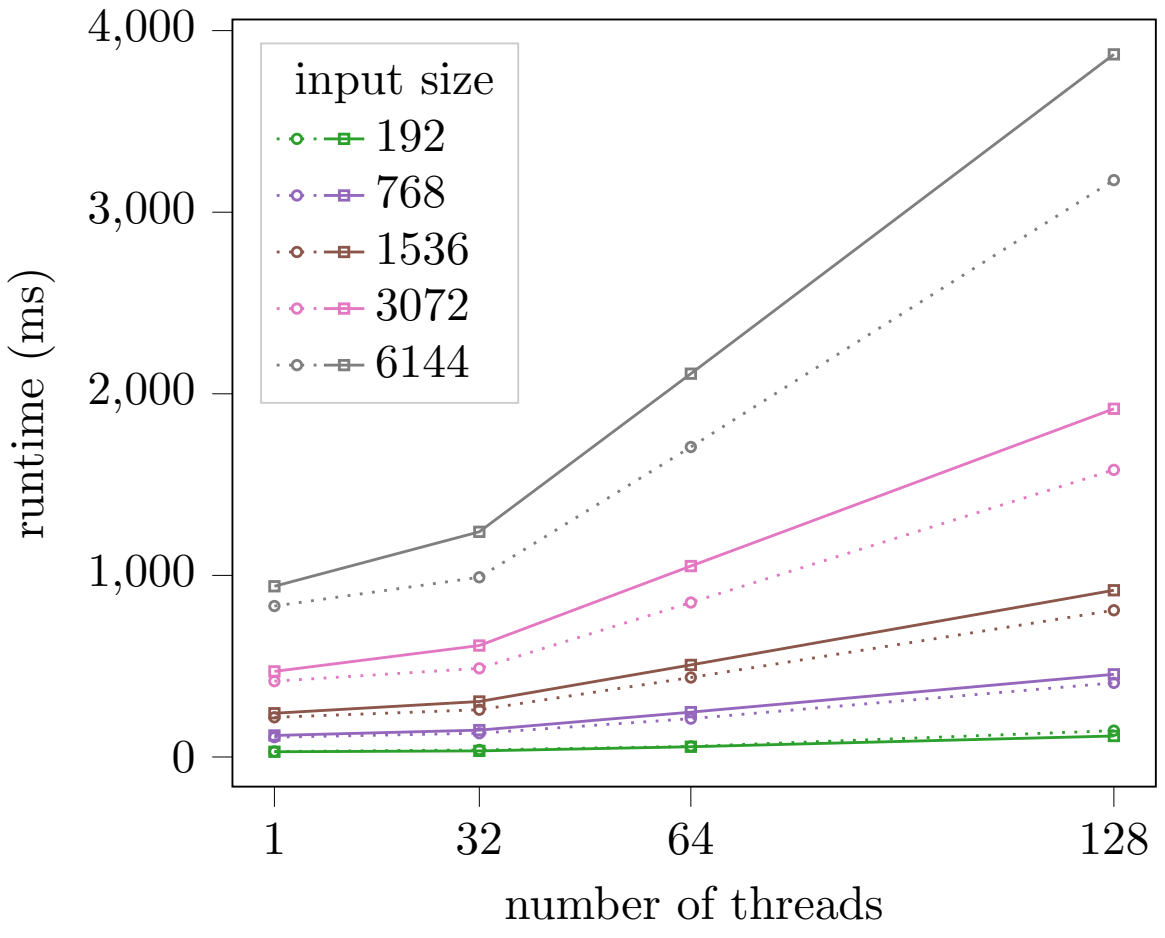

(a) Actual runtimes.

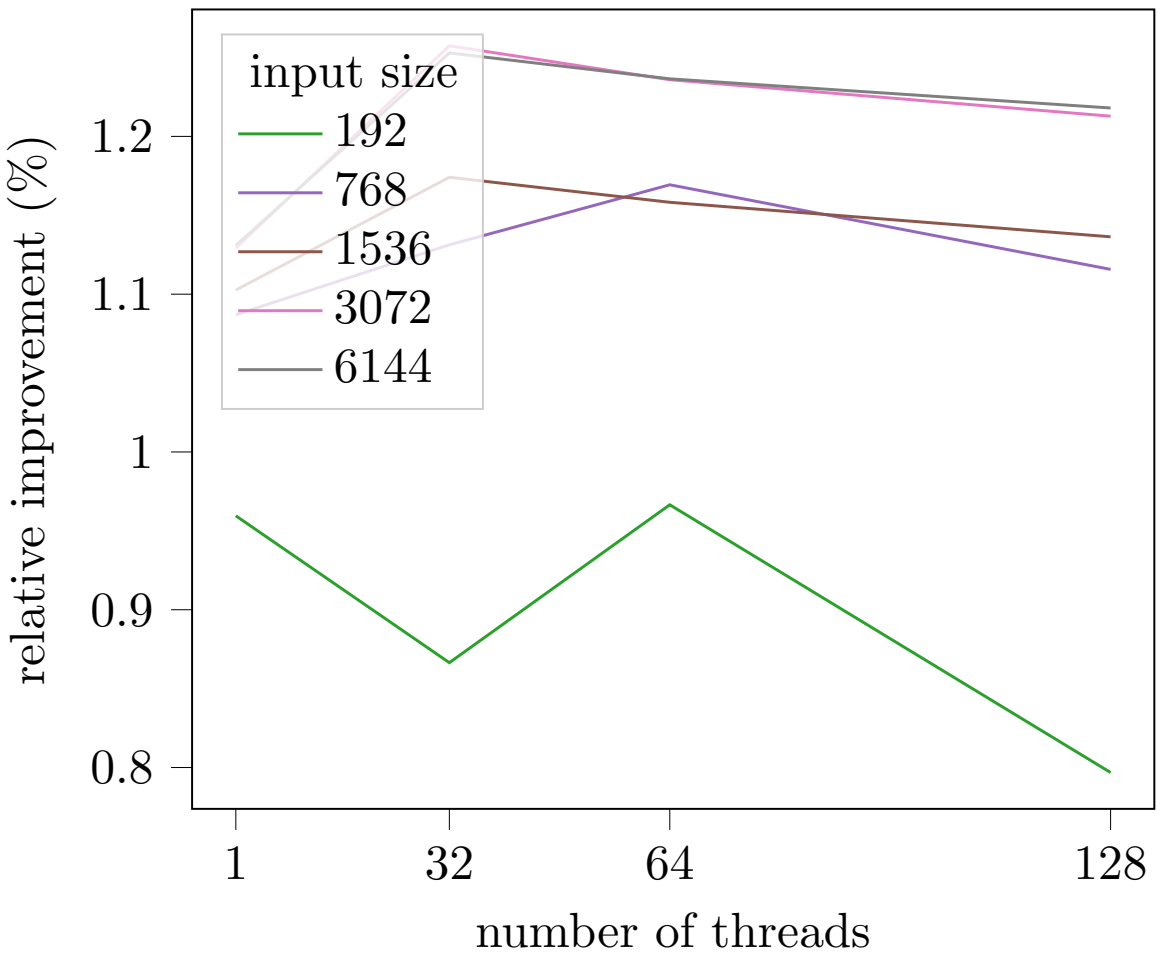

(b) Relative runtimes.

Figure 4.3: Evaluation on `dcgan` on platform 1.

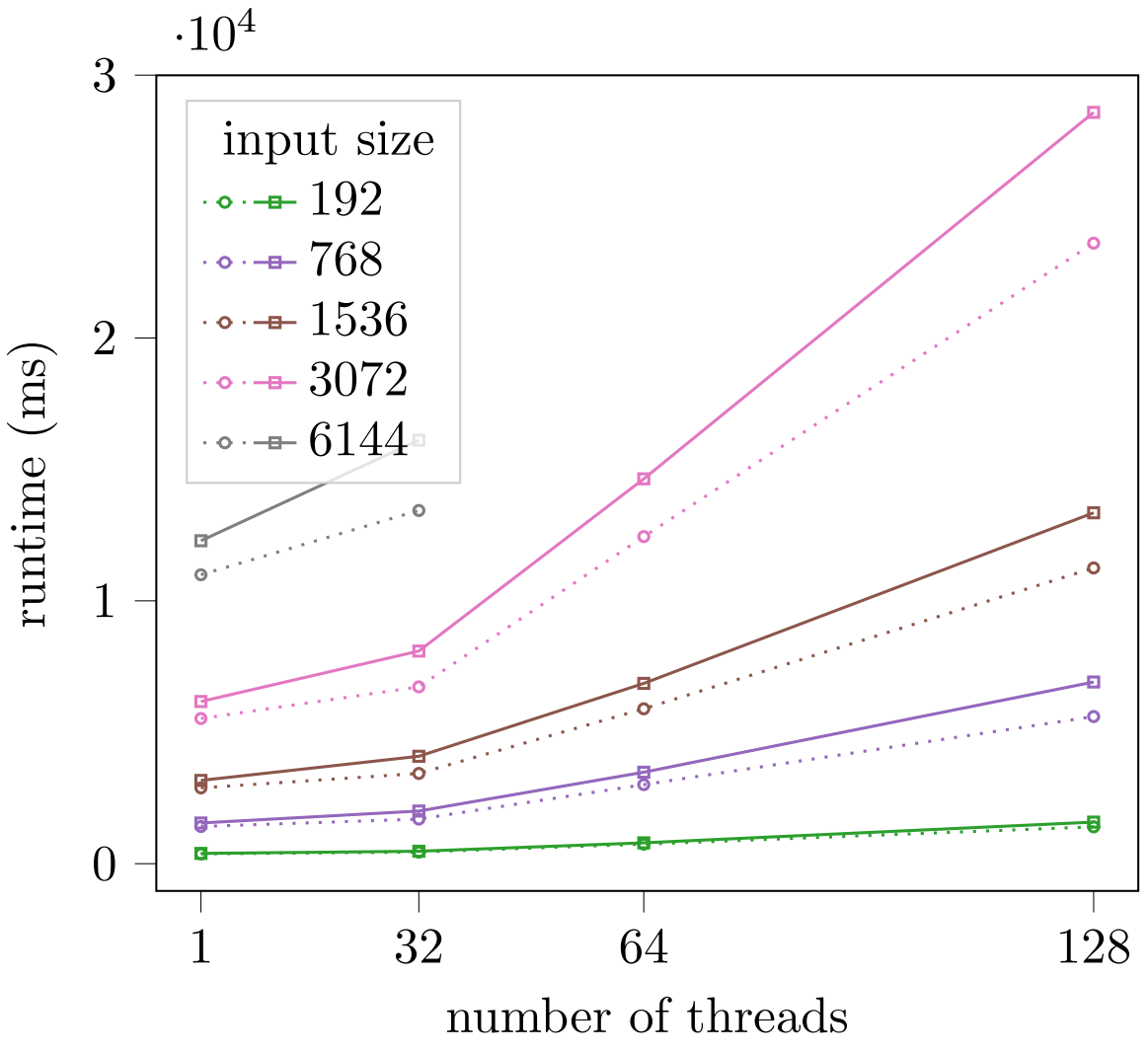

(a) Actual runtimes.

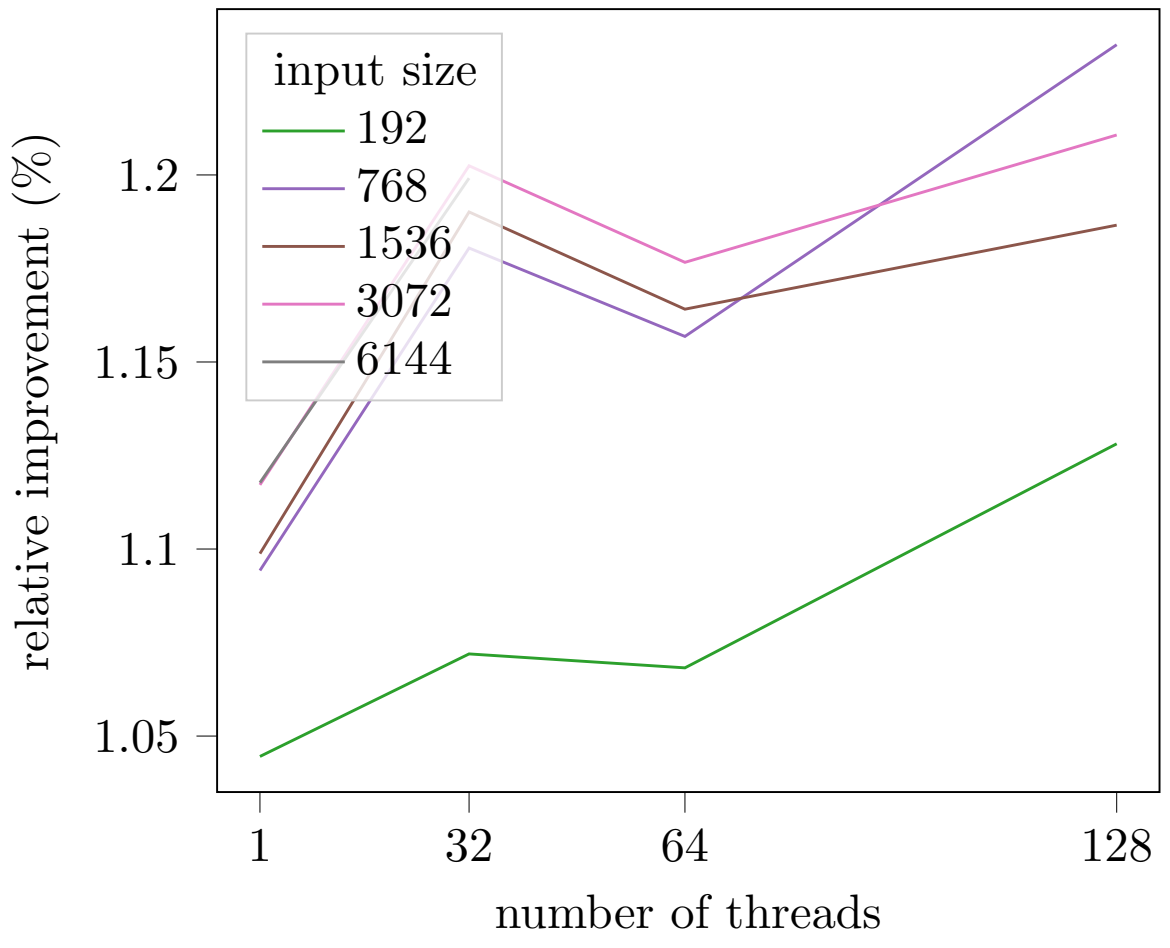

(b) Relative runtimes.

Figure 4.4: Evaluation on `fcn_resnet50` on platform 1.

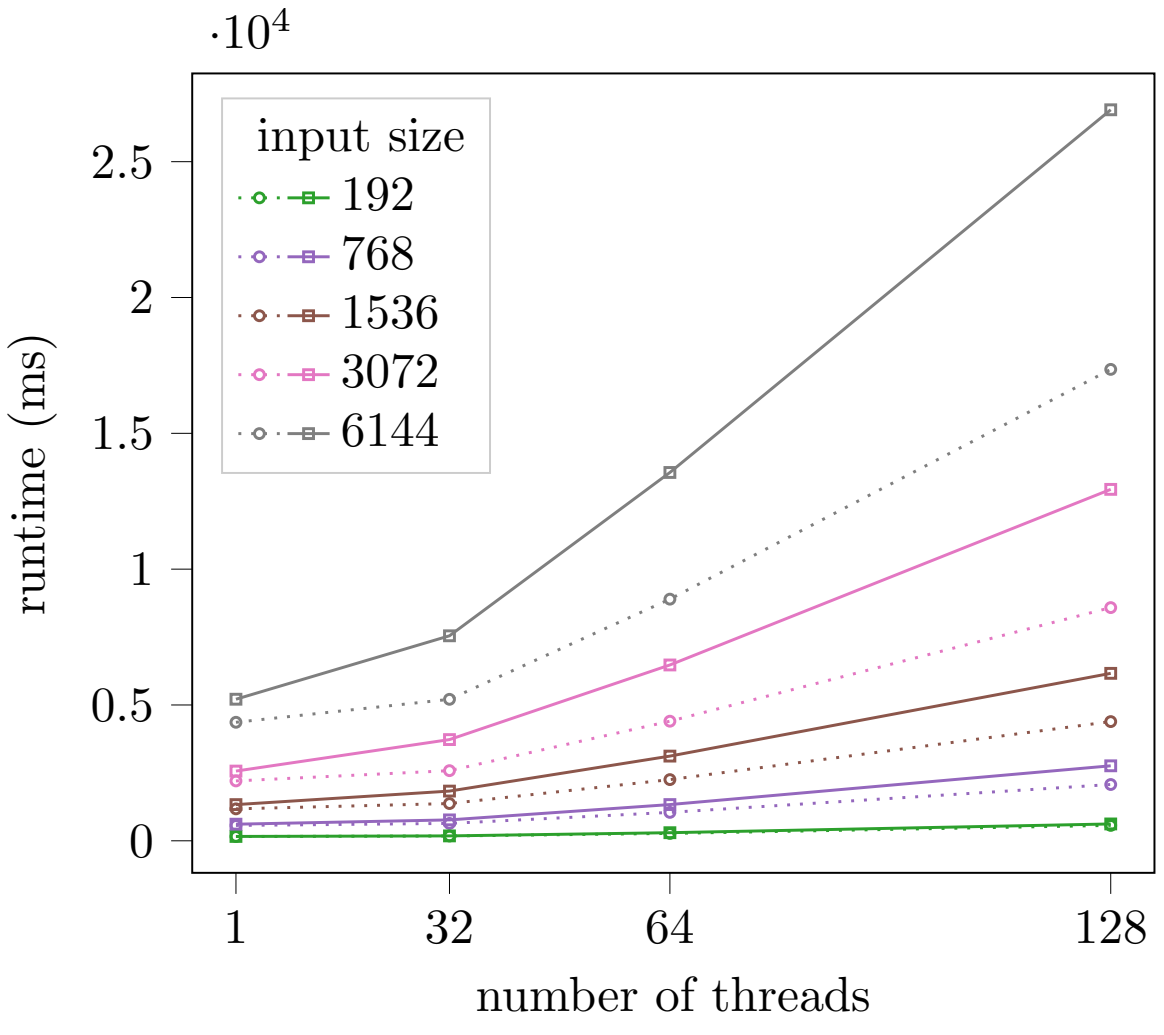

(a) Actual runtimes.

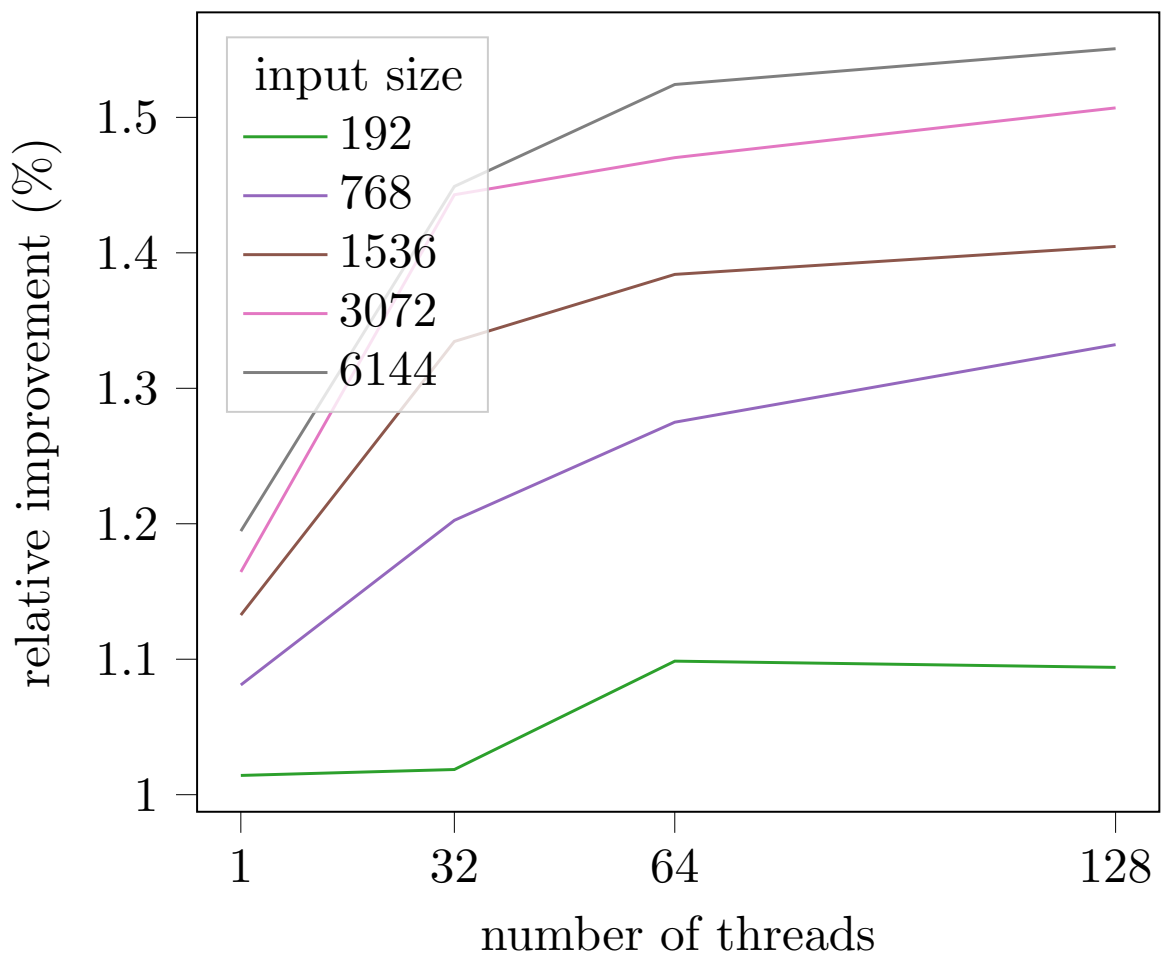

(b) Relative runtimes.

Figure 4.5: Evaluation on `regnet_x_8gf` on platform 1.

We tabulate performance statistics for the remaining DNNs in appendix 8.1.

# CHAPTER 5
# DISCUSSION

We observe that PyTorch+MemoMalloc robustly performs better than PyTorch+jemalloc, in terms of latency, for almost all input sizes and thread counts. How large that performance advantage is, varies amongst the networks, most likely as a function of the arithmetic intensity of the kernels of those networks. In the instances that MemoMalloc performs worse, it is the case that most allocations made by those networks fall below the 1MB oversize threshold (see Figure 5.1) and thereby have allocations serviced primarily by jemalloc's thread cache. See Appendix 8.3 for allocation distribution plots for the remaining DNNs. That is to say, those allocations can be performed with low latency overheads by jemalloc's thread cache, and thus jemalloc does not incur any overhead relative to MemoMalloc.

One notable feature of the performance trends is the reduction in relative performance with increasing thread count. That is to say, MemoMalloc performs well at 32 threads on platform 1 and 24 threads on platform 2, but then that relative performance slowly decays. This is most likely because the processors on our test platforms in fact possess fewer physical cores than reported to the operating system (due to hyperthreading). The limited number of cores (relatively speaking) acts as a natural "speed bump" on the number of operations a given thread can perform over the course of executing the DNN (thus constraining the maximum amount of mutex contention in the PyTorch+jemalloc configuration). This is evident from the overall increase in runtime experienced for all input sizes as a function of thread count.

Finally, it is important to consider the tradeoffs made in deploying MemoMalloc over jemalloc. MemoMalloc trades latency for, potentially, higher average memory usage; while peak usage should be comparable (both allocators need to accommodate the maximum necessary memory at any given time), average usage should be higher with MemoMalloc because it does not perform any frees over the course of the forward pass. To investigate this tradeoff, we collect statistics on the total number of bytes in active extents actually mapped by jemalloc (gathered using mallctl). Note that jemalloc always allocates aligned memory, while MemoMalloc only sometimes allocates aligned memory (depending on adjacent allocations), and thus the comparison is only approximate. Consider googlenet for input size = 128 (see Figure 5.2). Indeed, we observe that peak usage by MemoMalloc is comparable to that of jemalloc, average usage is higher (see in Appendix 8.2 for the same comparison for other DNNs). This internal fragmentation is acceptable in environments that have ample memory, or in instances where DNN processes take priority, but could prevent the use of MemoMalloc in resource-constrained environments such as embedded devices (see Section 7).

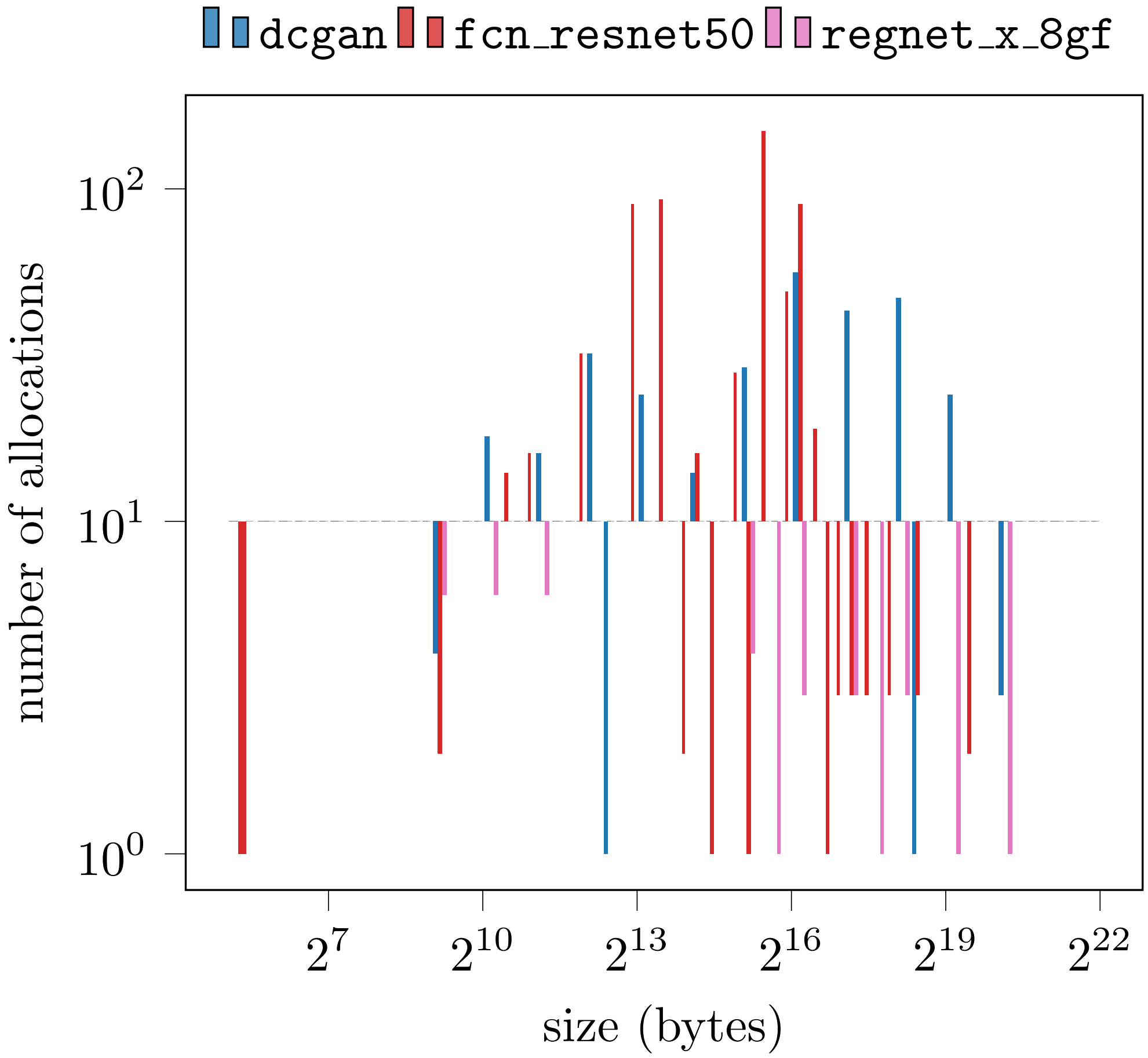

Figure 5.1: Distributions of intermediate allocations for DNNs for which PyTorch+`MemoMalloc` underperforms PyTorch+`jemalloc` at input size = 128.

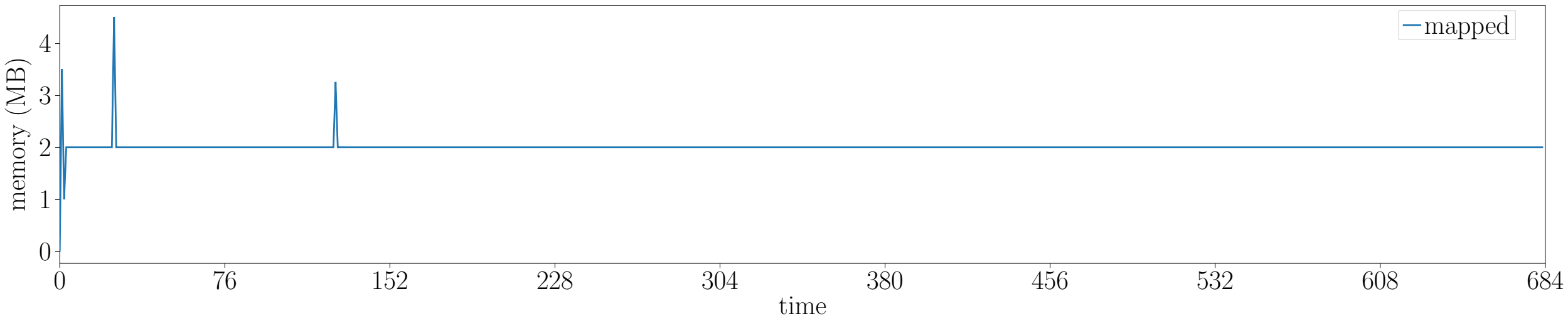

(a) Total number of bytes in active extents actually mapped by `jemalloc` for `googlenet` for input size = 128.

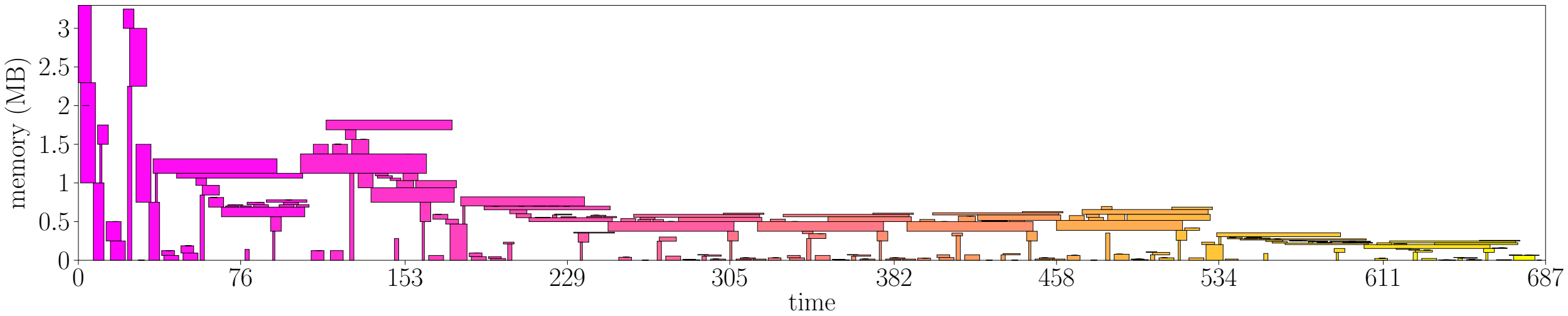

(b) Heap map for `MemoMalloc` with `greedy_by_size` strategy for `googlenet` for input size = 128.

Figure 5.2: Comparing memory usage for `googlenet` by `jemalloc` versus `MemoMalloc`. Note that the entire ∼3.5MB is kept allocated for the duration of the forward pass.

# CHAPTER 6
# RELATED WORK

There is ample related work in this area. Sekiyama et al. [52] propose a profiling approach similar to ours. They formally define the offline DSA problem (we make use of their formalization in Section 2.3) and then solve it using a "Best-Fit" heuristic (from [12]) for a related problem (the *orthogonal strip-packing problem*). They observe a moderate reduction in intermediate memory allocations across batch sizes and a commensurate reduction in inference latency due to how their framework of choice (Chainer [65]) performs intermediate allocations. Their approach is distinct from ours in that it does not attempt to recover the structure of the DNN.

Lee et al. [36] study memory management for DNNs in the context of deployment to mobile devices. In this context, they aim to reduce peak memory usage such that networks may satisfy the memory constraints of on-device accelerators on various mobile phones. To this end, they describe two memory management algorithms: a greedy memory management algorithm that allocates a pool of shared objects on an operator-by-operator basis, and the `mincost_flow` strategy we described in Section 2.3. They report satisfactory performance improvements but primarily due to successfully migrating from CPU to the on-device accelerators. They do not attempt to capture allocations made by kernel implementations of operators (which do occur in their framework of choice, TensorFlow Lite).

Pisarchyk et al. [44] also study memory management in the context of DNNs but with respect to peak usage rather than execution latency. They evaluate the same set of memory planning strategies as us, in addition to a strategy called *Greedy by Breadth*. Greedy by Breadth operates under the assumption that intermediate tensors of large sizes typically cluster, on an operator-by-operator basis (i.e., large inputs to operators produce large outputs). Thus, they sort (in decreasing order) operators by a measure they define as *breadth* (the sum of sizes of input and output tensors) and assign offsets in this order. Pisarchyk et al. evaluate their strategies on various DNNs tailored to deployment on edge devices. While they observe that Greedy by Size achieves near optimal results (in concordance with our evaluation) they do not make any use of the additional structure of the DNN, nor do they attempt to perform alias analysis of tensors.

Nimble [55] does make use of the intermediate representation of the DNN and similarly inserts primitive allocation operations into the IR, but, critically, Nimble does not introspect into implementations of operators and therefore elides any implicit allocations. Notably, TVM (closely related to Nimble) began discussions[1] regarding static memory planning at approximately the same time as this project began.

One important body of work possessing high affinity with our own is the Multi-level Intermediate Representation (MLIR) project [35]. In the MLIR framework, there exist many intermediate representations (called *dialects*), that enable the specification of DNNs at various levels of abstraction. In particular, in the `linalg` dialect, sequences of DNN operators are decomposed in terms of the corresponding linear algebra; consider the representation of `conv` in Listing 6. The important feature of this representation to note is that the allocation

1. [Discussion/Alignment] Memory Planning

```
func @conv(%input: tensor<1x3x225x225xf32>, %filter: tensor<32x3x3x3xf32>,
           %output: tensor<1x32x112x112xf32>)
  -> tensor<1x32x112x112xf32> {
    %0 = bufferization.to_memref %input : memref<1x3x225x225xf32>
    %1 = bufferization.to_memref %filter : memref<32x3x3x3xf32>
    %2 = bufferization.to_memref %output : memref<1x32x112x112xf32>
    %3 = memref.alloc() : memref<1x32x112x112xf32>
    linalg.copy(%2, %3) : memref<1x32x112x112xf32>,memref<1x32x112x112xf32>
    linalg.conv_2d_nchw_fchw
      {
        dilations = dense<1> : tensor<2xi64>,
        strides = dense<2> : tensor<2xi64>
      }
      ins(%0, %1: memref<1x3x225x225xf32>, memref<32x3x3x3xf32>)
      outs(%3: memref<1x32x112x112xf32>)
    %4 = bufferization.to_tensor %3 : memref<1x32x112x112xf32>
    return %4 : tensor<1x32x112x112xf32>
  }
```

Listing 6: Representation of `conv` in the `linalg` dialect of MLIR.

`%3 = memref.alloc()` for the output of the convolution is explicitly represented, along with its shape `memref<1x32x112x112xf32>` (along with the shapes of all other tensors). This straightforwardly enables the writing of a compiler pass that implements static memory planning; indeed in MLIR this is called a "comprehensive bufferization"[2] and uses essentially the `mincost_flow` strategy.

2. mlir/lib/Dialect/Linalg/Transforms/ComprehensiveBufferizePass.cpp

# CHAPTER 7
# CONCLUSION

We studied the memory allocation patterns of DNNs, with respect to latencies incurred by synchronization mechanisms in conventional caching allocators. We then proposed and implemented a memory planning system for reducing such latencies (during inference) for DNNs. We evaluated our system and observed that it performs better than `jemalloc` for typical DNN workloads. Our implementation is open-source and in the process of being upstreamed to PyTorch.[1] In the future, we intend to factor out `MemoMalloc` into an independent module with a uniform API such that it can be plugged into any of the popular deep learning frameworks.

Future work in this area includes several directions:

- **Dynamics**. All of our work here assumes that there is no control flow and that all intermediate tensor sizes are fixed. In practice, this is only the case in certain environments and it would be preferable to be able to perform memory planning in the context of both control flow and dynamic intermediate tensor sizes. Our preliminary work indicates that in fact, this is possible; for DNNs where intermediate tensor sizes can be algebraically inferred from input shapes, it is possible to construct memory plans ahead-of-time (and to cache them) for common input shapes. Such a regime is called *symbolic memory planning*, owing to the employment of *symbolic shape inference* in order to derive algebraic relationships between input shapes and intermediate tensor sizes. The simplest example of this is symbolic memory planning in the context of a dynamic batch size; in this context it can be analytically proven that the MIP solution scales linearly with batch size, thus enabling amortized MIP memory planning.

- **Training**. Our work here has targeted primarily DNN inference, on the assumption that latency matters most in this context. While it is the case that service-level agreements and quality-of-service guarantees impose hard constraints on inference latencies, it is also the case that during training of DNNs, lower latencies could proportionally reduce costs (associated with the research process). The added complexities of training are twofold: firstly, the graph corresponding to backpropagation of gradients must be obtained (i.e., the *backwards graph*), and secondly, intermediate tensors must be kept alive (or stored) in order to be available during gradient computation. Both of these aspects present new challenges for static memory planning. Obtaining the backwards graph in TS IR is currently not possible but alternative tracing mechanisms, such as LazyTensor [60], could be used. Under current assumptions for heuristics memory planning strategies (such as `greedy_by_size`), intermediate tensors that need to be persisted or stored undoubtedly lead to highly fragmented memory plans. Thus, training necessitates a distinct set of heuristics for computing offsets.

- **GPUs**. Motivated by current deployment practices, we have only considered CPU deployment. But it is the case that GPUs are in fact, slowly being adopted as de-

1. https://github.com/pytorch/pytorch/pull/64347

ployment targets for inference. GPUs introduce many novel complications, due to exotic scheduling environments and complicated memory hierarchies; for example, on NVIDIA devices, execution of a group of threads will block on data being absent from shared memory. Despite such complications, there is reason to believe that static memory planning could be feasible on GPUs as well; NVIDIA has recently released an extension to the CUDA API called CUDA Graphs[2] whose use entails "freezing" and reusing fixed sets of memory addresses for multiple iterations of arbitrary sequences of kernels. Preliminary exploration of this API has shown that it does in fact reduce many of the latencies associated with allocation.

- **Edge Devices**. Recently edge platforms (mobile phones, wearables, IoT sensors) have also become feasible deployment targets for DNNs, owing to advances in research on DNN architectures that maintain accuracy while reducing resource consumption (such as quantized [68] and sparse networks [70]). These advances notwithstanding, those platforms reproduce many of the phenomena of their larger scale analogues [61]. Namely, memory consumption of DNNs on edge devices is of significant importance, due to proportionally scaled memories (i.e., relatively small), limited memory bandwidth capacities [67], and less powerful memory management units [19]. Simultaneously, limited threading capabilities impose constraints on the complexity (and therefore sophistication) of possible memory management schemes, such as dynamic allocators [47] and software virtual memory [7]. We are investigating deploying `MemoMalloc` on such platforms.

2. https://developer.nvidia.com/blog/cuda-graphs/

# CHAPTER 8
# APPENDIX

## 8.1 Runtime comparison

Evaluation of `MemoMalloc` vs. `jemalloc` with respect to inference latency. Note:

- Dotted lines correspond to `MemoMalloc` and solid lines correspond to `jemalloc`.
- Input sizes are measured in kilobytes.
- Platforms 1 and 2 are described in Tables 4.1, 4.2.

### *8.1.1 Platform 1*

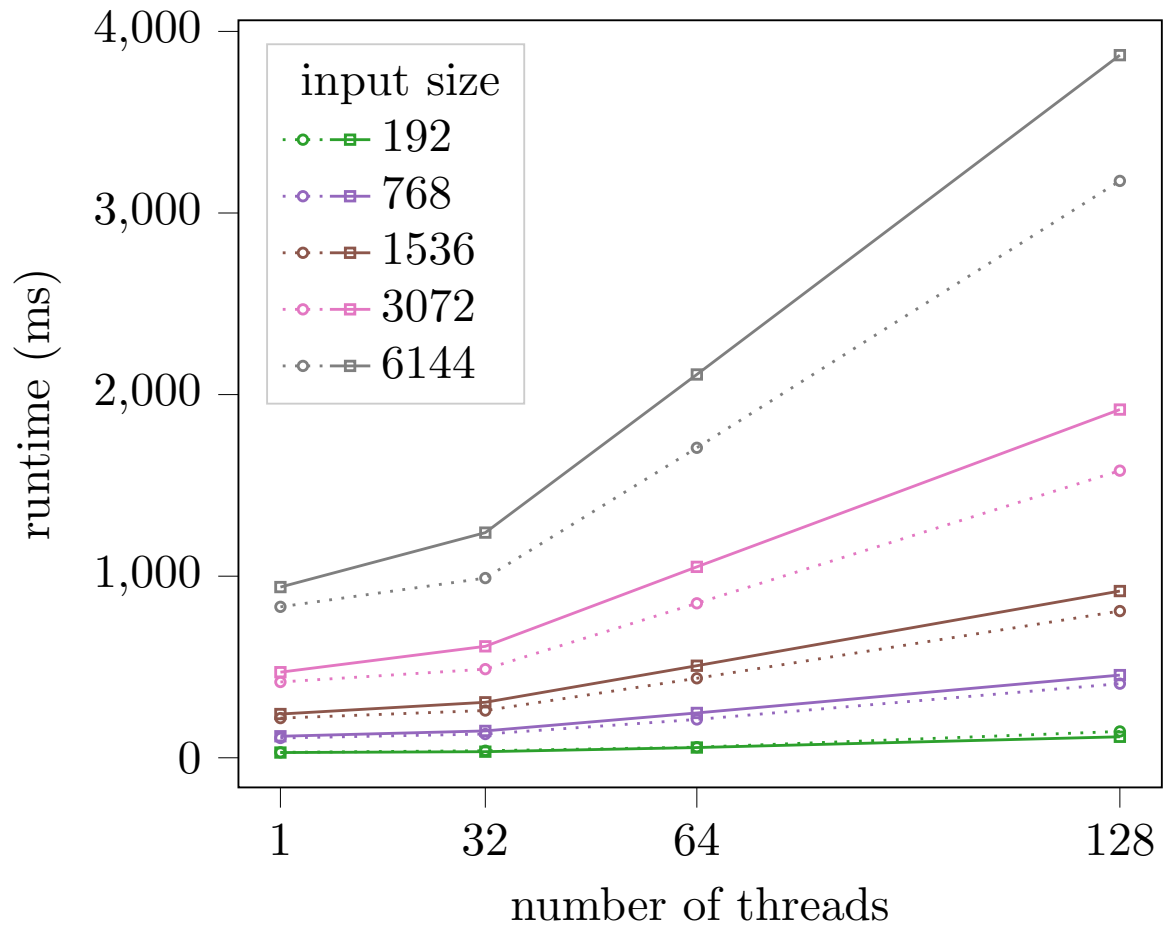

(a) Actual runtimes.

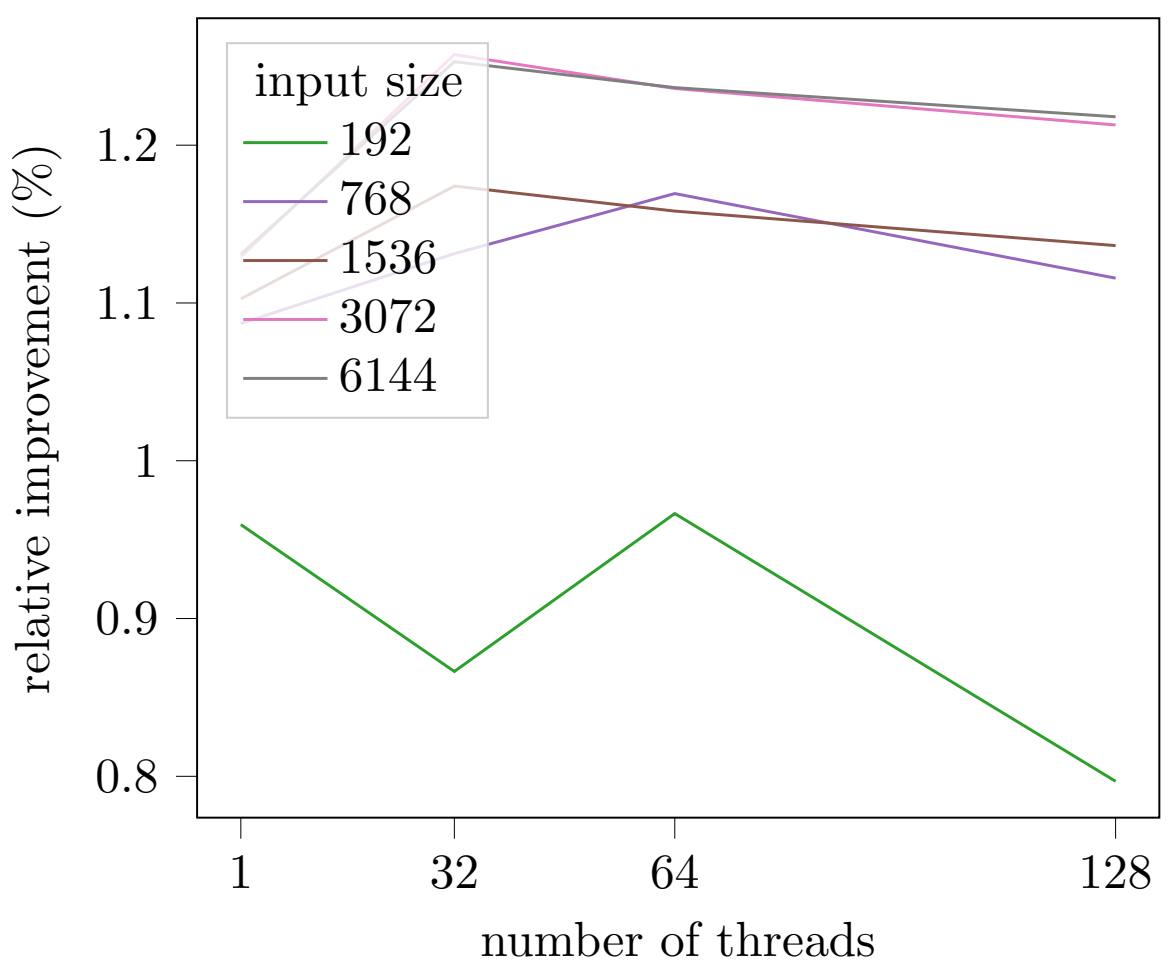

(b) Relative runtimes.

Evaluation on `dcgan`.

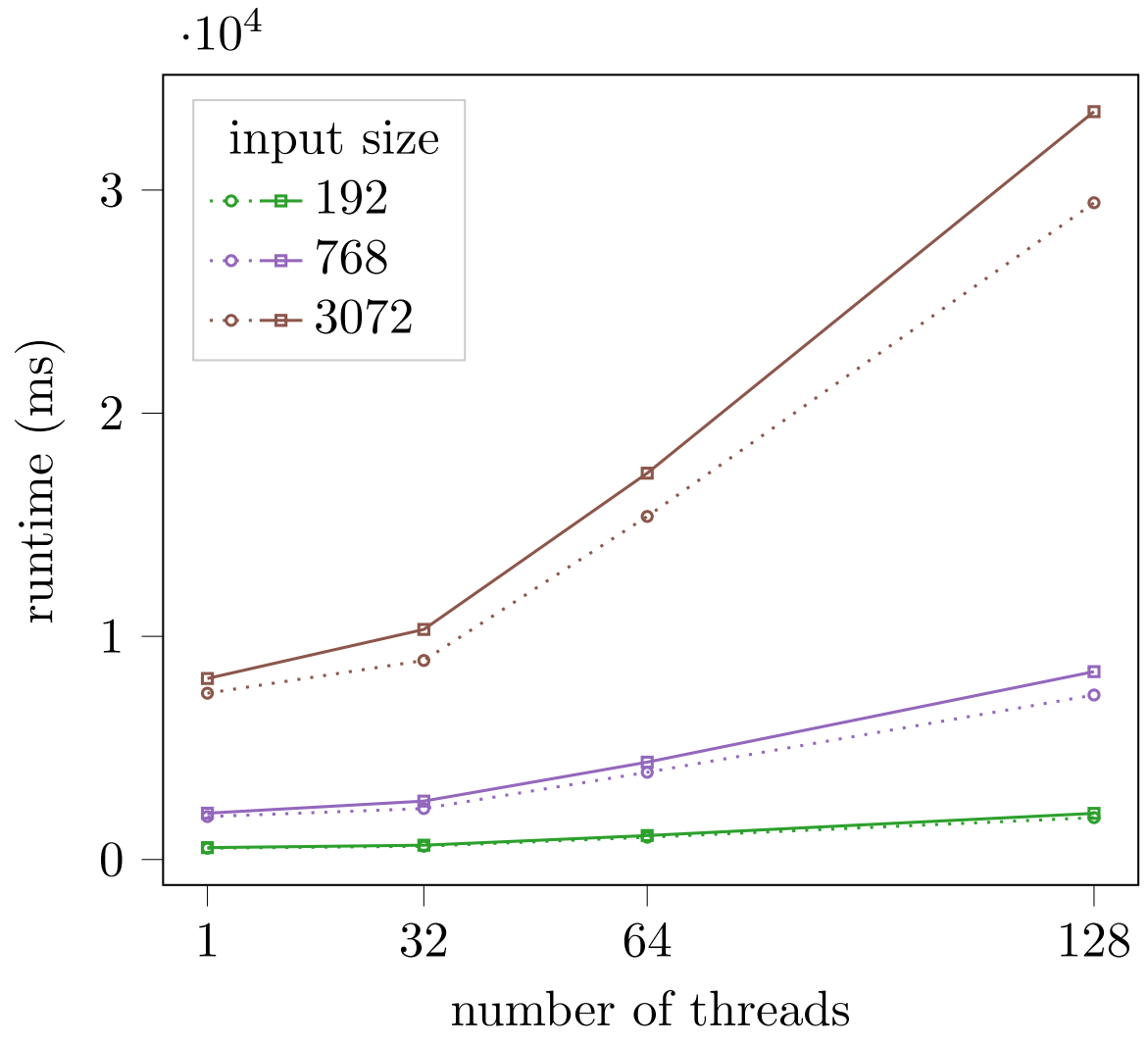

(a) Actual runtimes.

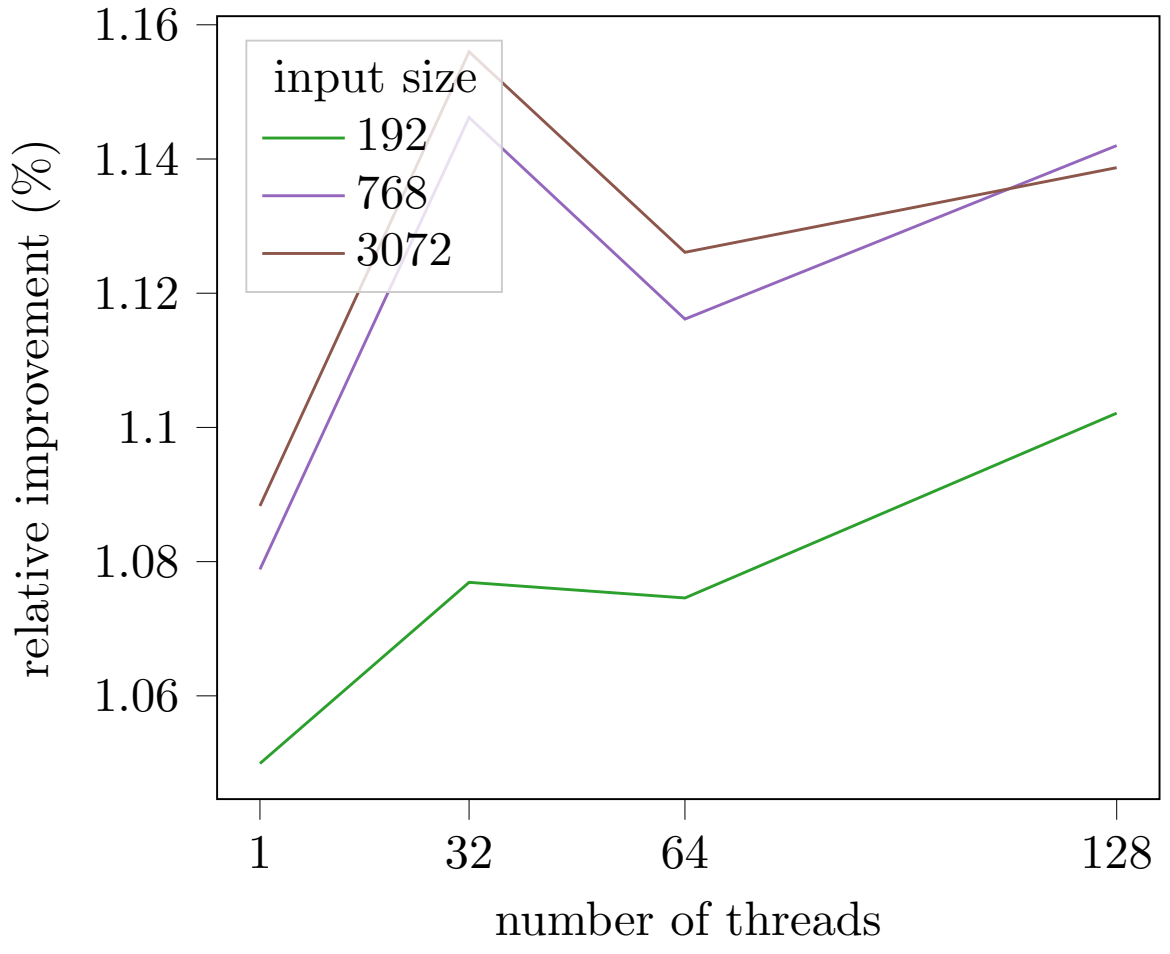

(b) Relative runtimes.

Evaluation on `deeplabv3_resnet50`.

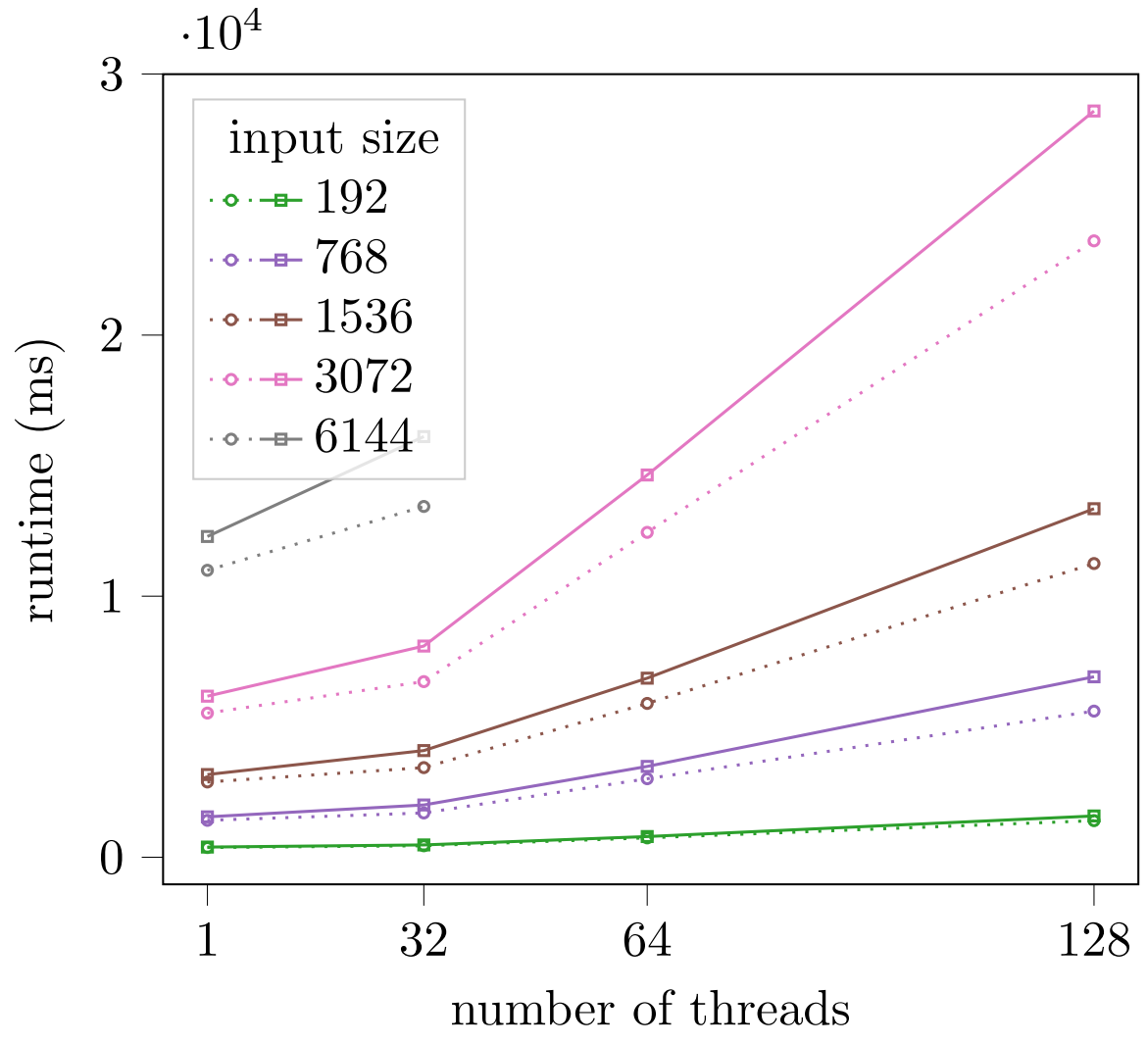

(a) Actual runtimes.

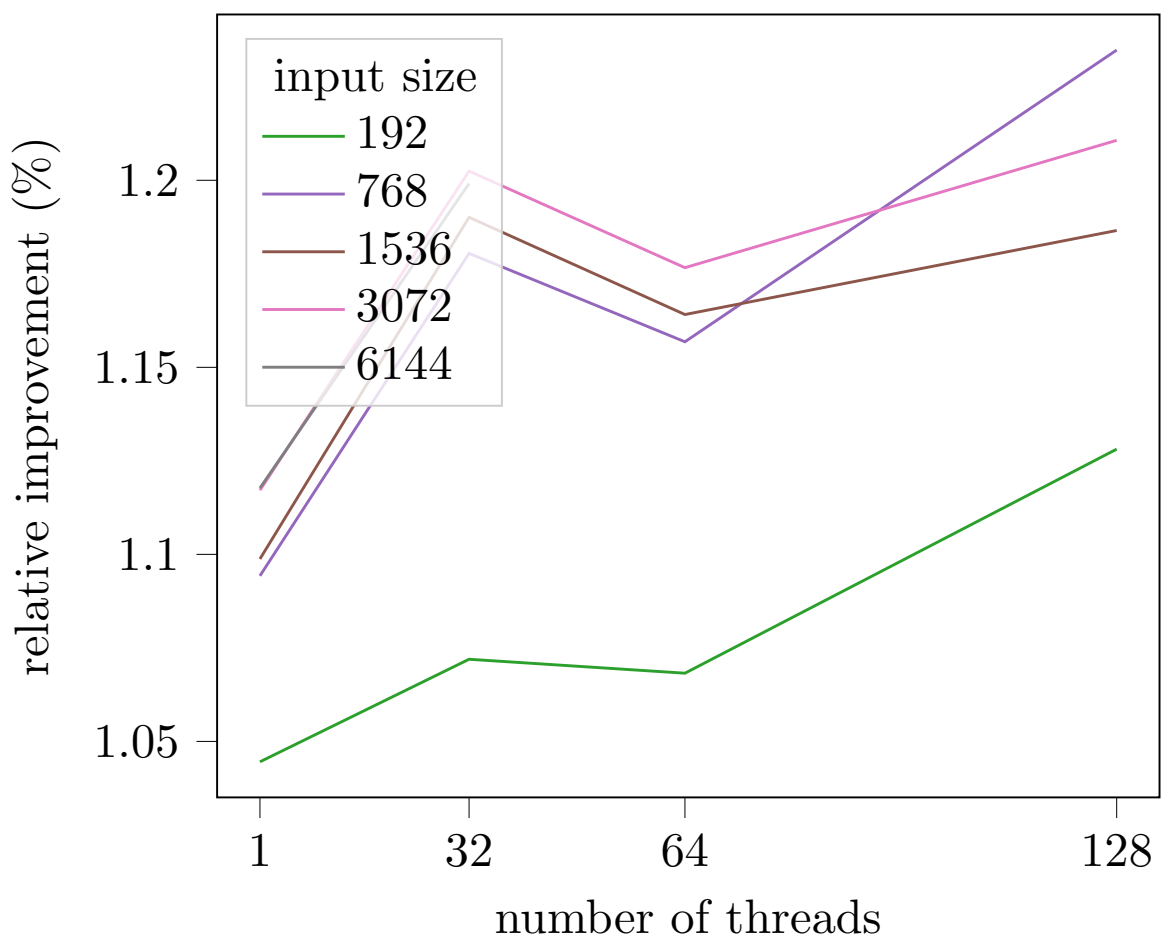

(b) Relative runtimes.

Evaluation on `fcn_resnet50`.

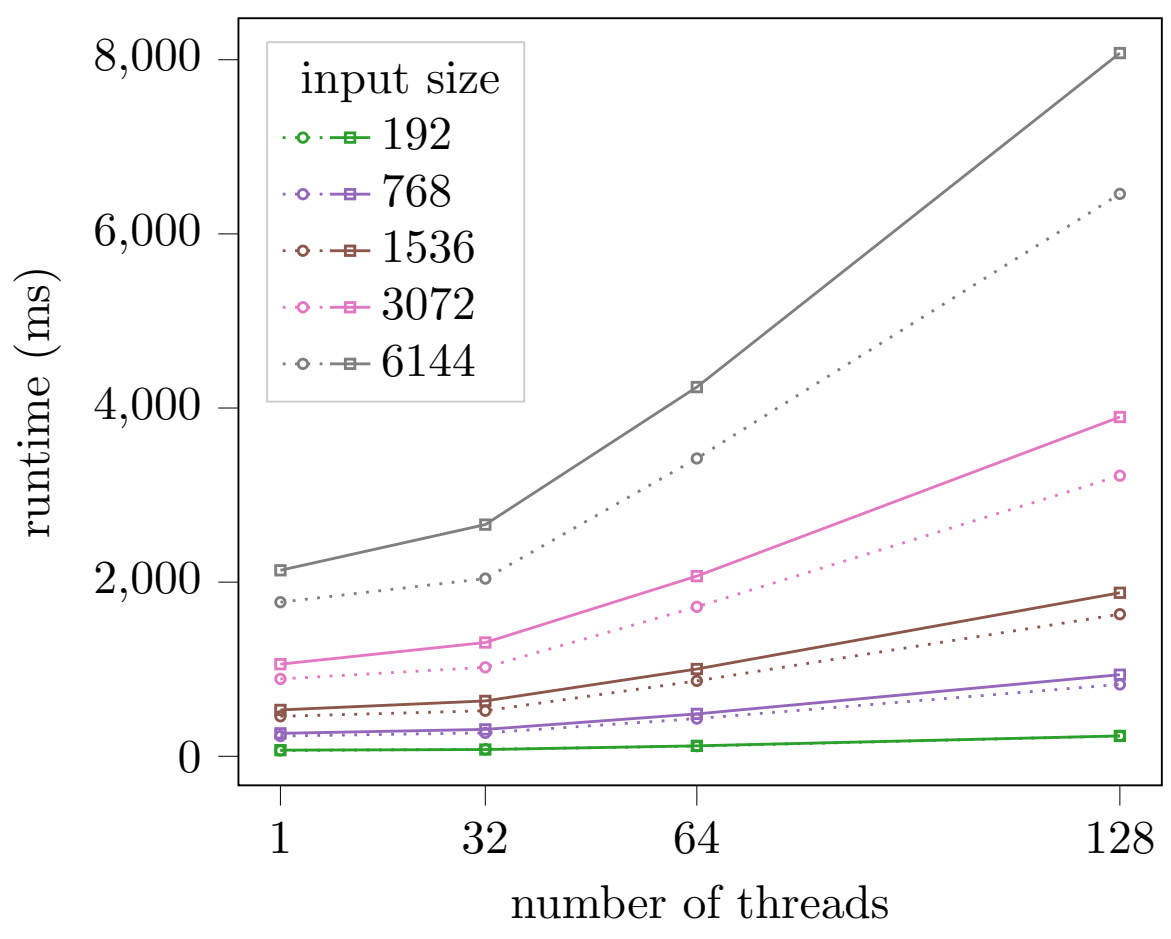

(a) Actual runtimes.

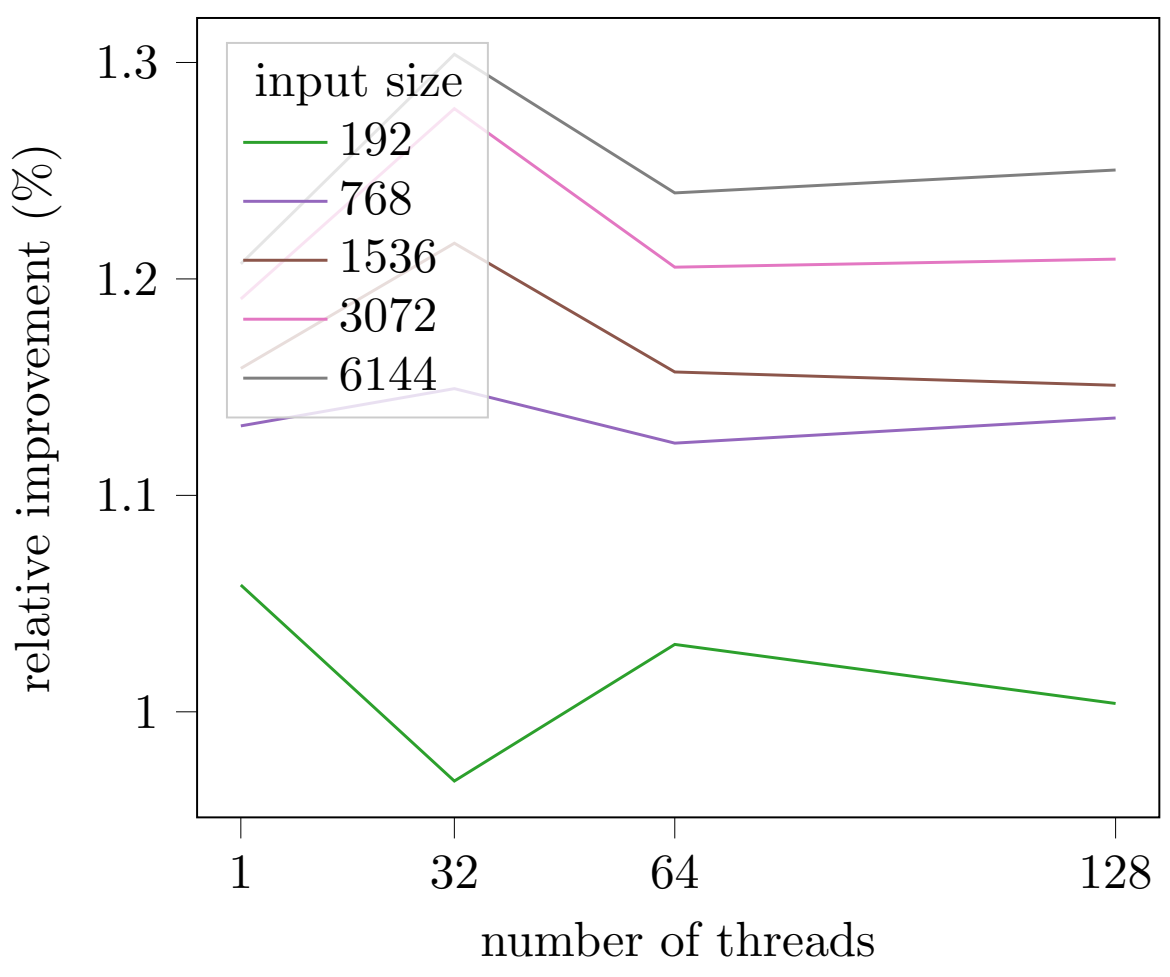

(b) Relative runtimes.

Evaluation on `googlenet`.

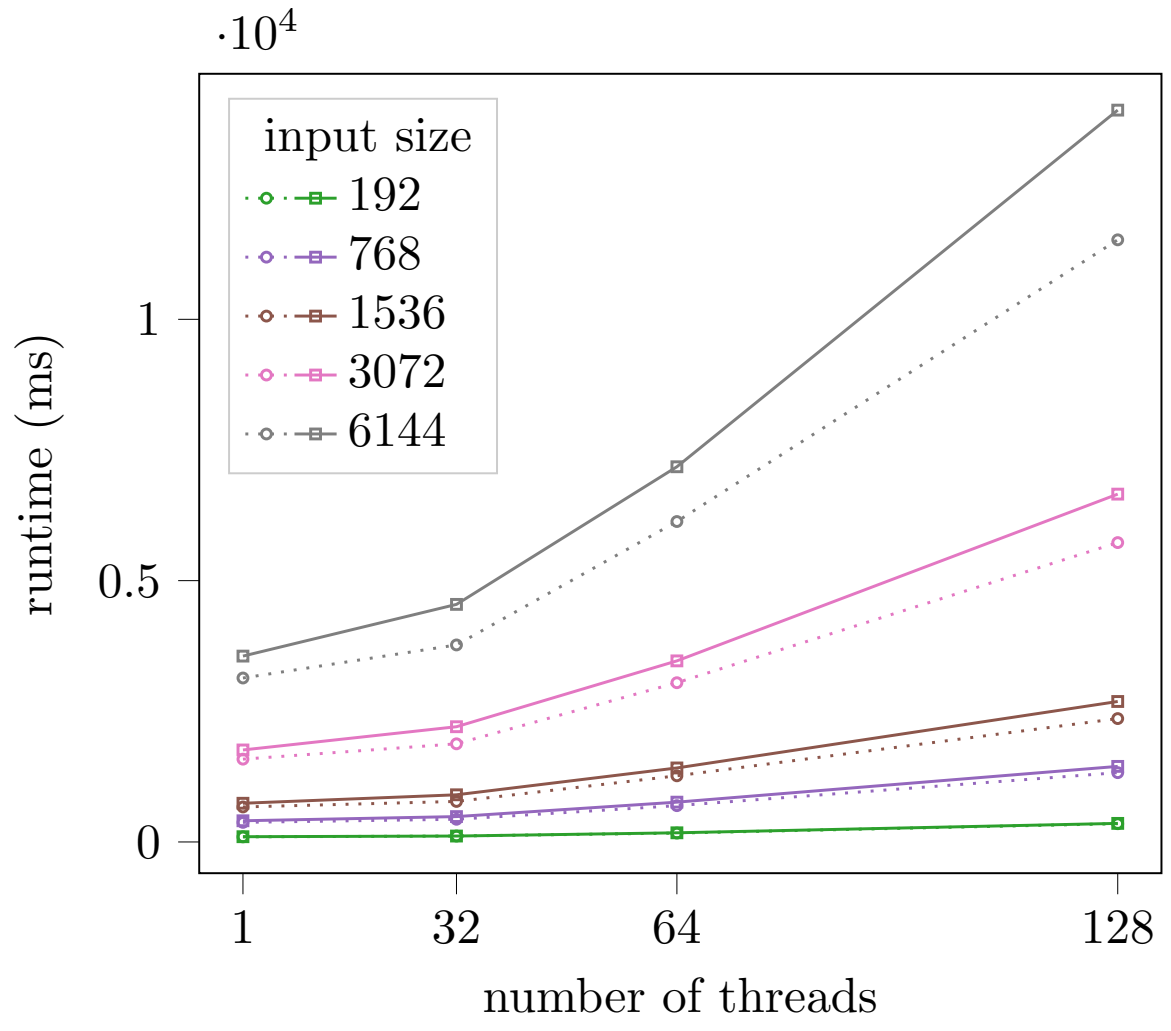

(a) Actual runtimes.

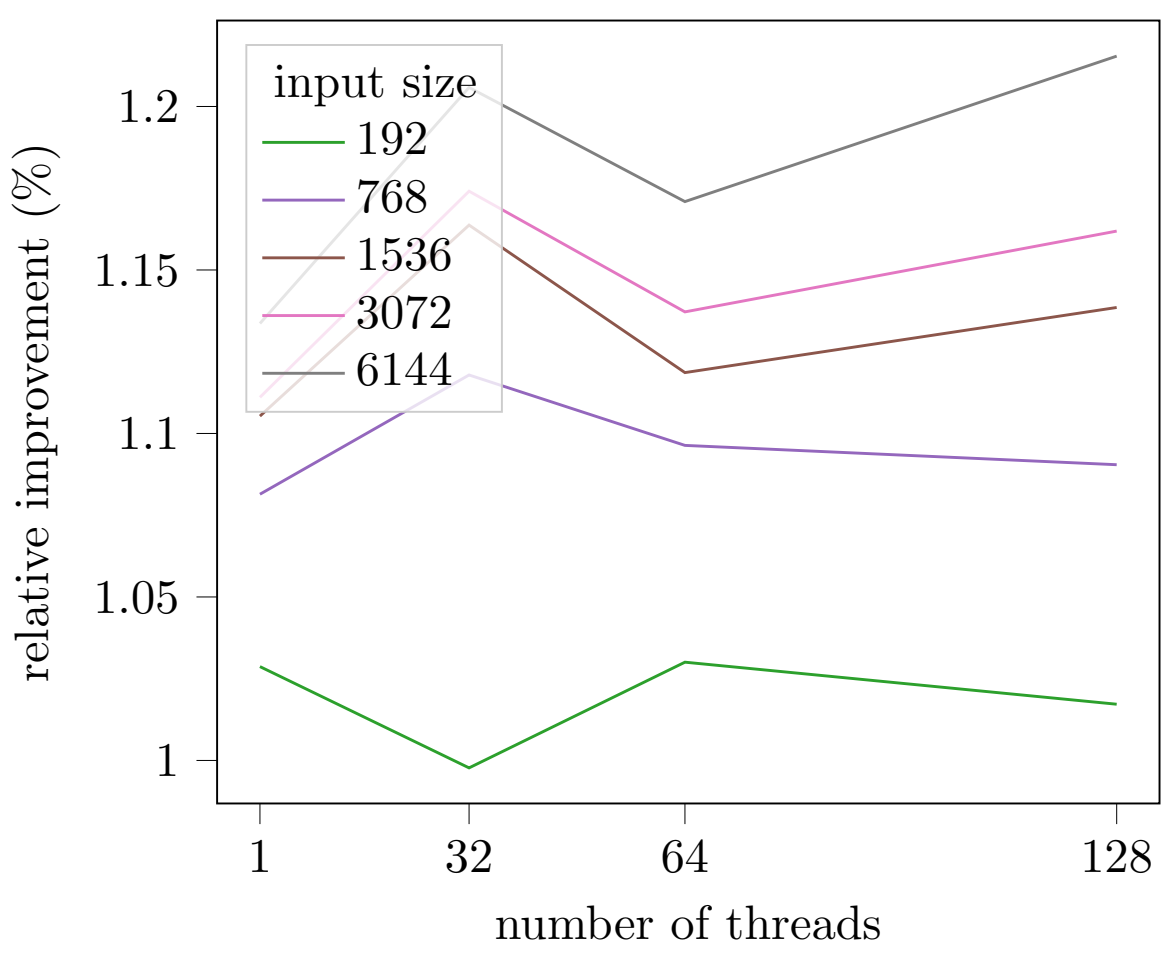

(b) Relative runtimes.

Evaluation on `inception_v3`.

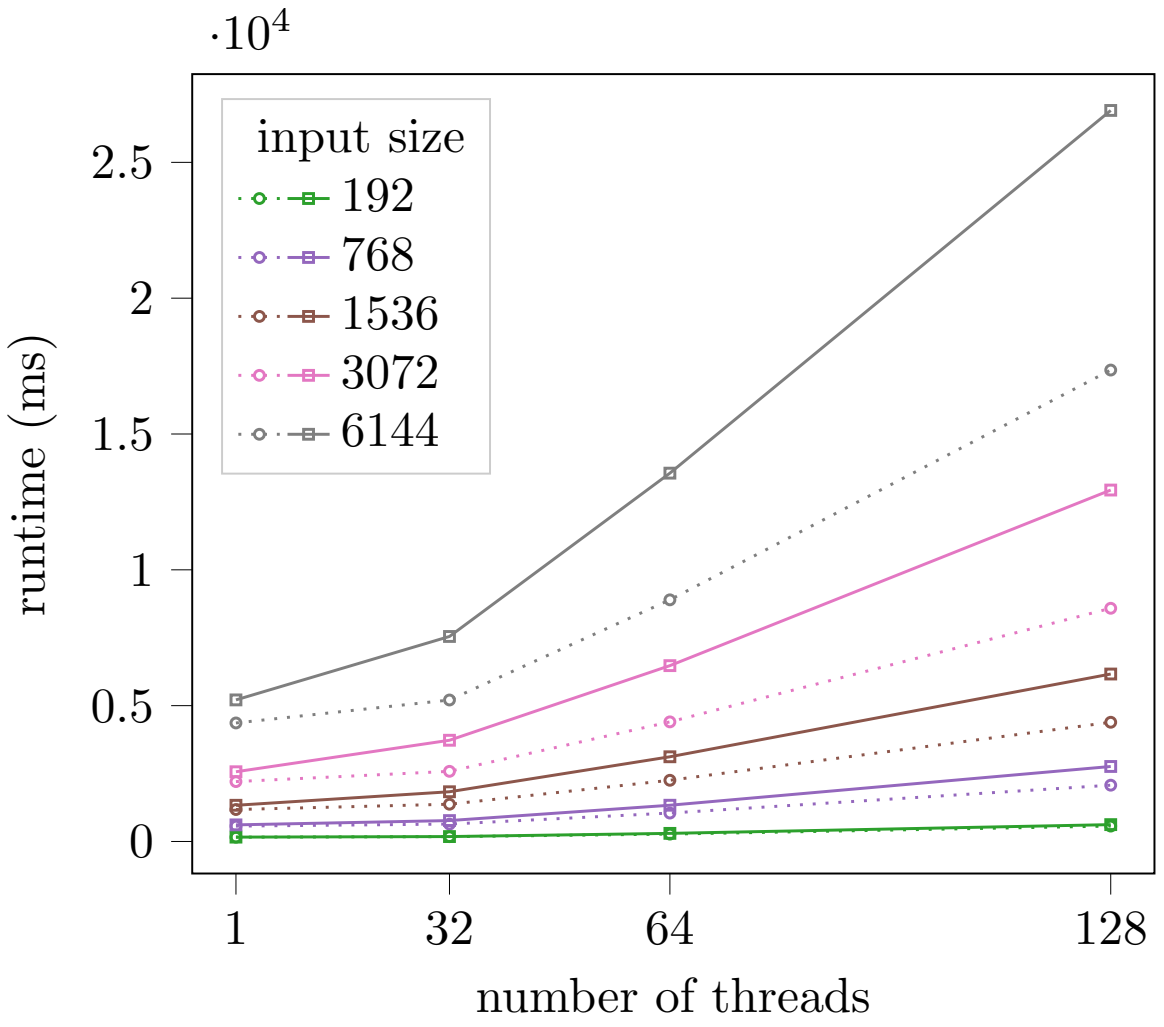

(a) Actual runtimes.

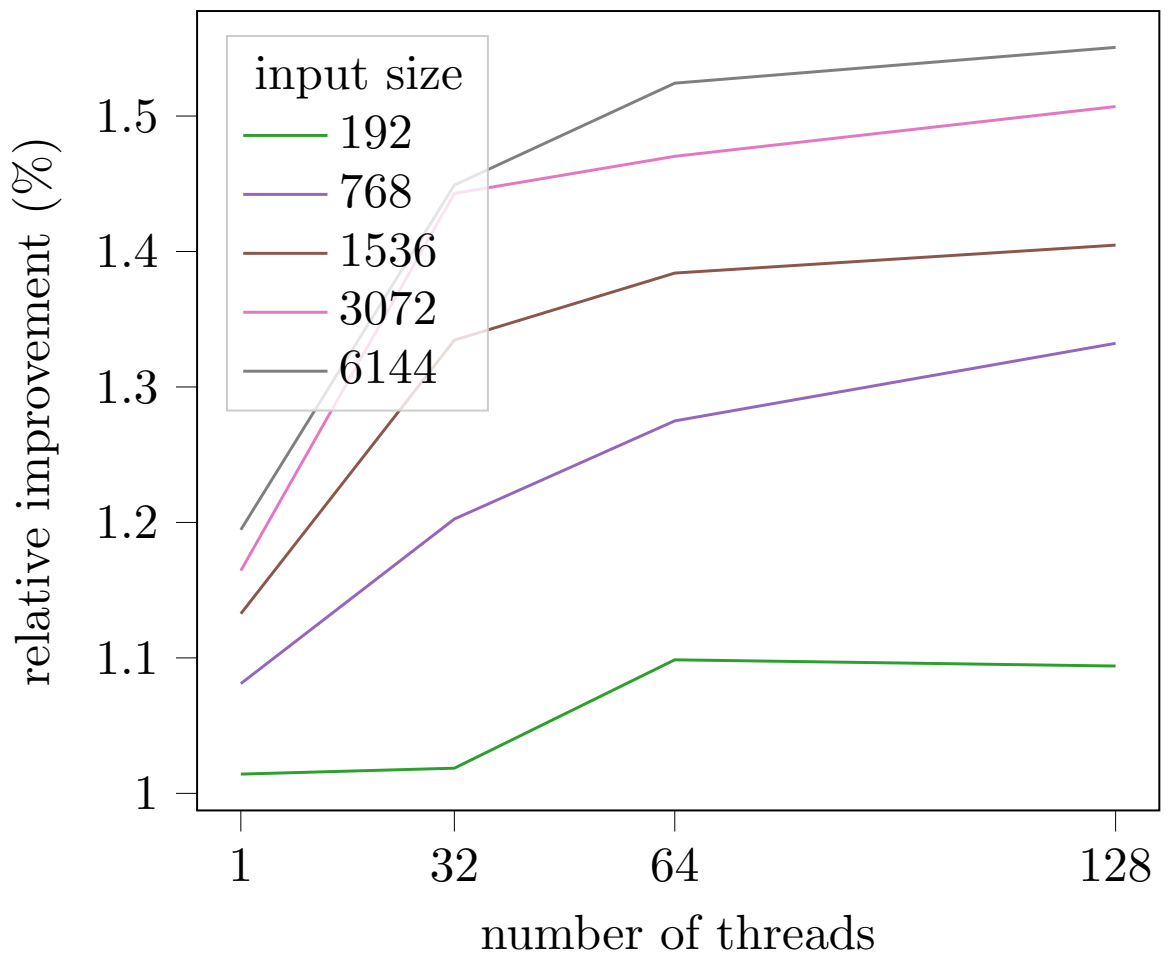

(b) Relative runtimes.

Evaluation on `regnet_x_8gf`.

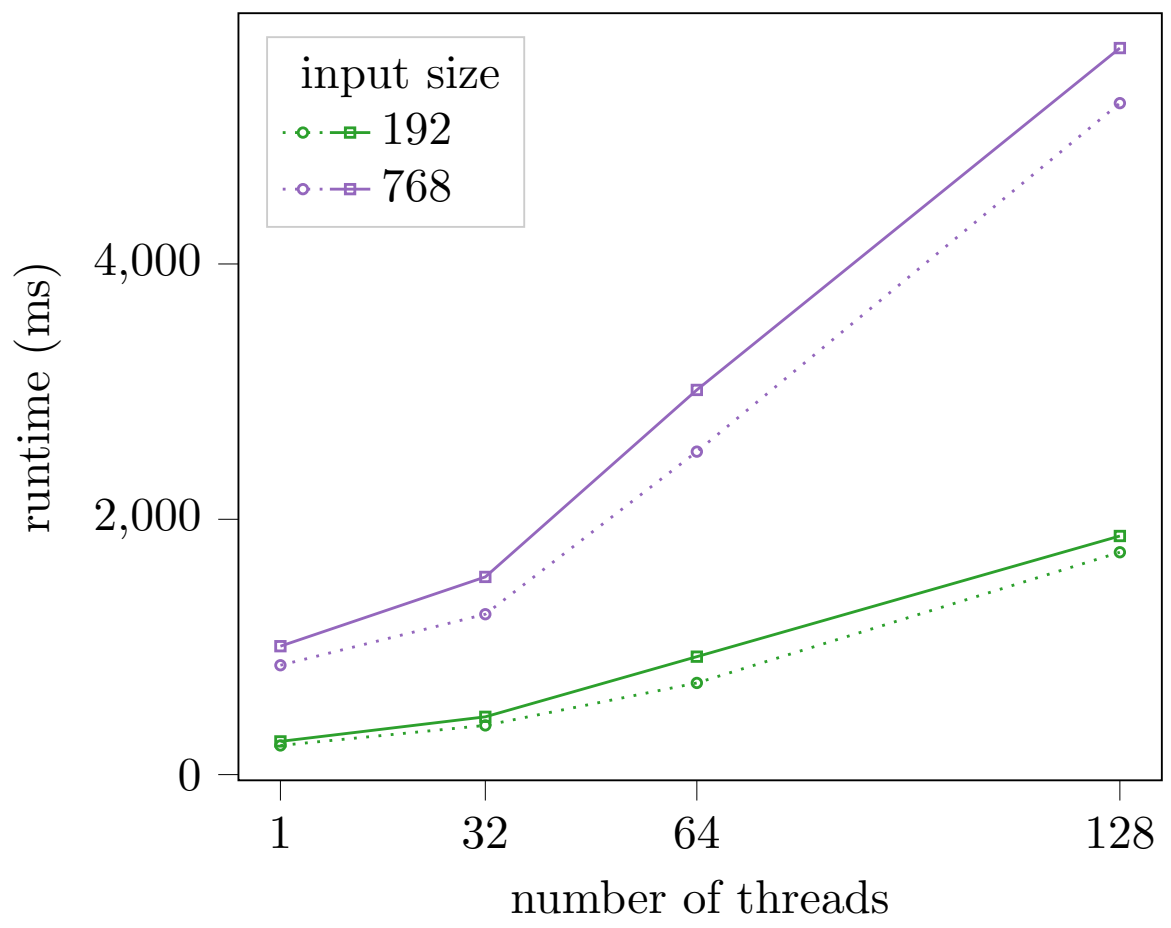

(a) Actual runtimes.

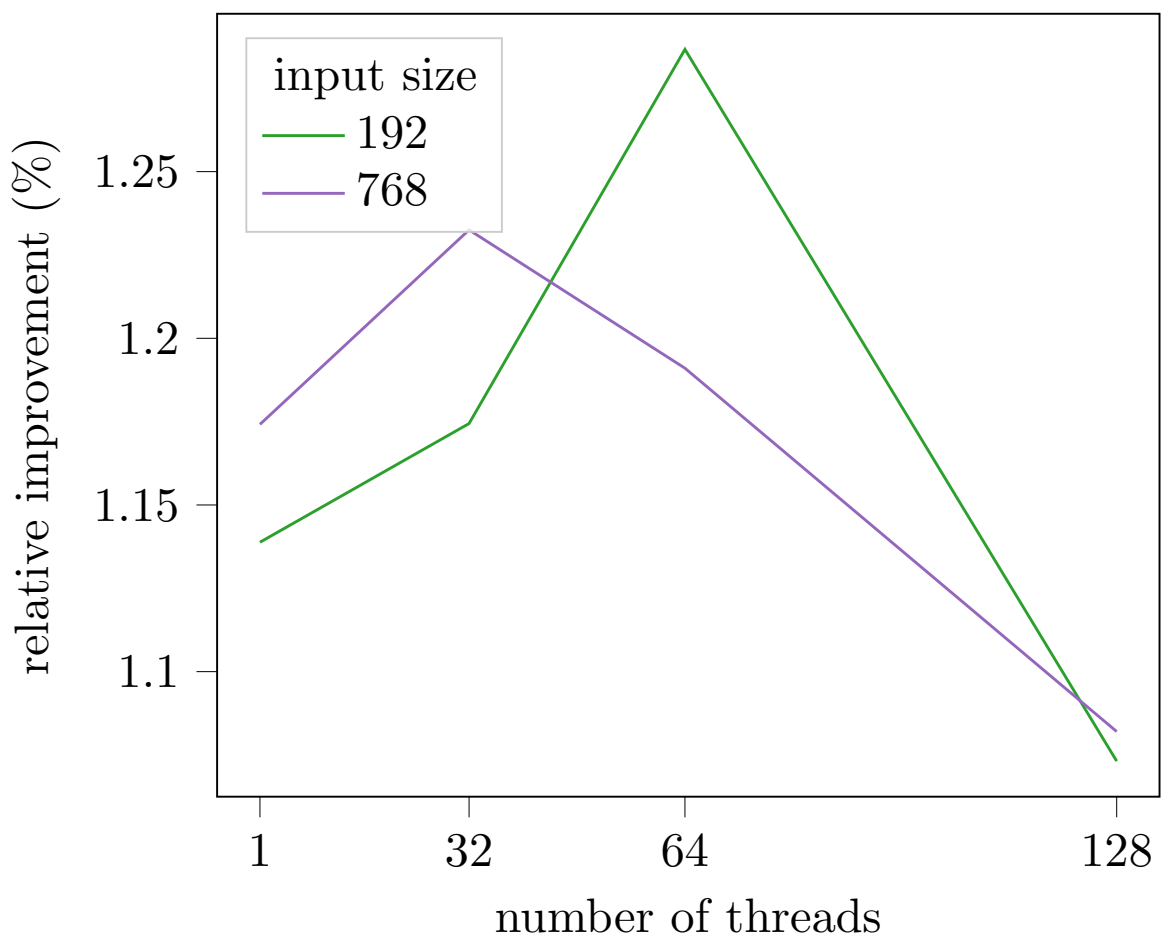

(b) Relative runtimes.

Evaluation on `vgg16`.

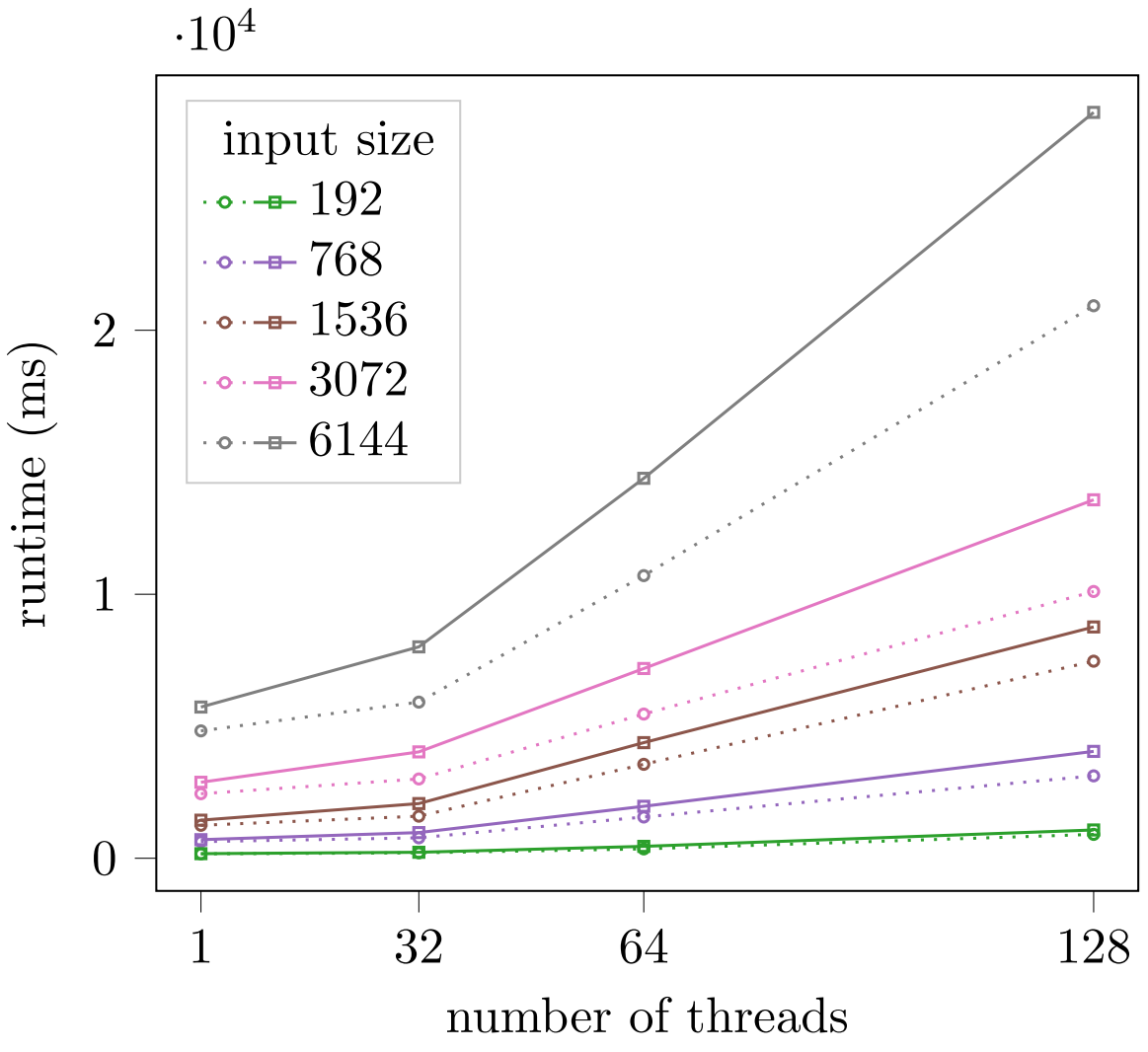

(a) Actual runtimes.

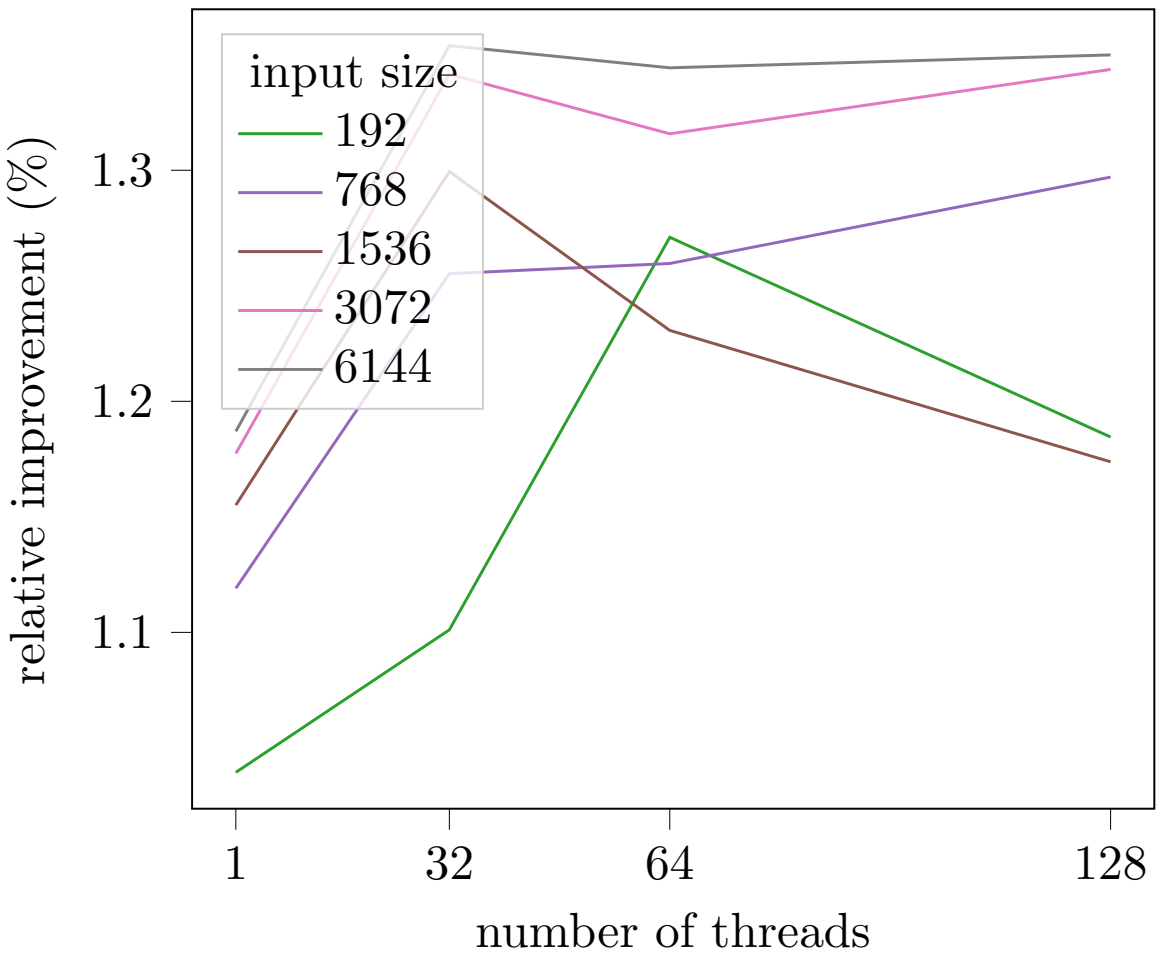

(b) Relative runtimes.

Evaluation on `wide_resnet50_2`.

### *8.1.2 Platform 2*

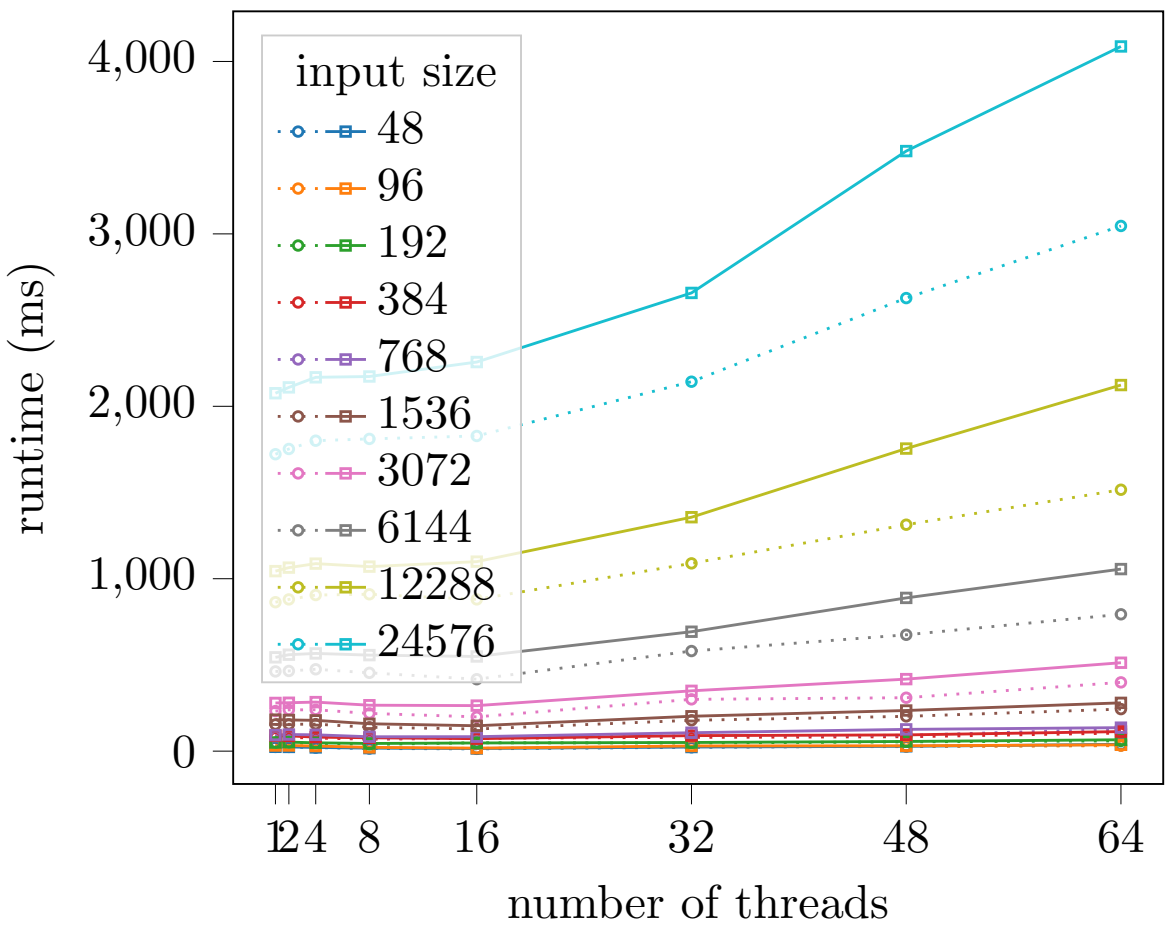

(a) Actual runtimes.

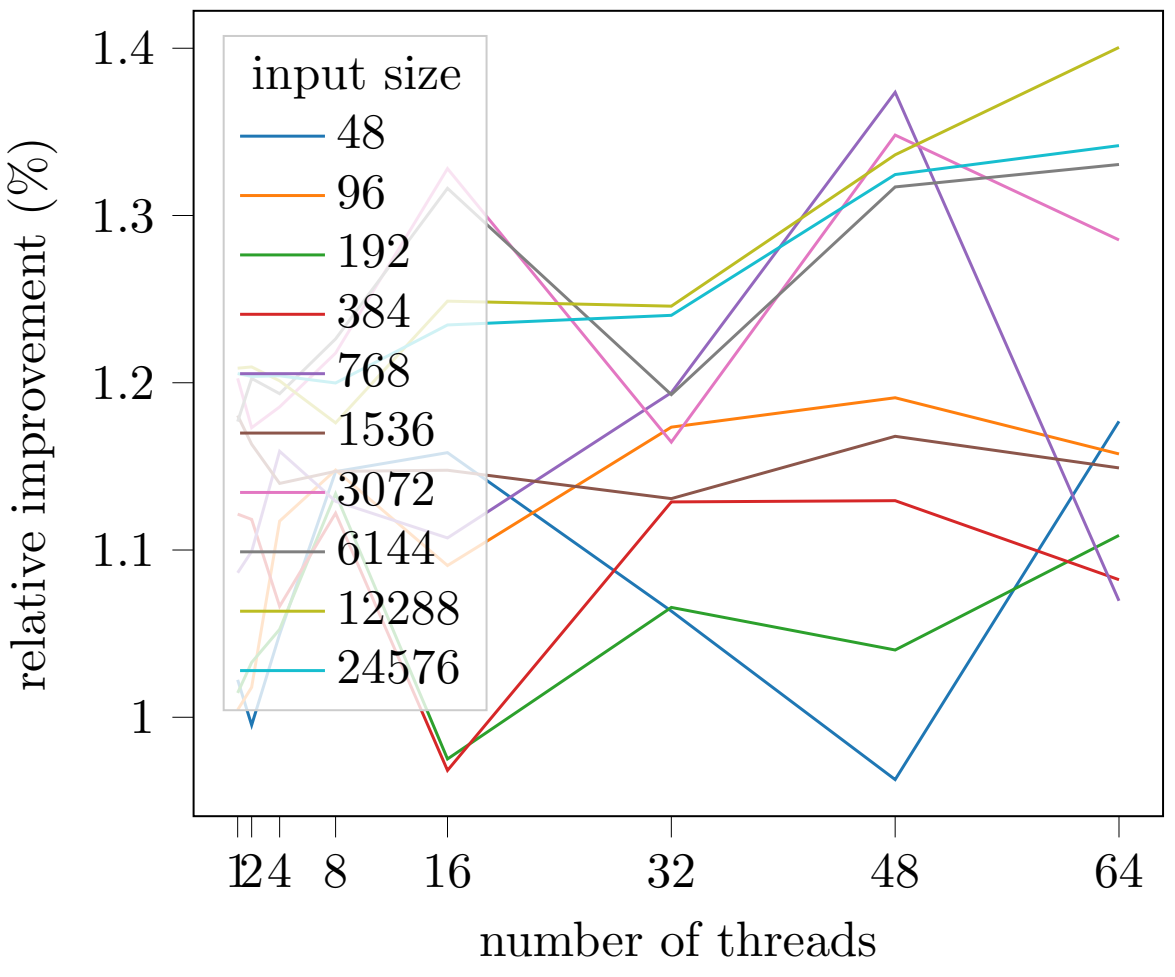

(b) Relative runtimes.

Evaluation on `alexnet`.

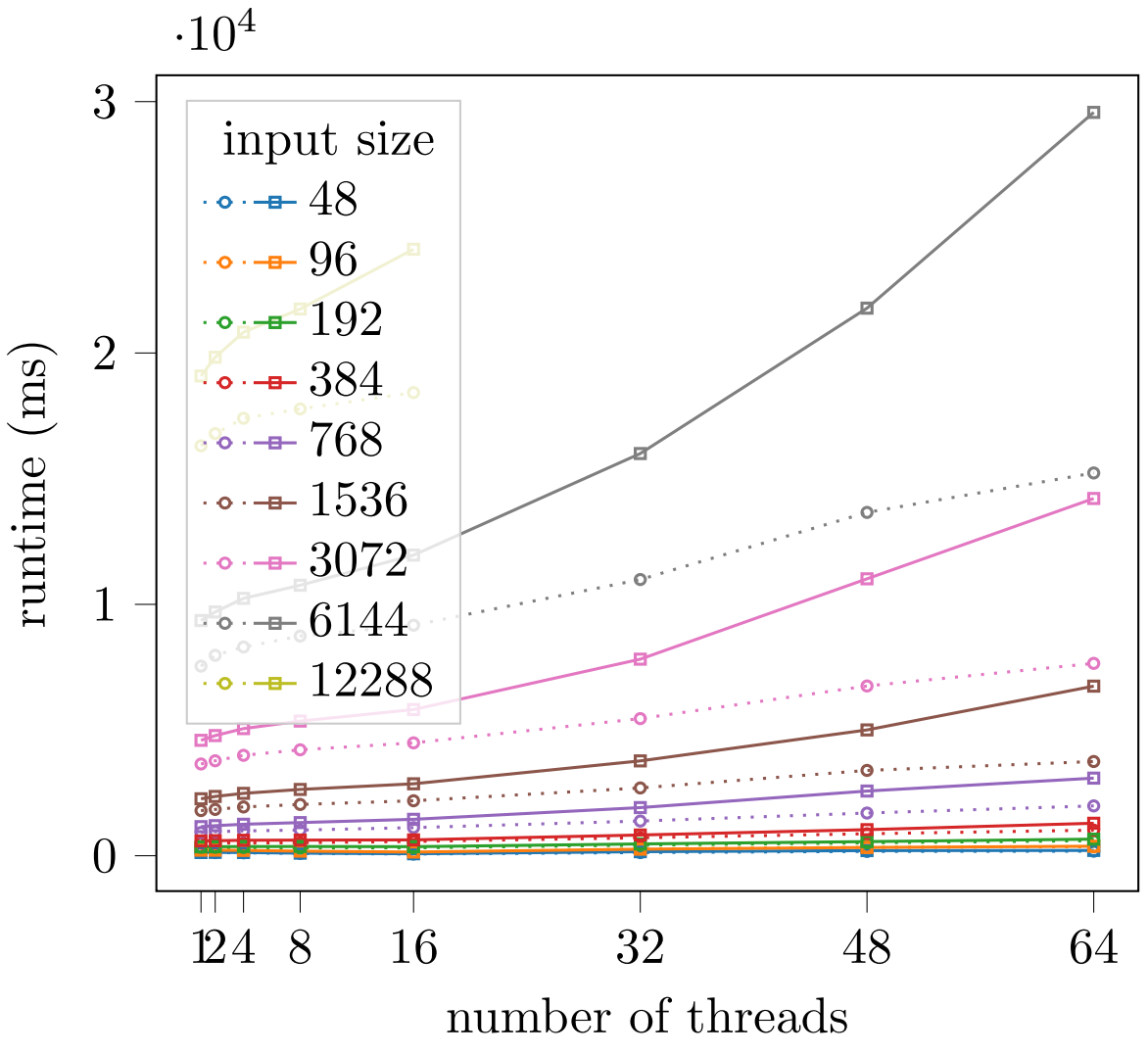

(a) Actual runtimes.

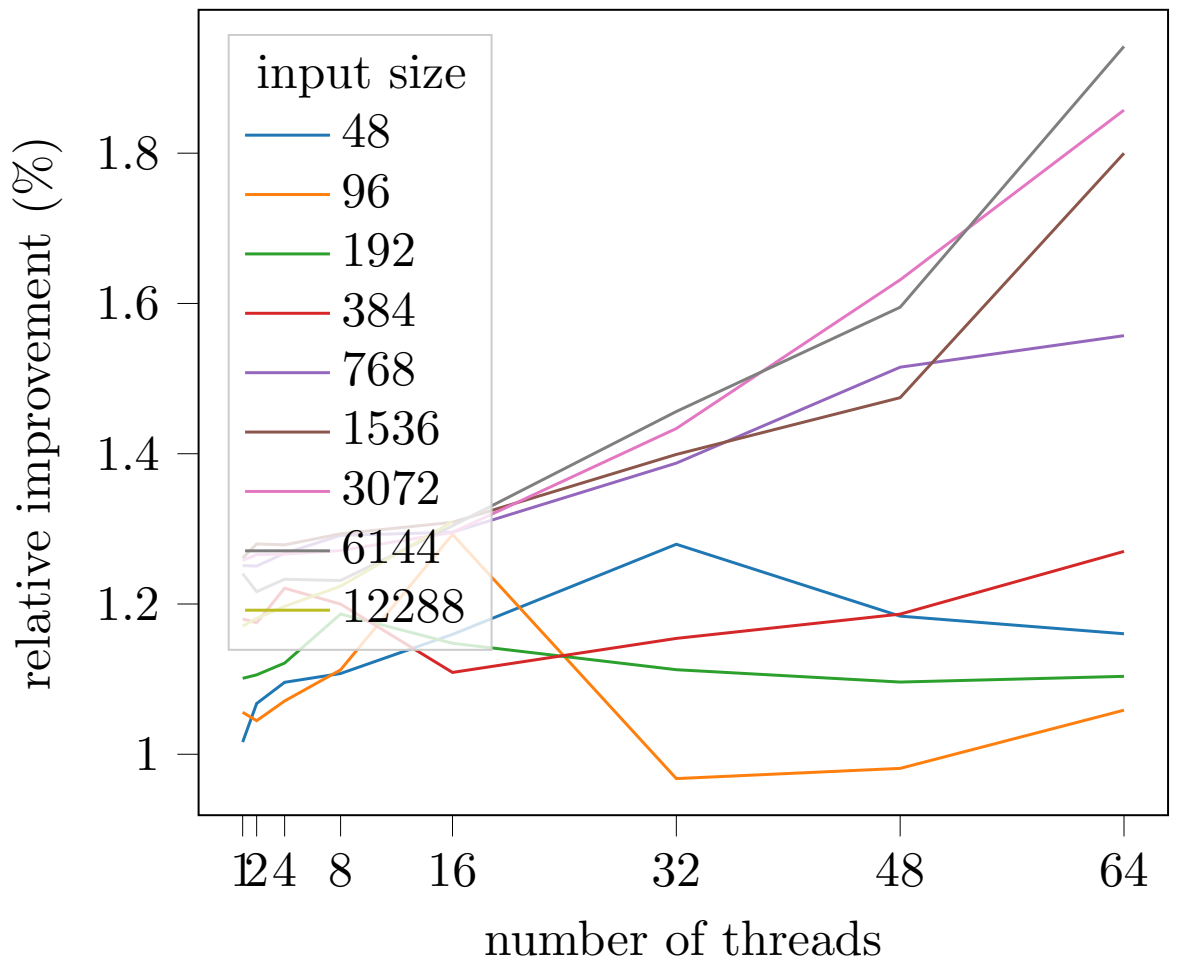

(b) Relative runtimes.

Evaluation on `densenet161`.

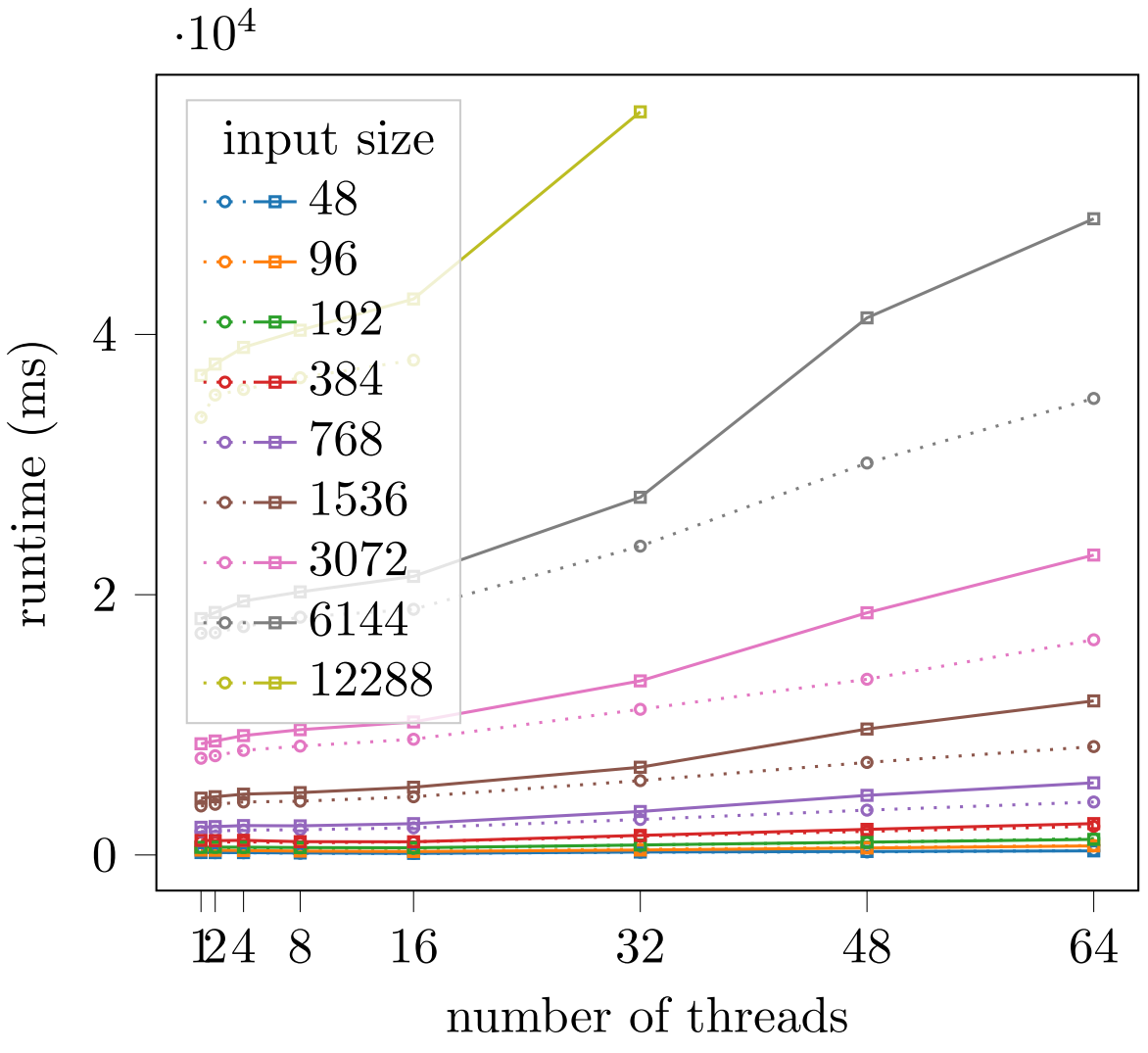

(a) Actual runtimes.

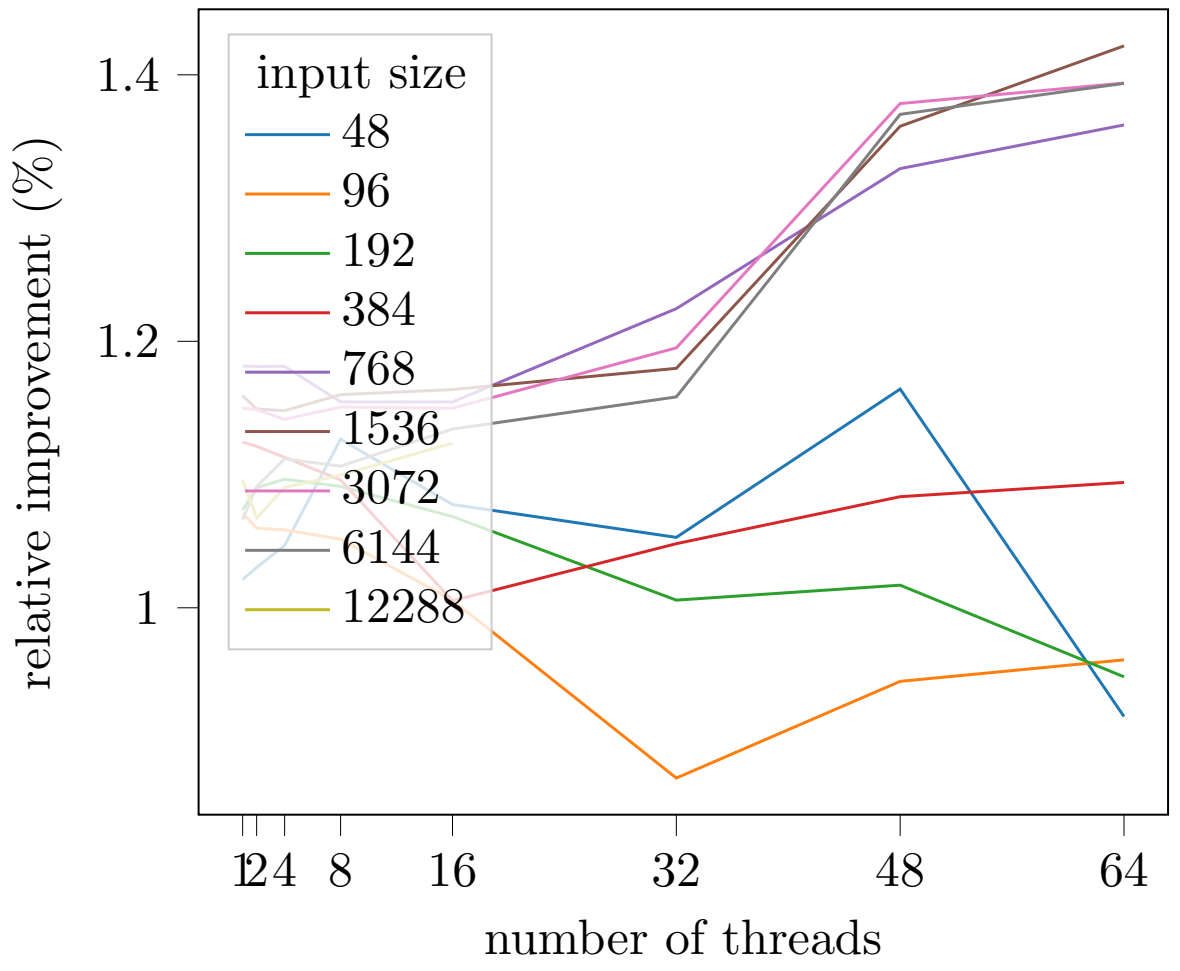

(b) Relative runtimes.

Evaluation on `regnet_x_32gf`.

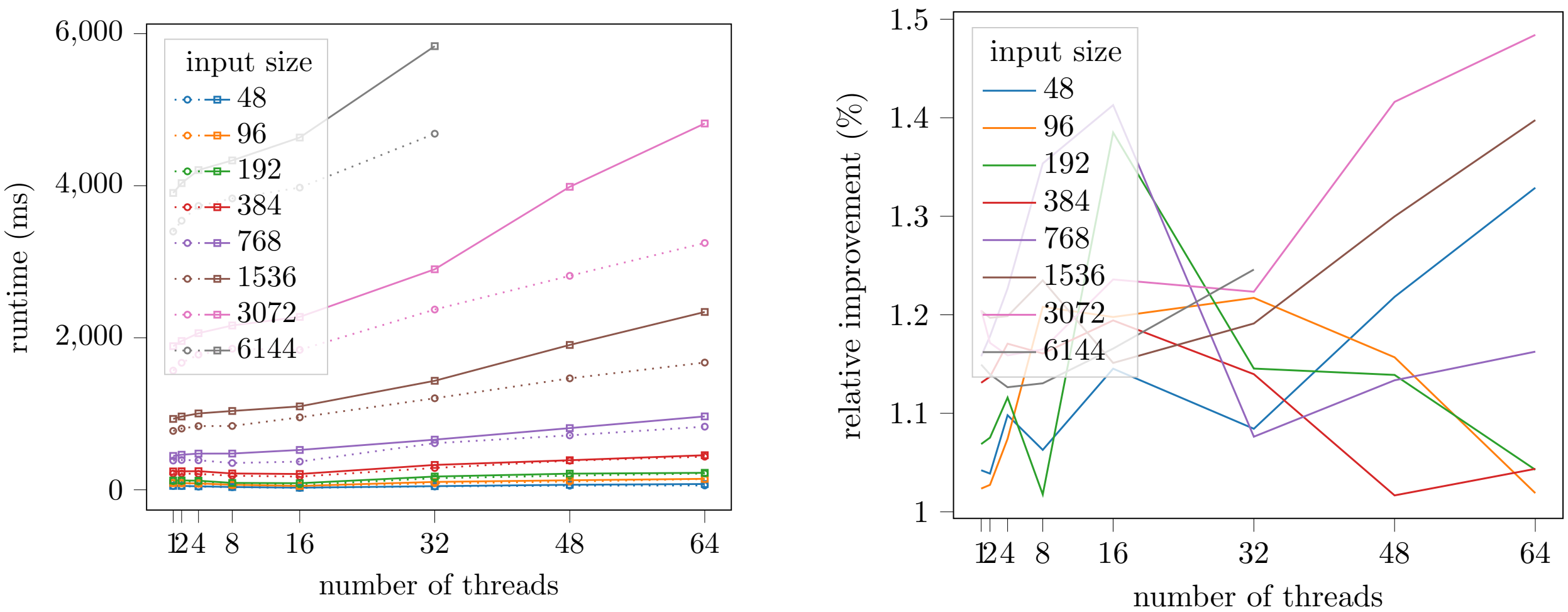

(a) Actual runtimes.

(b) Relative runtimes.

Evaluation on `resnet50`.

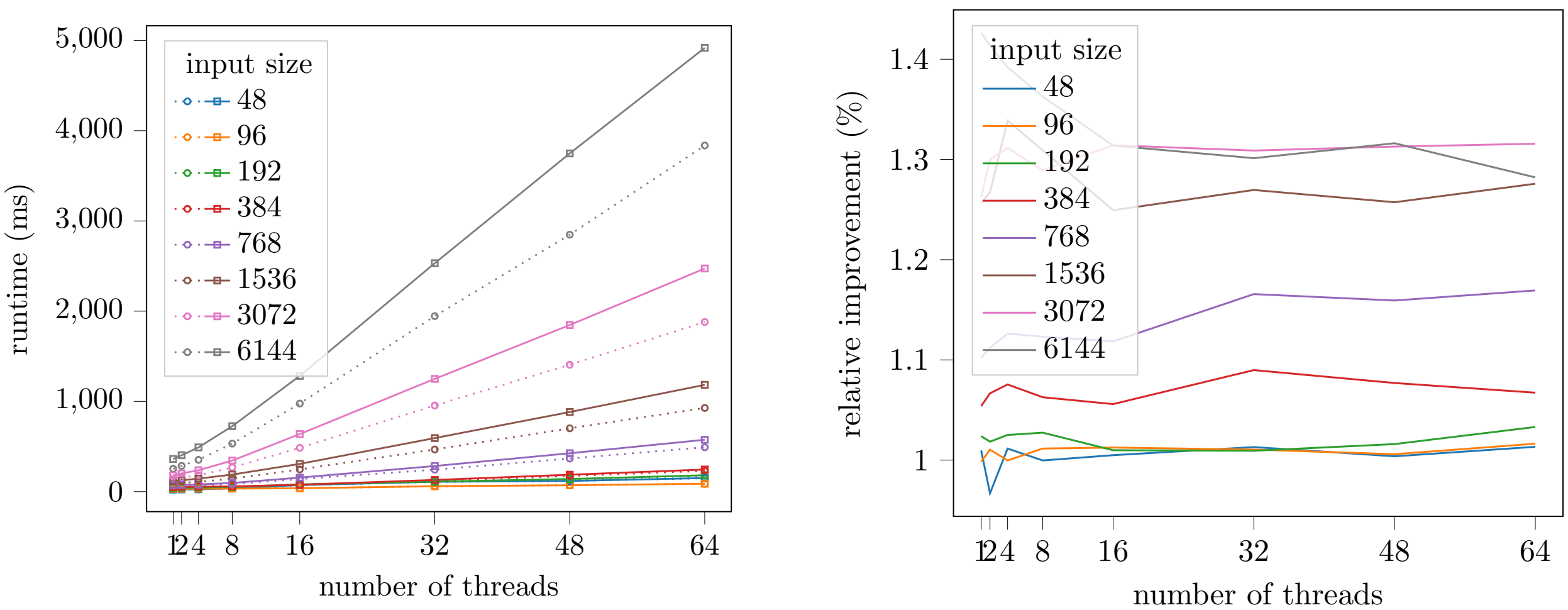

(a) Actual runtimes.

(b) Relative runtimes.

Evaluation on `squeezenet1_0`.

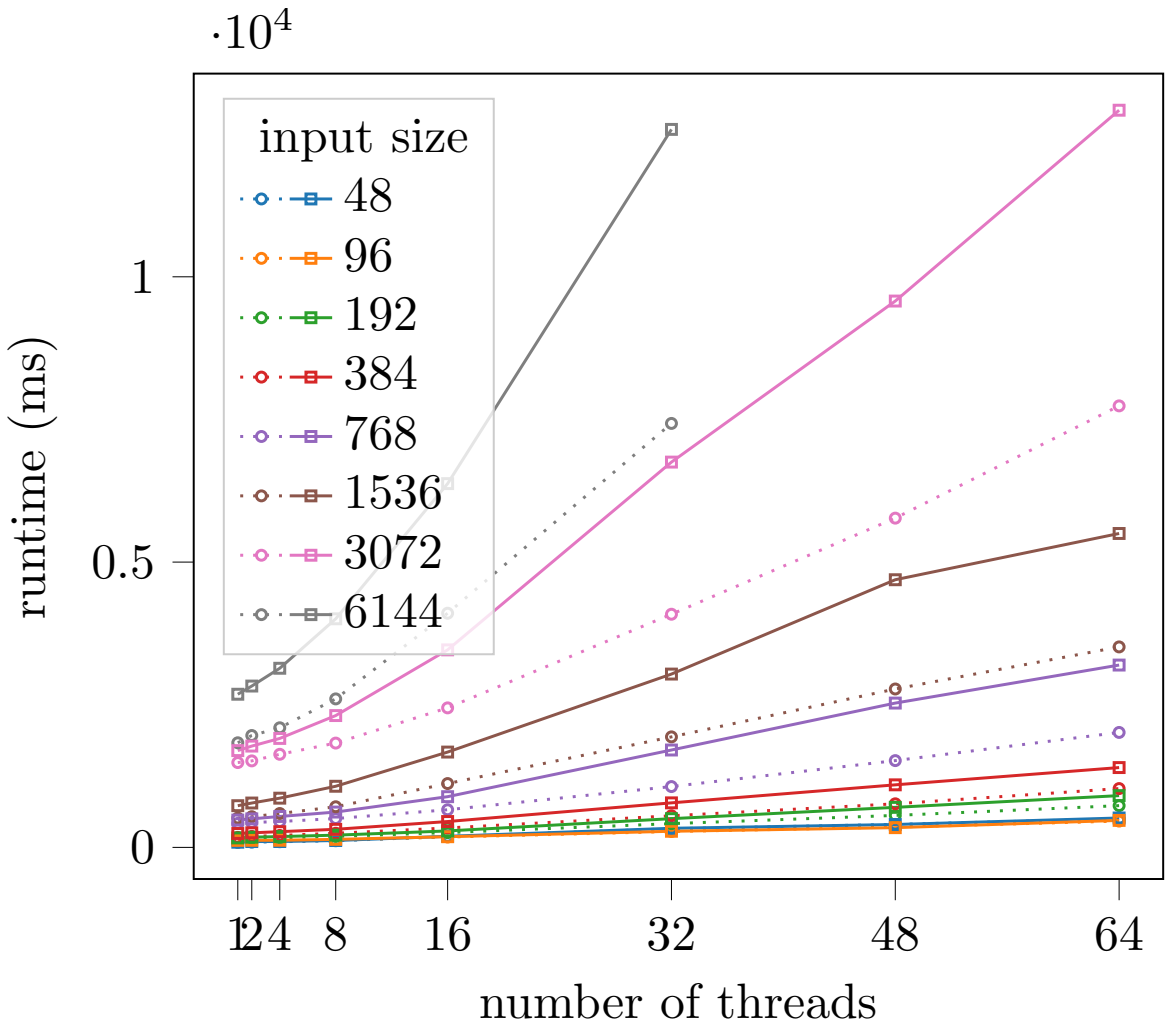

(a) Actual runtimes.

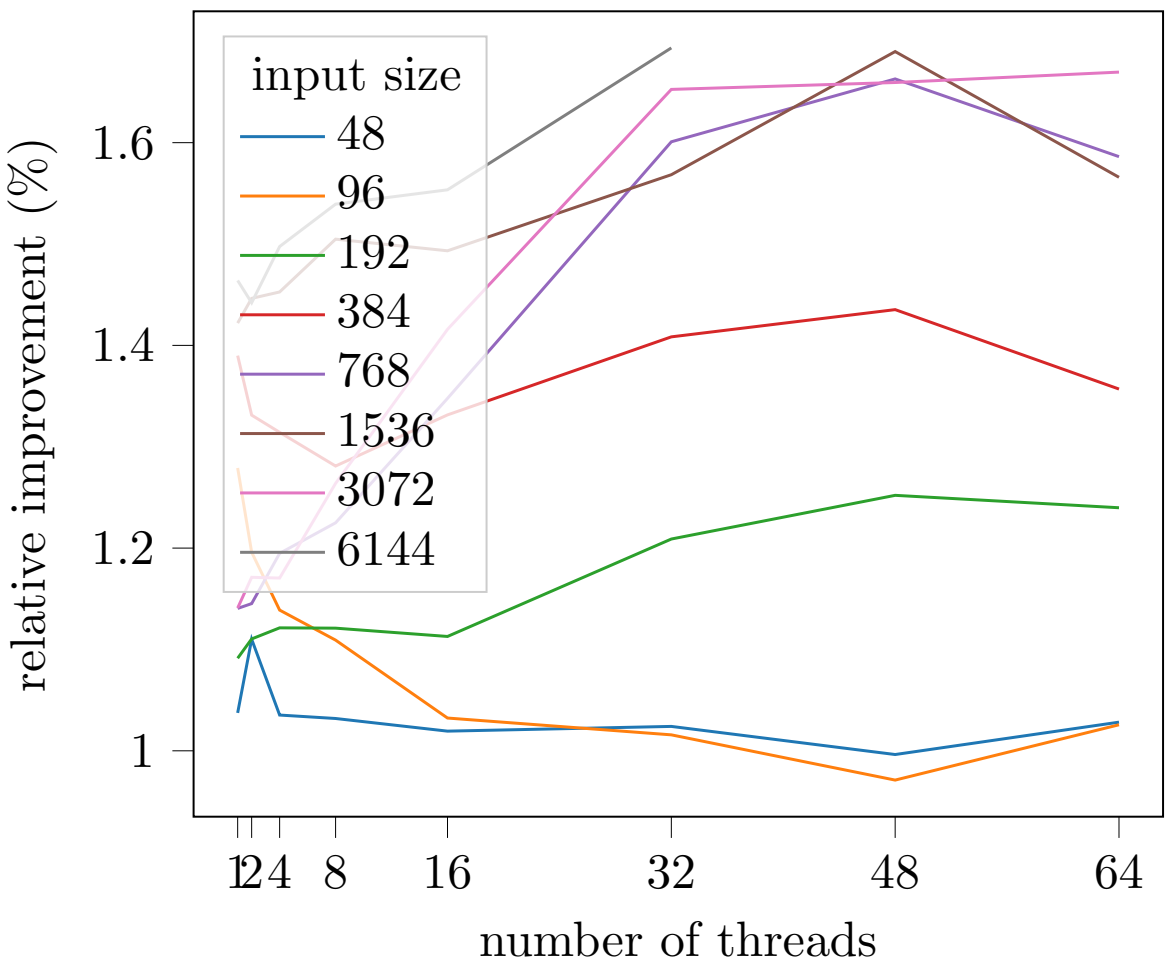

(b) Relative runtimes.

Evaluation on `vgg13_bn`.

## 8.2 Heap Maps for Memory Planning Strategies

We present "heap maps" generated by memory planning strategies for input size = 128. We pair these with the `mapped` statistics reported by `jemalloc` for the same configuration.

### *alexnet*

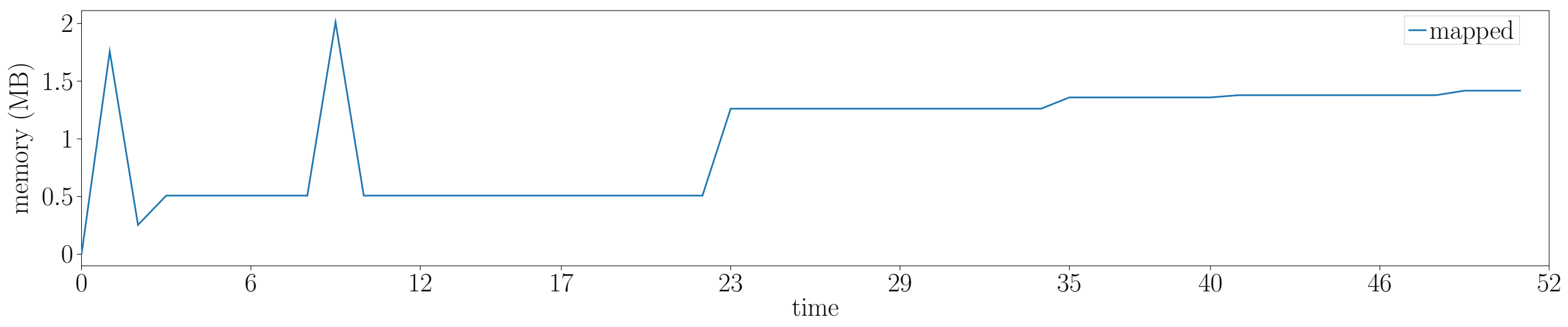

Figure 8.15: `jemalloc`

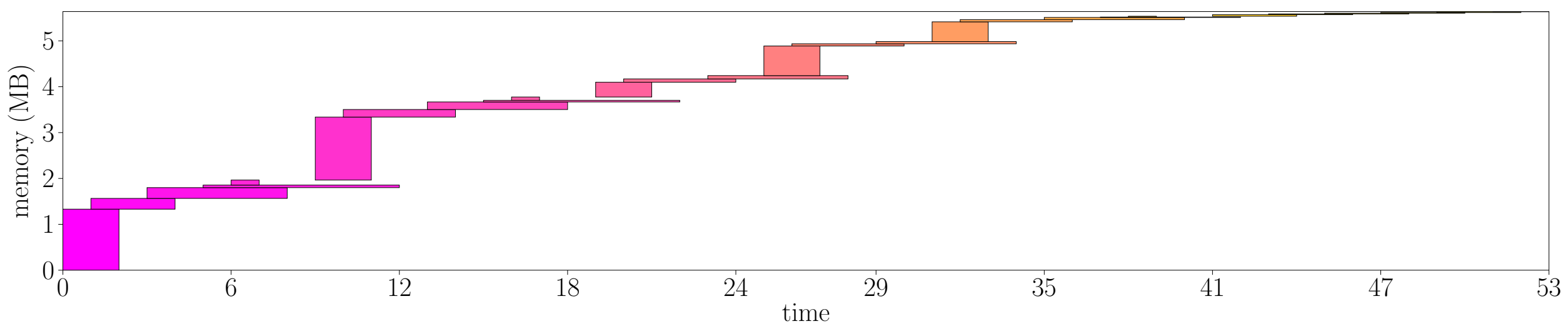

Figure 8.16: `bump_allocation`

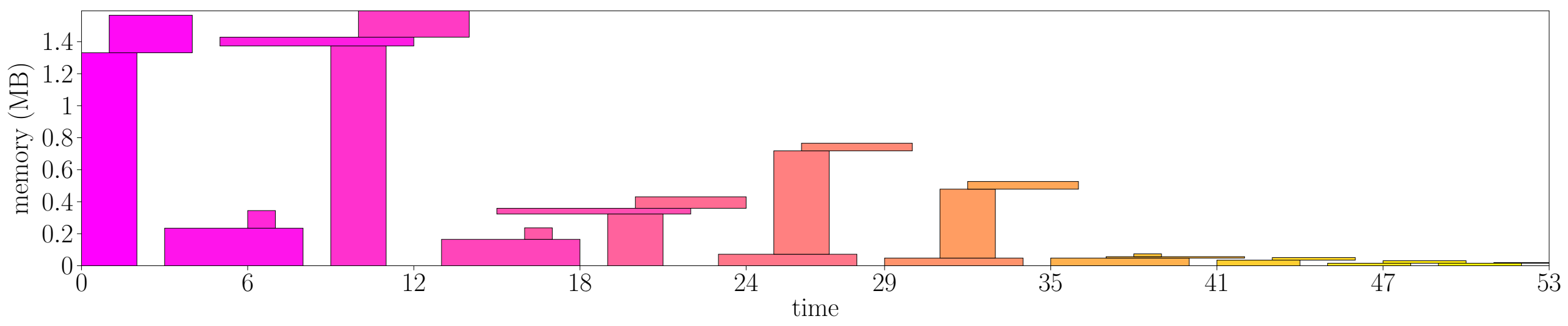

Figure 8.17: `gergov`

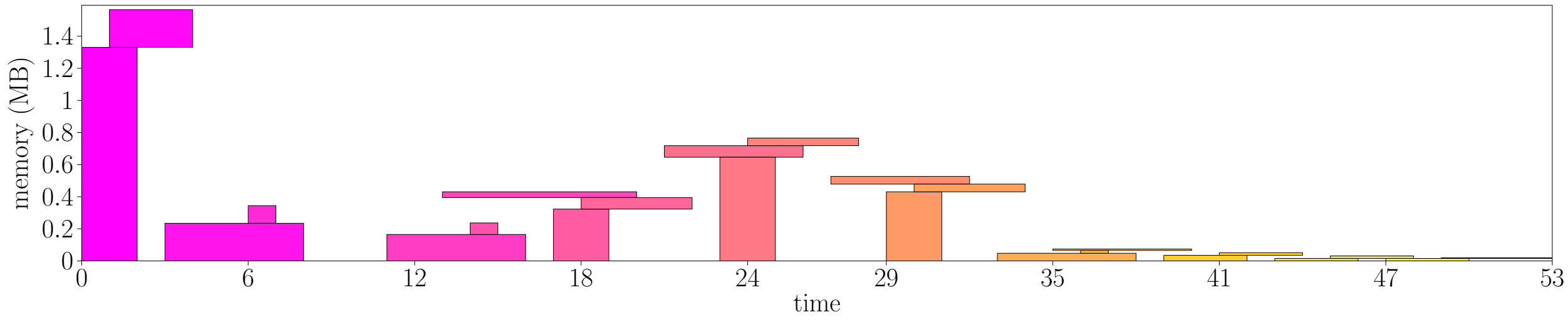

Figure 8.18: `greedy_by_size`

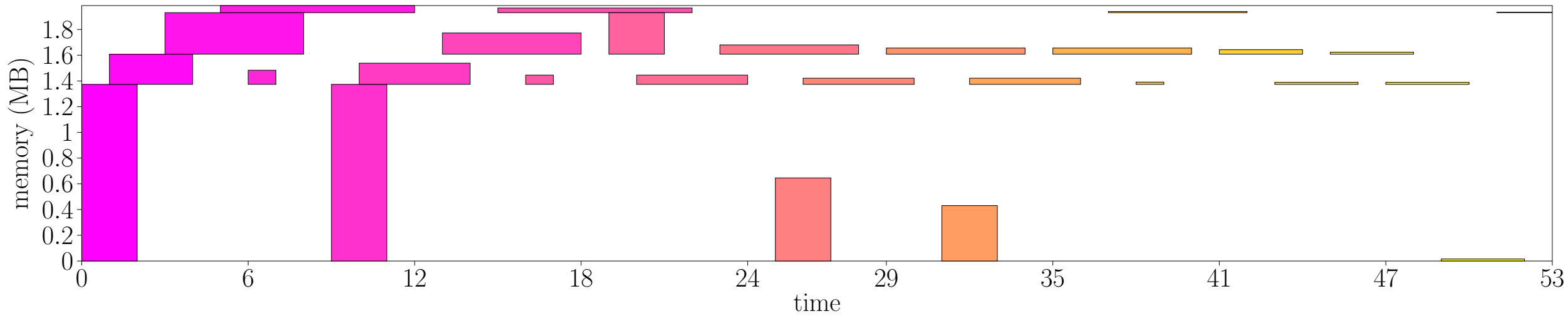

Figure 8.19: `mincost_flow`

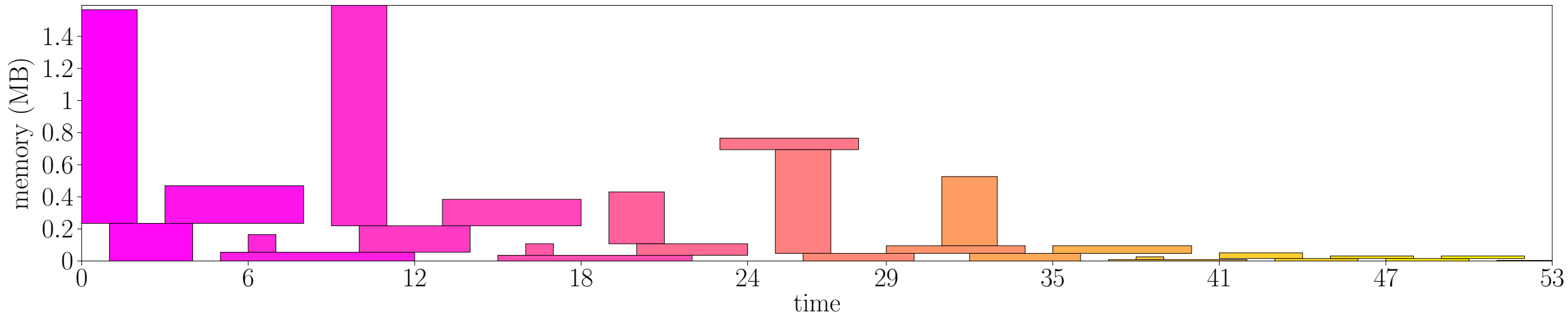

Figure 8.20: `mip`

## *dcgan*

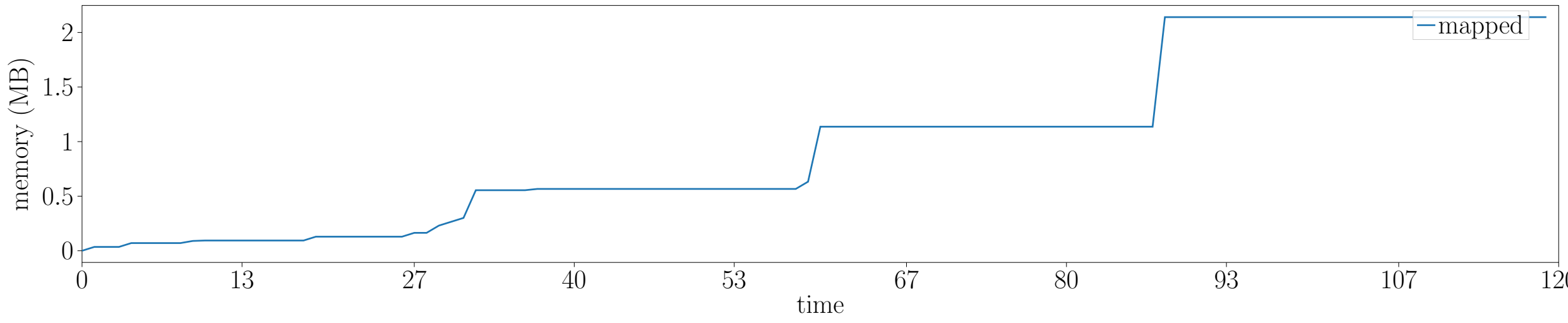

Figure 8.21: `jemalloc`

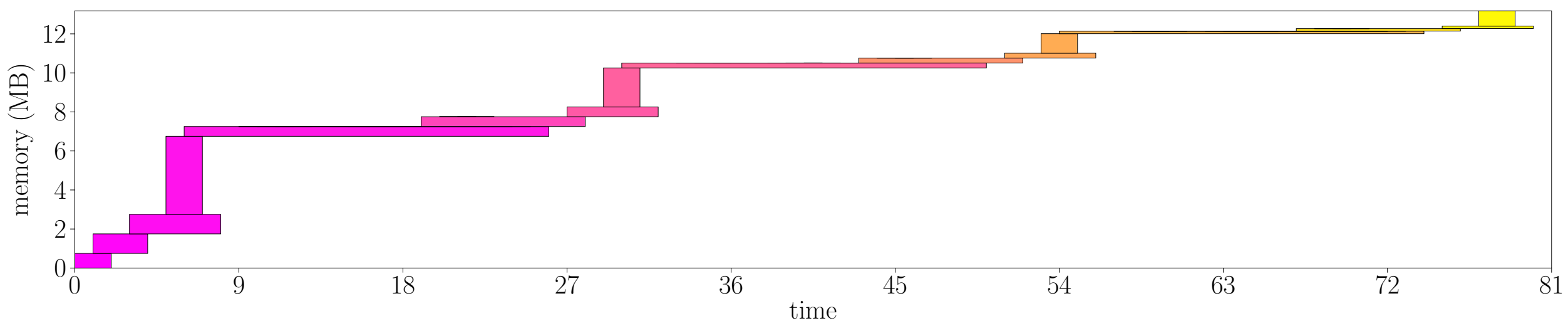

Figure 8.22: `bump_allocation`

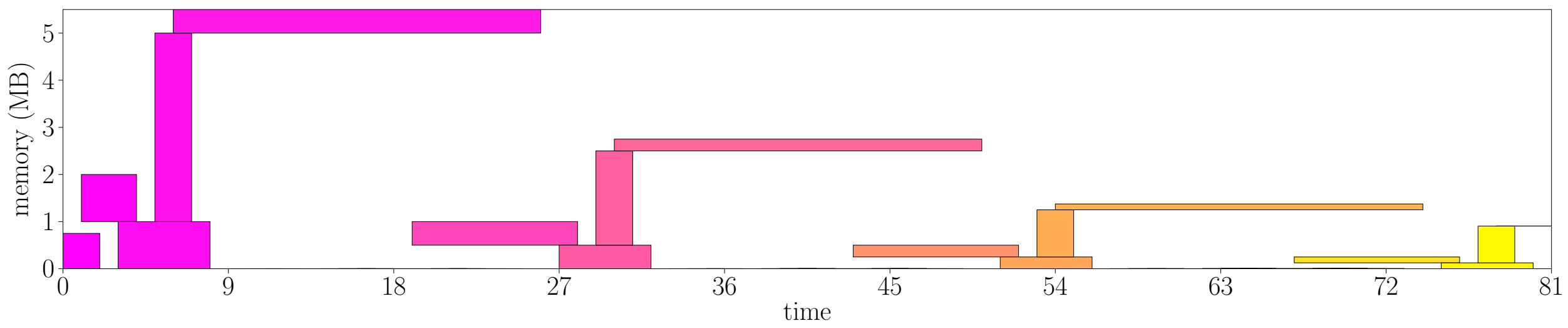

Figure 8.23: `gergov`

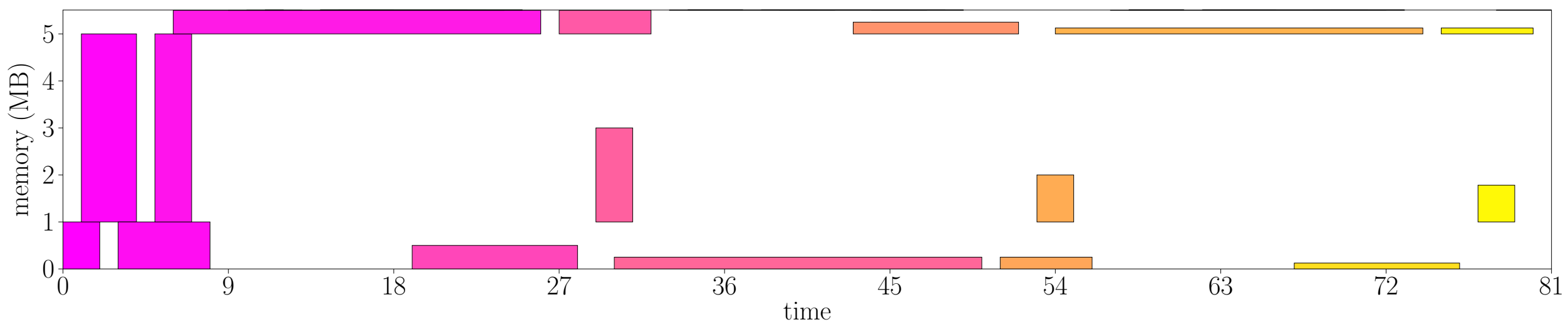

Figure 8.24: `mincost_flow`

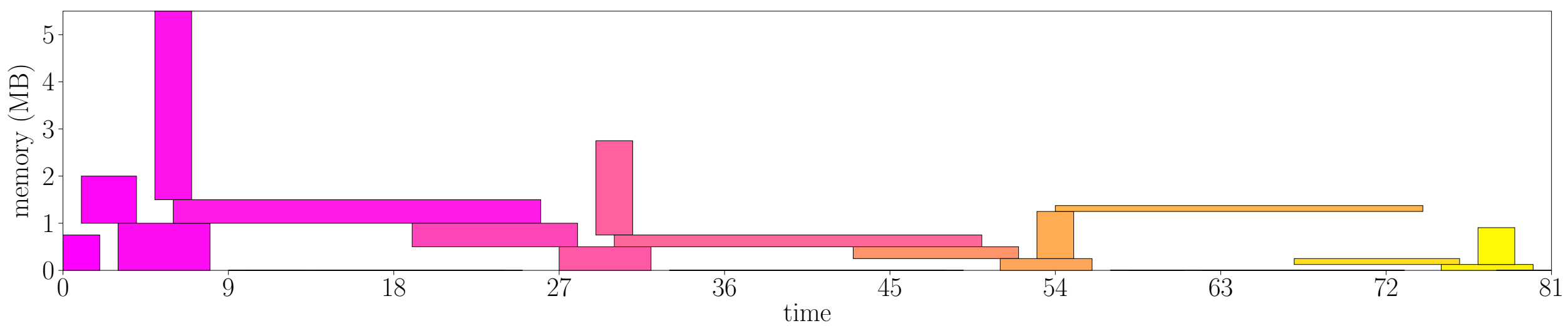

Figure 8.25: `mip`

## *deeplabv3_resnet50*

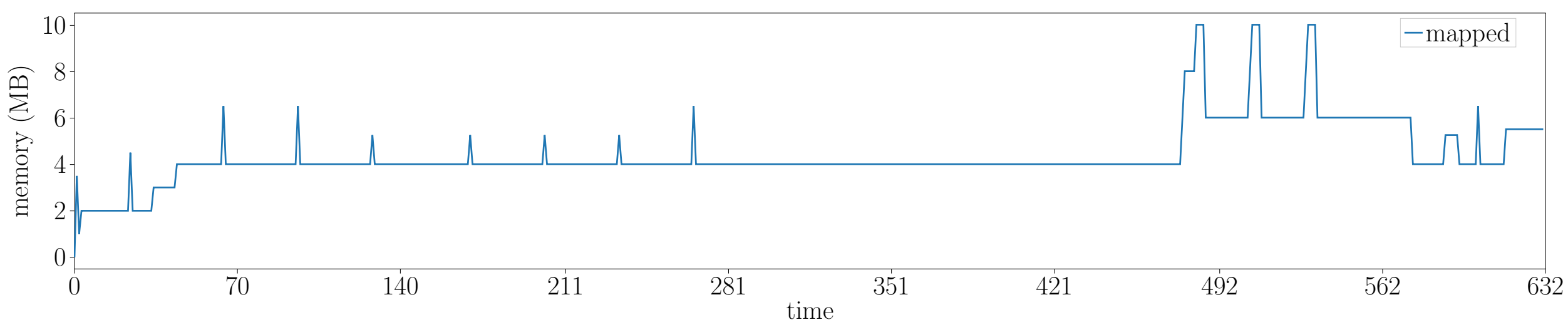

Figure 8.26: `jemalloc`

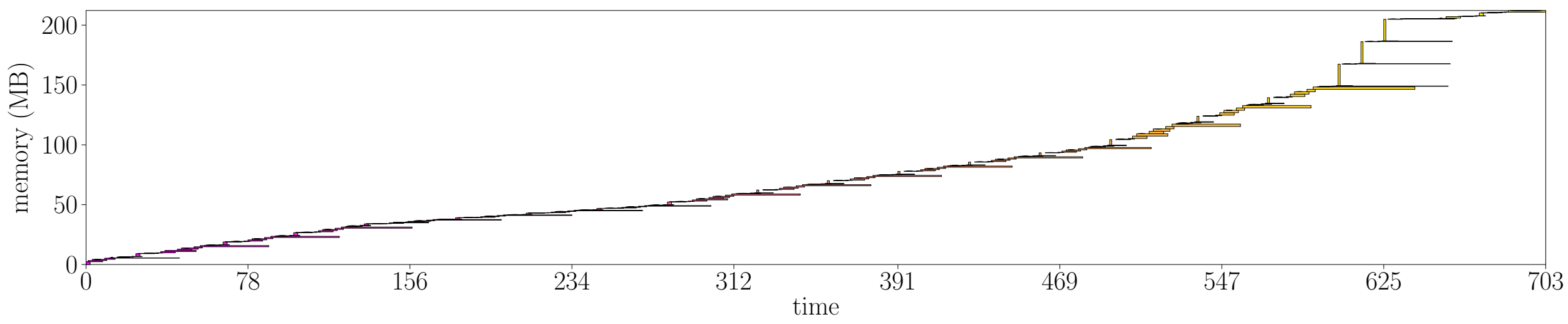

Figure 8.27: `bump_allocation`

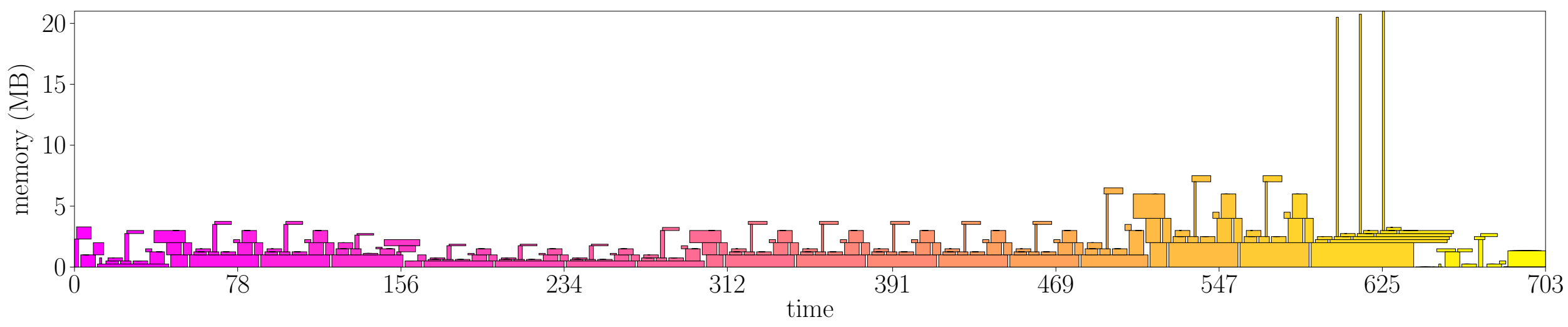

Figure 8.28: `gergov`

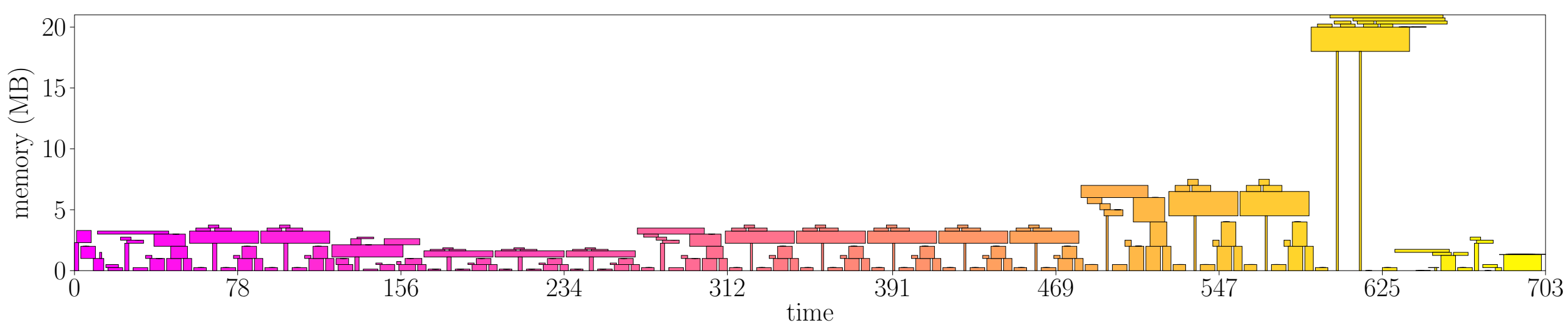

Figure 8.29: `greedy_by_size`

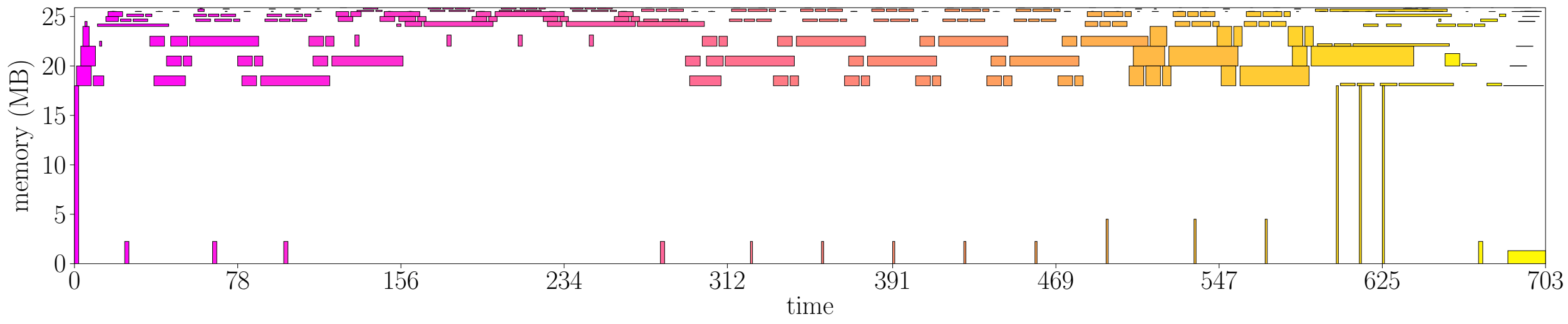

Figure 8.30: `mincost_flow`

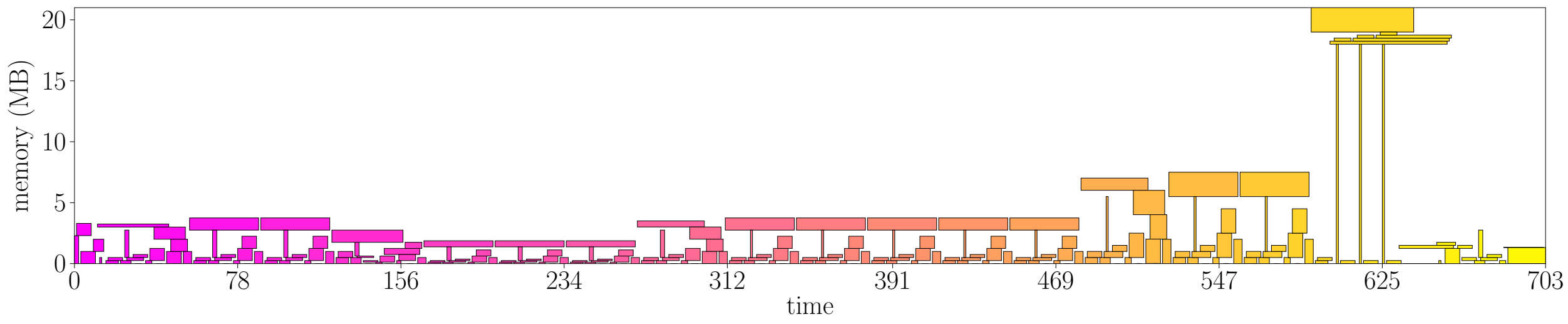

Figure 8.31: `mip`

## *densenet161*

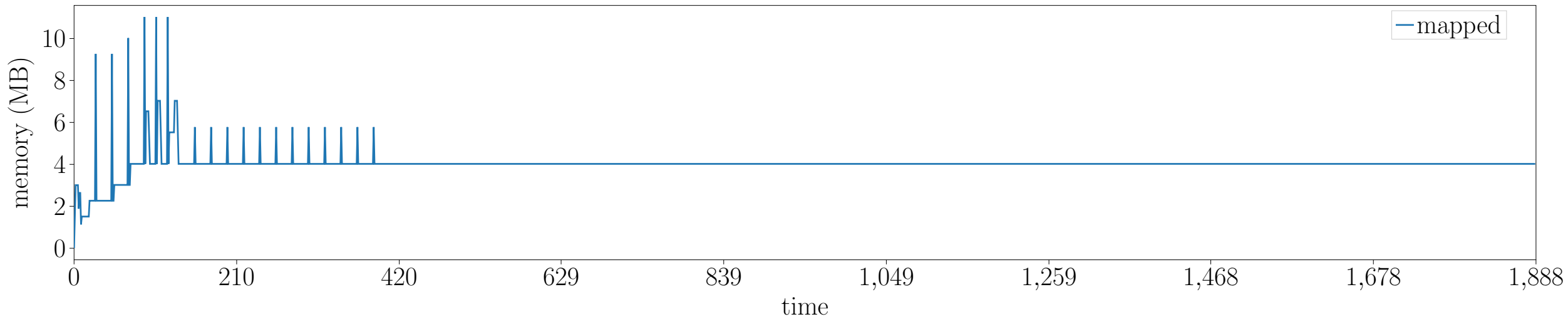

Figure 8.32: `jemalloc`

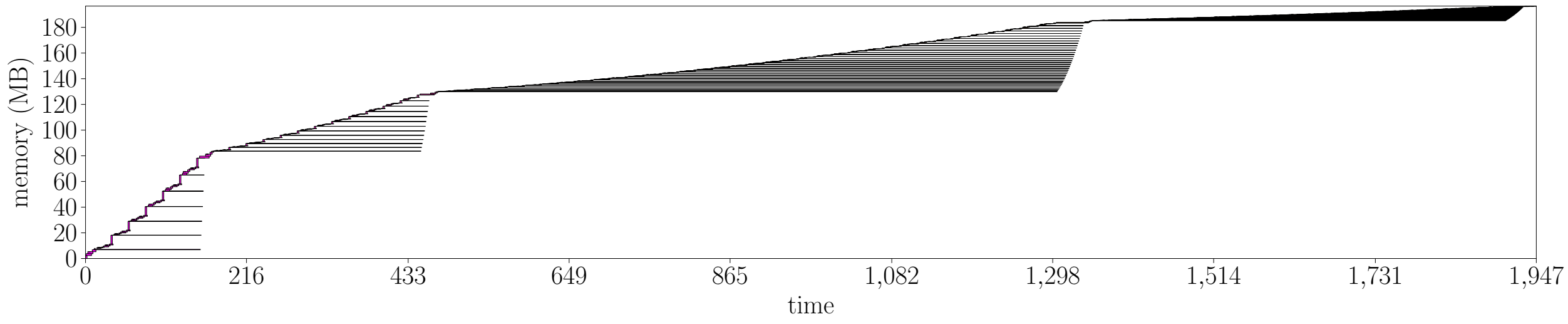

Figure 8.33: `bump_allocation`

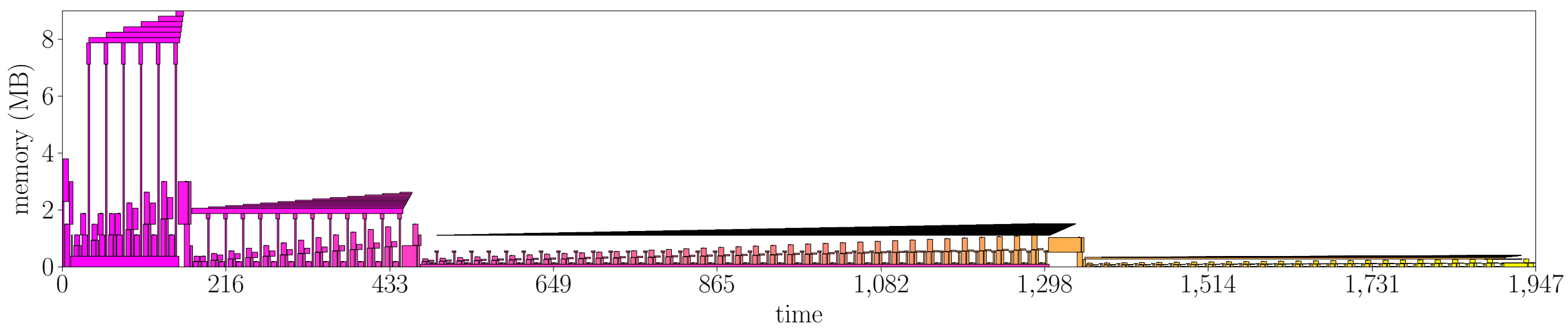

Figure 8.34: `gergov`

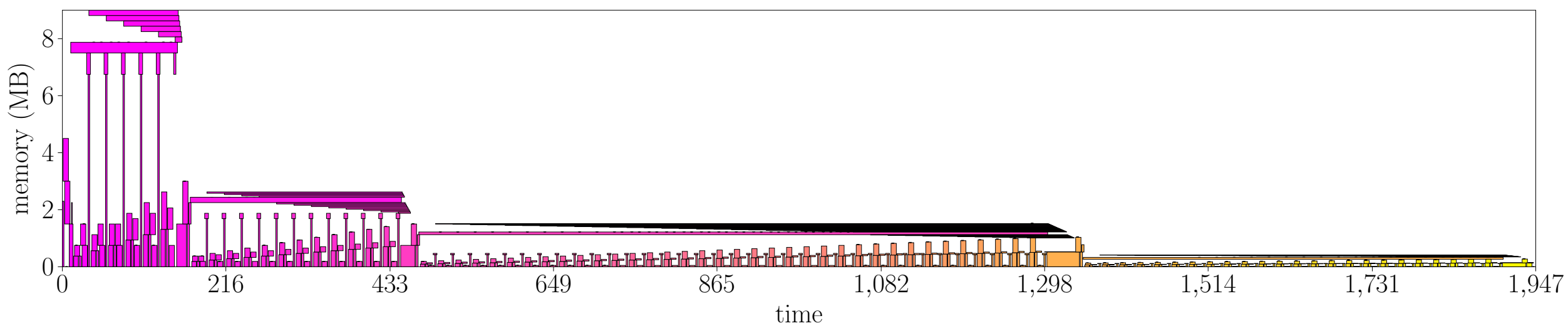

Figure 8.35: `greedy_by_size`

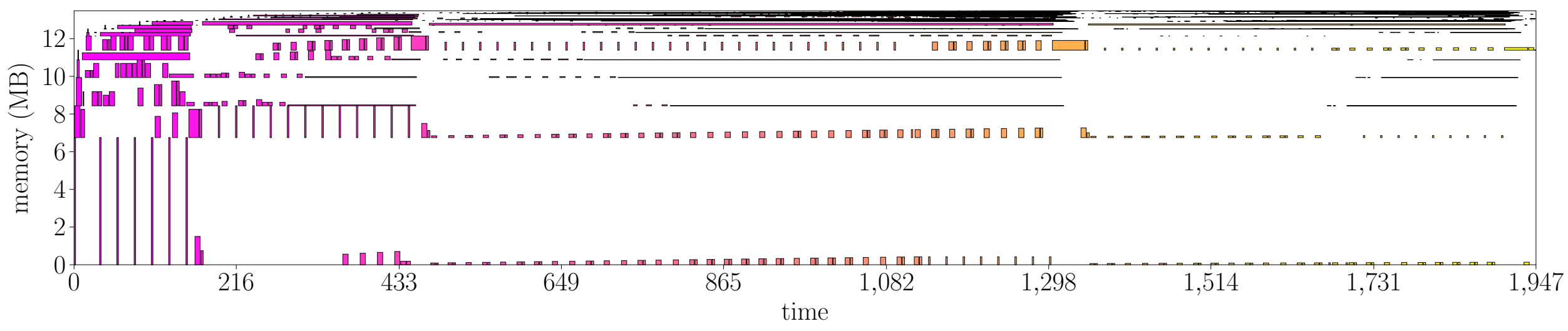

Figure 8.36: `mincost_flow`

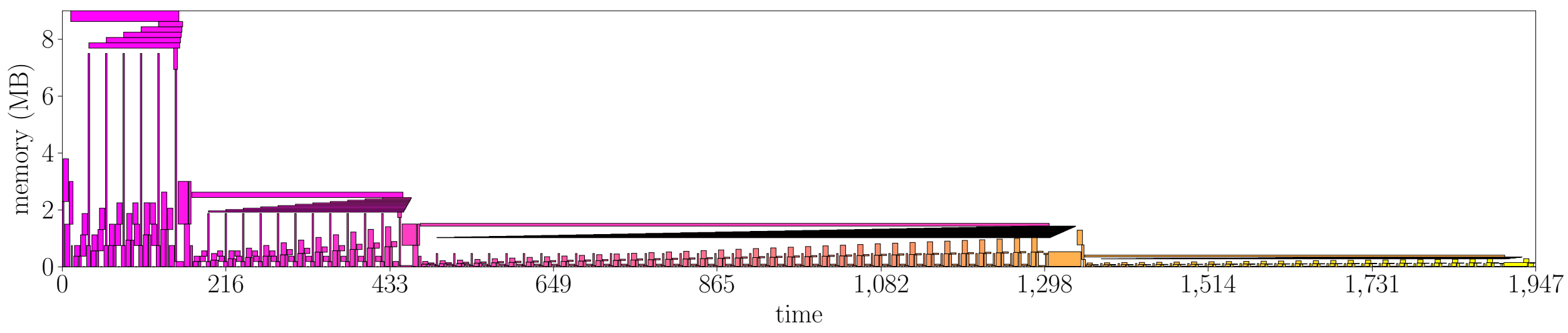

Figure 8.37: `mip`

# *fcn_resnet50*

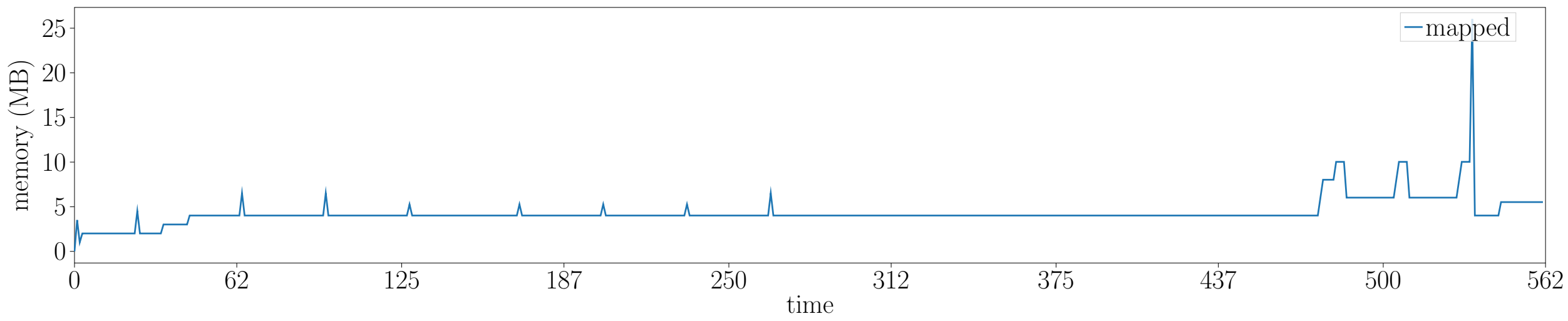

Figure 8.38: `jemalloc`

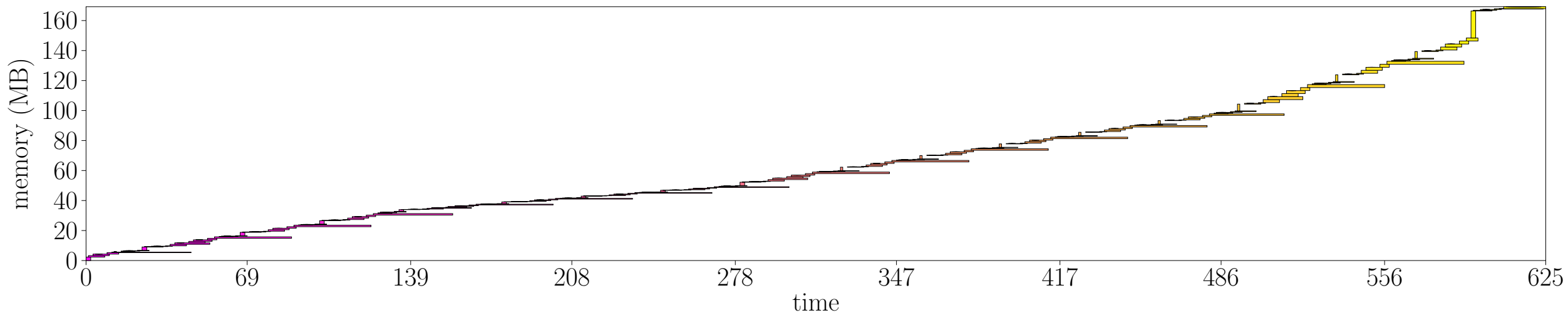

Figure 8.39: `bump_allocation`

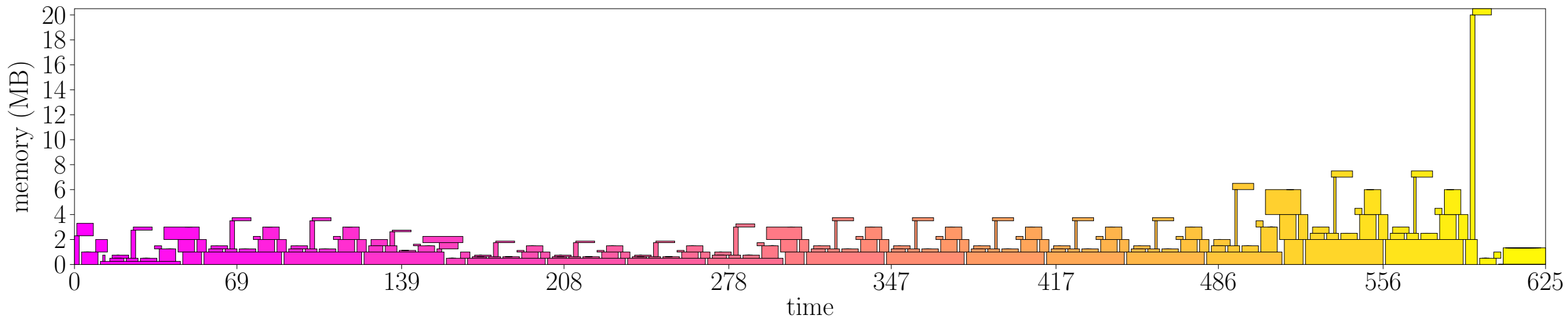

Figure 8.40: `gergov`

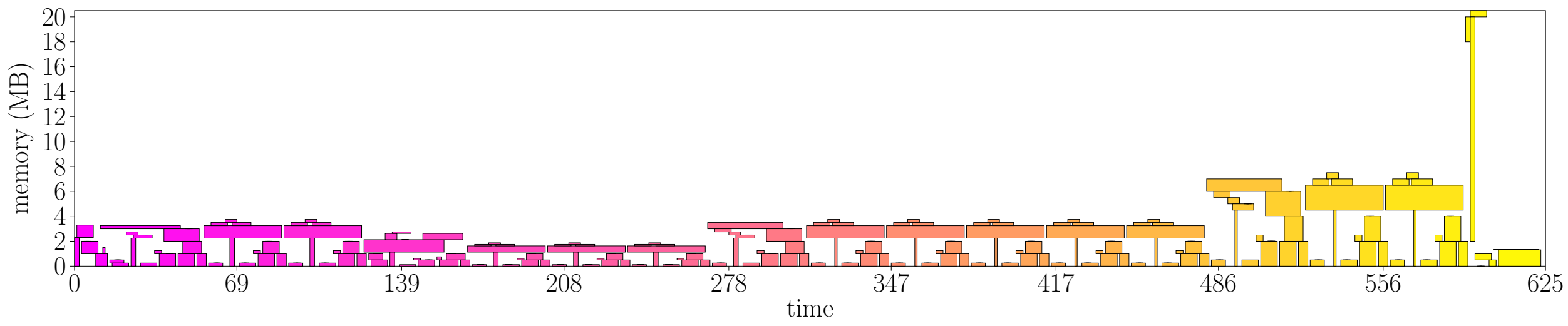

Figure 8.41: `greedy_by_size`

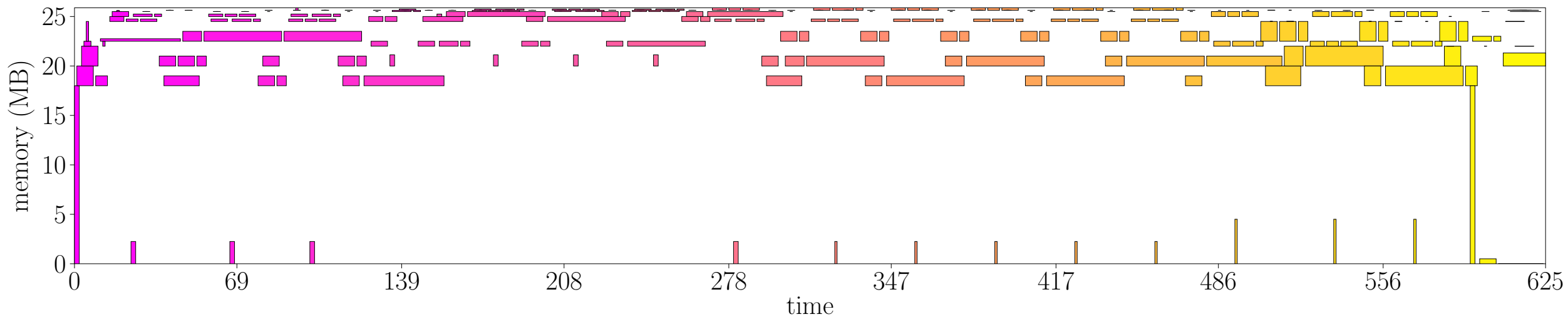

Figure 8.42: mincost_flow

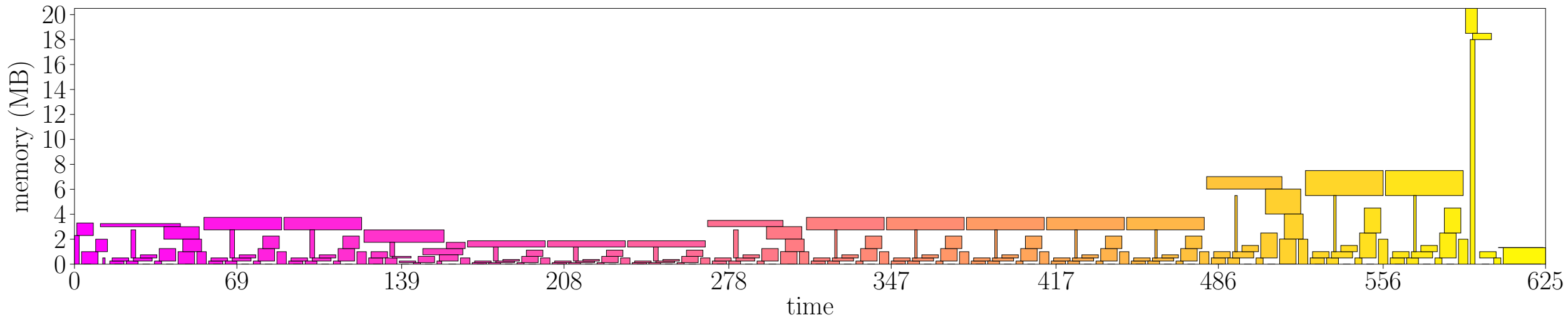

Figure 8.43: mip

## *googlenet*

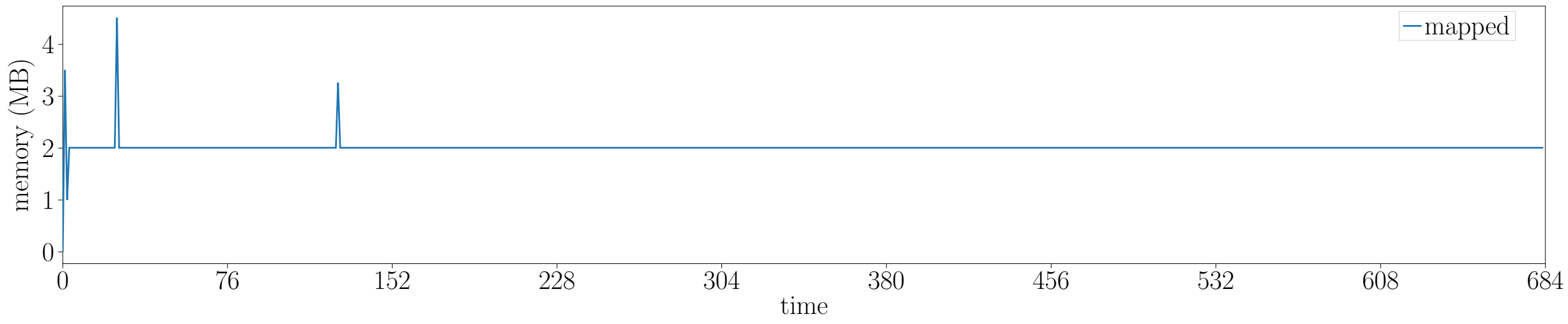

Figure 8.44: jemalloc

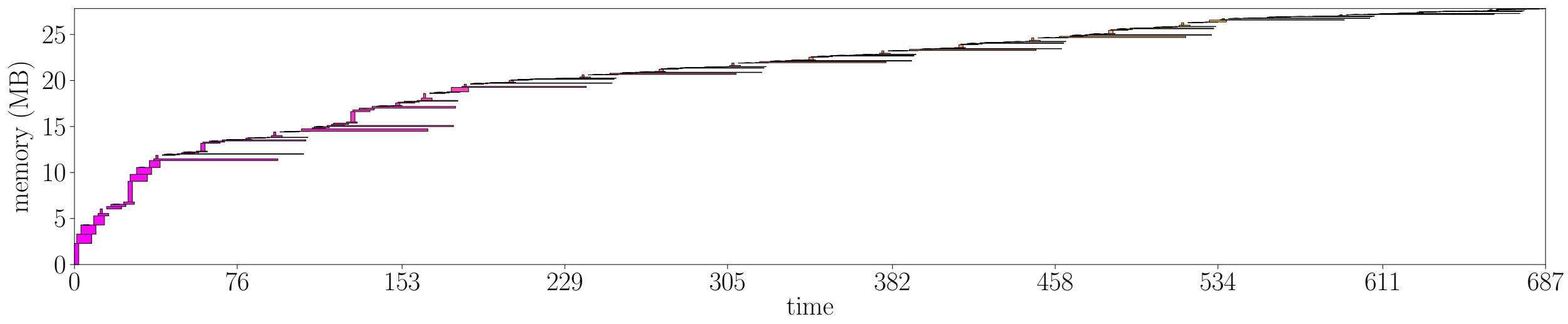

Figure 8.45: bump_allocation

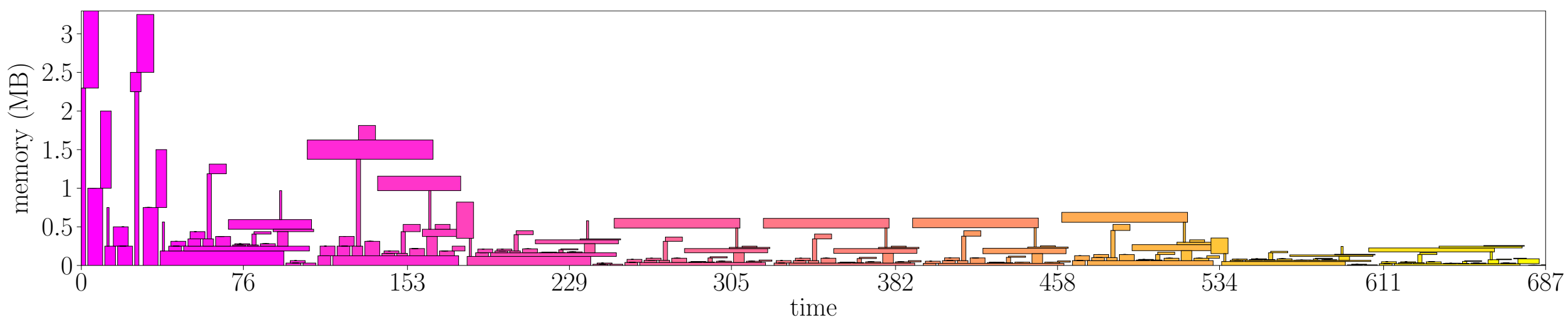

Figure 8.46: `gergov`

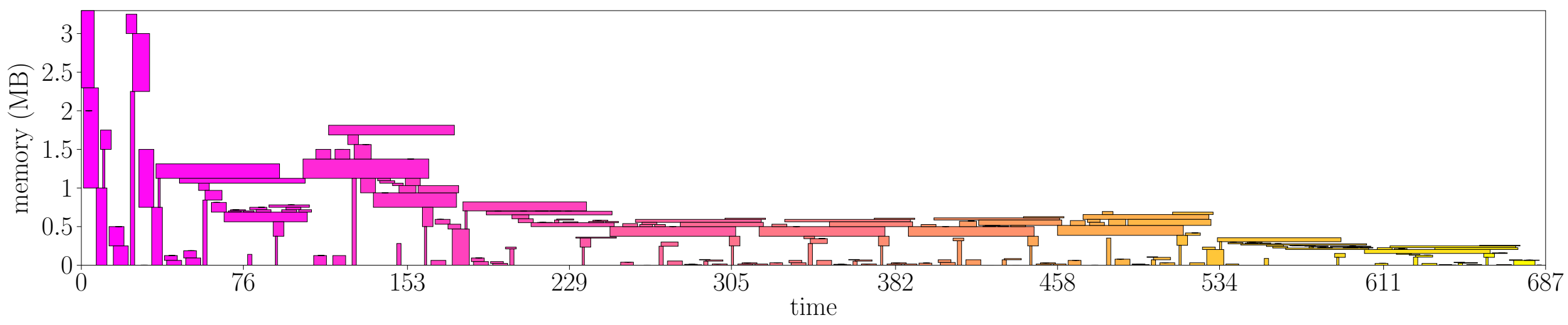

Figure 8.47: `greedy_by_size`

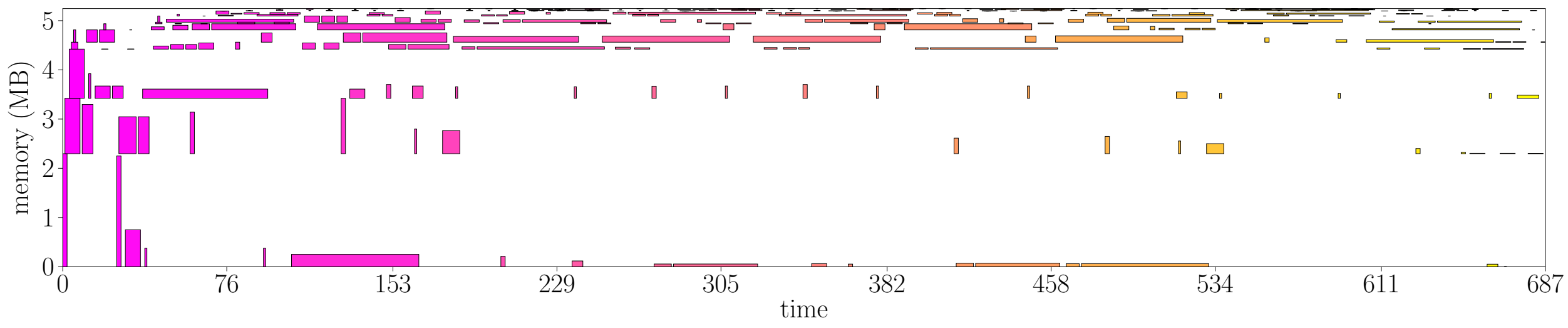

Figure 8.48: `mincost_flow`

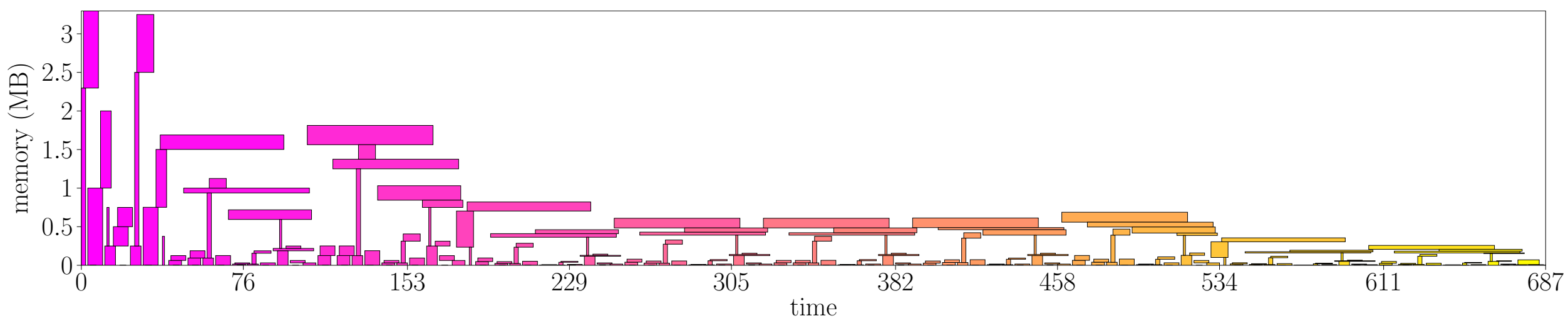

Figure 8.49: `mip`

## *regnet_x_8gf*

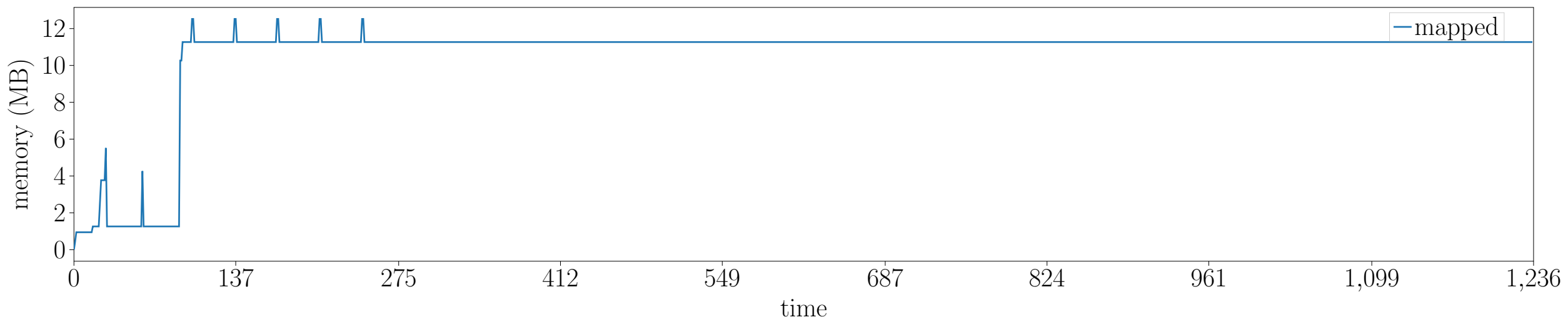

Figure 8.50: `jemalloc`

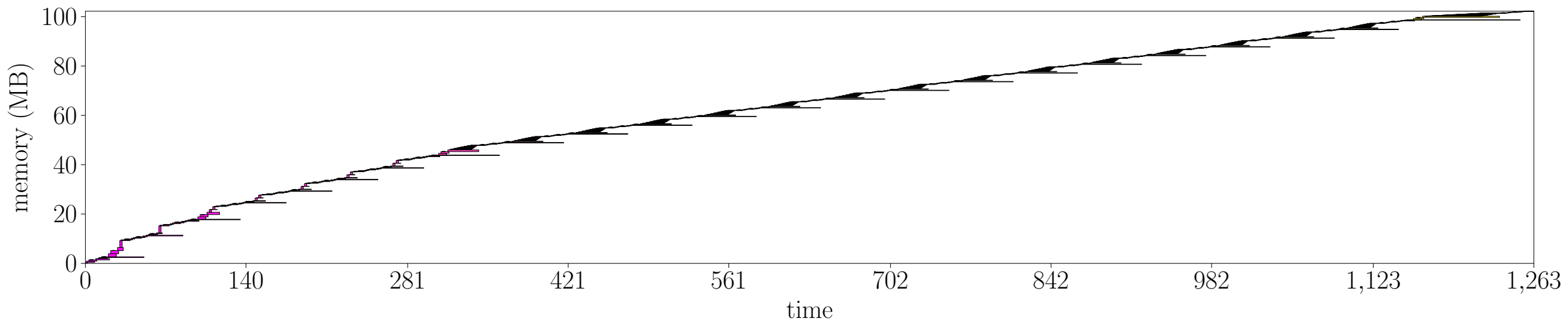

Figure 8.51: `bump_allocation`

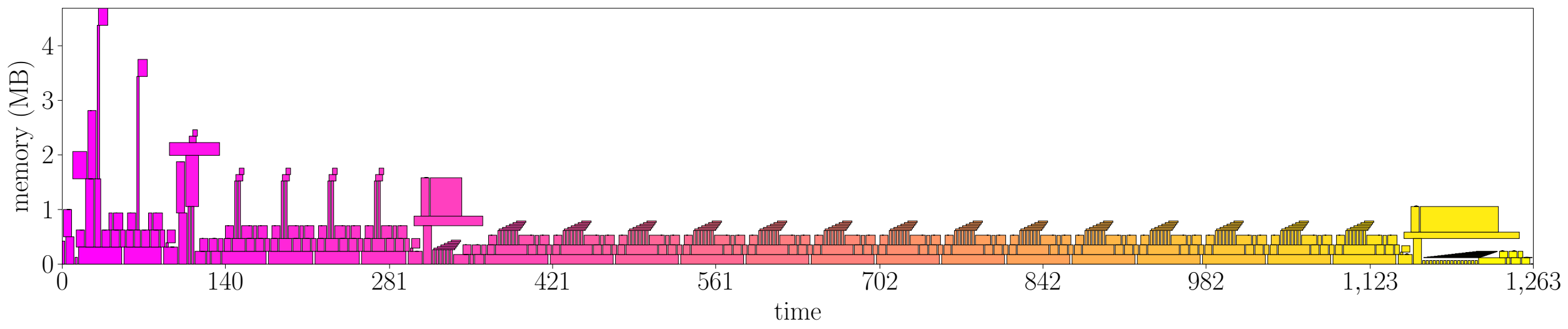

Figure 8.52: `gergov`

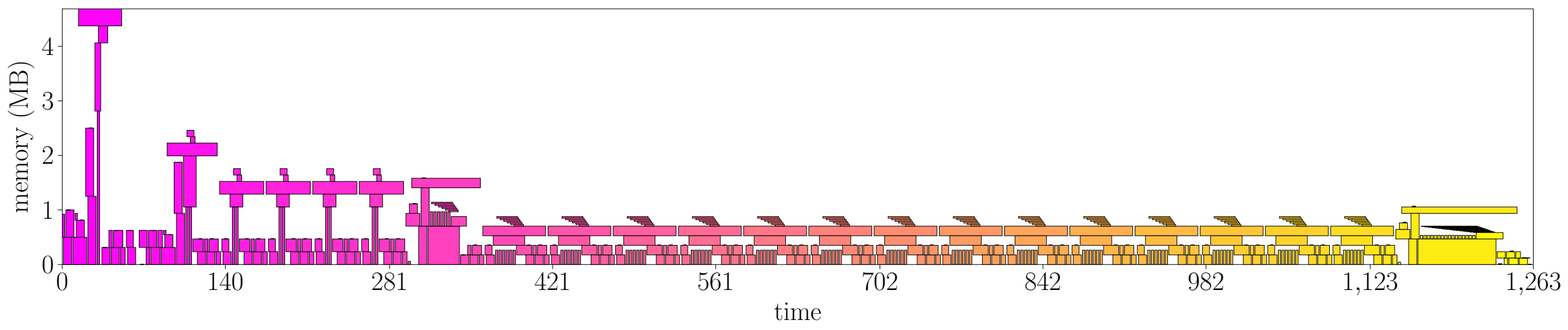

Figure 8.53: `greedy_by_size`

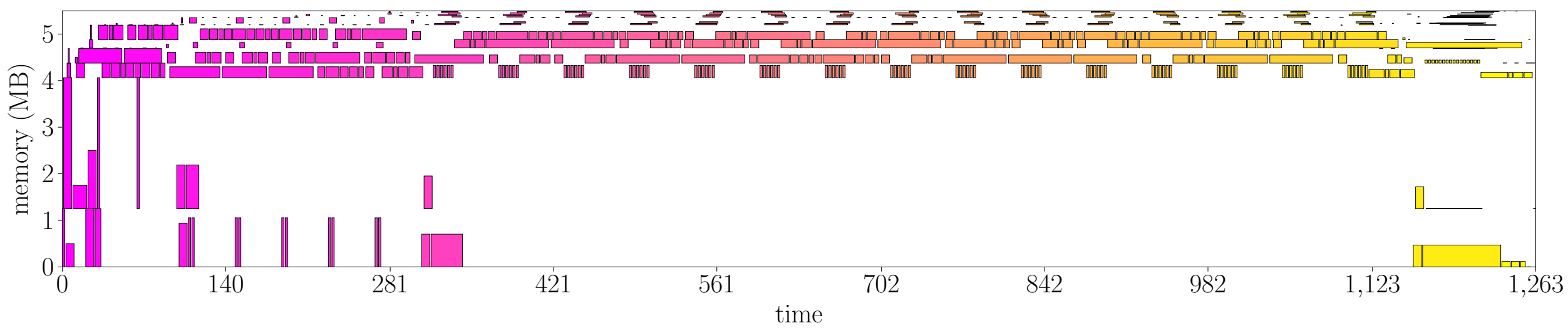

Figure 8.54: `mincost_flow`

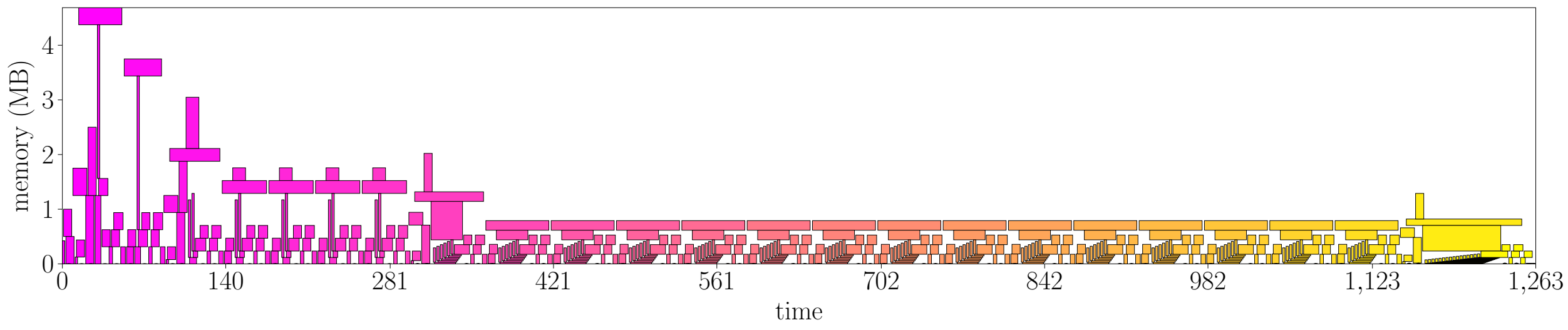

Figure 8.55: `mip`

## *squeezenet1_0*

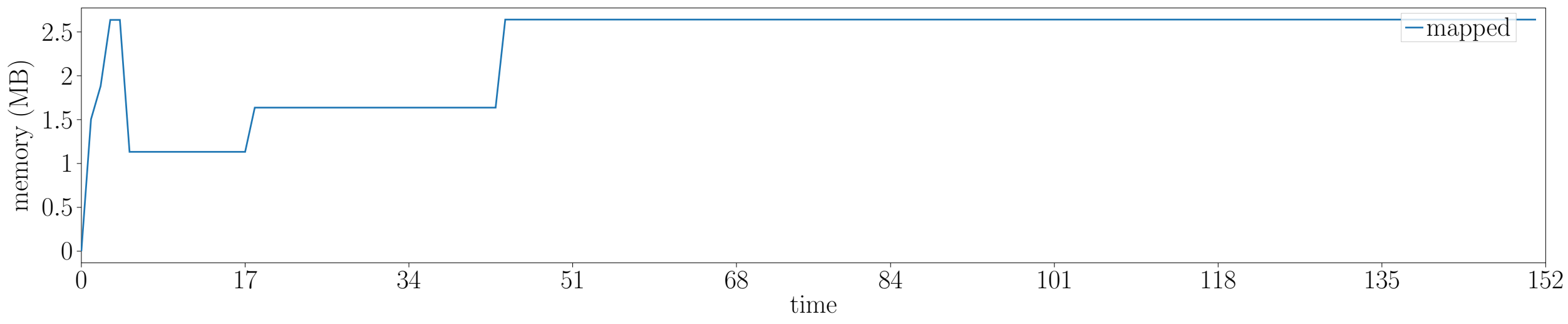

Figure 8.56: `jemalloc`

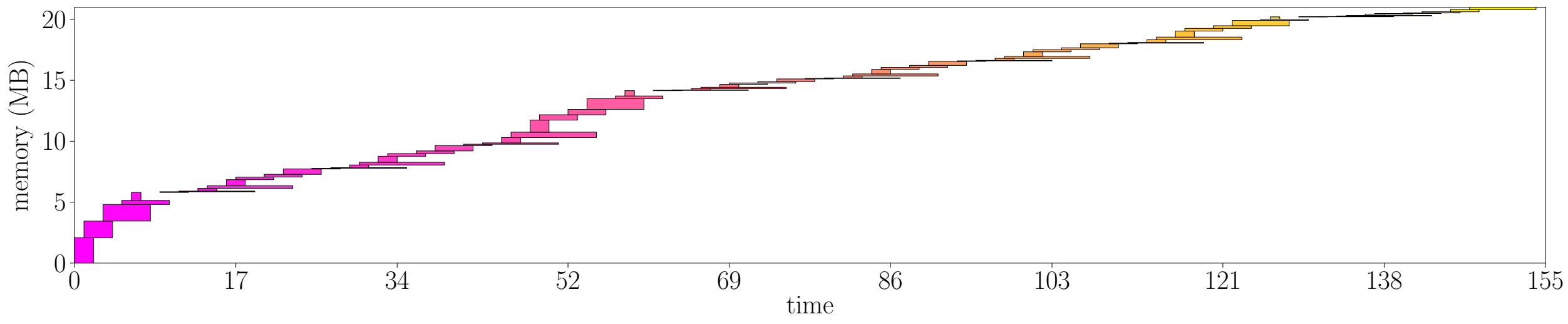

Figure 8.57: `bump_allocation`

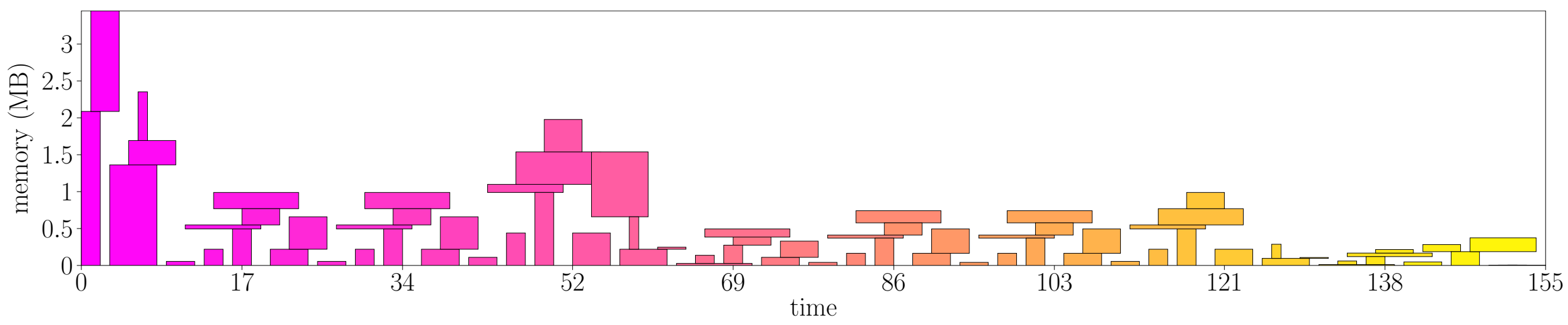

Figure 8.58: `gergov`

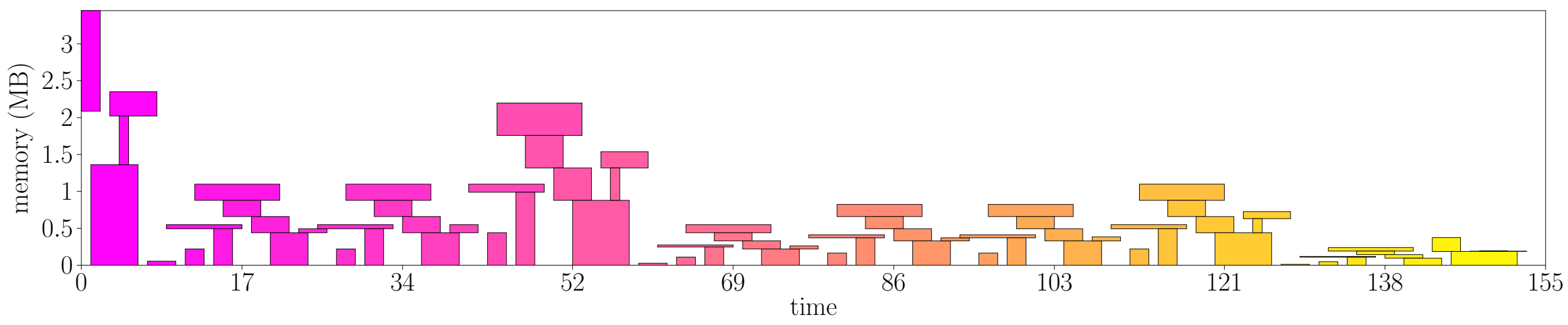

Figure 8.59: `greedy_by_size`

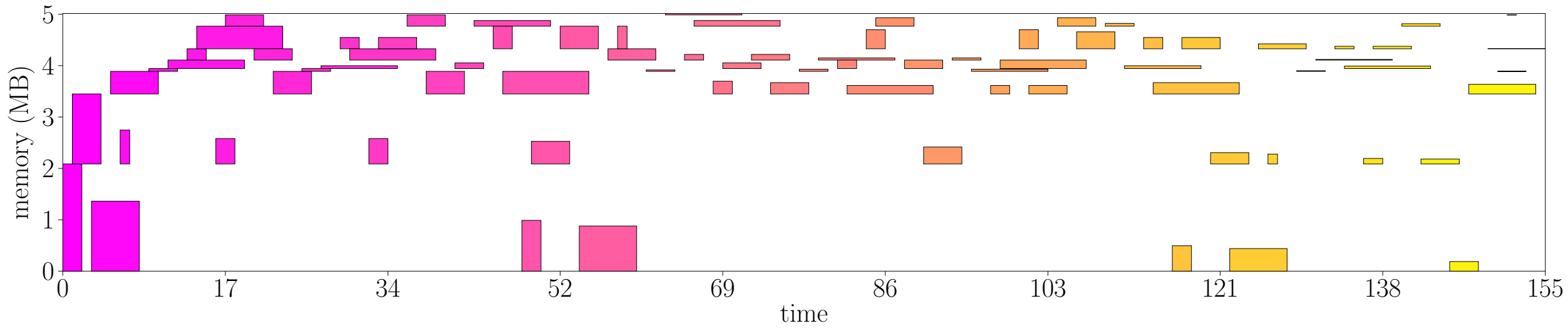

Figure 8.60: `mincost_flow`

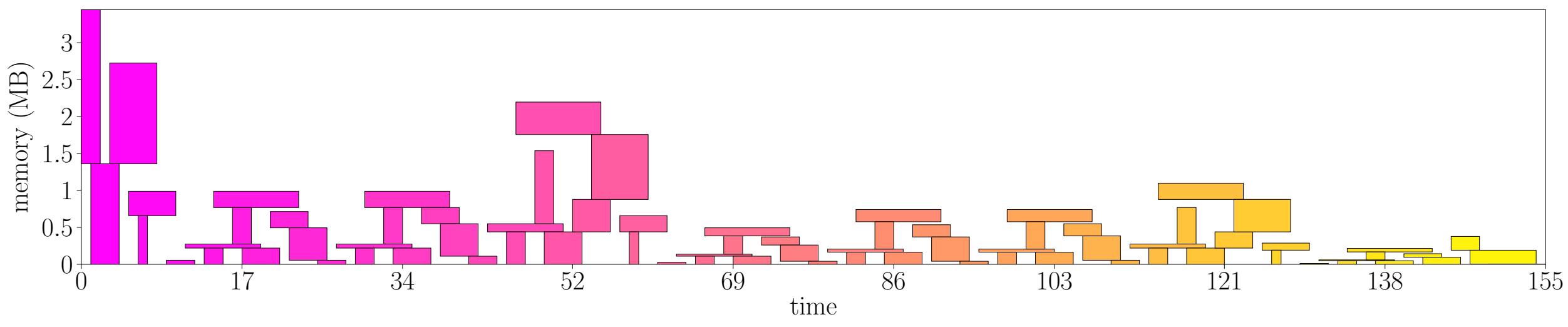

Figure 8.61: `mip`

*vgg16*

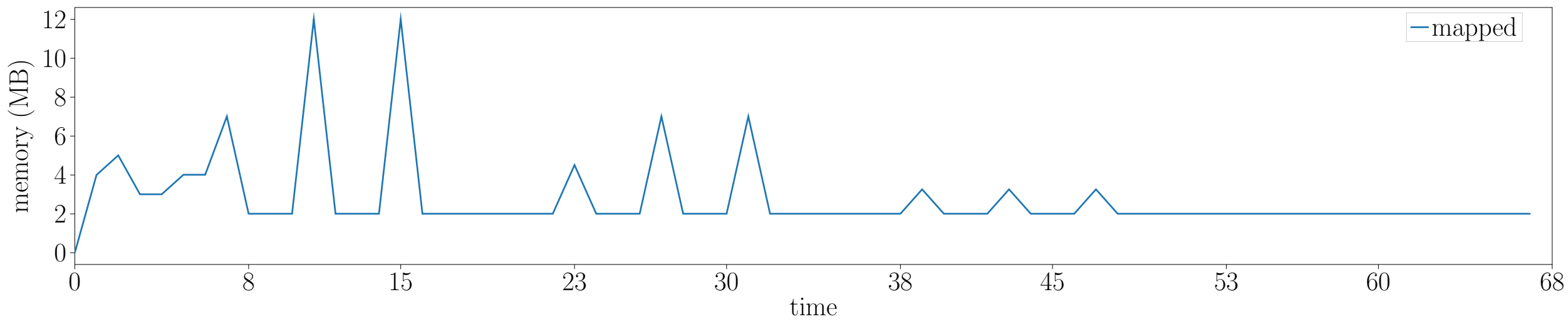

Figure 8.62: `jemalloc`

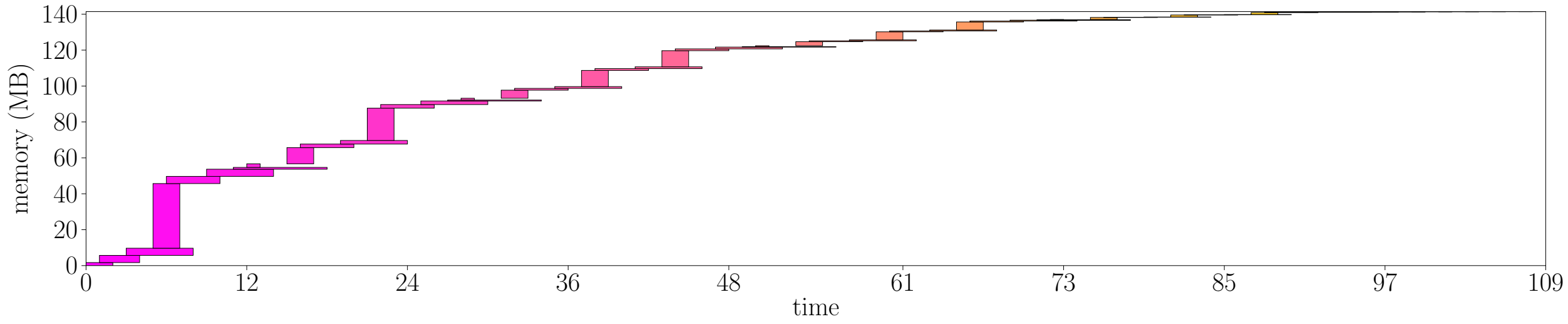

Figure 8.63: `bump_allocation`

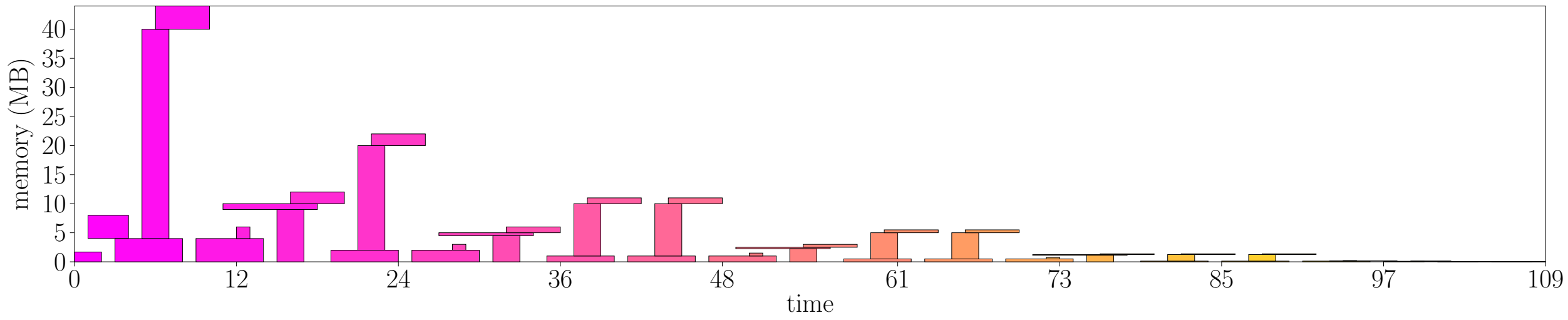

Figure 8.64: `gergov`

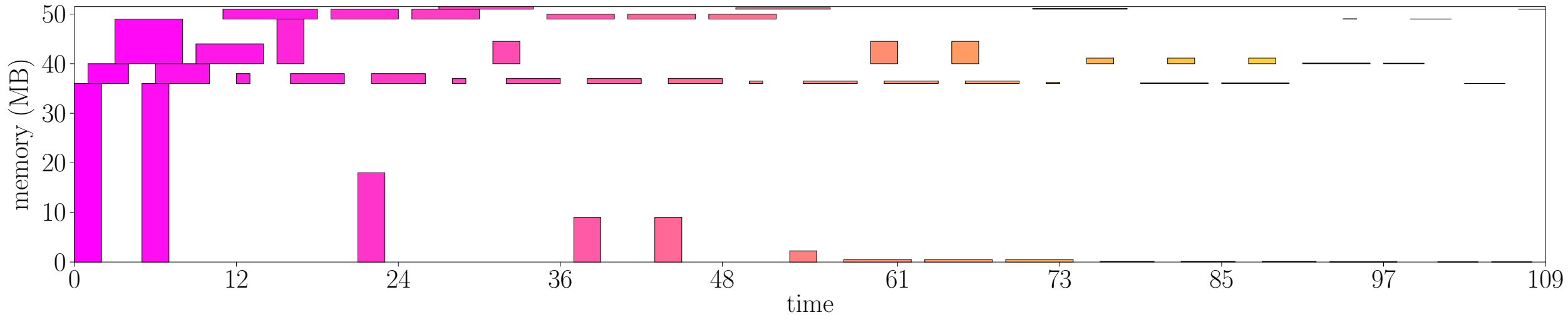

Figure 8.65: `mincost_flow`

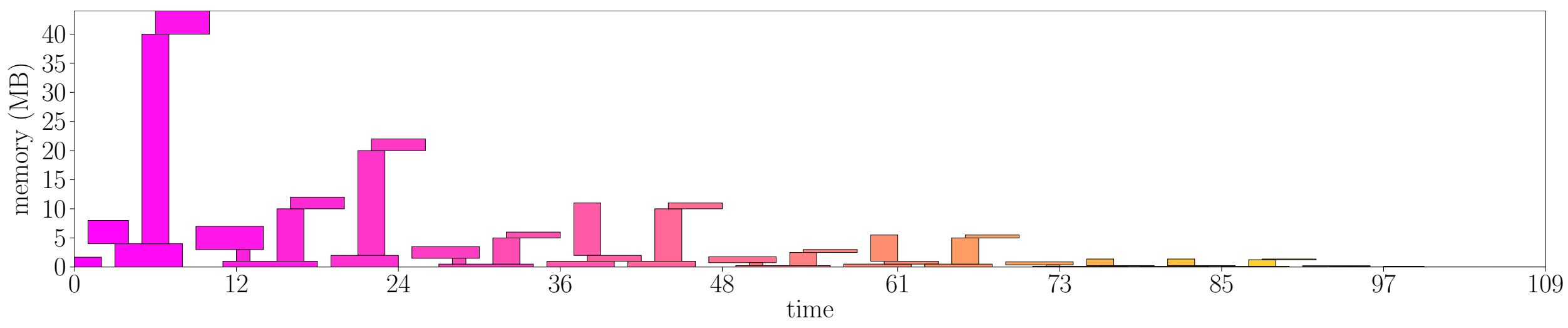

Figure 8.66: `mip`

## *wide_resnet50_2*

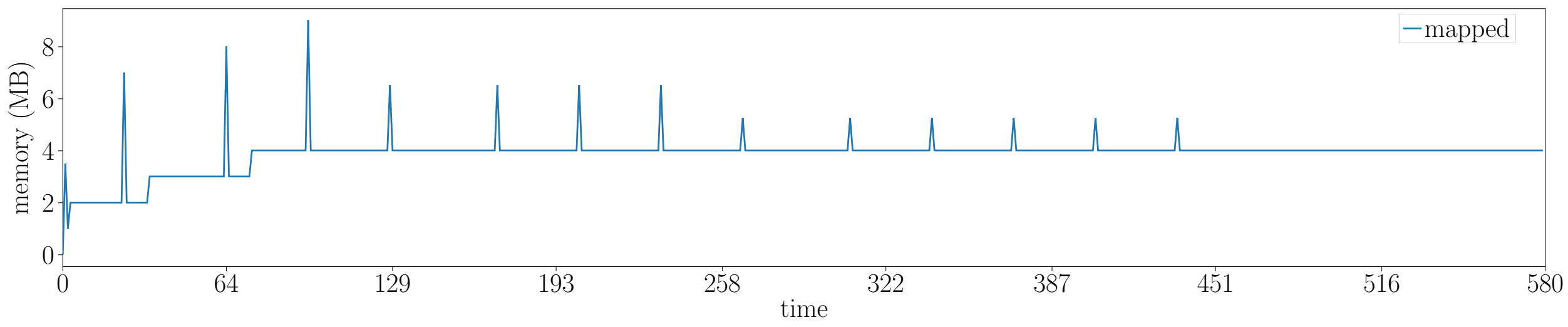

Figure 8.67: `jemalloc`

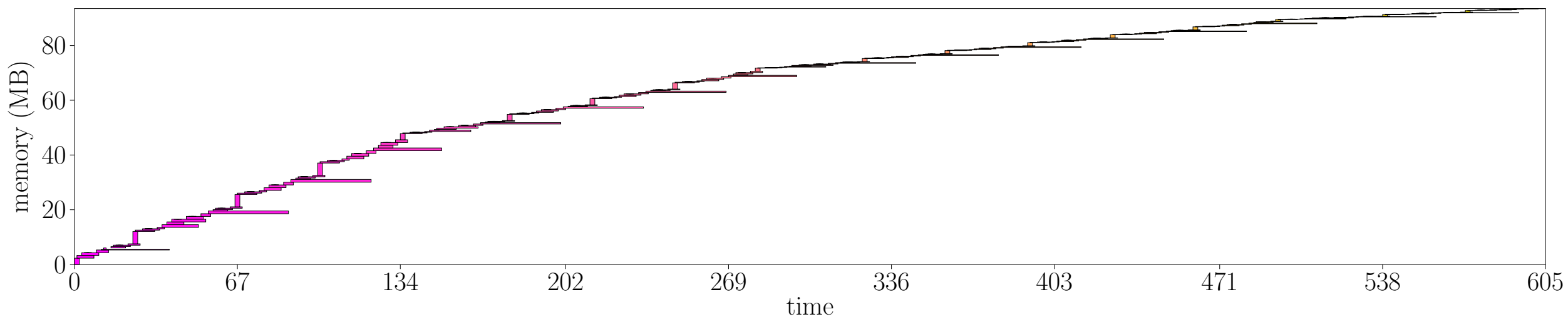

Figure 8.68: `bump_allocation`

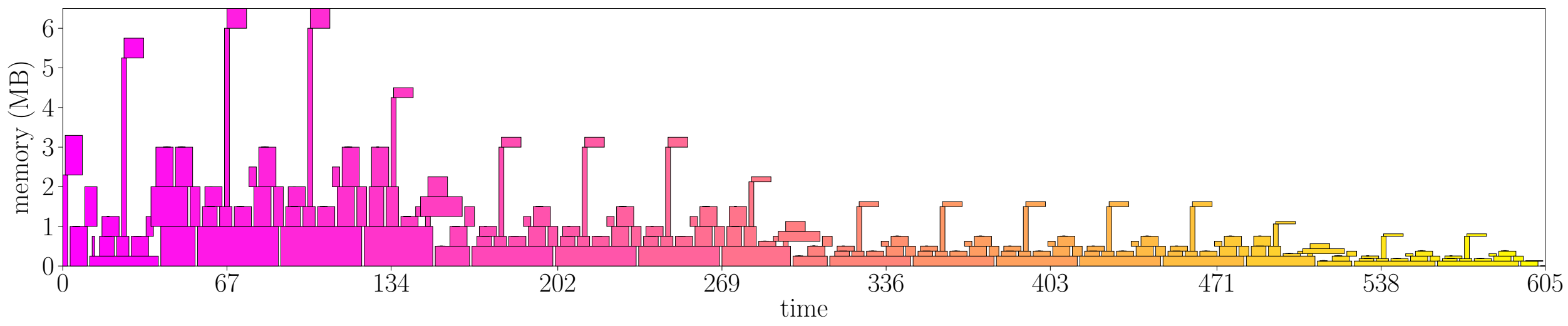

Figure 8.69: `gergov`

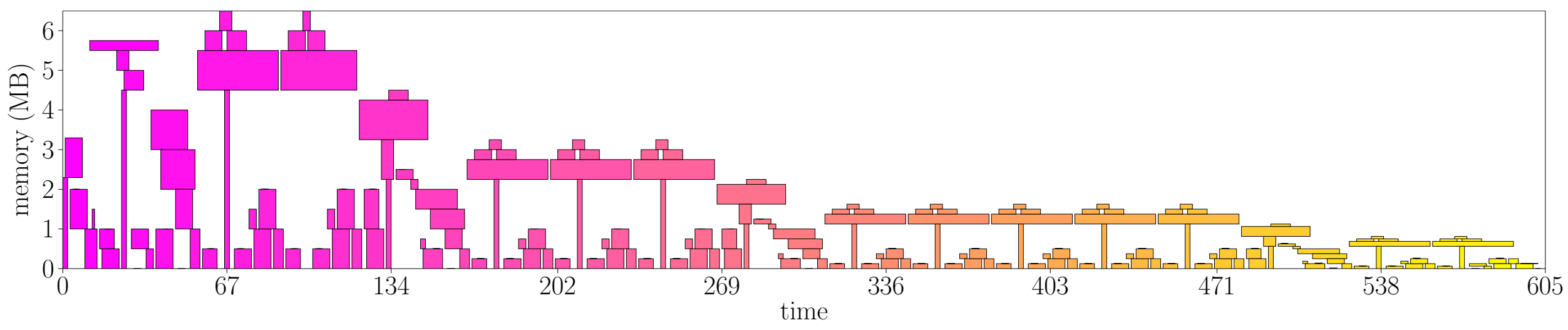

Figure 8.70: `greedy_by_size`

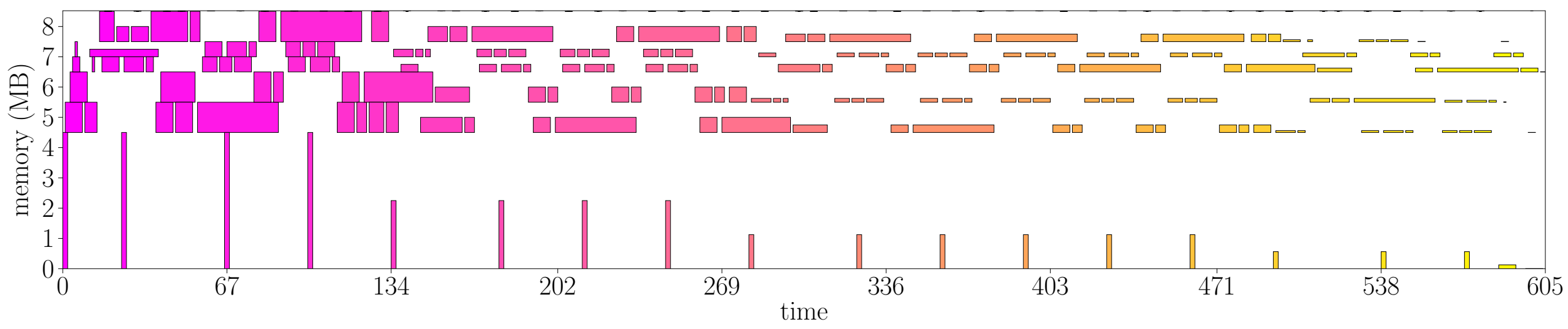

Figure 8.71: `mincost_flow`

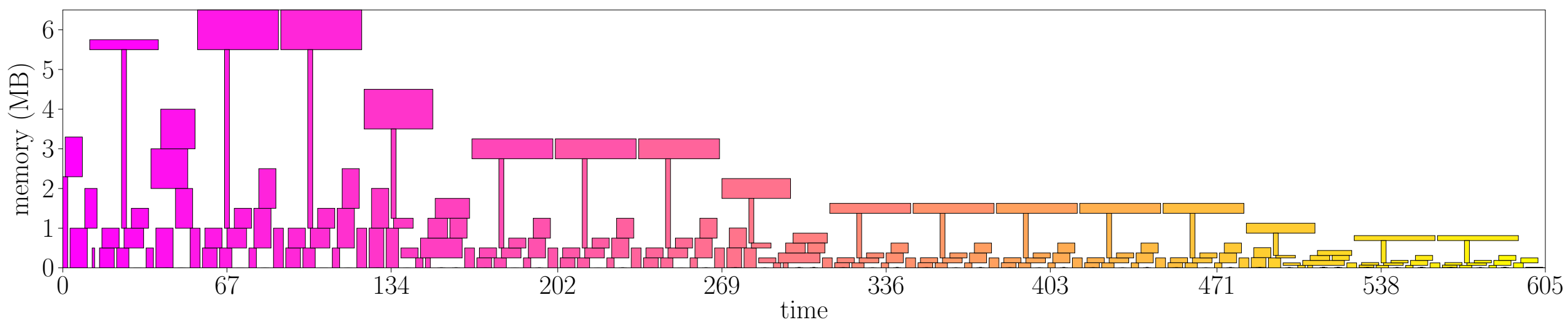

Figure 8.72: `mip`

## 8.3 Allocation Distributions for DNNs

We present allocation distributions for `batch_size` = 1 at `input_size` = 64, 128, 256, 512. Note that the $y$-axis on the following plots is symlog scaled (i.e., bars representing allocations that number less than 10 grow *down*).

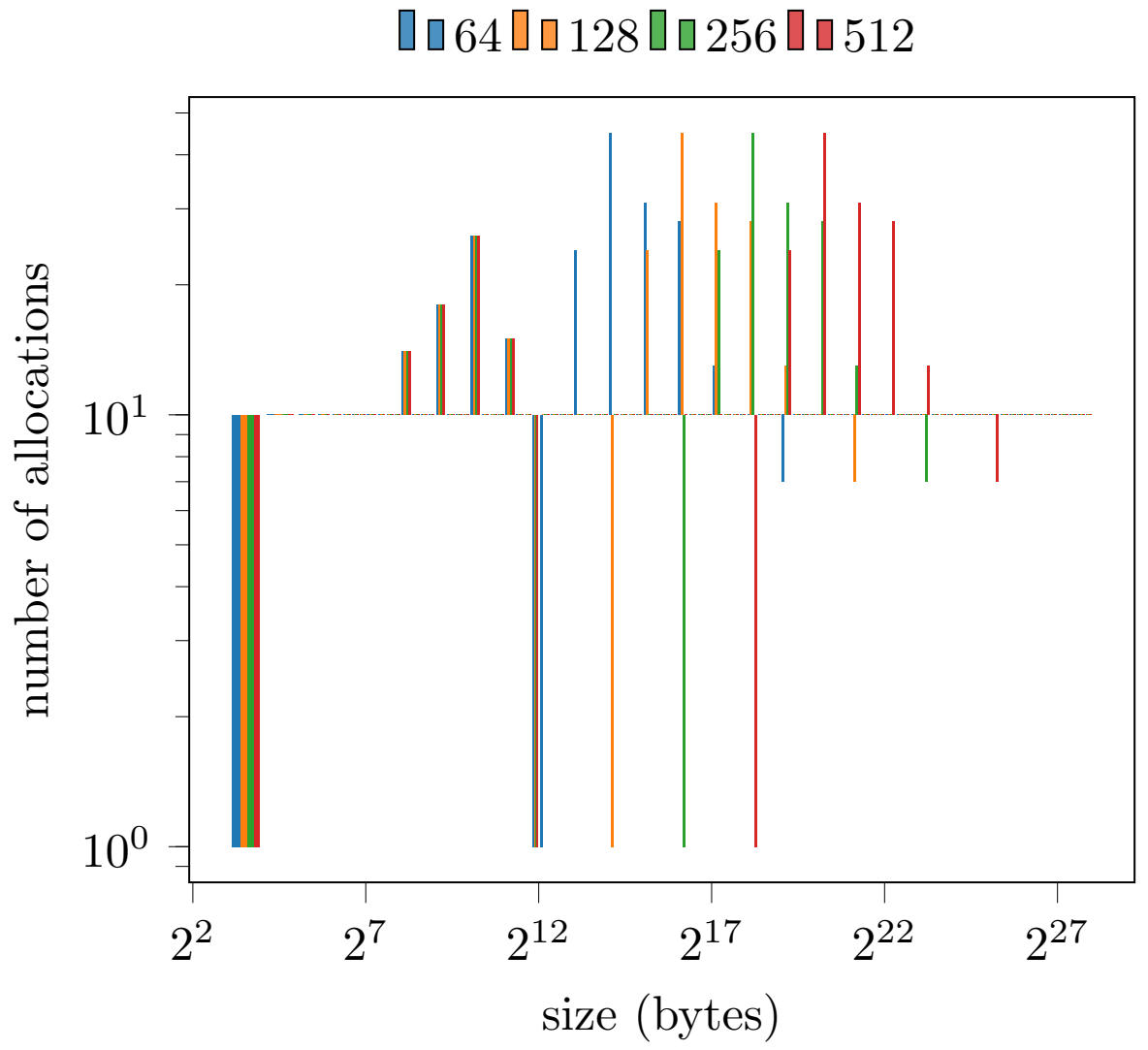

Allocation distributions for `resnet34`

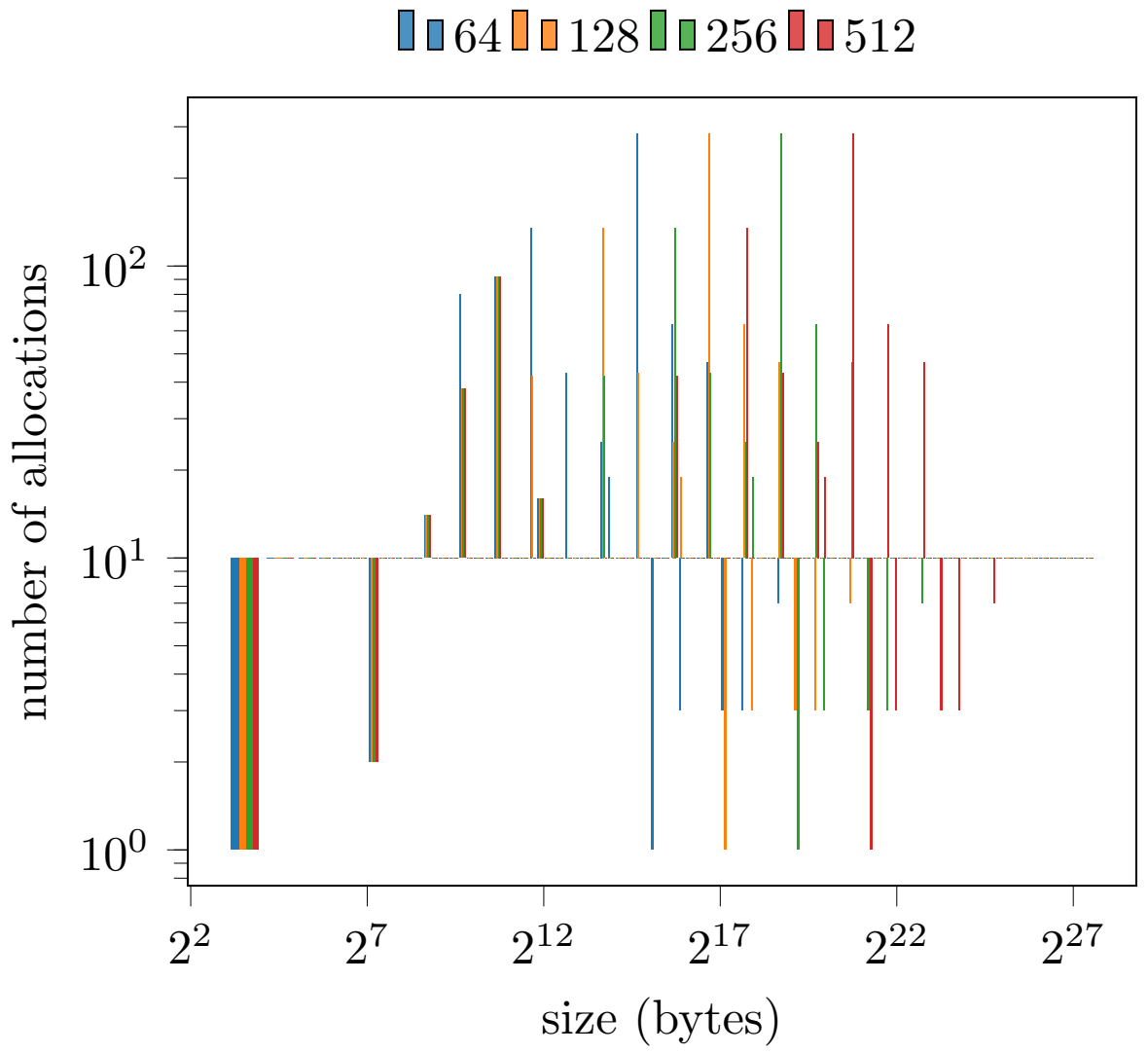

Allocation distributions for `regnet_x_3_2gf`

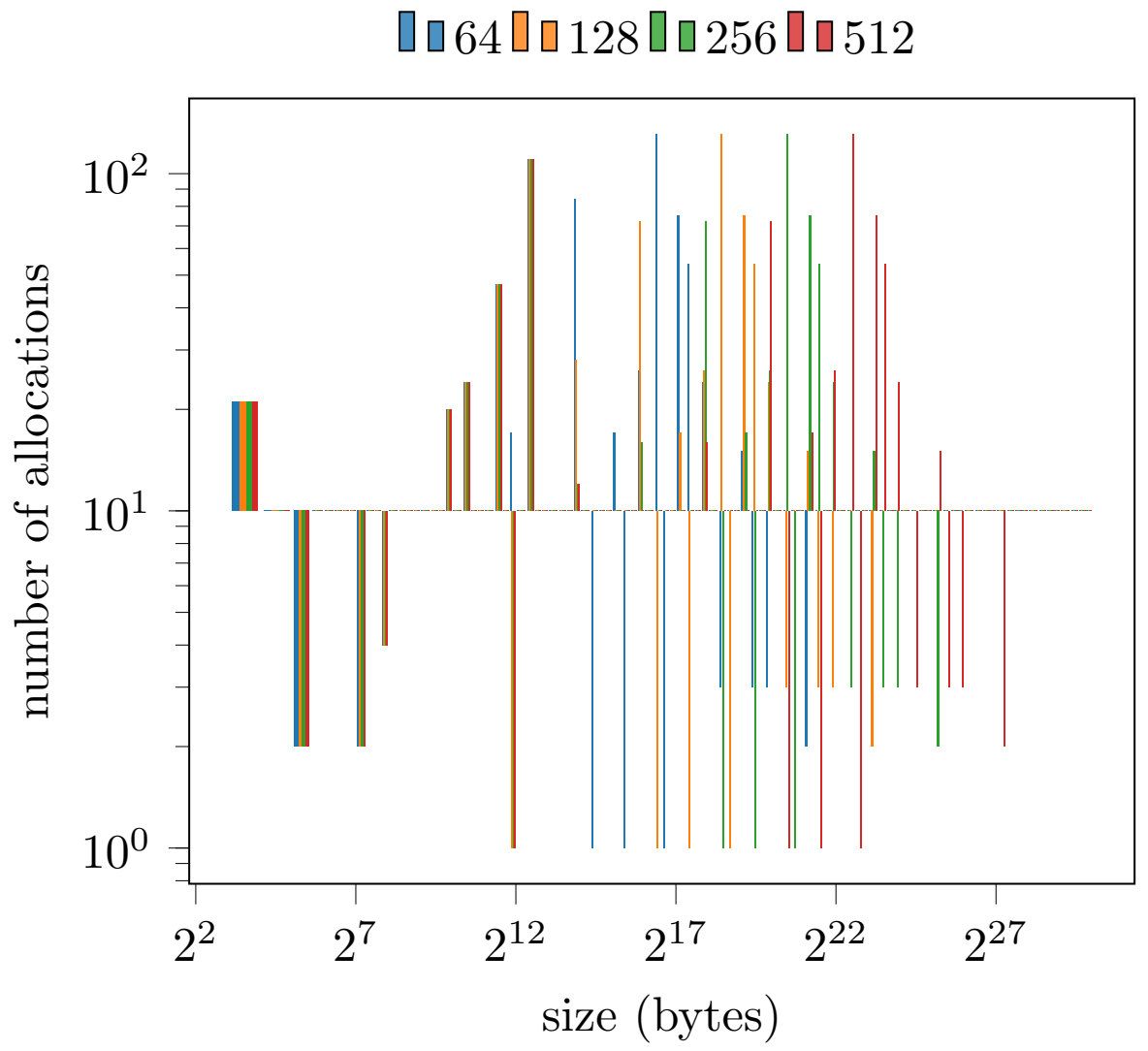

Allocation distributions for `regnet_y_32gf`

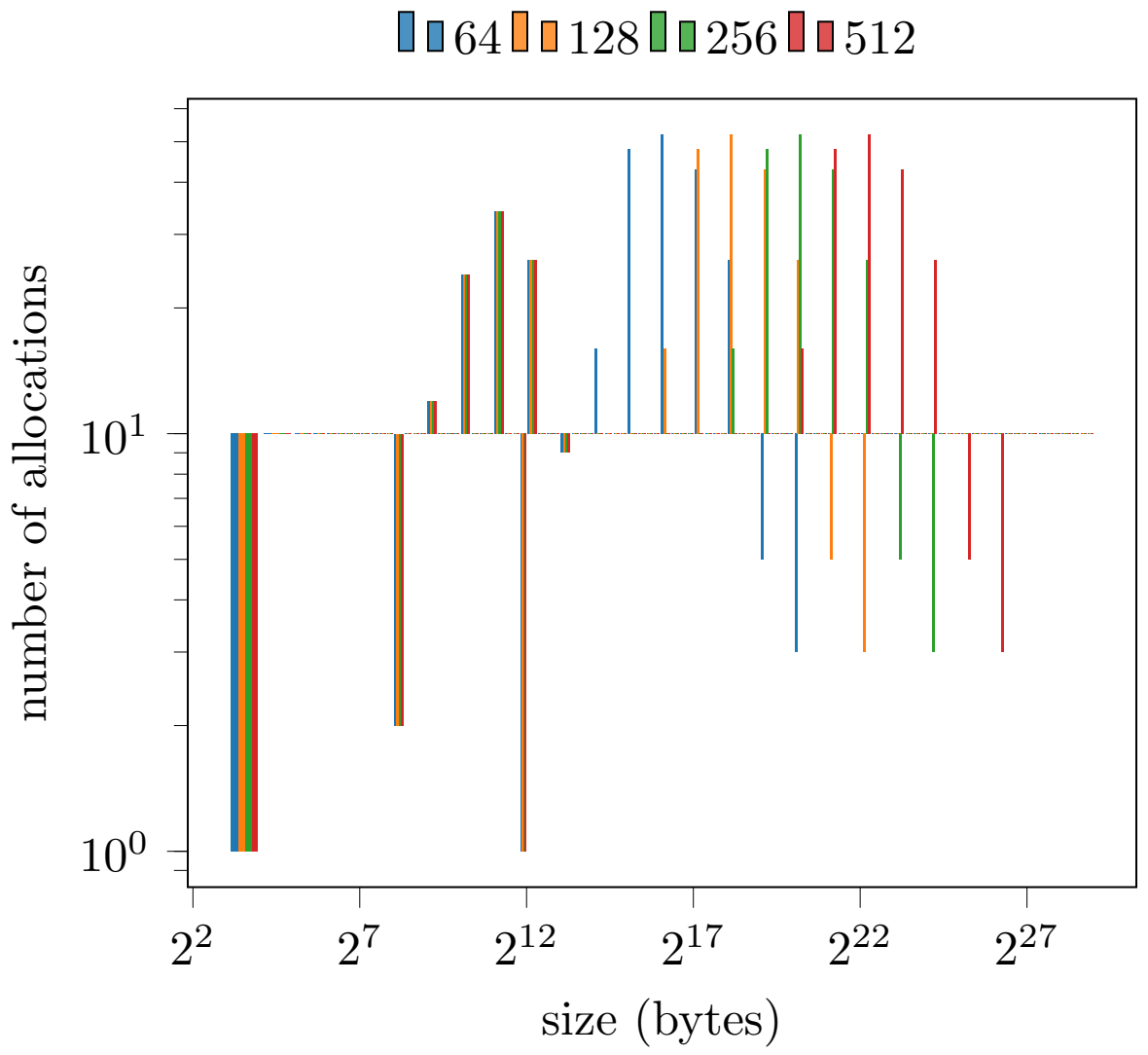

Allocation distributions for `wide_resnet50_2`

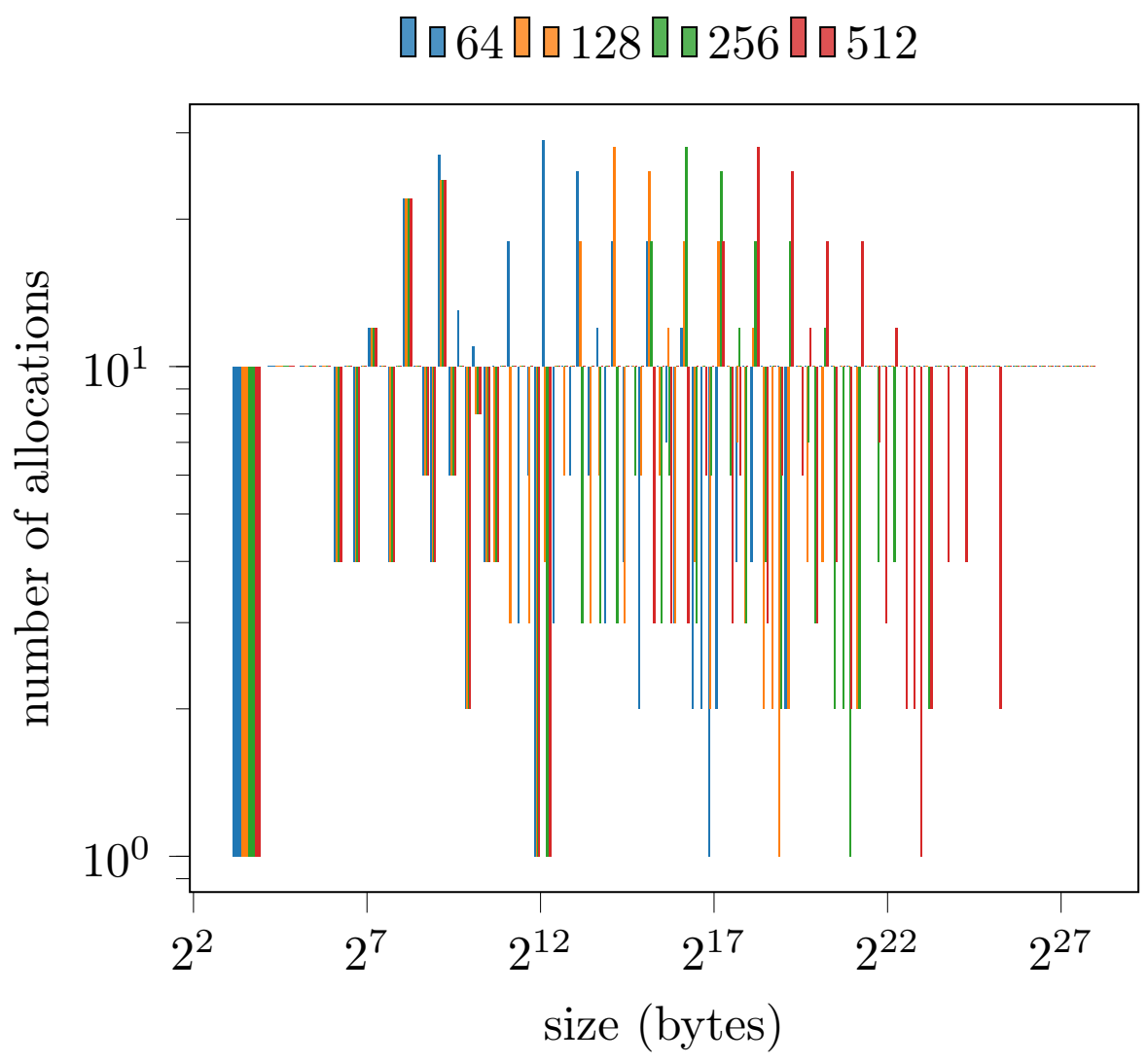

Allocation distributions for `googlenet`

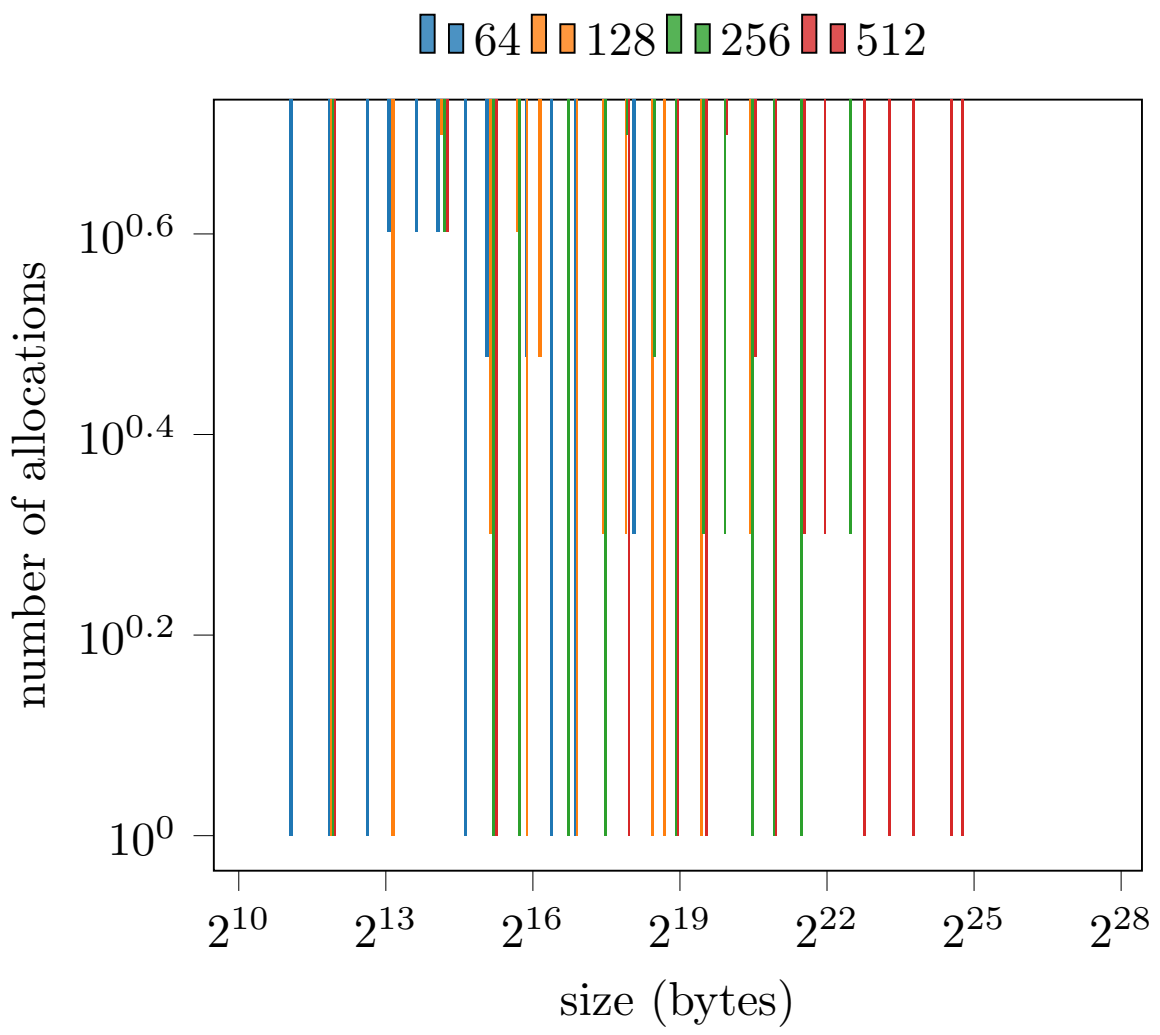

Allocation distributions for `alexnet`

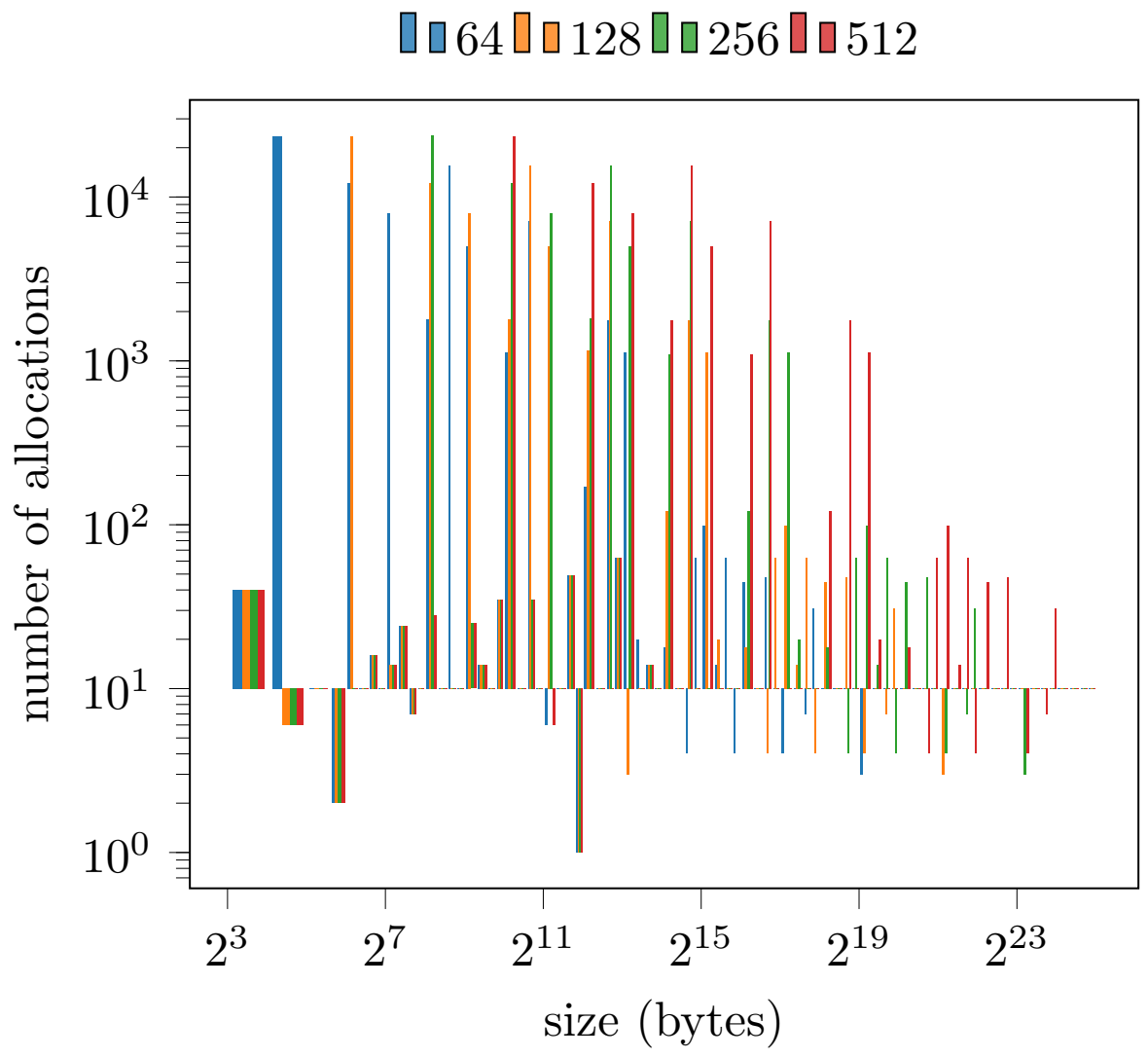

Allocation distributions for `efficientnet_b5`

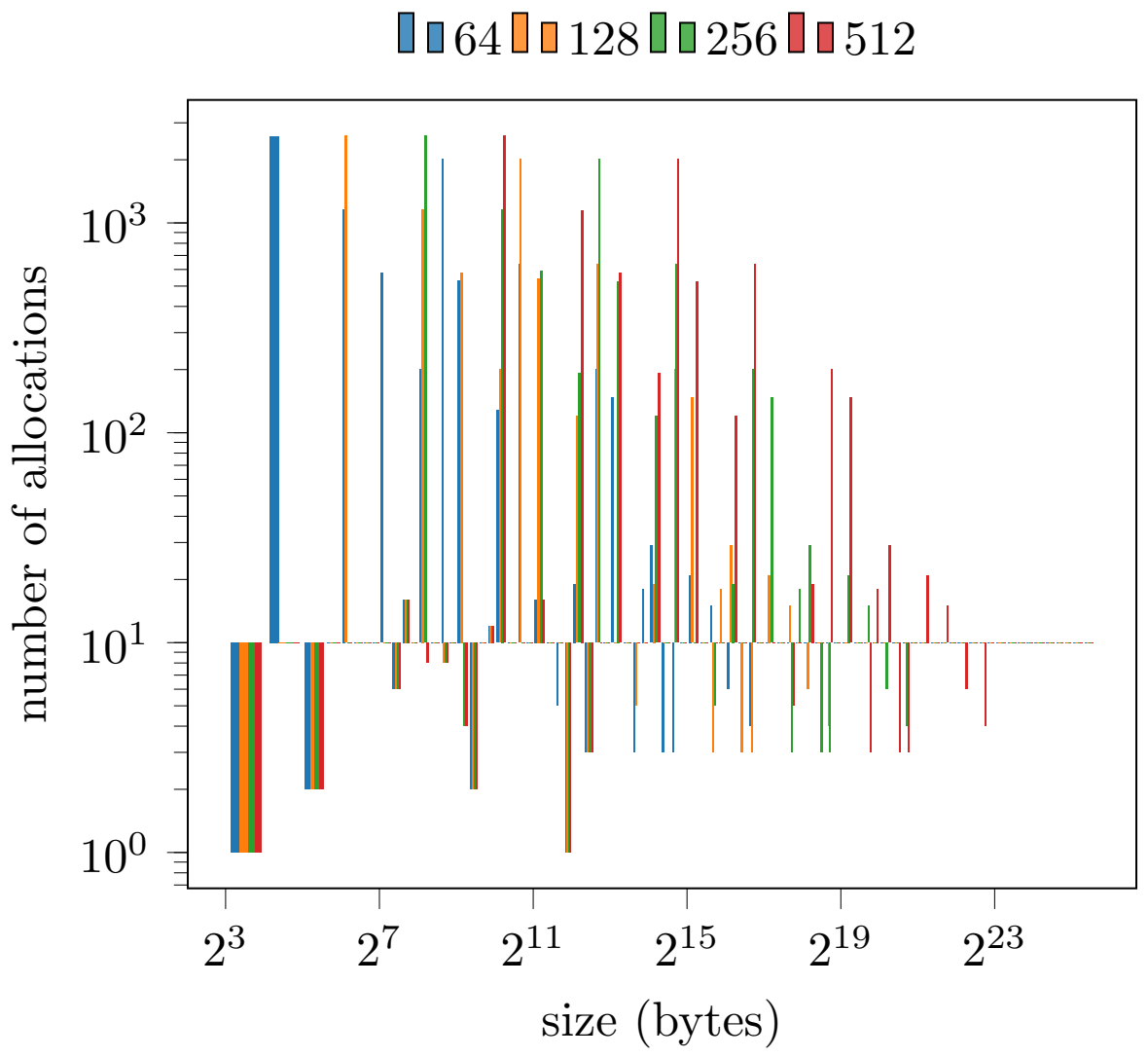

Allocation distributions for `mnasnet0_5`

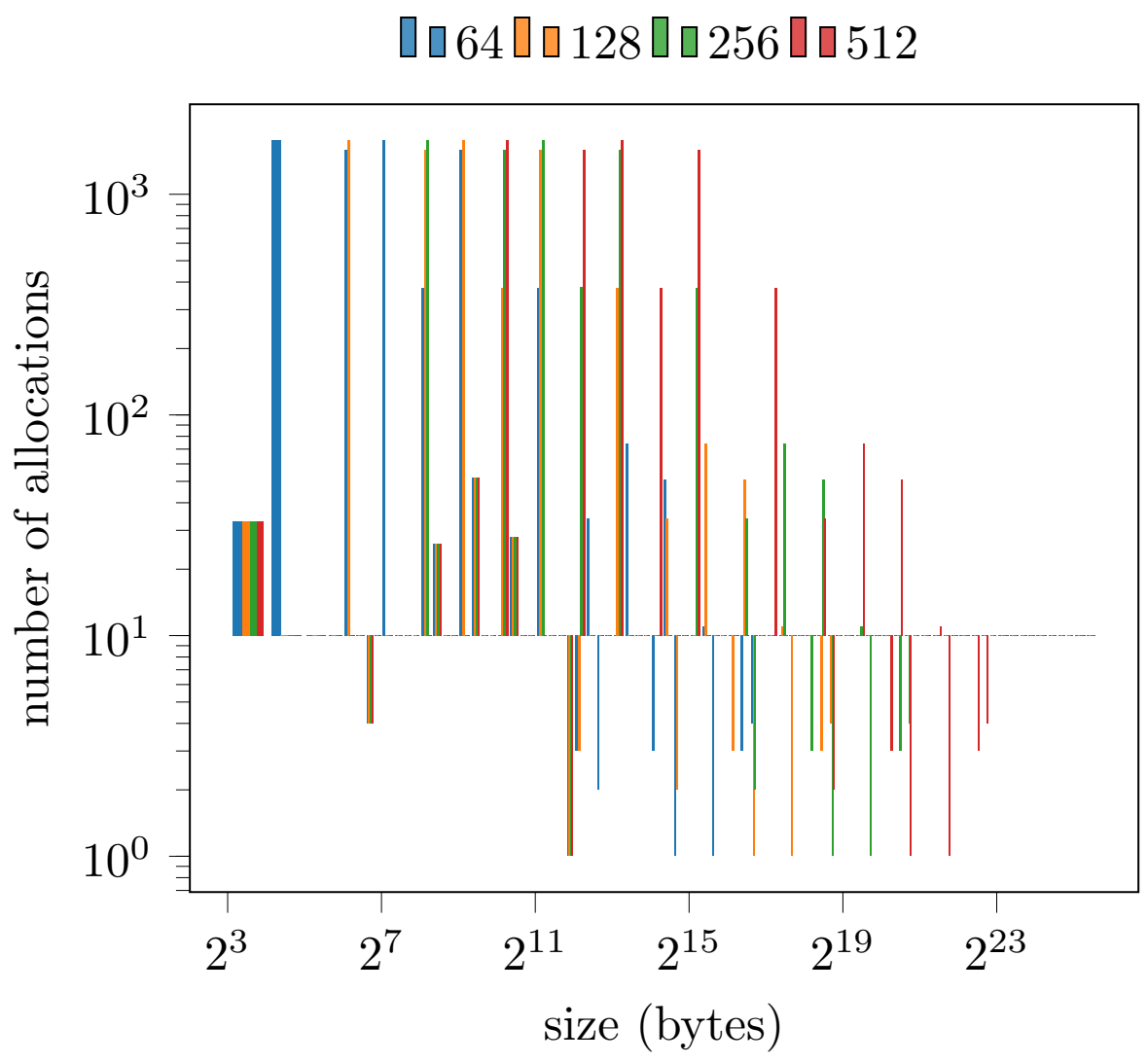

Allocation distributions for `shufflenet_v2_x1_5`

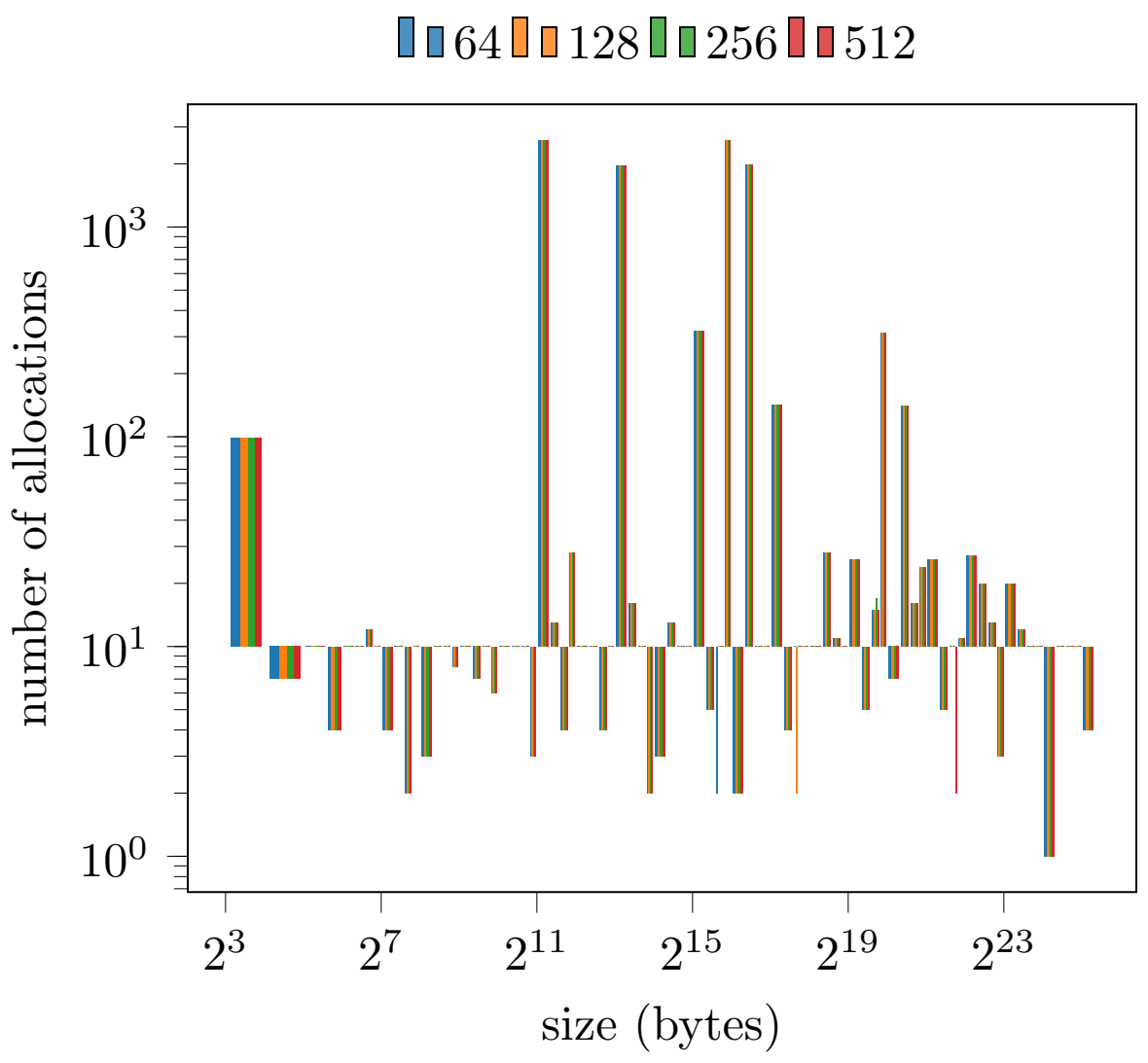

Allocation distributions for `fasterrcnn_mobilenet_v3_large_fpn`

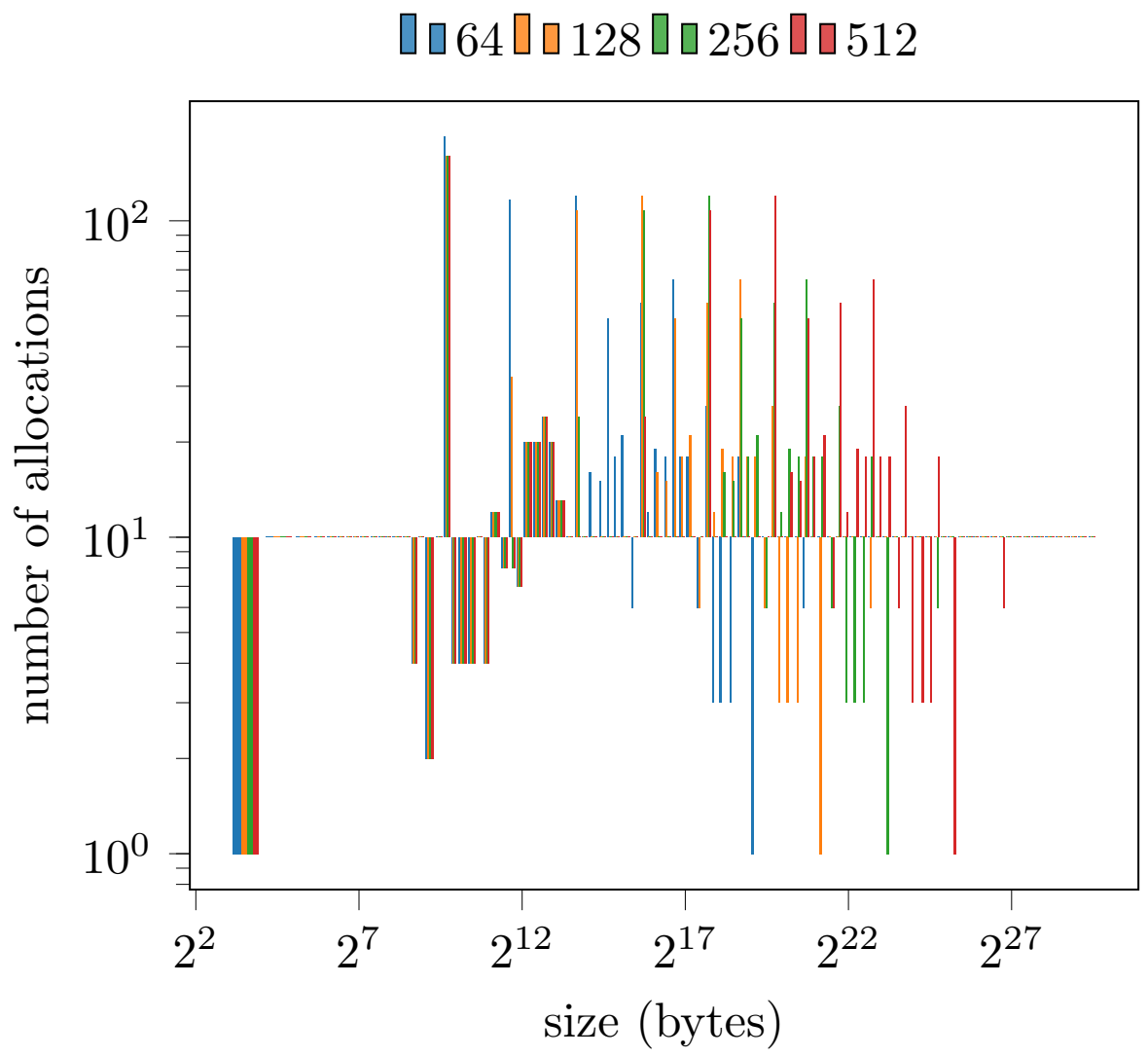

Allocation distributions for `densenet161`

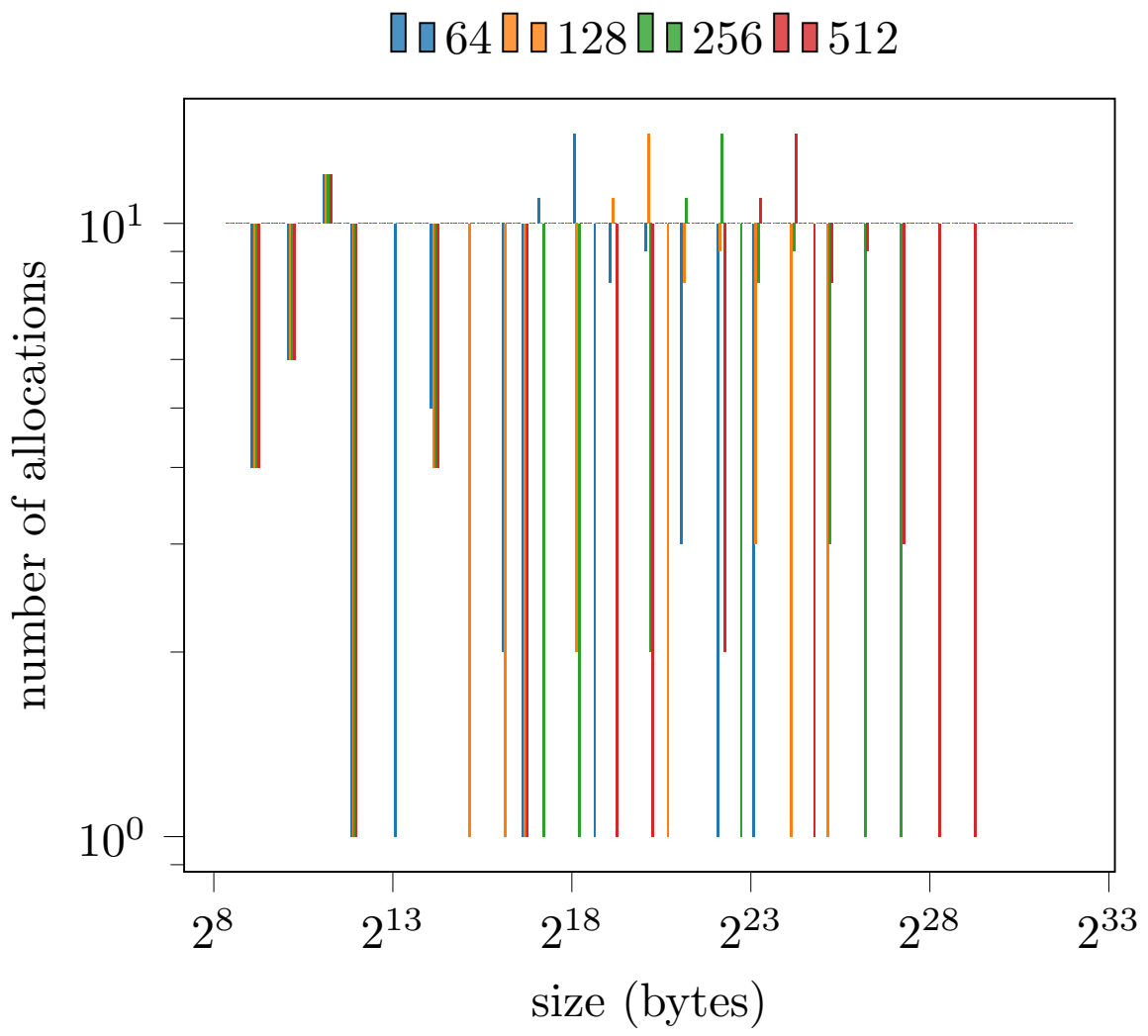

Allocation distributions for `vgg16_bn`

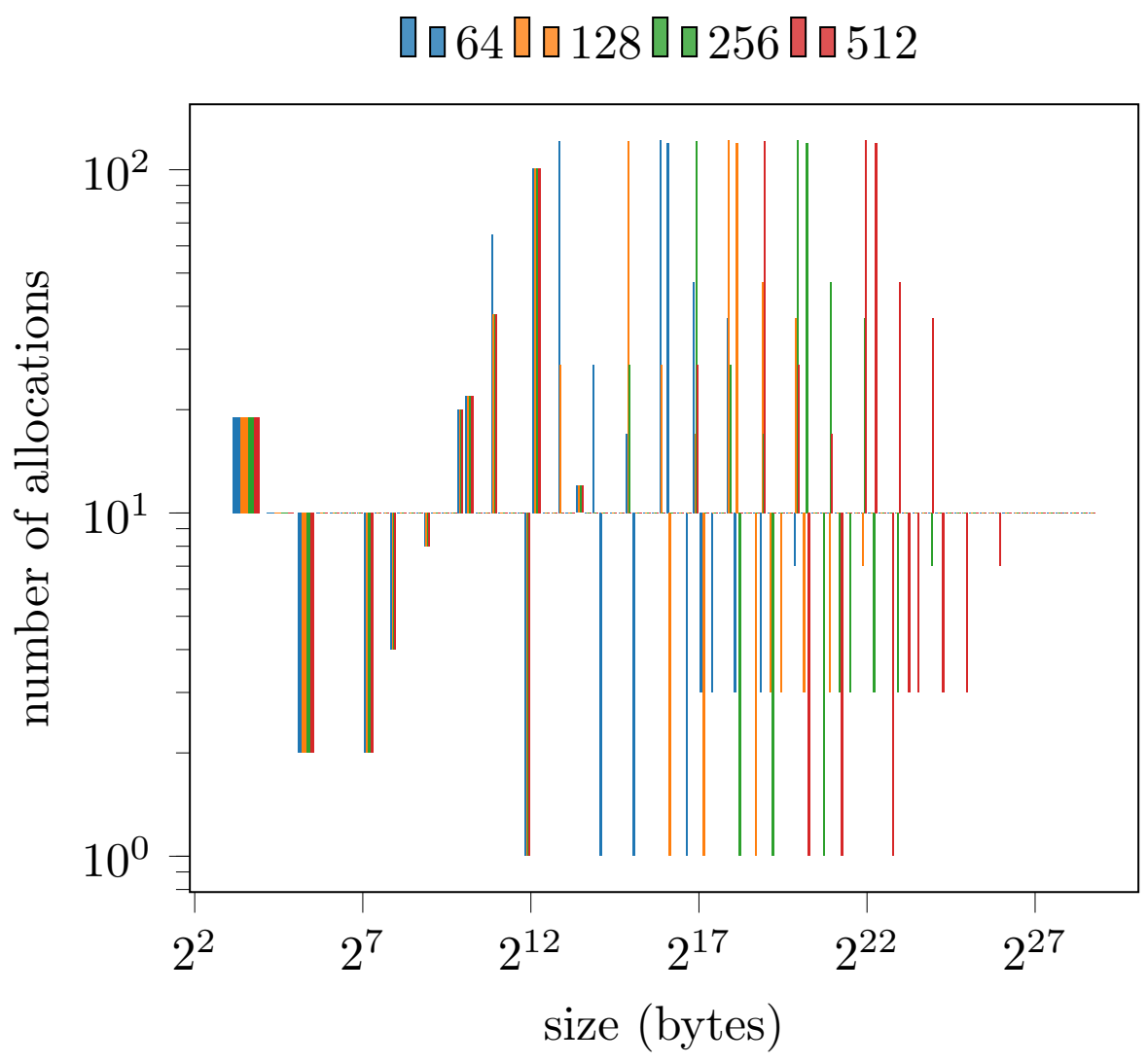

Allocation distributions for `regnet_y_16gf`

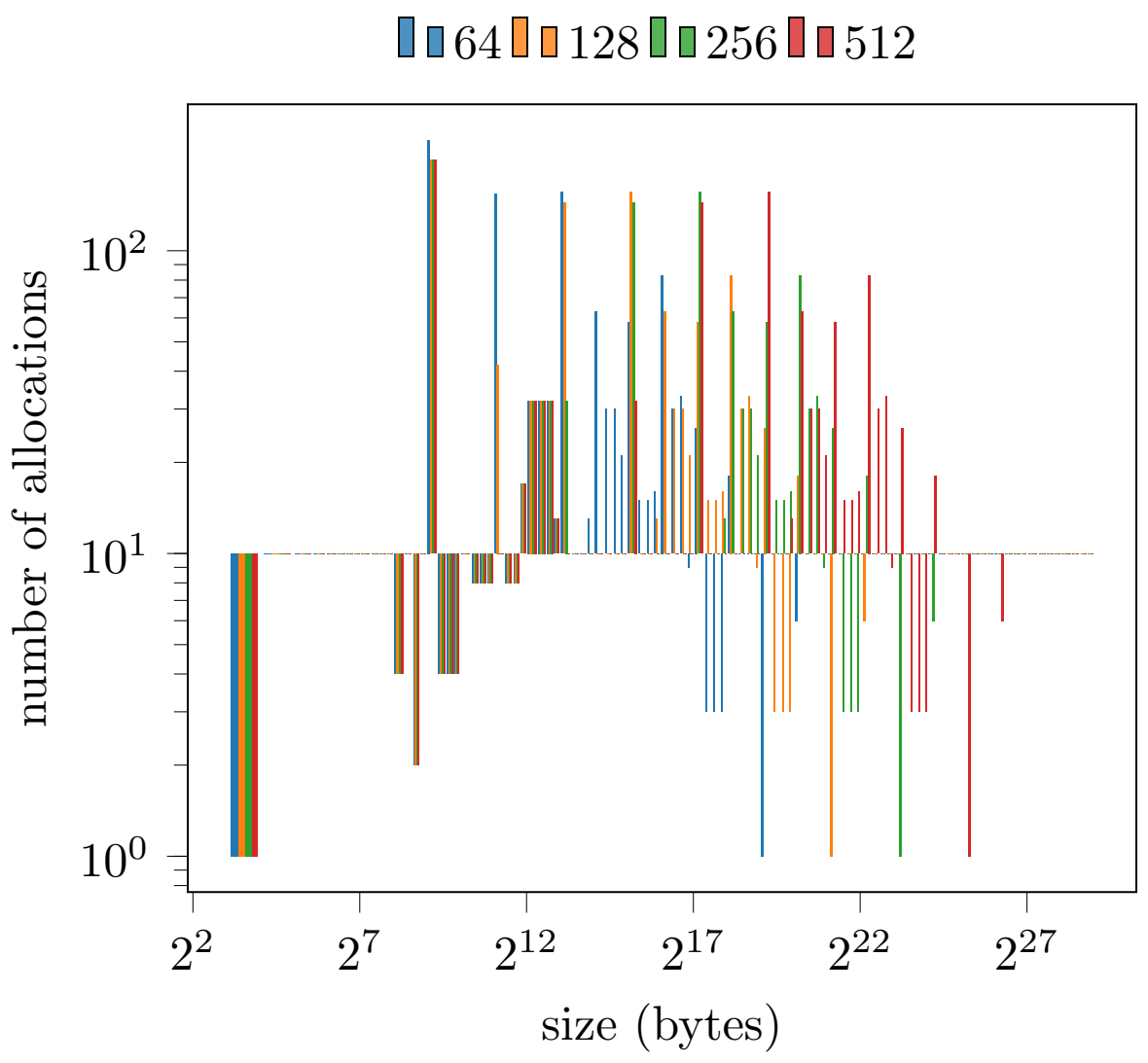

Allocation distributions for `densenet201`

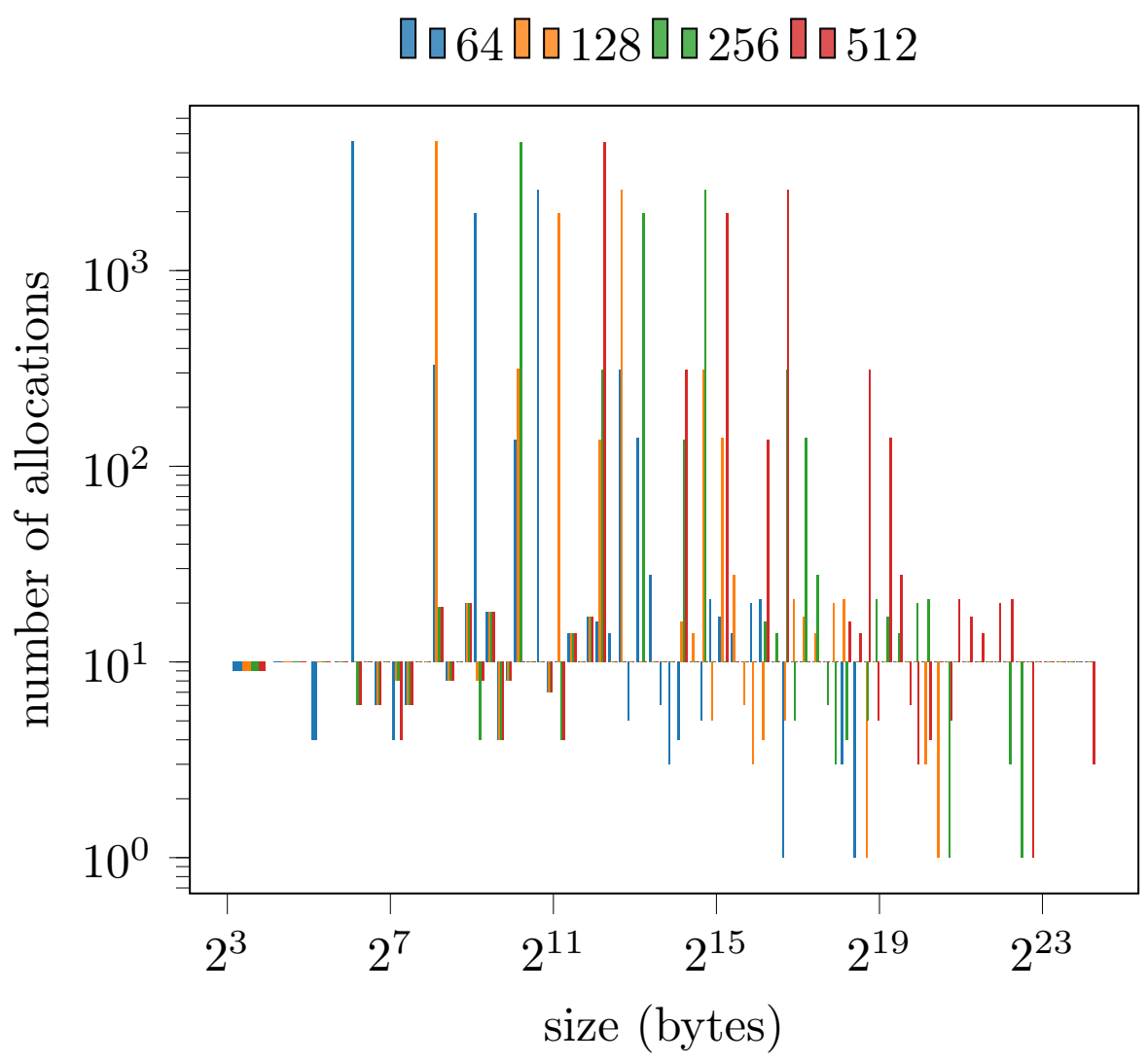

Allocation distributions for `lraspp_mobilenet_v3_large`

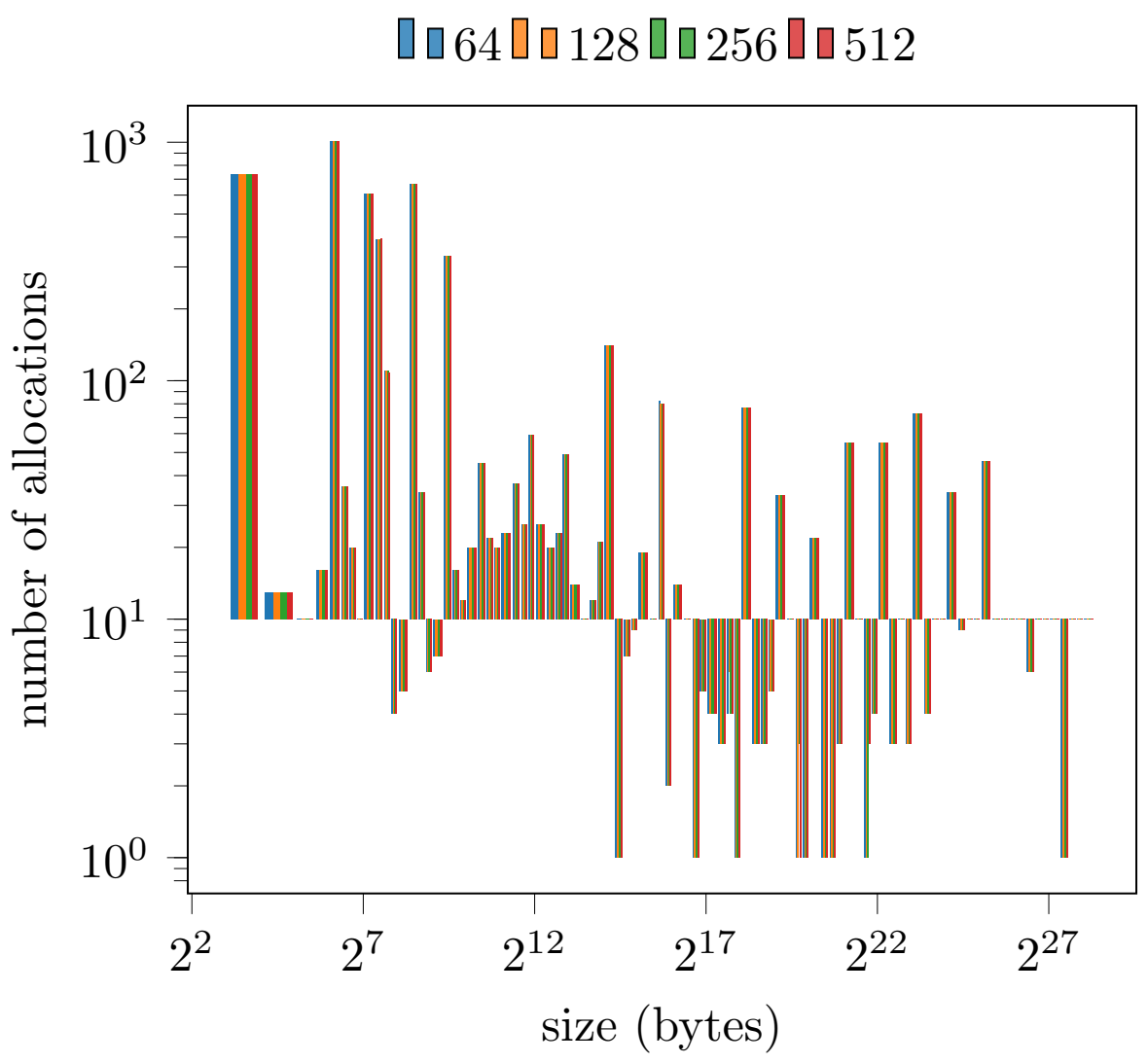

Allocation distributions for `keypointrcnn_resnet50_fpn`

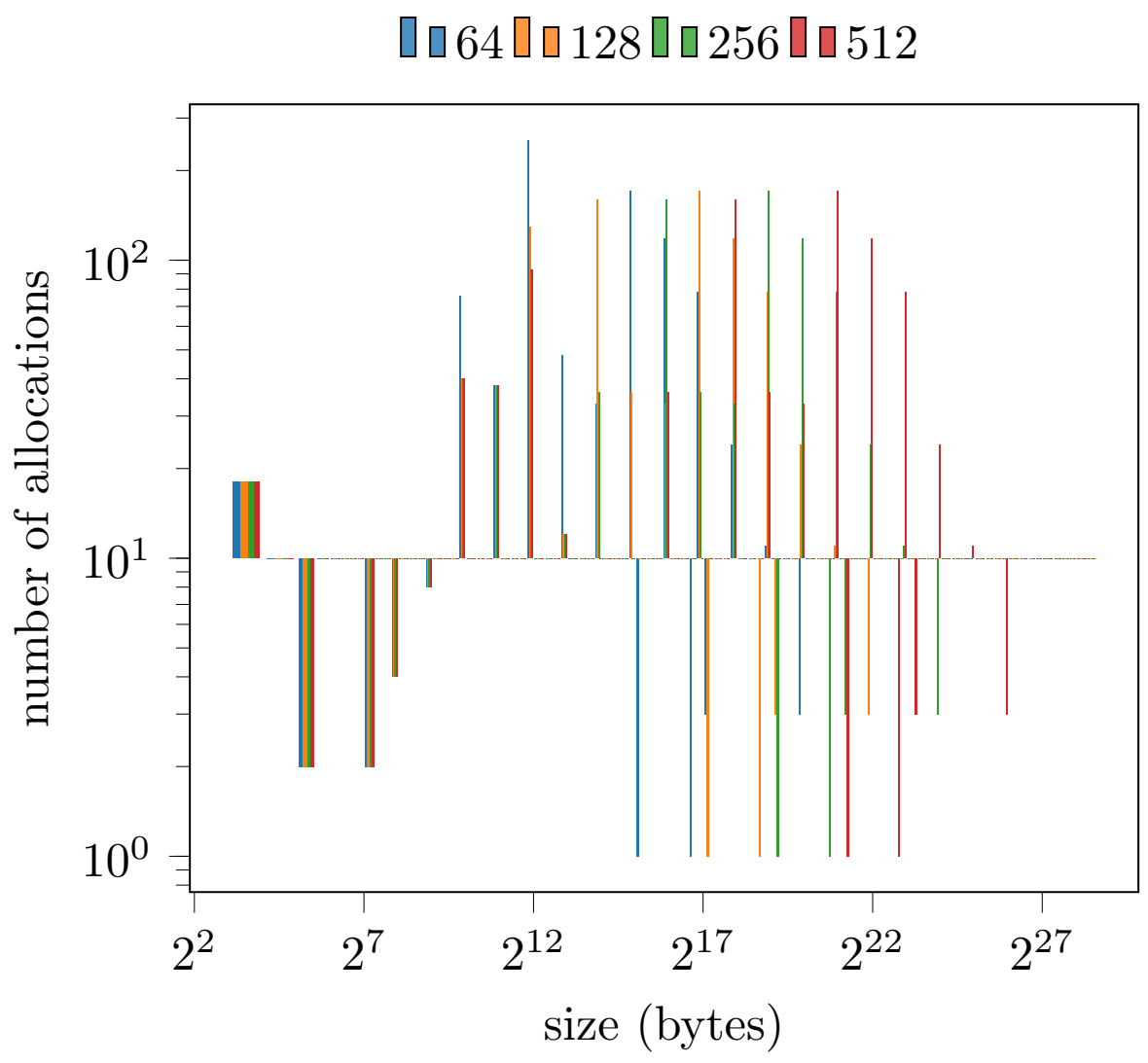

Allocation distributions for `regnet_y_8gf`

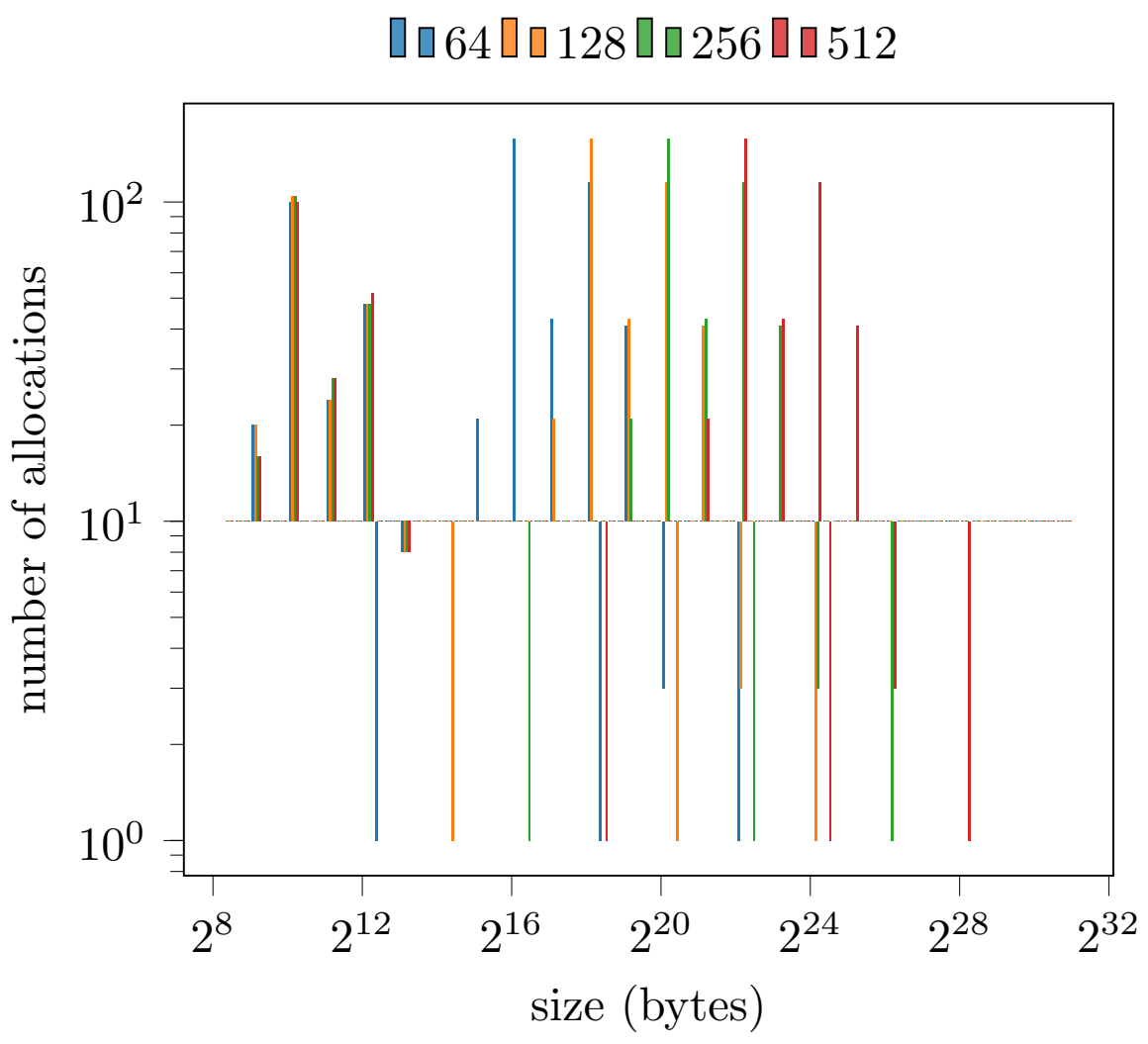

Allocation distributions for `fcn_resnet101`

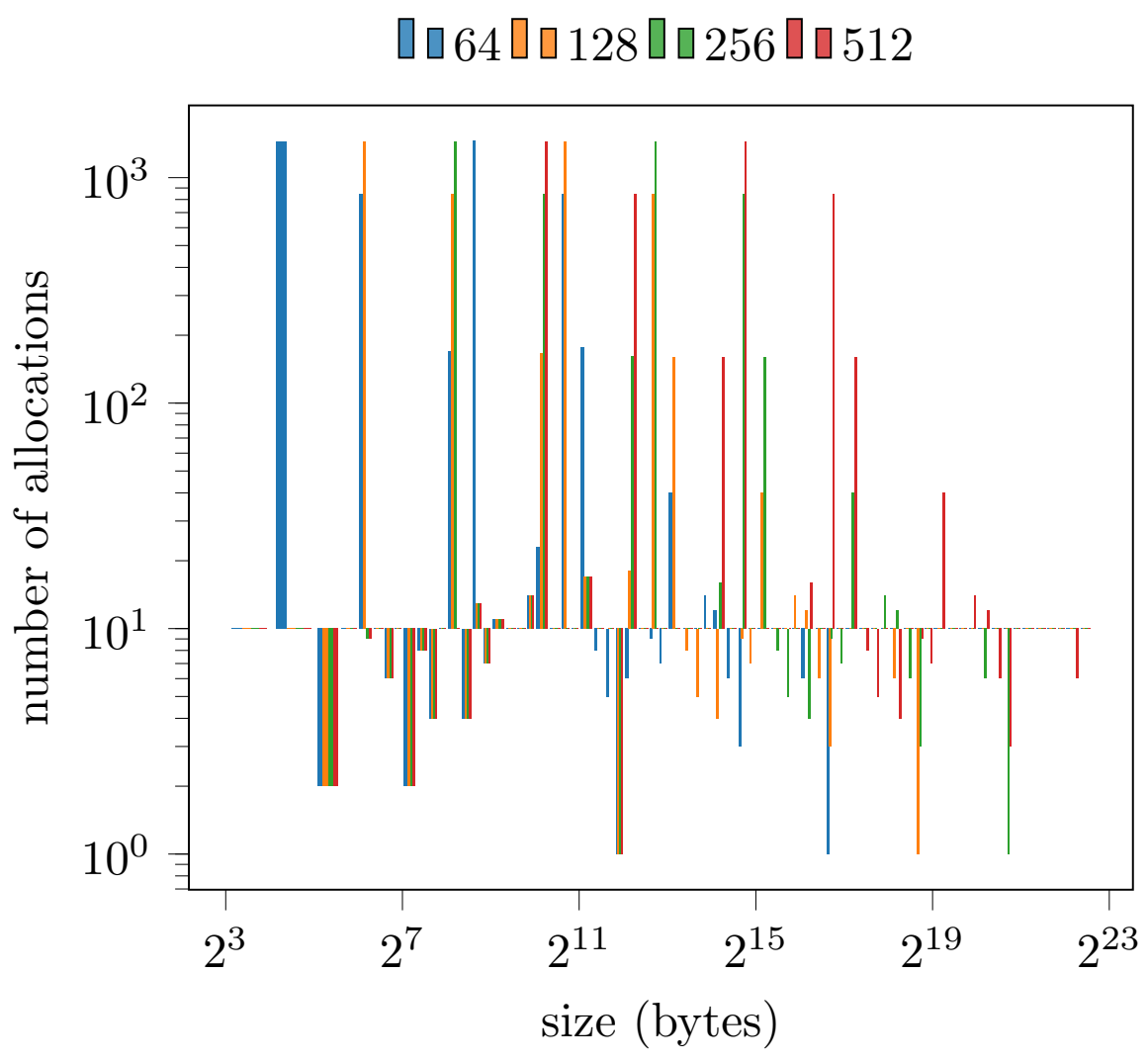

Allocation distributions for `mobilenet_v3_small`

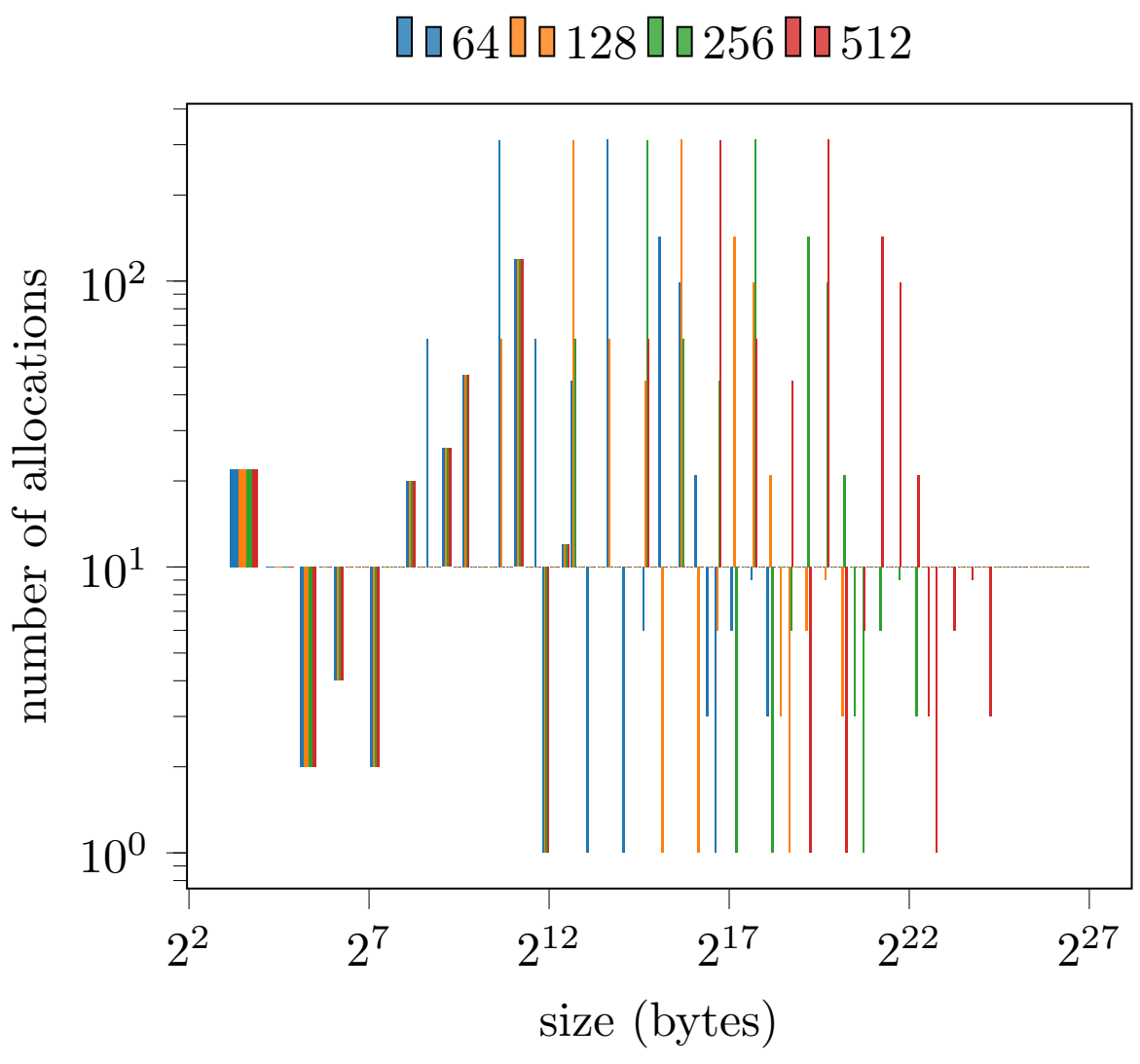

Allocation distributions for `regnet_y_3_2gf`

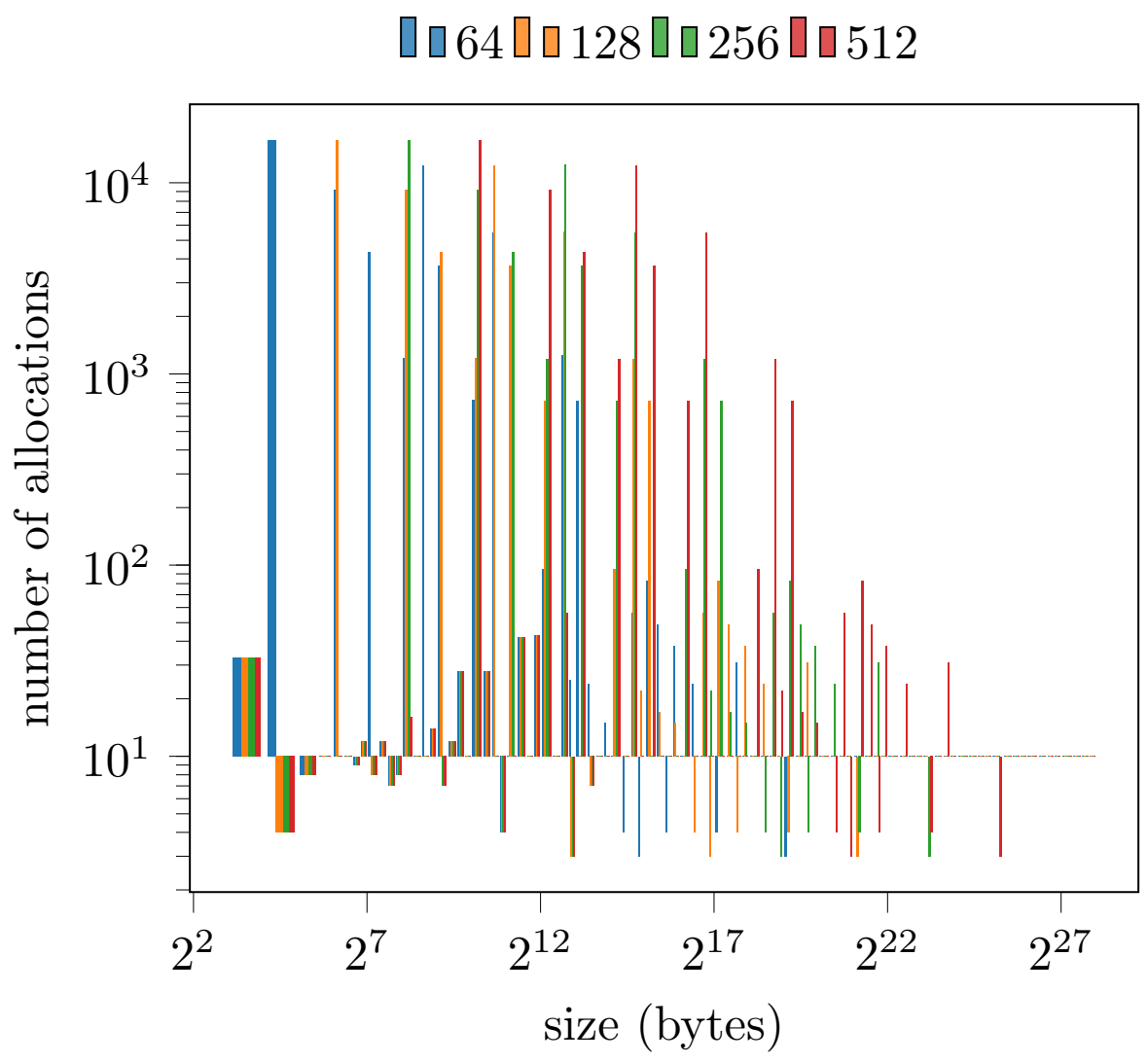

Allocation distributions for `efficientnet_b4`

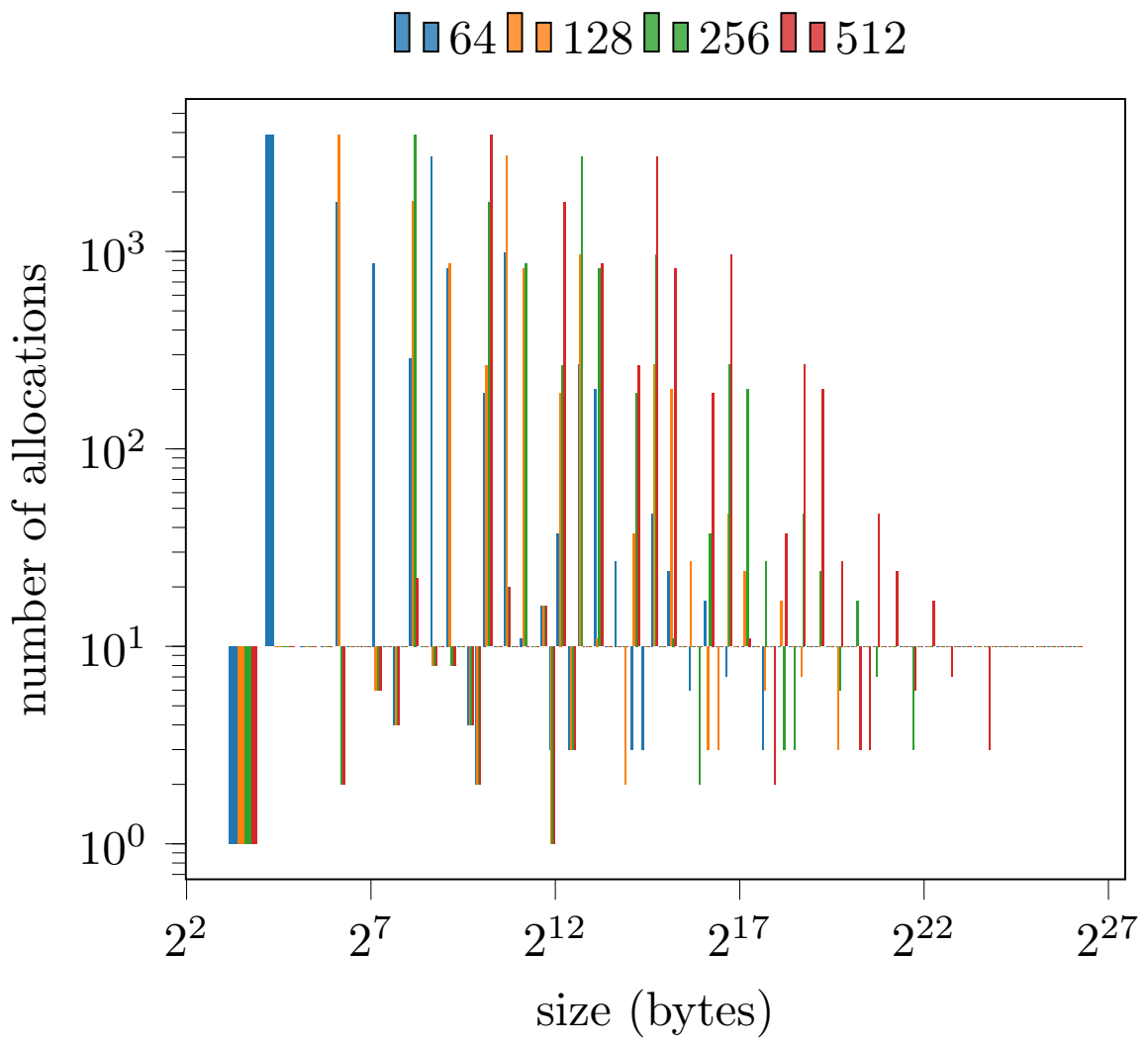

Allocation distributions for `mnasnet0_75`

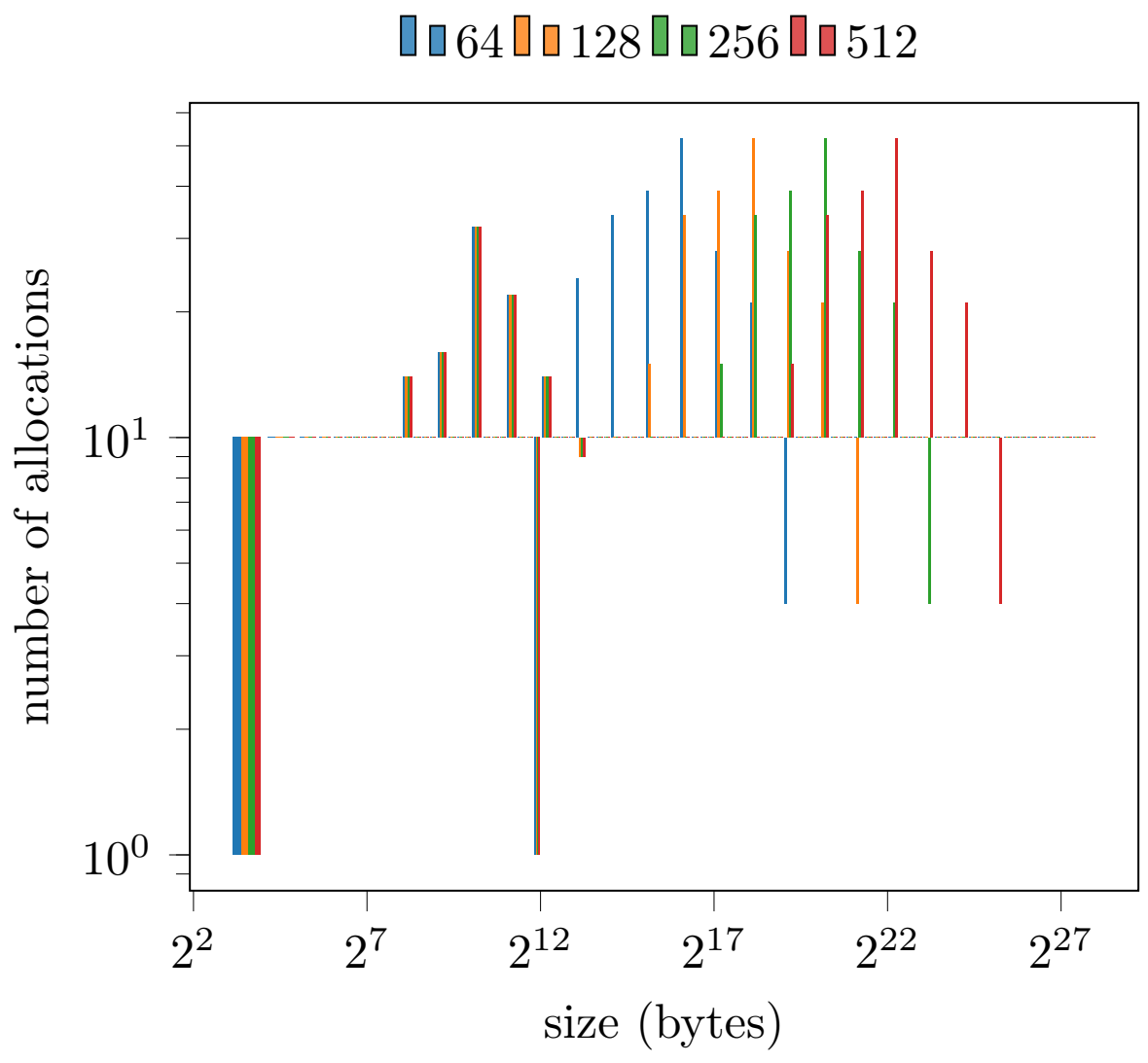

Allocation distributions for `resnet50`

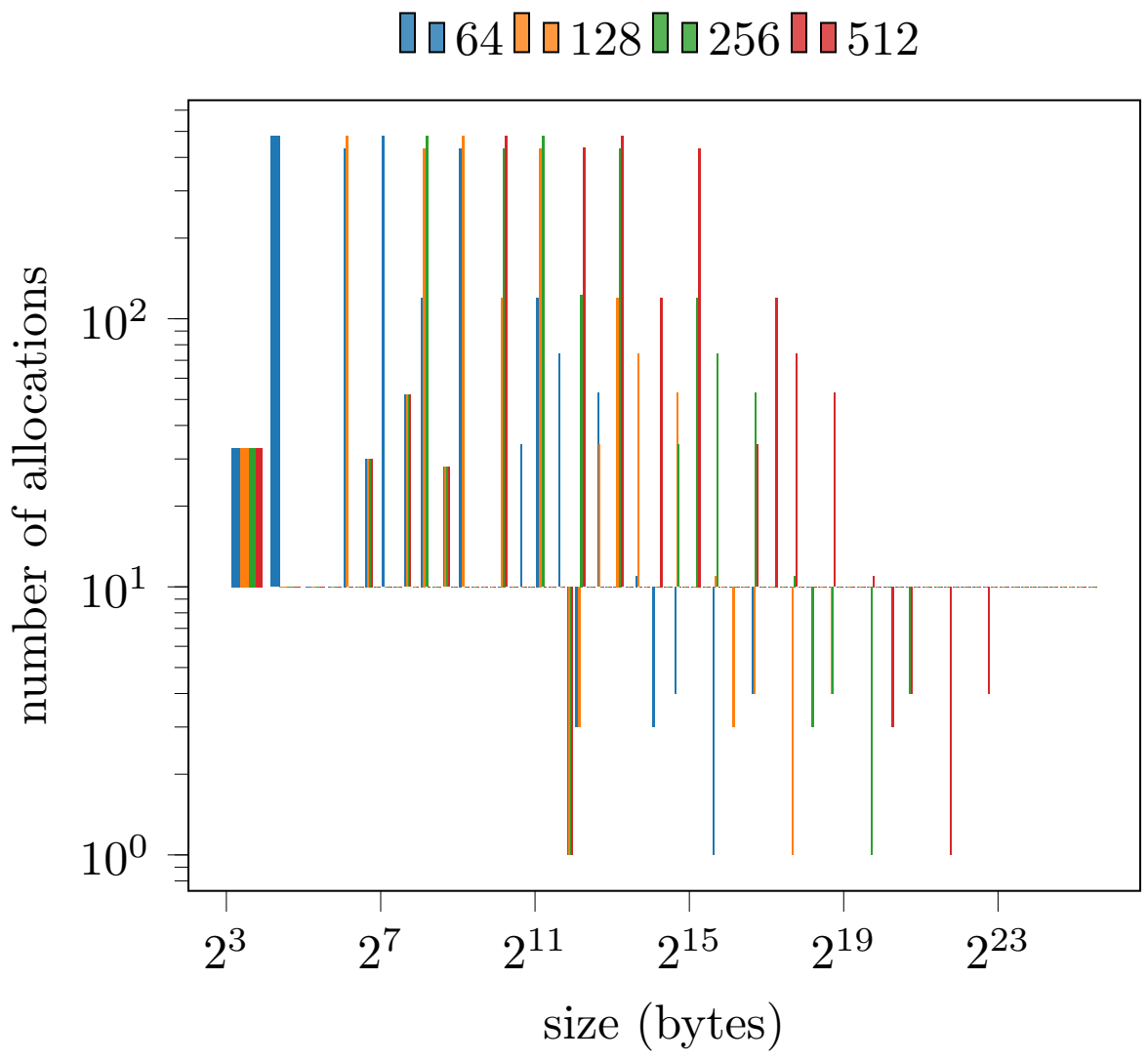

Allocation distributions for `shufflenet_v2_x0_5`

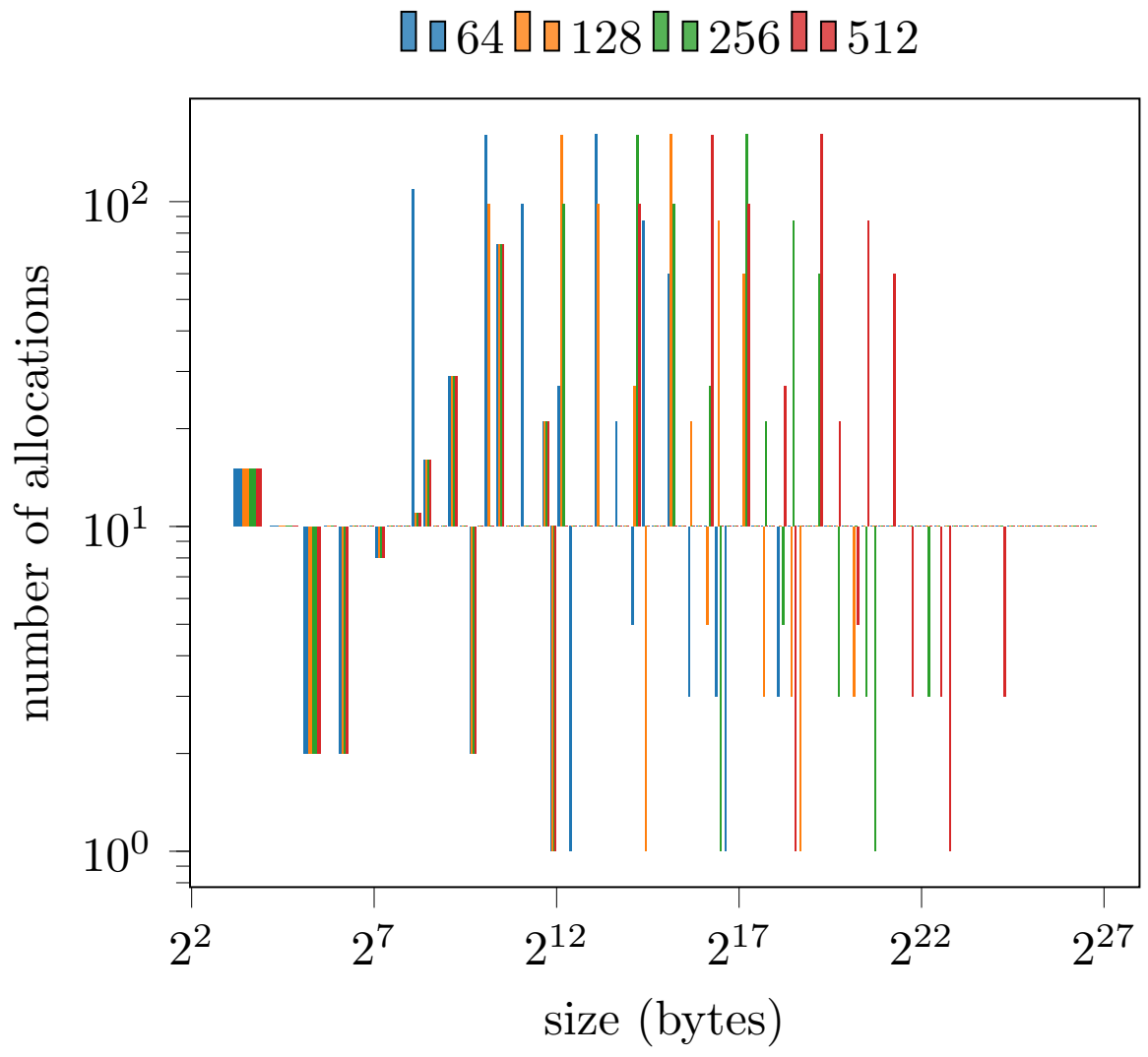

Allocation distributions for `regnet_y_800mf`

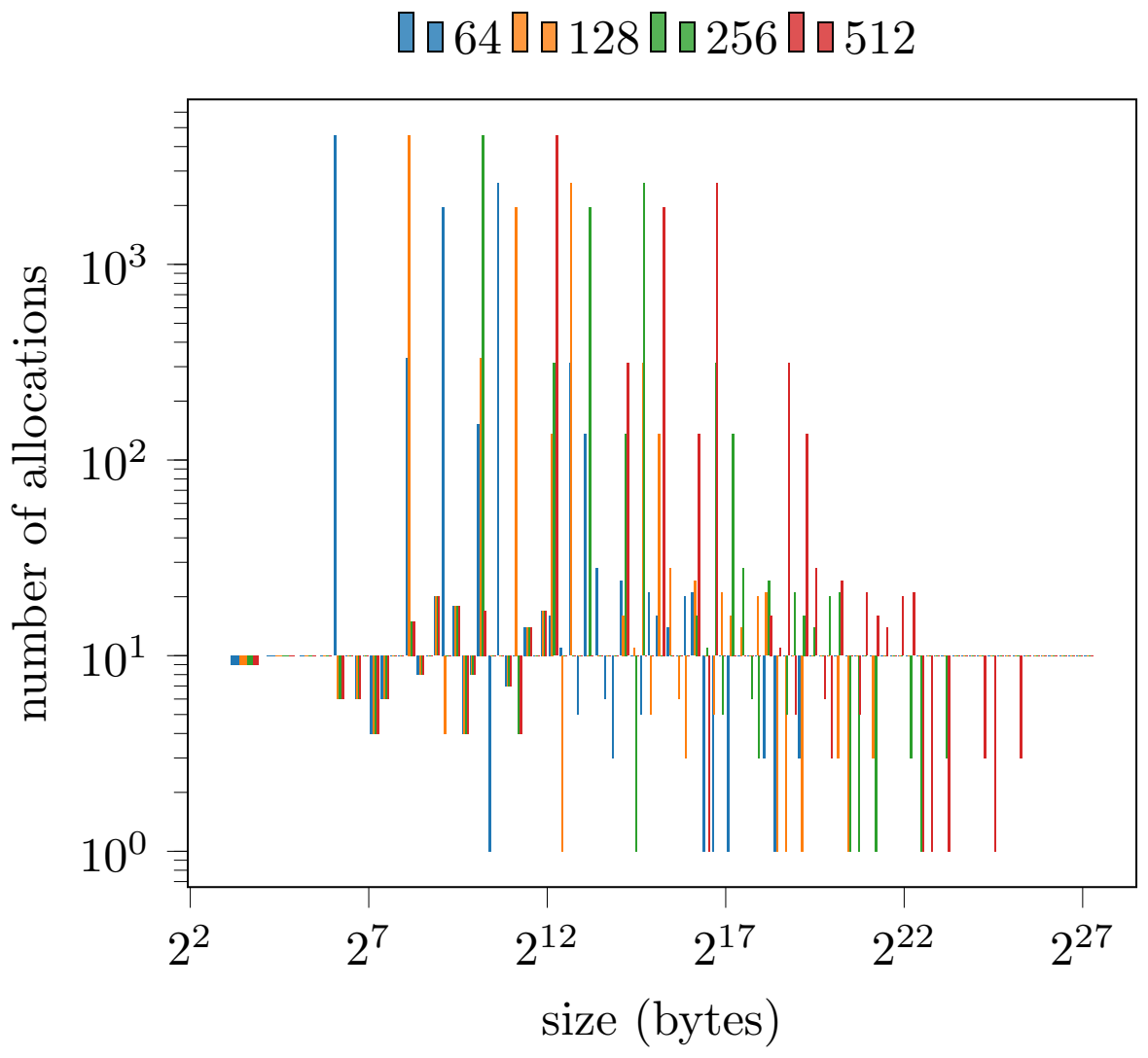

Allocation distributions for `deeplabv3_mobilenet_v3_large`

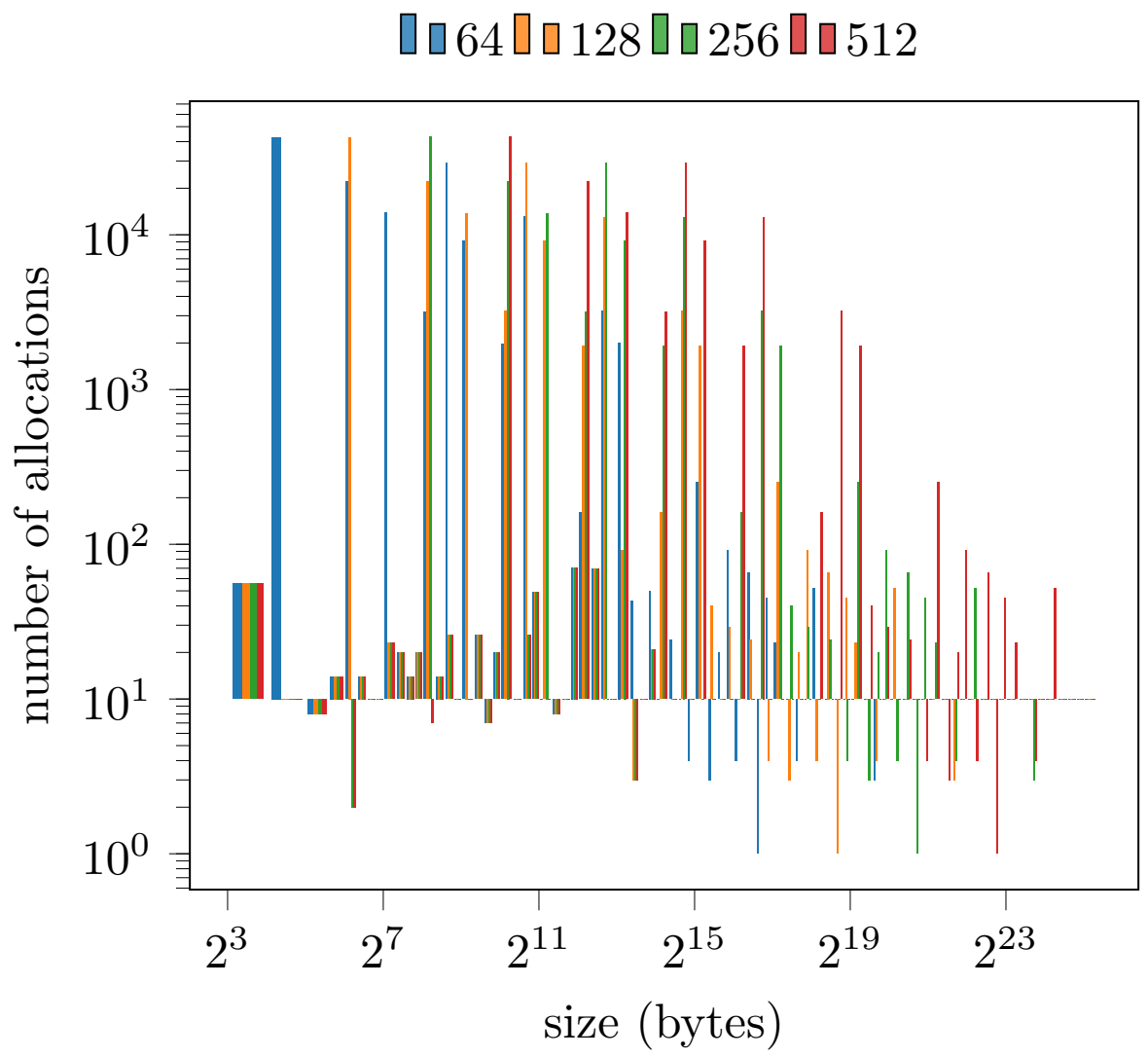

Allocation distributions for `efficientnet_b7`

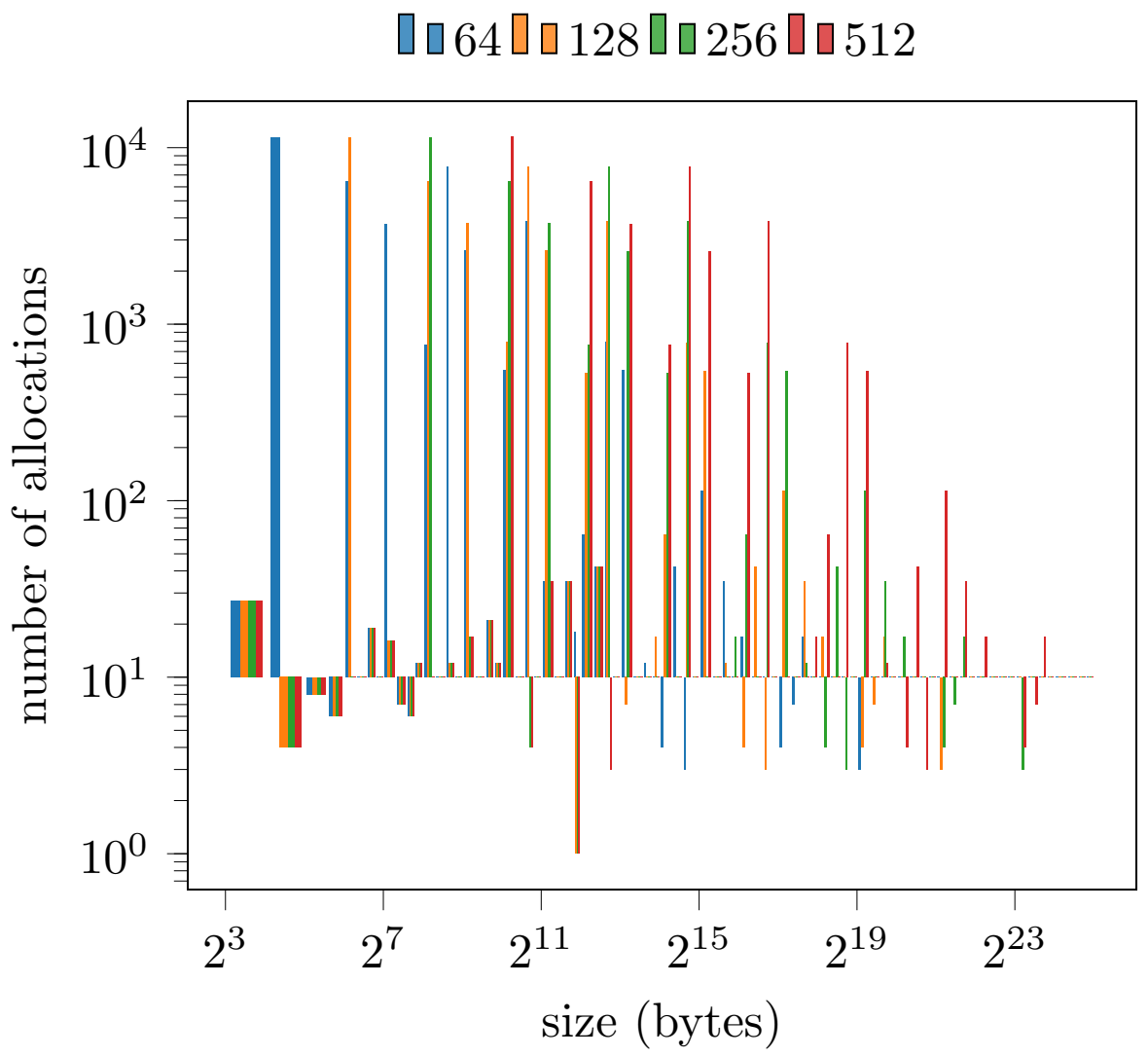

Allocation distributions for `efficientnet_b3`

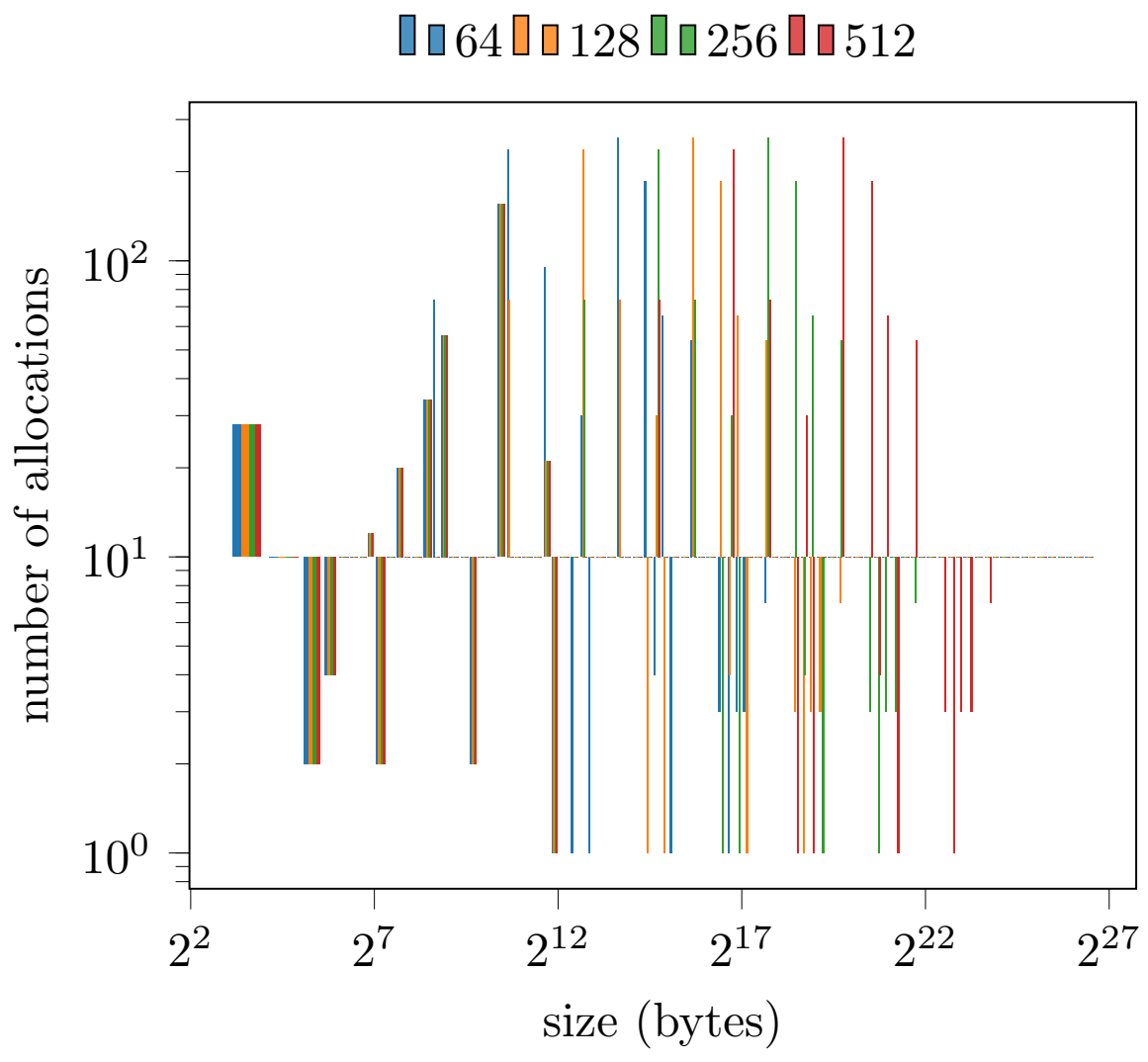

Allocation distributions for `regnet_y_1_6gf`

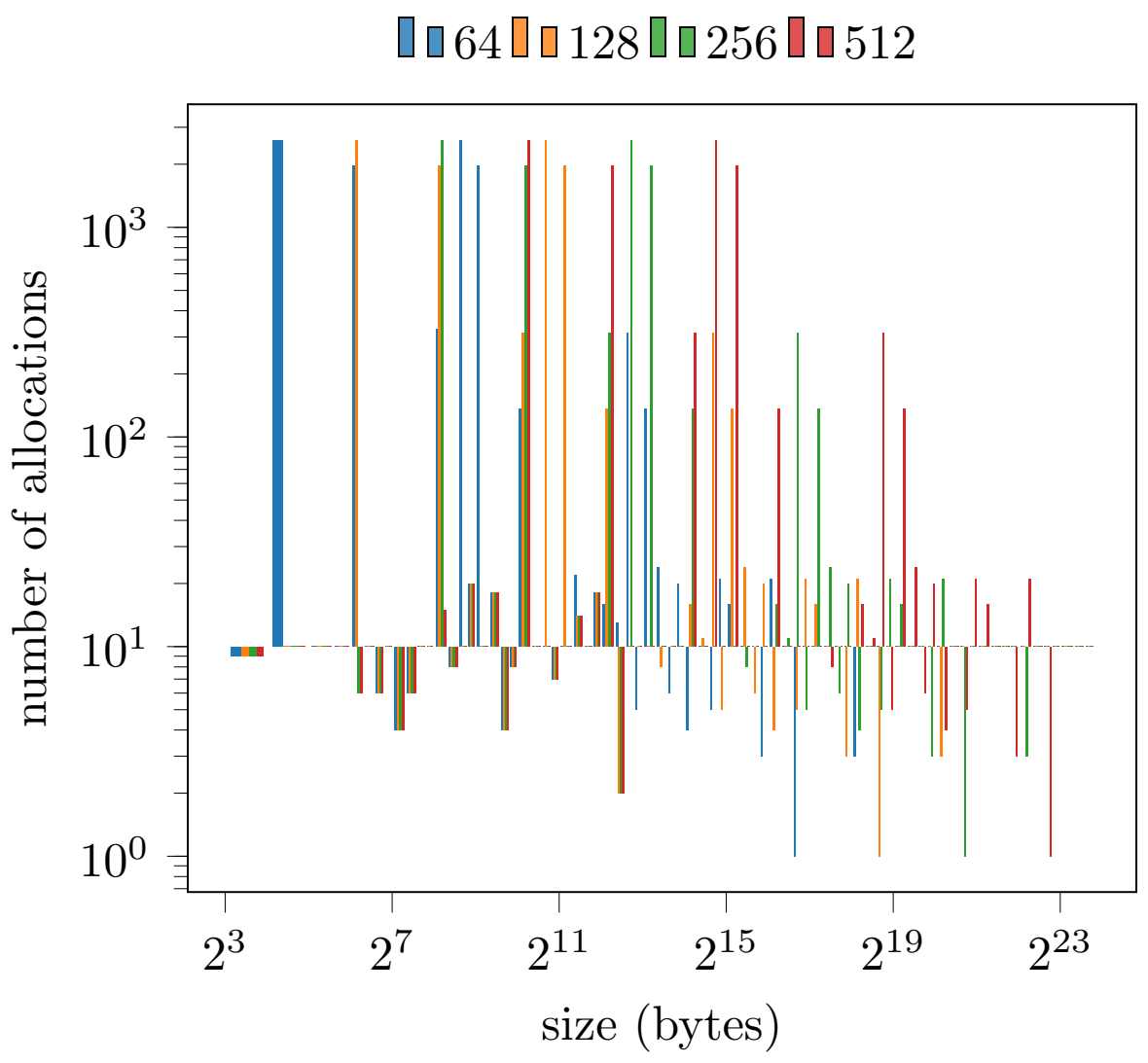

Allocation distributions for `mobilenet_v3_large`

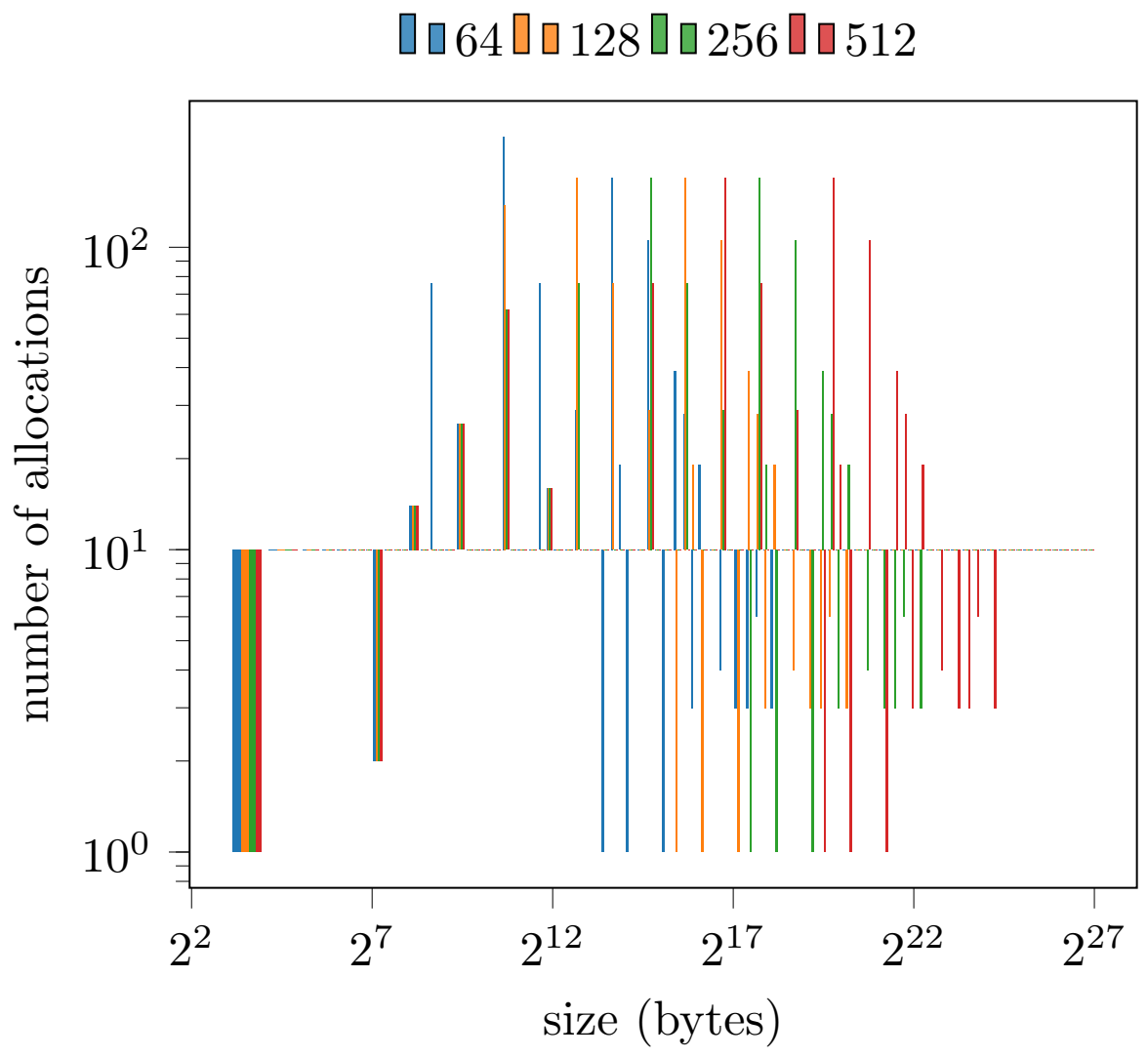

Allocation distributions for `regnet_x_1_6gf`

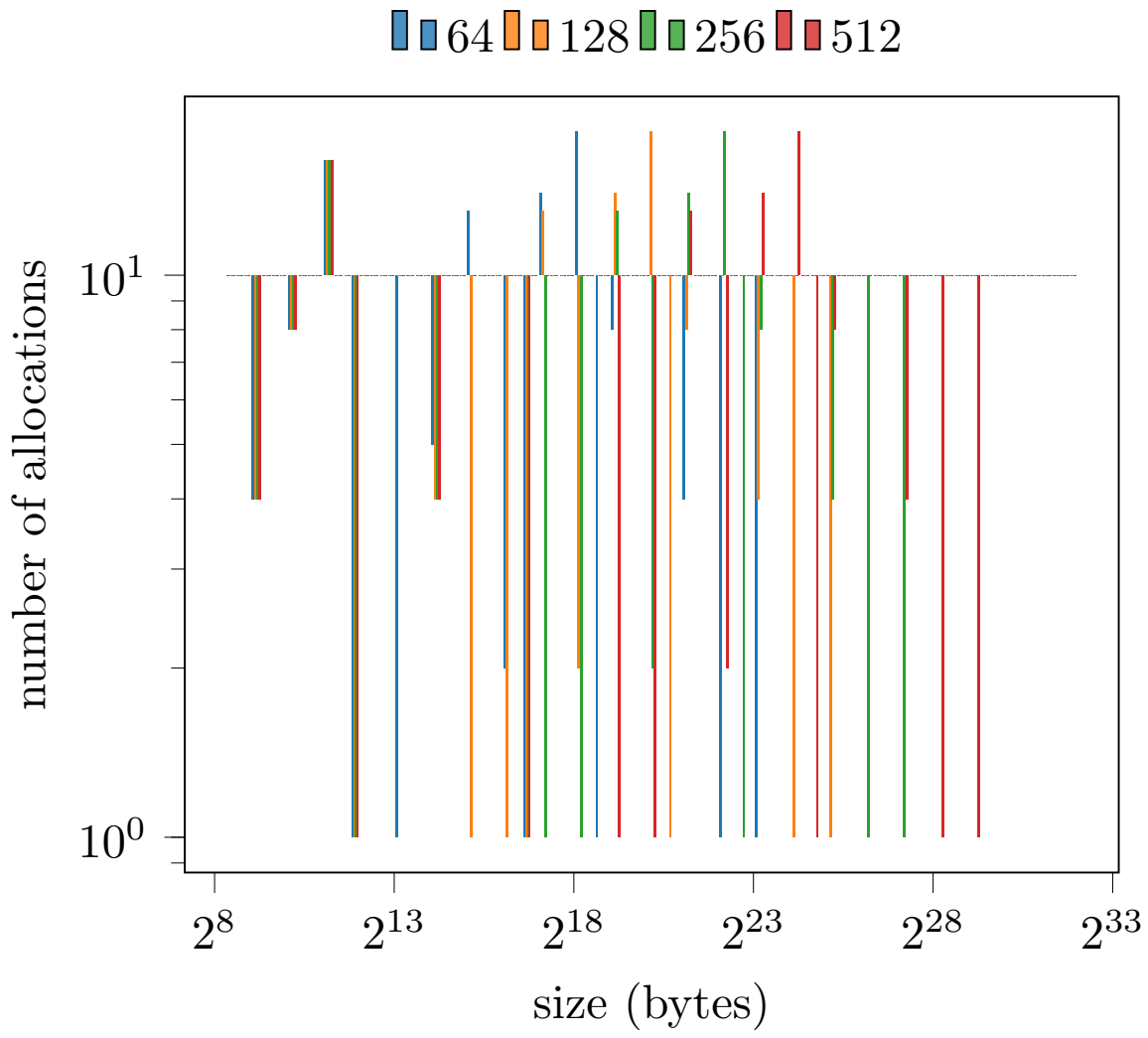

Allocation distributions for `vgg19_bn`

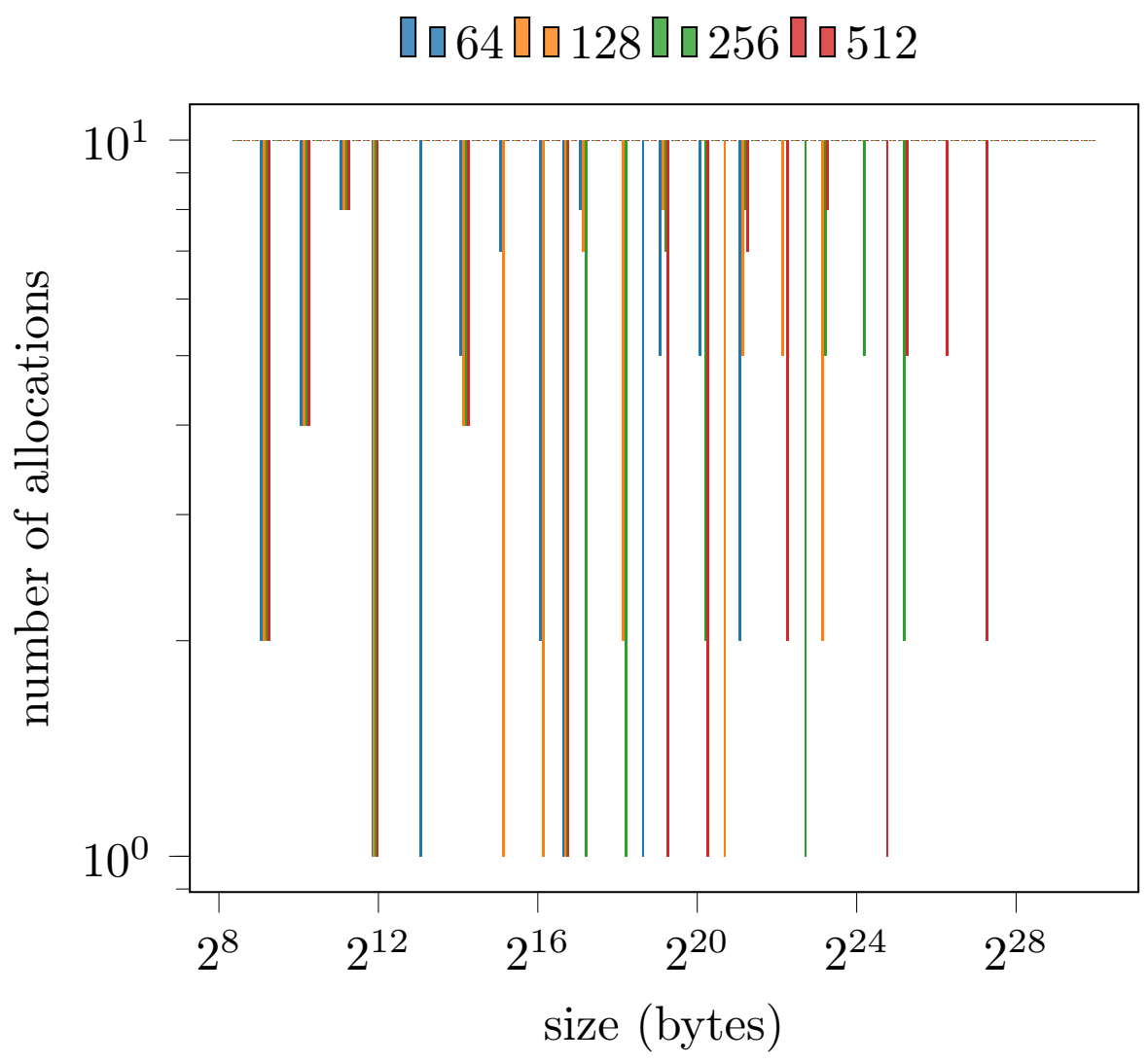

Allocation distributions for `vgg11_bn`

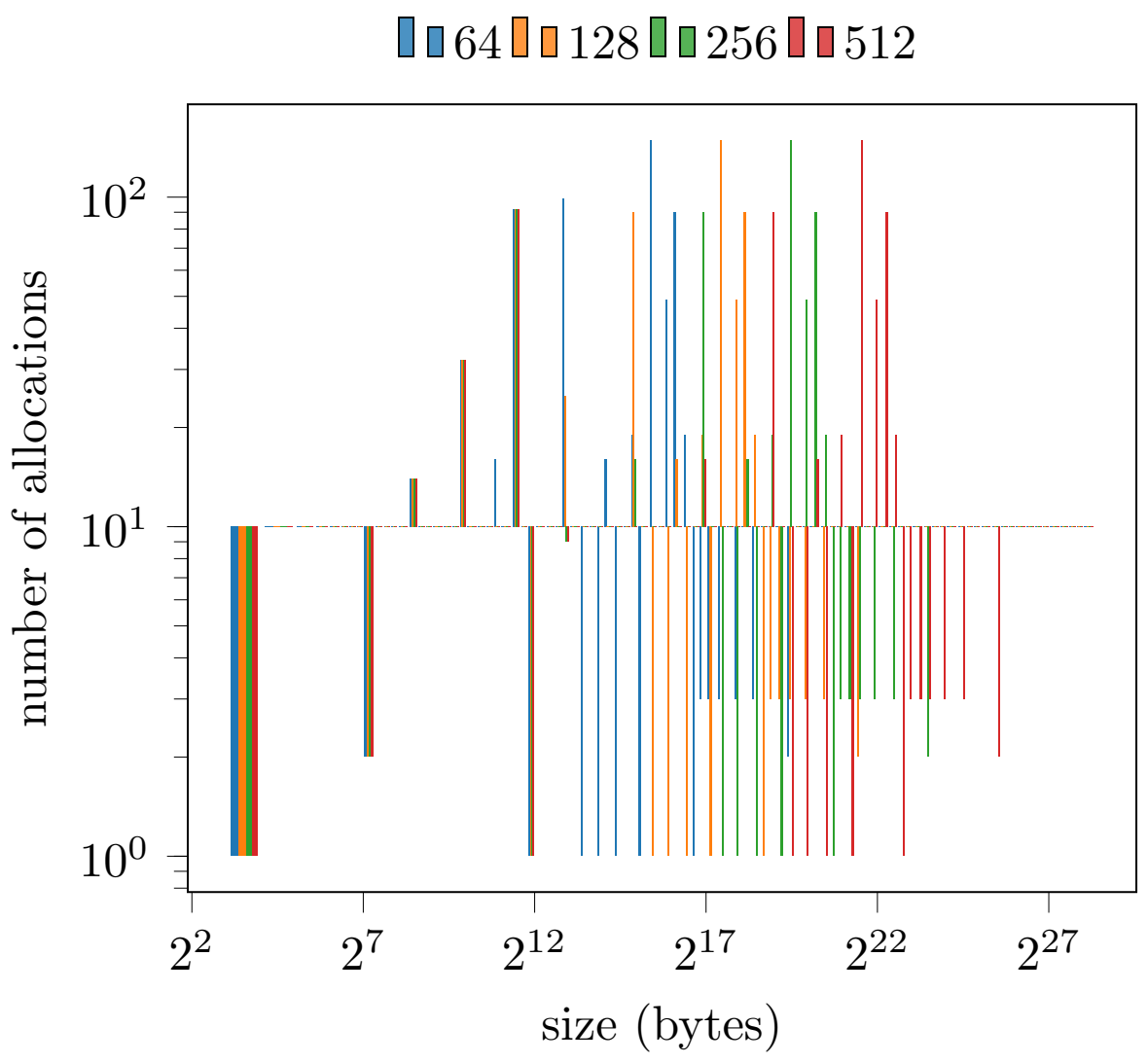

Allocation distributions for `regnet_x_8gf`

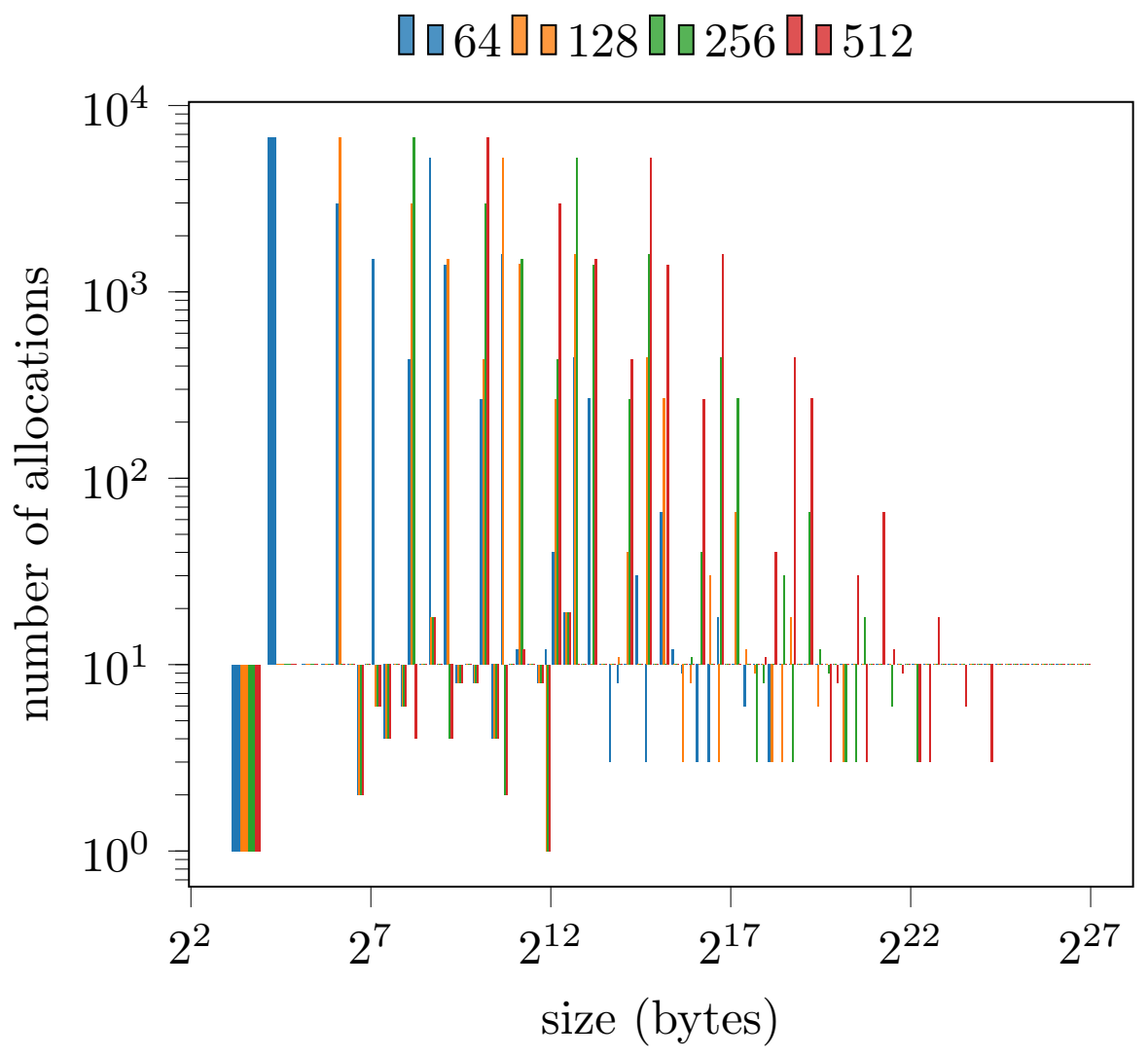

Allocation distributions for `mnasnet1_3`

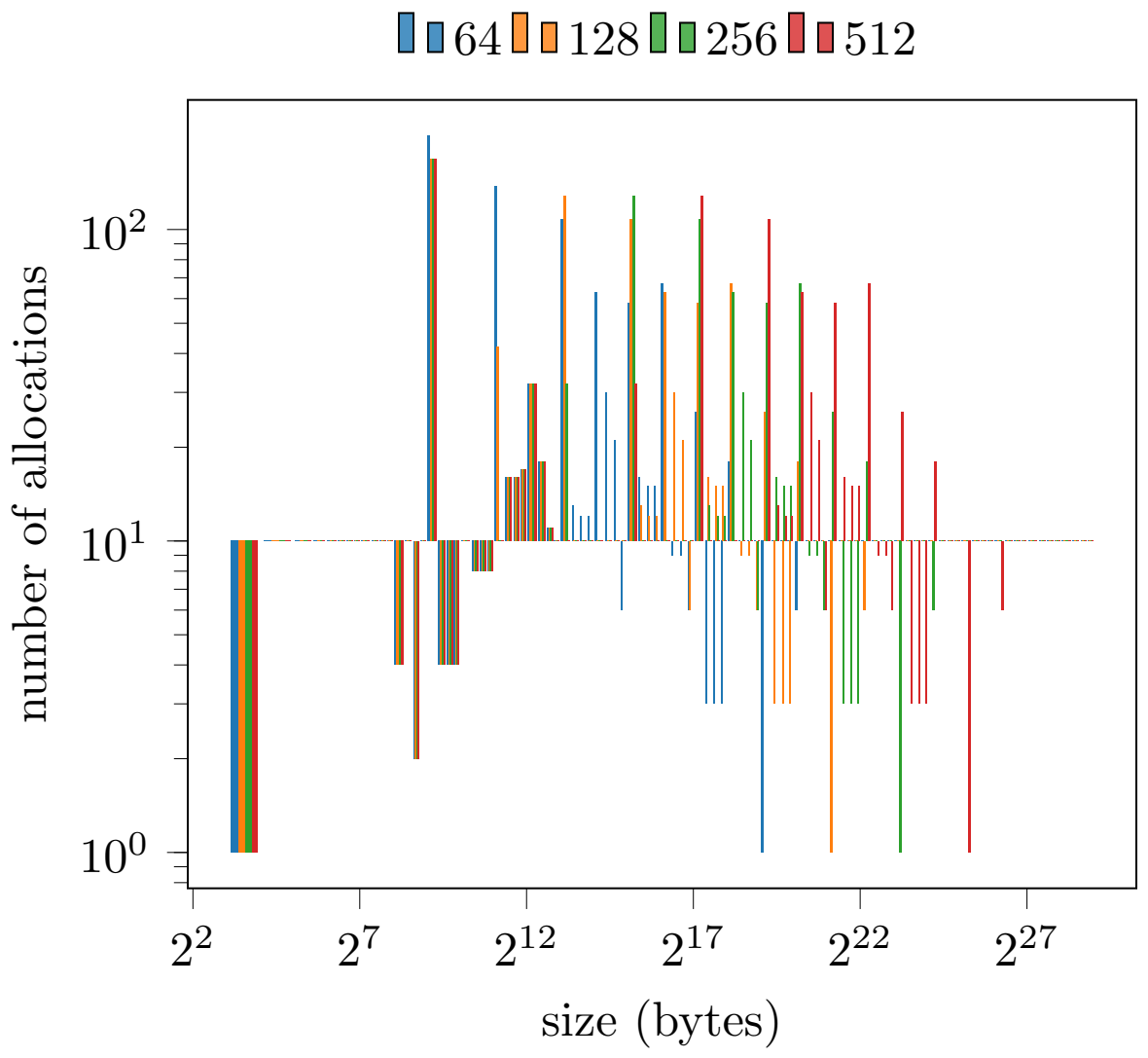

Allocation distributions for `densenet169`

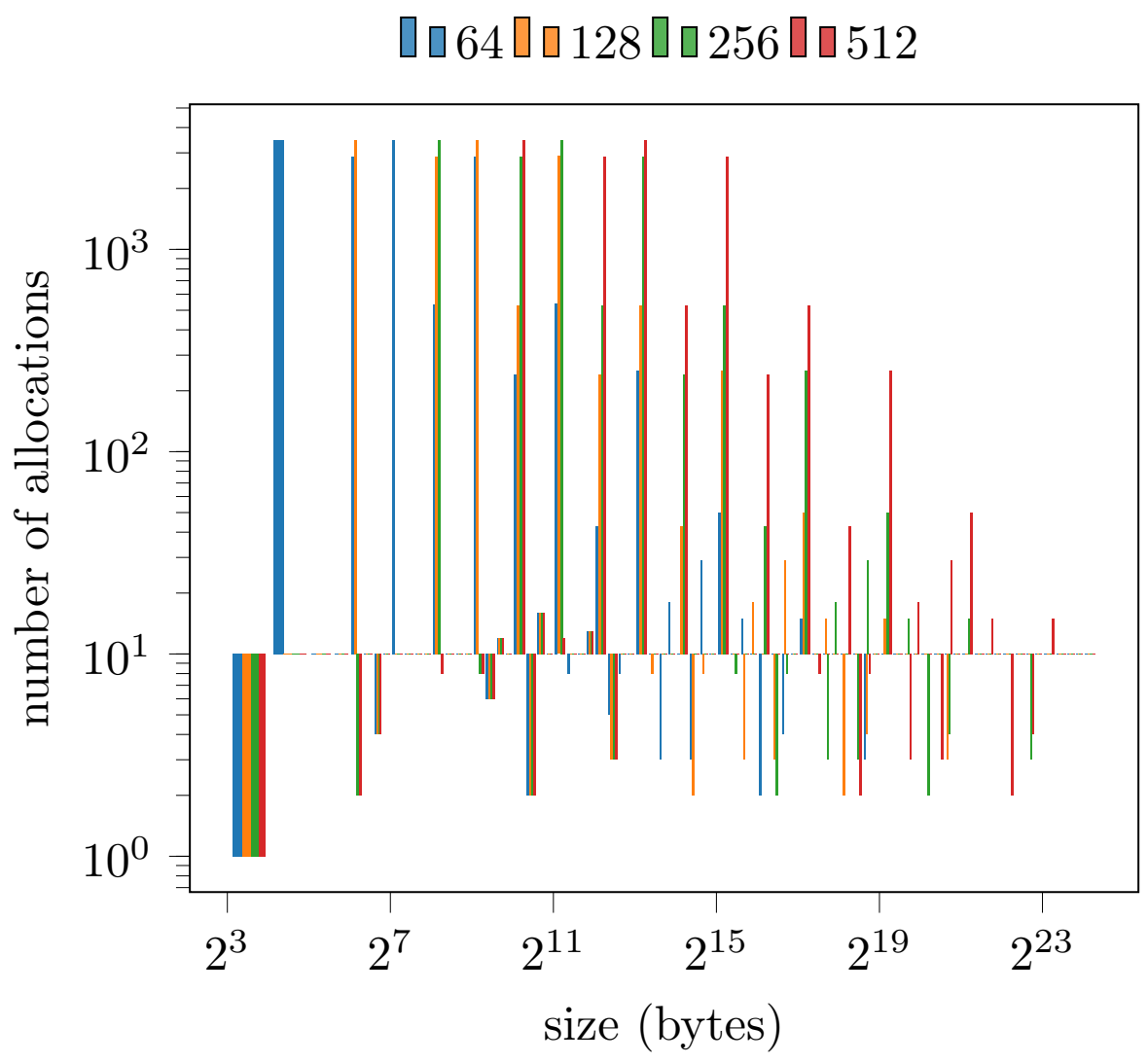

Allocation distributions for `mobilenet_v2`

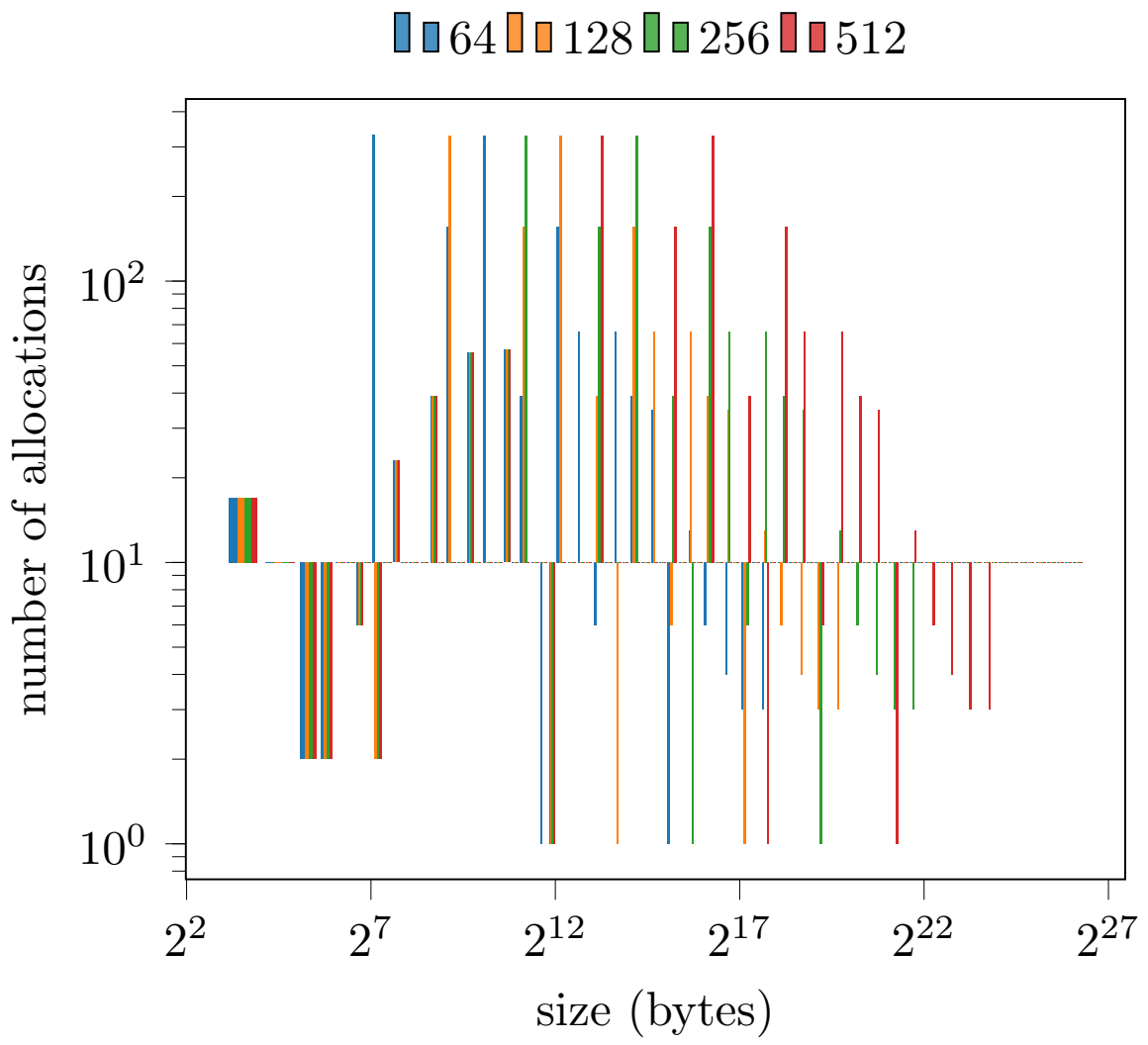

Allocation distributions for `regnet_y_400mf`

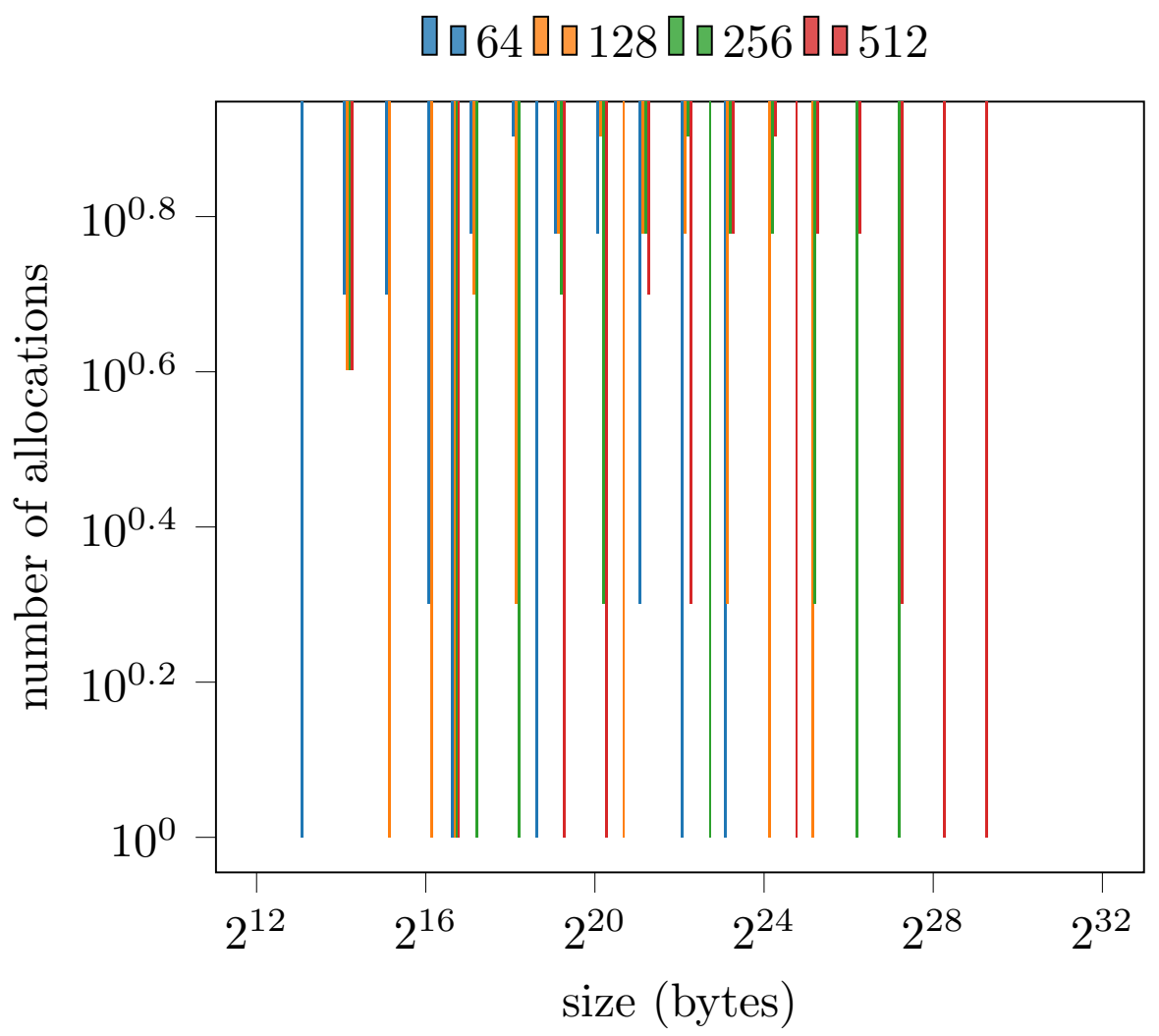

Allocation distributions for `vgg13`

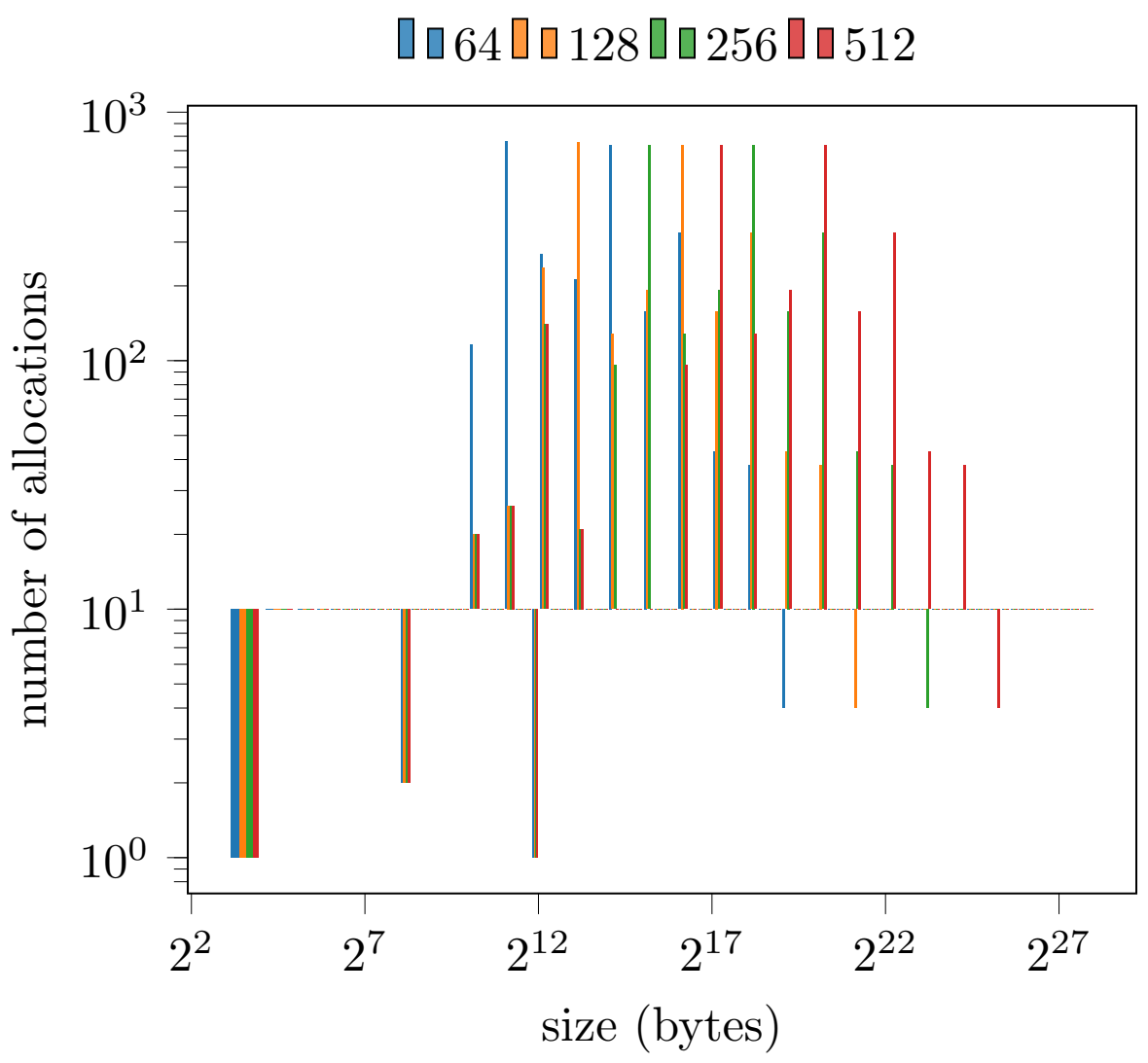

Allocation distributions for `resnext101_32x8d`

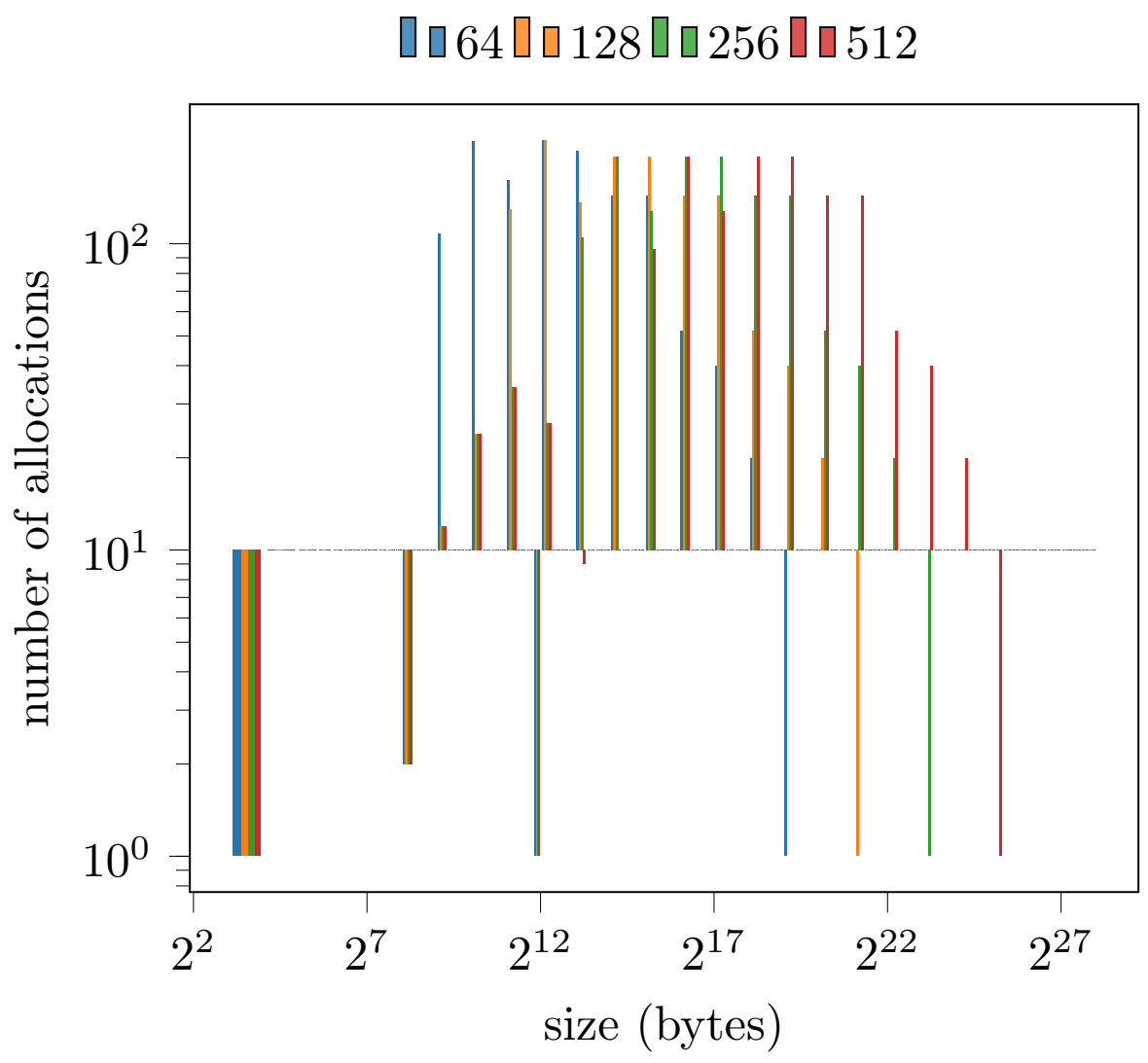

Allocation distributions for `resnext50_32x4d`

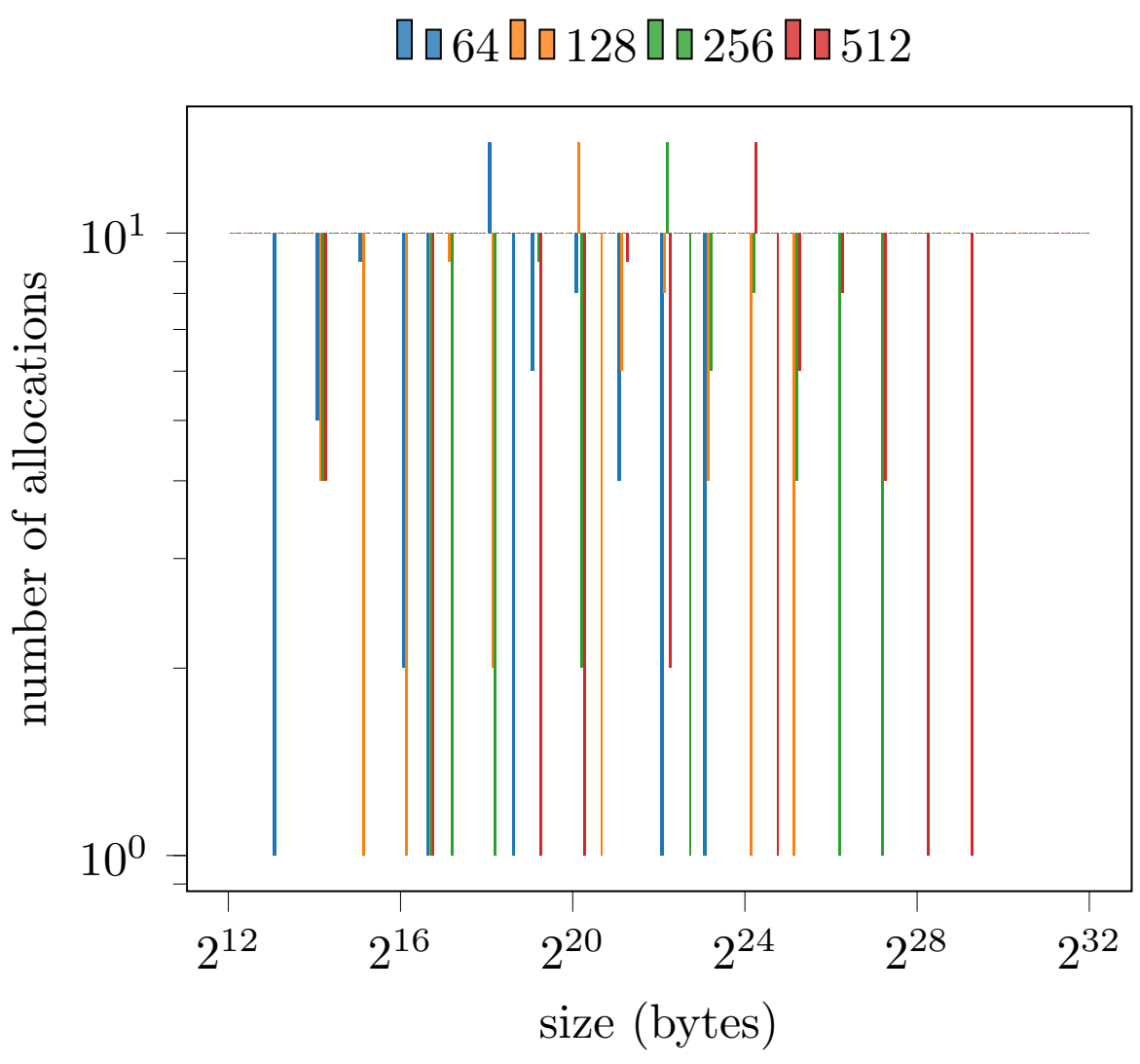

Allocation distributions for `vgg19`

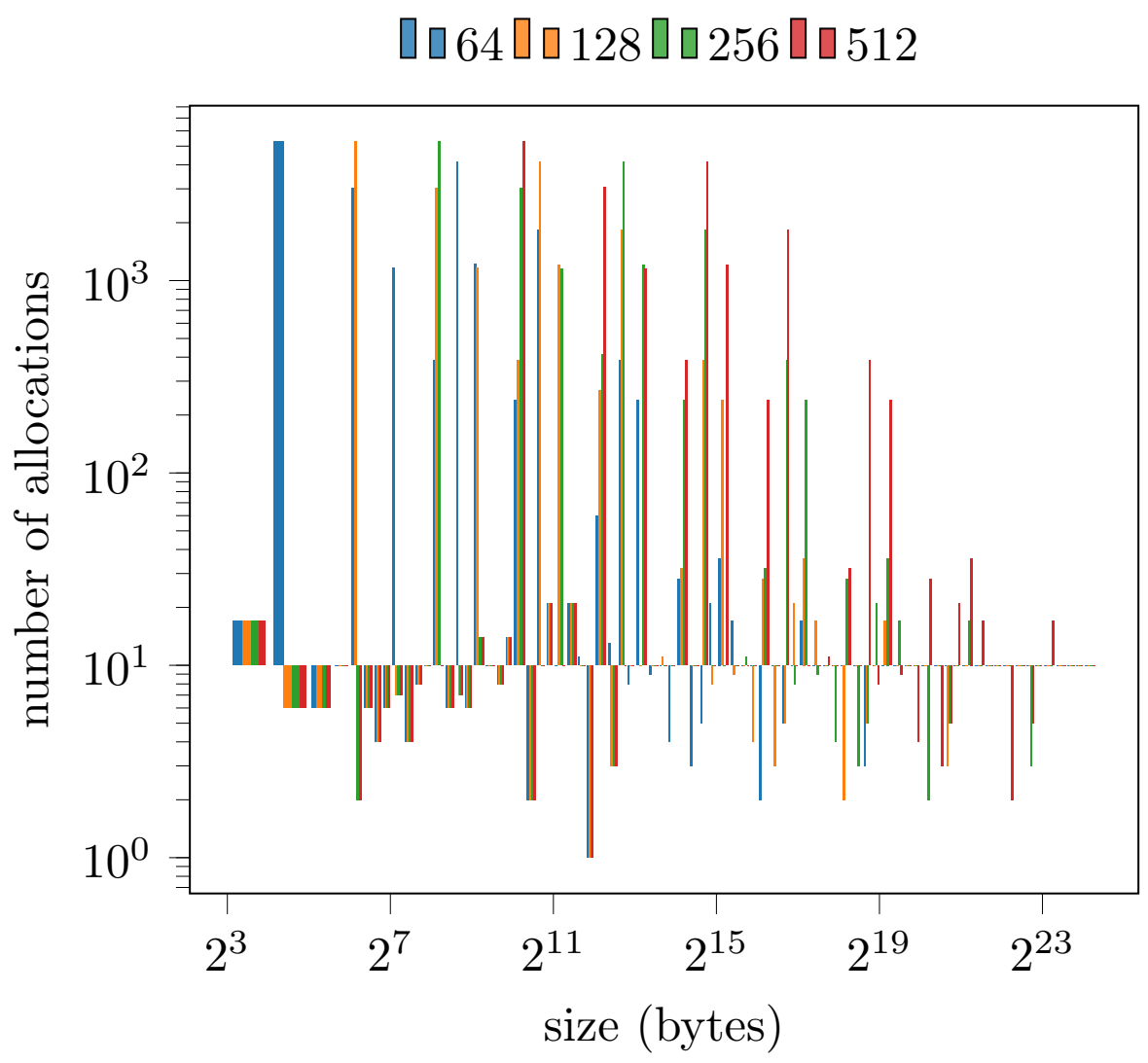

Allocation distributions for `efficientnet_b0`

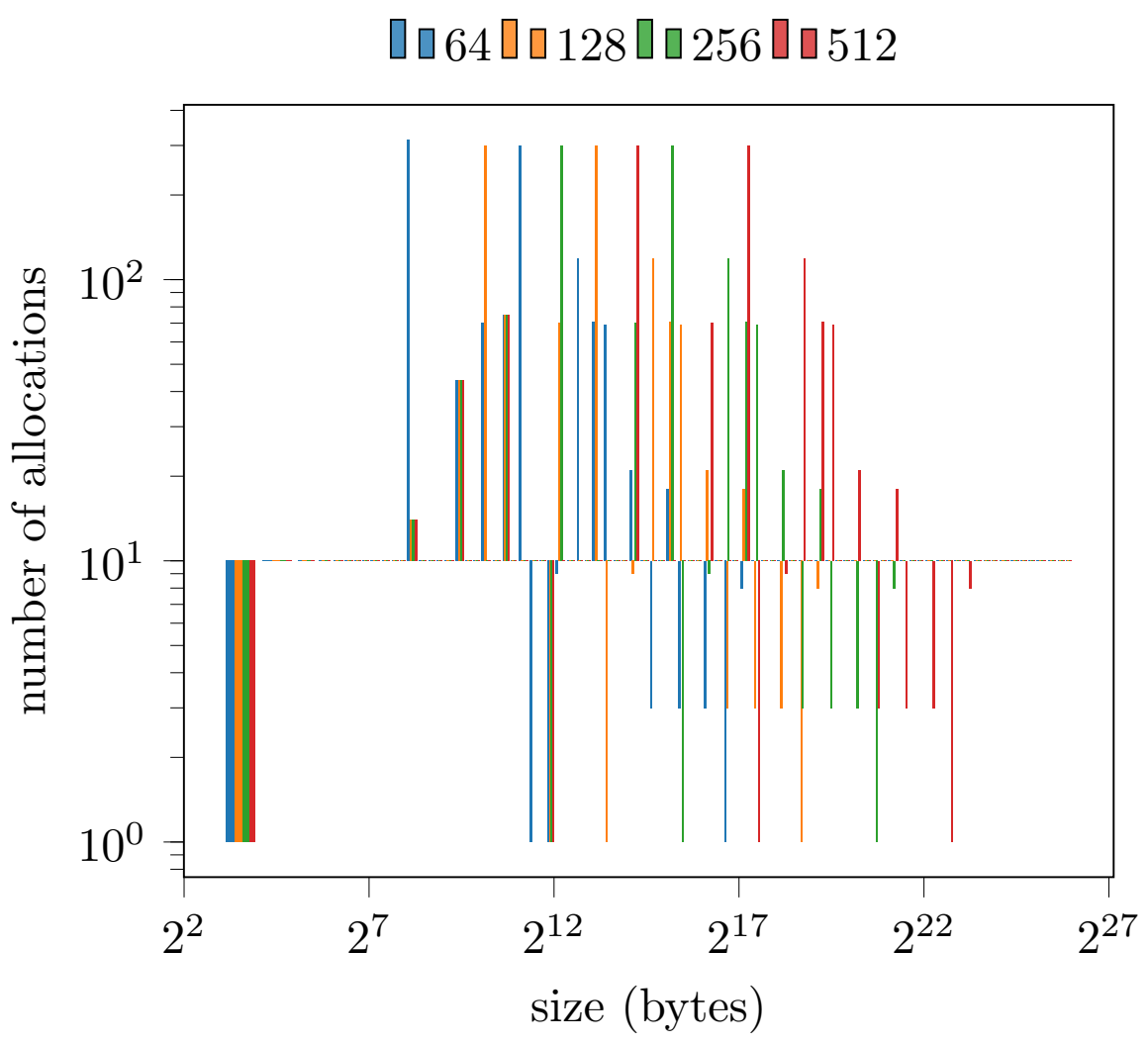

Allocation distributions for `regnet_x_400mf`

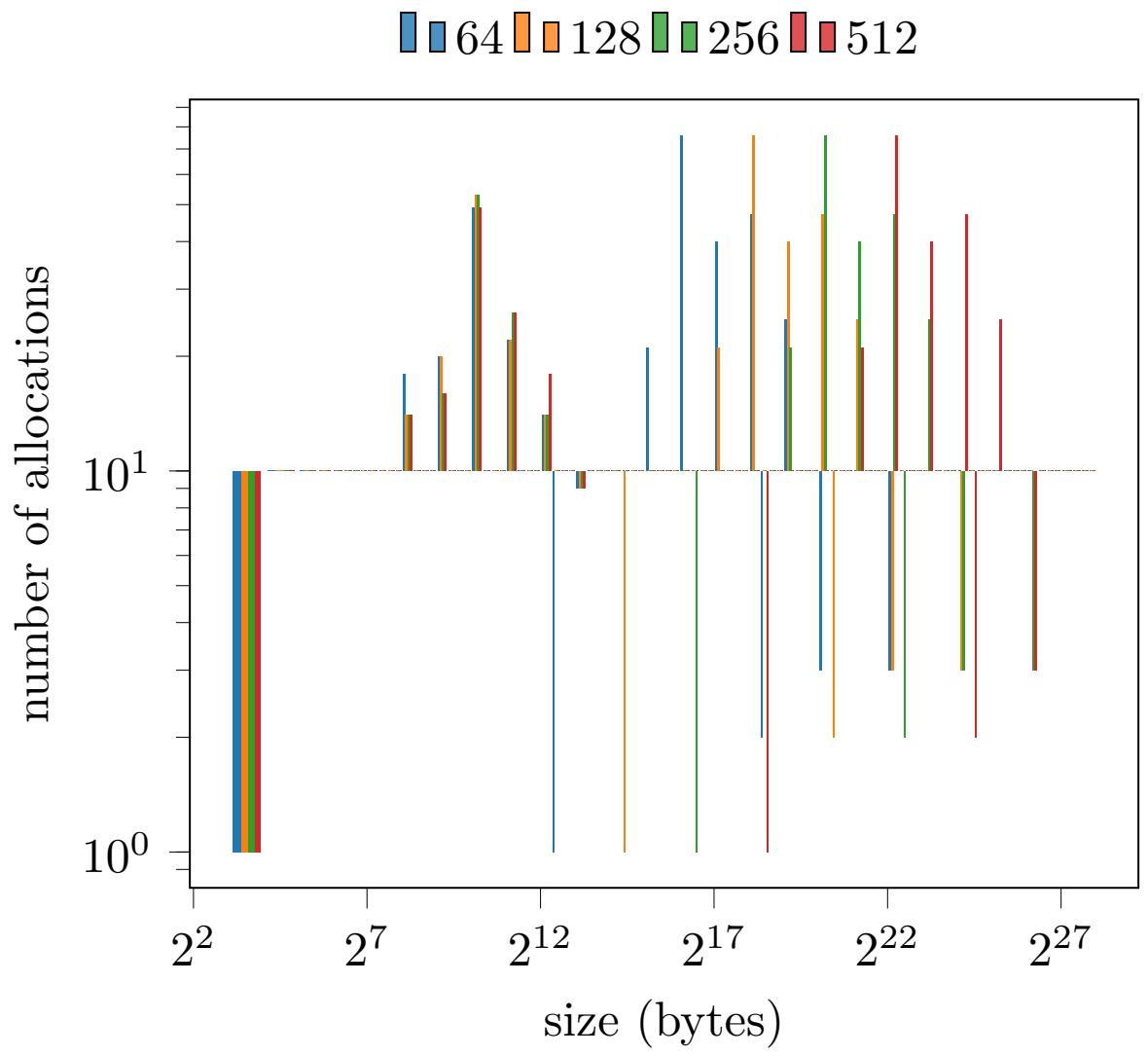

Allocation distributions for deeplabv3_resnet50

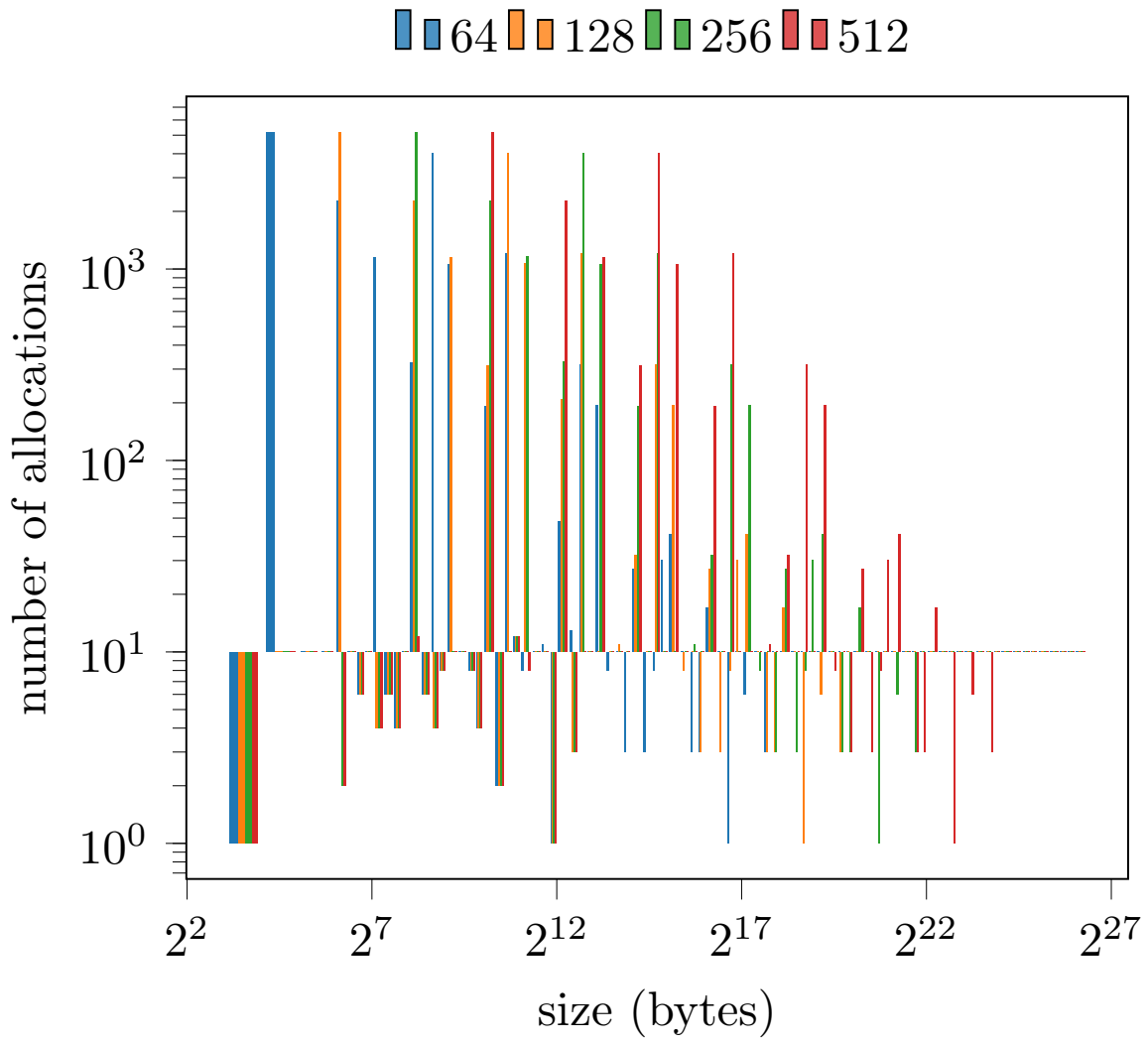

Allocation distributions for mnasnet1_0

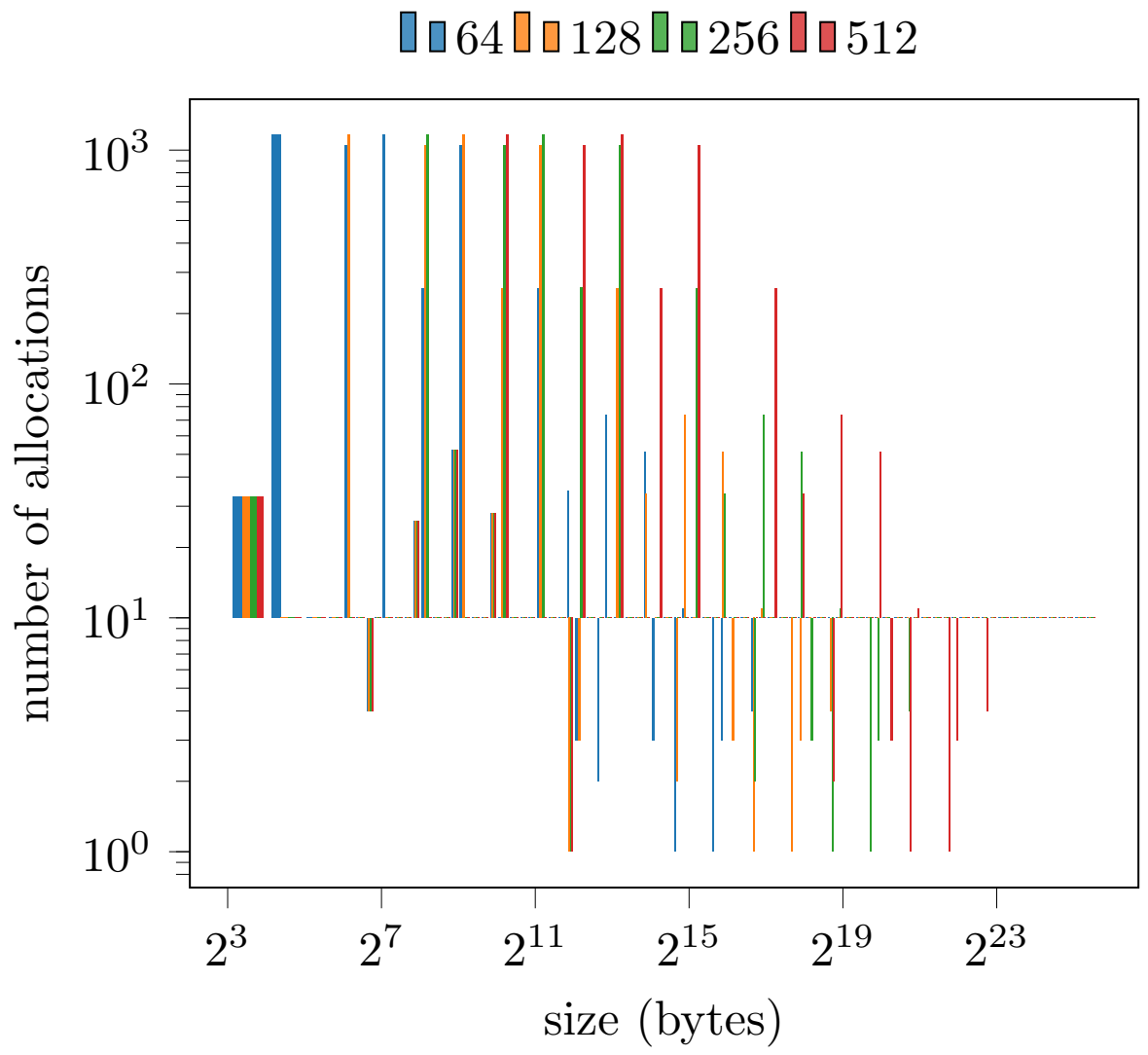

Allocation distributions for shufflenet_v2_x1_0

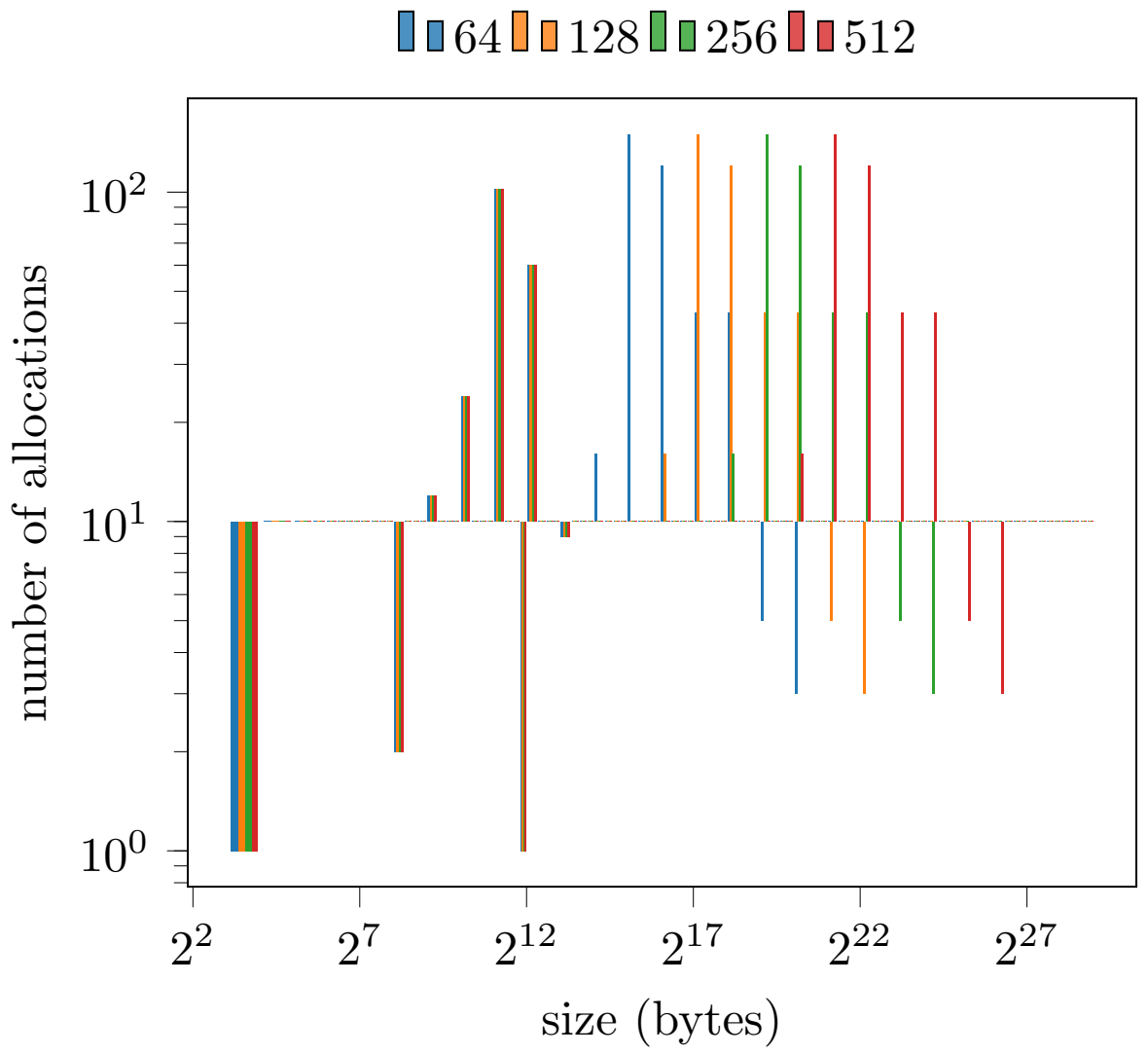

Allocation distributions for wide_resnet101_2

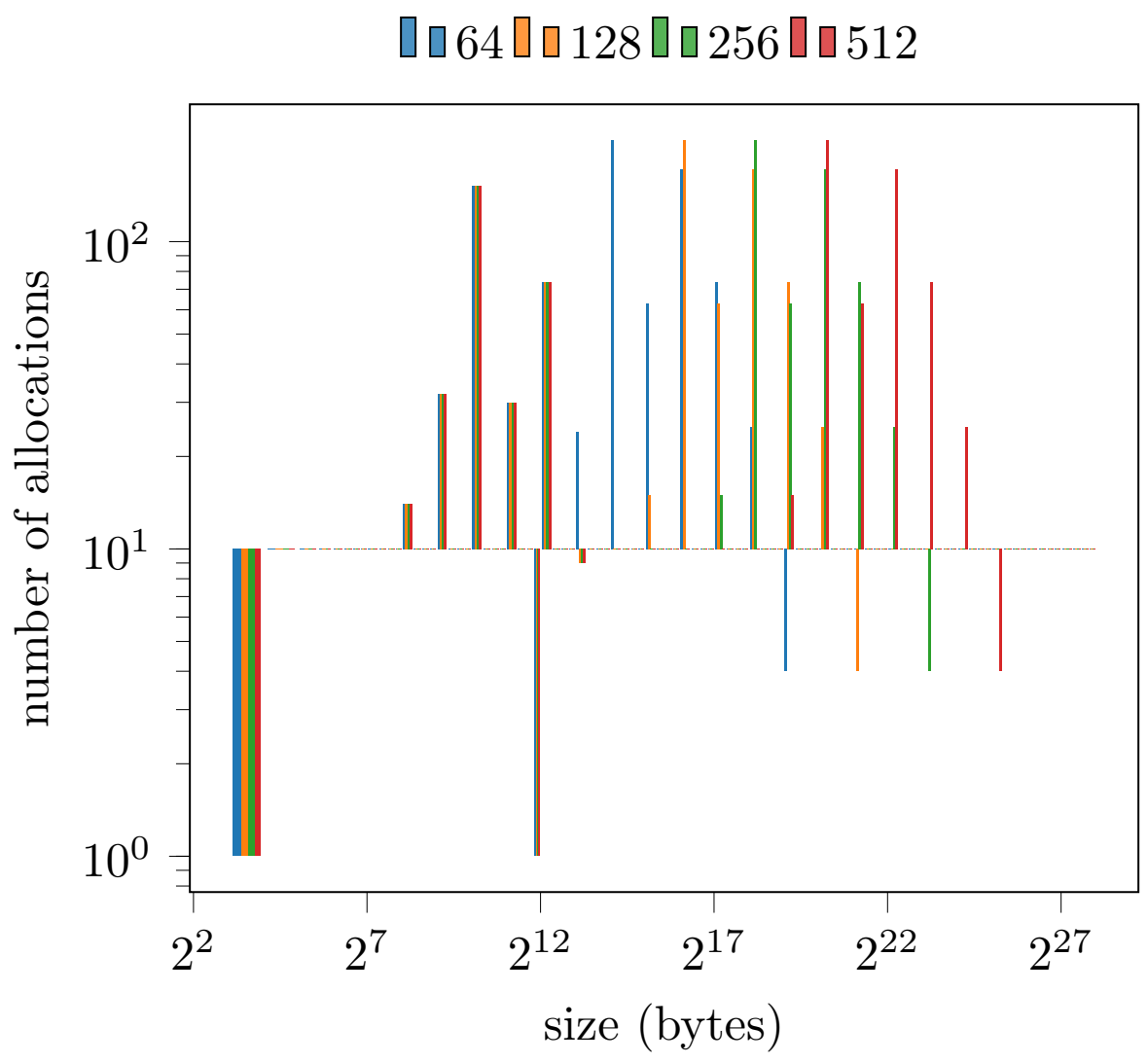

Allocation distributions for `resnet152`

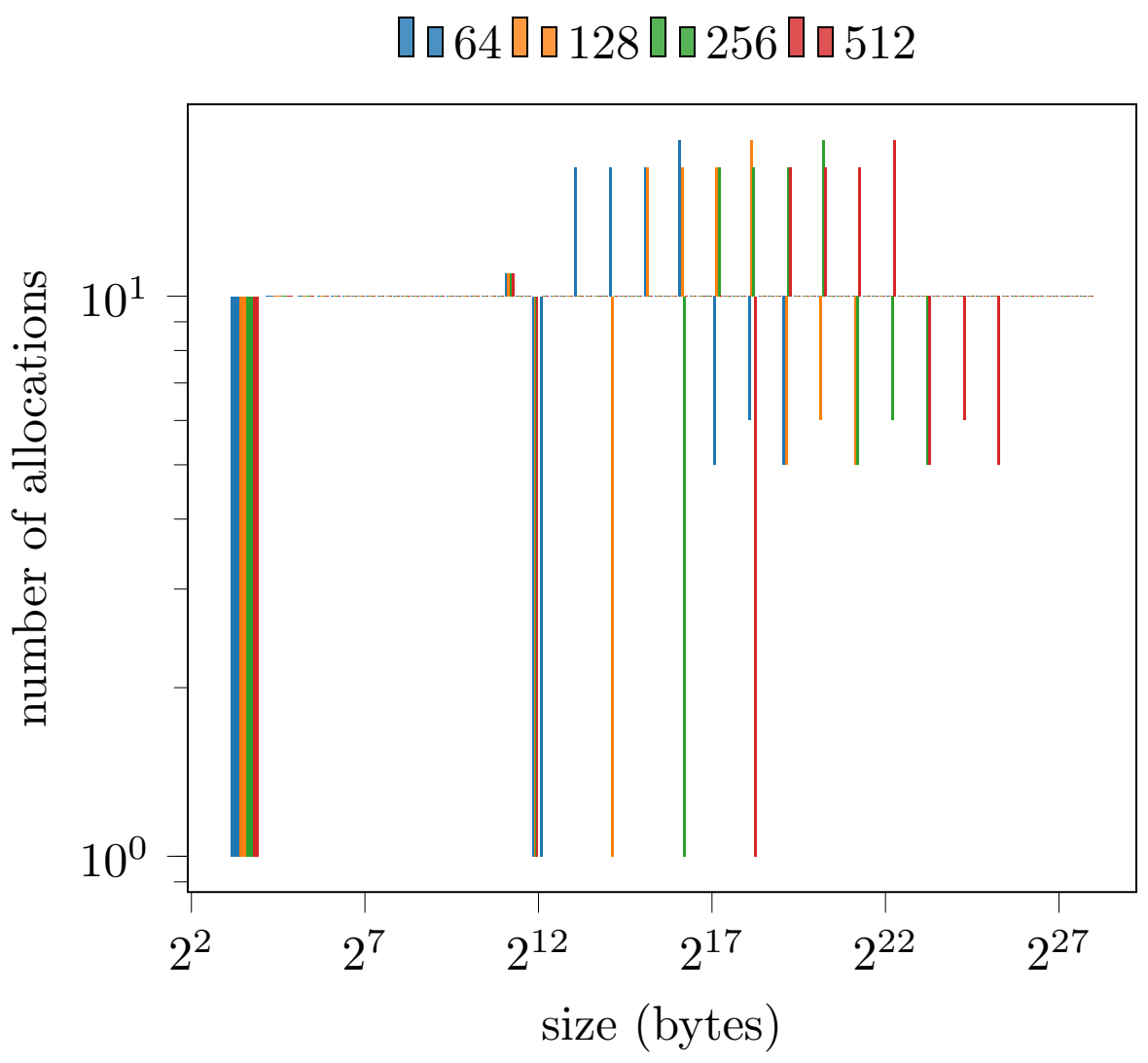

Allocation distributions for `resnet18`

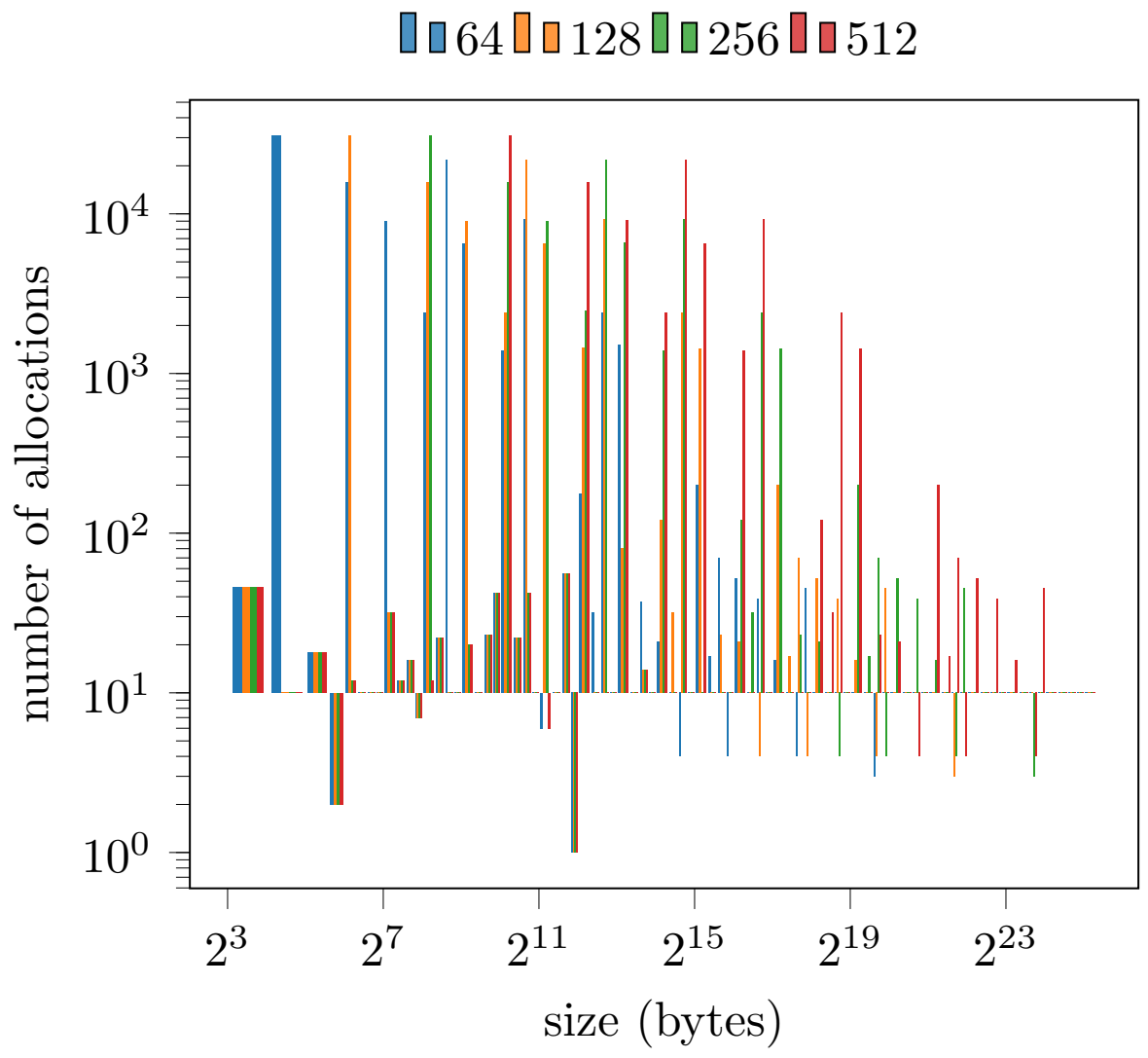

Allocation distributions for `efficientnet_b6`

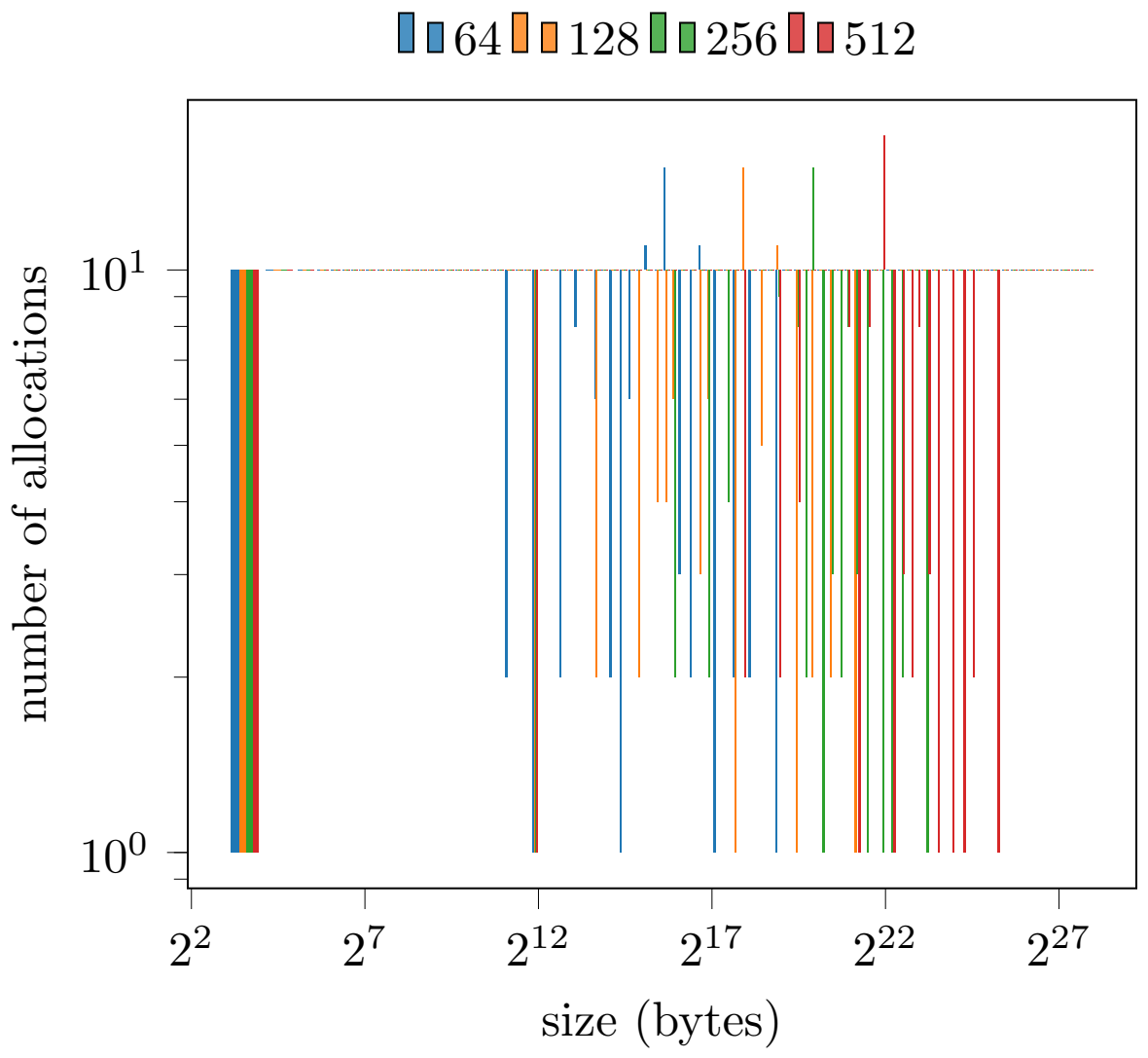

Allocation distributions for `squeezenet1_0`